\documentclass[a4paper, 12pt]{article}
\usepackage{graphicx}
\usepackage{subfig}
\usepackage{float} 
\usepackage{hyperref}
\usepackage{setspace}
\usepackage{url}
\usepackage{amssymb} 
\usepackage{xcolor}
\usepackage{fancyvrb}
\usepackage{adjustbox}
\usepackage{array}     
\usepackage{fbox}
\usepackage{caption}
\usepackage{subcaption} 
\usepackage{amsmath}
\usepackage[numbers]{natbib}

\usepackage{placeins}  
\usepackage{url}
\usepackage[utf8]{inputenc}
\usepackage{listings}
\usepackage{paracol}
\usepackage[utf8]{inputenc}
\usepackage[most]{tcolorbox}
\usepackage{titlesec} 
\definecolor{mygray}{gray}{0.95}
\usepackage{wrapfig}
\usepackage{capt-of}
\usepackage{authblk}
\usepackage[left=1in, right=1in, top=1in, bottom=1in]{geometry}
\title{\textsc{Impact of Noise on LLM-Models Performance in Abstraction and Reasoning Corpus (ARC) Tasks with Model Temperature Considerations}}

\begin{document}
\author[1]{Nikhil Khandalkar}
\author[1]{Krishna Shinde}
\author[1]{Pavan Yadav} 
\author[1]{\\Lokesh B. Ramegowda}
\author[2]{Rajarshi Das}
\affil[1]{Enkefalos Technologies} 
\affil[2]{MQube Cognition} 
\affil[1]{\texttt{\{nikhil.khandalkar, krishna.shinde, pavan.yadav, lokeshbr\}@enkefalos.com}}
\affil[2]{\texttt{rajarshi.das@mqube.ai}}

\date{}
\maketitle
\begin{abstract} 
\noindent Recent advancements in Large Language Models (LLMs) have sparked interest in their structured reasoning capabilities, particularly in abstraction and pattern recognition tasks. The Abstraction and Reasoning Corpus (ARC) benchmark serves as a key evaluation tool for assessing AI models' ability to generalize and solve novel reasoning tasks. While GPT-4o successfully solves all ARC tasks at zero noise, models such as DeepSeek R1 and LLaMA 3.2 fail to solve any, raising questions about their abstraction and generalization capabilities beyond pattern matching. To investigate this further, we evaluate these models under varying noise levels and temperature settings. Our findings indicate that introducing noise significantly degrades performance across all models, underscoring their fragility under uncertain conditions. This suggests that while some models demonstrate reasoning abilities, they remain highly sensitive to input perturbations, limiting their robustness. By analyzing how different architectures handle noise and uncertainty, we provide insights into the limitations of current AI systems in structured reasoning. Our study highlights the need for more resilient AI models that can adapt to real-world complexity, informing future research on improving generalization, robustness, and alignment with human cognitive flexibility.
\footnote{code: \url{https://github.com/Enkefalos-Technologies/arc-challenge}}
\end{abstract} 
 
\section{Introduction}
As AI systems advance in solving complex reasoning tasks, evaluating their ability to generalize and align with human-like problem-solving strategies becomes crucial. The Abstraction and Reasoning Corpus (ARC) Challenge, introduced by Franois Chollet serves as a benchmark for assessing an AI model’s capacity to perform abstract reasoning, a skill fundamental to human intelligence. Unlike traditional machine learning tasks that rely heavily on pattern recognition over large datasets, the ARC Challenge emphasizes few-shot learning, generalization, and abstraction, requiring models to infer underlying rules from minimal examples ~\cite{chollet2019}.
The ARC Challenge consists of diverse problem-solving tasks that demand conceptual reasoning, pattern recognition, and rule abstraction skills that closely resemble human cognitive processing. Each task presents input-output examples demonstrating a transformation rule, and the AI model must deduce the correct rule to generalize to new, unseen test cases. Since human cognition excels at such tasks through structured representation and high-level abstractions, ARC provides an ideal testbed for evaluating the alignment between AI models and human-like reasoning mechanisms.
Early approaches to solving ARC primarily relied on symbolic AI and program synthesis  using manually defined heuristics to infer rules from example pairs~\cite{johnson2020}. However, these methods faced scalability issues, as they required handcrafted representations for each task. With the rise of deep learning, researchers explored the application of convolutional neural networks (CNNs), transformers, and large language models (LLMs) to ARC~\cite{mitchell2021abstraction}. While some studies demonstrated moderate success using pretrained language models, they often relied on pattern recognition rather than true abstraction, failing to solve tasks that required extrapolation beyond their training distribution~\cite{tervo2022}. More recent research evaluated and GPT-4o on ARC showing improved performance but still highlighting significant limitations, particularly in handling noise and reasoning under uncertainty.
A major challenge identified in prior work is the lack of robust generalization. Studies have shown that LLMs trained on large text corpora struggle to infer abstract rules in a structured reasoning setting. Additionally, research on model robustness suggests that introducing noise into input examples leads to significant performance drops, indicating that current models are highly sensitive to minor perturbations~\cite{iteration_assay}.
Building upon these findings, our study conducts a comprehensive evaluation of GPT-4o, and LLaMA 3.2 on ARC tasks under different noise levels and temperature settings. By analyzing model performance in these varied conditions, we aim to provide deeper insights into the role of structured reasoning in LLMs and their limitations in handling uncertainty.~\cite{mitchell2023models}. Our work extends prior research by systematically quantifying how noise affects reasoning performance and identifying the architectural differences that contribute to these limitations.
To address these concerns, we conduct a comprehensive evaluation of state-of-the-art models, including GPT-4o. LLaMA 3.2, under varying noise levels and model temperature settings.
\section{Motivation}
The Abstraction and Reasoning Corpus (ARC) benchmark serves as a fundamental testbed for evaluating an AI model’s ability to infer abstract patterns and solve problems requiring human-like reasoning. Unlike traditional machine learning benchmarks that rely on large-scale data-driven pattern recognition, ARC challenges models to generalize from a limited number of examples using conceptual reasoning ~\cite{chollet2019}. This ability to generalize is a hallmark of human intelligence but remains a significant challenge for modern AI systems ~\cite{mitchell2021abstraction}.  
In real-world applications, AI systems frequently encounter noisy, ambiguous, or incomplete data. However, current models, including large language models (LLMs) and deep learning-based approaches, often struggle to maintain robust performance under such conditions. Their reliance on statistical correlations rather than genuine abstraction makes them vulnerable to disruptions caused by small perturbations. Recent studies have demonstrated that AI models, including GPT-4o and multimodal models, perform inconsistently when abstract reasoning is required. These findings highlight the need for more structured approaches that enhance conceptual understanding and generalization~\cite{illusions_of_understanding}. By systematically introducing noise into ARC tasks, we aim to evaluate the extent to which AI models can adapt to uncertainty and maintain reasoning accuracy. Research on ARC and related benchmarks suggests that human cognition is significantly more robust when handling abstraction and generalization compared to existing AI models ~\cite{mitchell2023gpt4}. This analysis not only highlights the weaknesses of existing approaches but also helps guide future advancements in AI architectures that are more resilient to real-world imperfections . The findings from this study will contribute to the broader goal of developing AI systems that can reason abstractly, much like humans, while also improving their robustness in dynamic environments.  

\raggedbottom 
\linespread{0.92} 


\subsection{The Importance of Studying Noise in ARC Tasks}
\noindent
Human cognition is inherently resilient to small distortions in information, allowing us to infer missing details and adapt to uncertainty. In contrast, AI models often struggle with unexpected variations in data. The ARC dataset consists of grid-based tasks, where each grid is a rectangular grid (list of lists) of integers between 0 and 9, representing different patterns and structures. Grid sizes range from 1×1 to 30×30, presenting a diverse set of reasoning challenges.

For our research, we specifically analyze 2-shot and 3-shot examples, where the model is given two, three input-output pairs as demonstrations before making predictions on a test case. However, despite selecting a fixed number of demonstrations (2-shot, 3-shot), our approach is designed to handle ARC tasks of any dimension, ensuring that our findings apply broadly across different task complexities.

By introducing controlled noise in the input or output grids, we systematically evaluate how AI models maintain their reasoning ability under increasing levels of uncertainty. This allows us to quantify the extent to which GPT-4o can generalize beyond clean examples and whether it can infer logical rules even when data is perturbed.

\vspace{1em} 
\subsection{Investigating the Impact of Model Temperature on Performance}
\noindent
The temperature parameter in AI models plays a critical role in determining prediction variability. A lower model temperature setting encourages deterministic outputs, ensuring stability and consistency in responses. Conversely, increasing the model temperature introduces greater randomness, which can enhance diversity in generated outputs but may also lead to instability and erratic behavior.

By exploring how varying temperature settings interact with different noise levels, we aim to understand whether increased randomness aids or hinders AI performance in structured reasoning tasks. This analysis is essential for optimizing model configurations to balance accuracy, creativity, and robustness.

\vspace{1em} 
\subsection{Implications for AI Development}
\noindent
Understanding how AI models perform under noisy conditions and varying model temperatures has significant implications for real-world AI applications. AI systems must be capable of handling imperfect data, adapting to uncertainty, and generalizing beyond seen examples capabilities that remain challenging for many models~\cite{mitchell2023models}.

In our study, we evaluate the performance of GPT-4o, LLaMA 3.2, and DeepSeek-R1 on ARC tasks. While GPT-4o demonstrates the ability to solve these reasoning-based tasks, both LLaMA 3.2 and DeepSeek-R1 fail to solve any task, highlighting a significant gap in grid-based reasoning capabilities across different model architectures. This suggests that models trained primarily on language-based tasks (like LLaMA 3.2 and DeepSeek-R1) may struggle with pattern-based and spatial reasoning, which require structured abstraction processes beyond text-based learning.

By analyzing GPT-4o’s performance on 2-shot, and 3-shot ARC tasks across different noise levels and model temperatures, we gain deeper insights into the strengths and limitations of different model architectures. These findings can inform future AI research, particularly in enhancing structured reasoning abilities, improving robustness to noise, and developing models that better align with human-like problem-solving skills.

\section{Methodology}
To evaluate the impact of noise and temperature variations on AI models in solving ARC tasks, we conducted a structured experiment using GPT-4o, and LLaMA 3.2.

\subsection{Selection of Tasks}

The ARC dataset consists of 400 diverse tasks, each requiring varying levels of pattern recognition, abstraction, and reasoning. Given the extensive number of tasks, evaluating all of them under different conditions (such as model variations, noise levels, and temperature settings) would be computationally expensive and time-consuming. Therefore, a subset of 7 tasks it contains three 2-shots and four 3-shots was carefully selected based on the following criteria. We selected these tasks from the dataset source \href{https://arcprize.org/}{https://arcprize.org/}.\\ \newline
\textbf{Selection Process:} We follow the below process to choose above 7 tasks.
\begin{enumerate}   

\item \textbf{Baseline Solvability by Strong Models:} We selected tasks that were successfully solved at noise level 0 by at least one state-of-the-art model, such as GPT-4o.
\item \textbf{Empirical Task Selection via Trial Analysis:} Each task was evaluated over 30 trials using the GPT-4o model. Only those tasks which the model solved correctly in more than 50\% of the trials were considered. This criterion ensures that the selected tasks demonstrate consistent solvability by a strong model, providing a reliable basis for comparative evaluation under noisy and multi-shot conditions.
\item \textbf{Challenge for Weaker Models:} LLaMA 3.2 and DeepSeek-R1 failed to solve any of the selected tasks at noise level 0, highlighting the performance gap between different model architectures. This contrast helps in understanding differences in structured reasoning across models.
\item \textbf{Controlled Experimentation:} Selecting a small but representative set of tasks allows for systematic noise introduction and analysis without overwhelming computational resources. This enables deeper insights into how each model responds to structured perturbations.
\item \textbf{Relevance to ARC’s Core Challenges:} The selected tasks align with the fundamental principles of ARC, including abstract reasoning, pattern recognition, and logical generalization. This ensures that the results remain relevant to evaluating AI’s reasoning abilities.

\end{enumerate}
By selecting these 7 tasks based on the above factors, our study provides meaningful insights into how different models handle structured reasoning and noise, without the impracticality of evaluating all 400 tasks.
Here are the ids of the selected tasks: \texttt{272f95fa, 539a4f51, aabf363d, bda2d7a6-a, bda2d7a6, bdad9b1f, cbded52d}.
The pictorial representation of task id \texttt{272f95fa} without noise is shown in Figure~\ref{fig:figure0}.

\begin{center}
        \centering
        
        \vspace{1em}
        \includegraphics[width=1.0\textwidth]{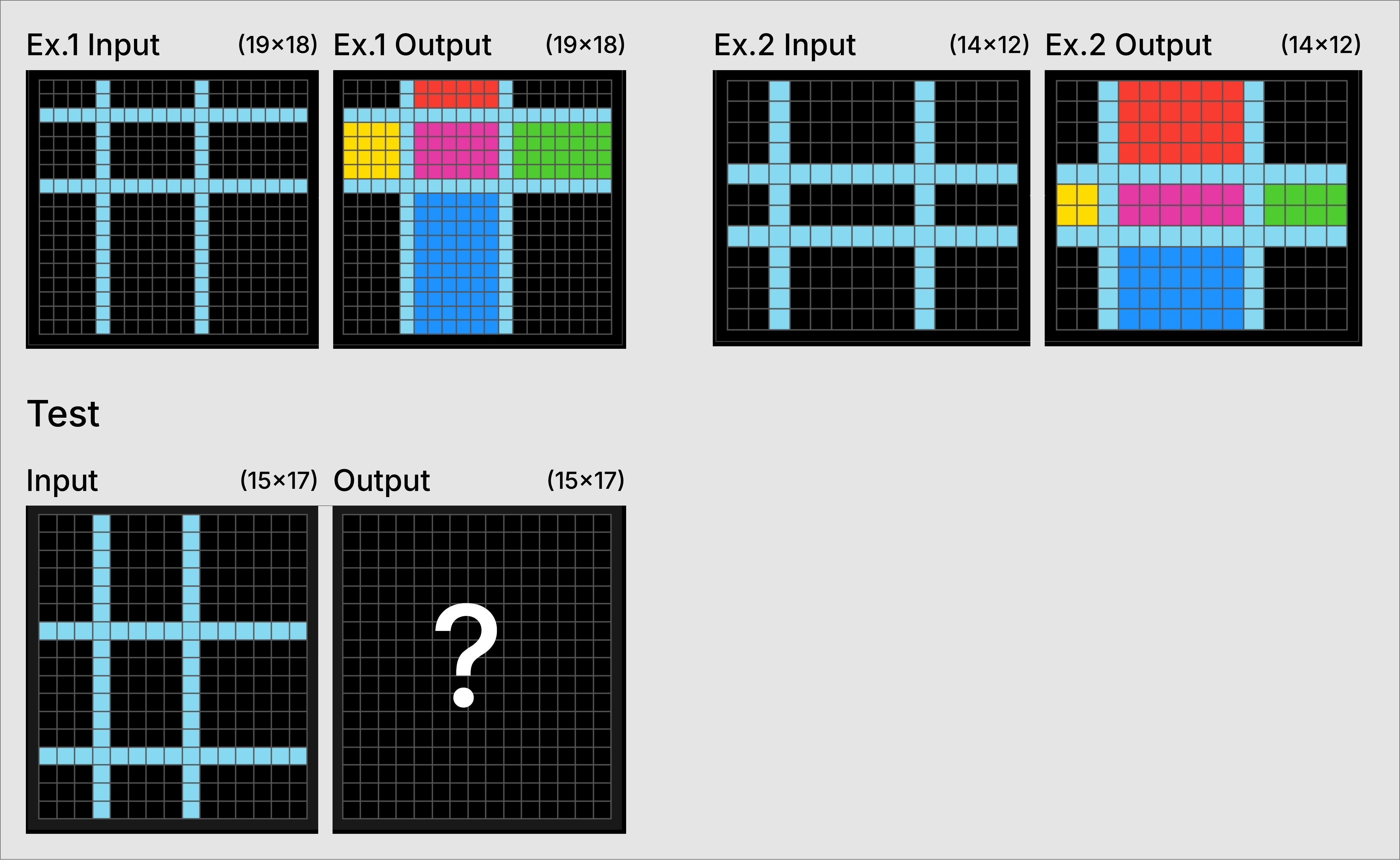}
        \captionof{figure}{Shows pictorial representation of task 272f95fa with 2-shot examples. Without introdusing the noise. }
        \label{fig:figure0}
        
        \vspace{1em} 
        
        
\end{center}

\subsection{Introducing Noise to ARC Tasks}

To systematically evaluate the robustness of AI models in reasoning-based tasks, we introduced structured noise into the ARC dataset. This noise was applied only to tasks that were correctly solved by GPT-4o at noise level 0 (i.e., in their original, unaltered form). Our objective was to assess how increasing noise affects model performance and whether it degrades the model’s ability to infer correct logical transformations.

We introduced two types of noise:
(a) Noise added to the input grid (while keeping the output grid unchanged).
(b) Noise added to the output grid (while keeping the input grid unchanged).

Each approach serves a unique purpose in analyzing the robustness of transformations between input and output grids. The following subpoints describe the noise injection process in detail.
\begin{enumerate}
\item\textbf{Noise added to the input grid in the prompt:} 
    In this scenario, noise is introduced only in the input grid, while the output grid remains unchanged.

\textbf{Process}

\begin{enumerate}
    \item Randomly select \(\mathbb{T} \) positions in the input grid.
\item For each selected position \( (i, j) \), representing the \(i^{\text{th}}\) row and \(j^{\text{th}}\) column of the grid:
    \begin{itemize}
        \item Replace the original value with a different number from the set of unique values from the input or output grid.
        \item Ensure that the new value is not identical to the original.
    \end{itemize}
    \item The output grid is kept unchanged to analyze the system’s ability to handle noisy inputs while maintaining correct outputs.
\end{enumerate}
 
    \item \textbf{Noise added to the output grid in the prompt:} 
    In this scenario, noise is introduced only in the output grid, while the input grid remains unchanged.

\textbf{Process}

\begin{enumerate}
    \item Randomly select \( \mathbb{T} \) positions in the output grid.
   \item For each selected position \( (i, j) \), representing the \(i^{\text{th}}\) row and \(j^{\text{th}}\) column of the grid:
    \begin{itemize}
        \item Replace the original value with a different number from the set of unique values from the input or output grid.
        \item Ensure that the new value is not identical to the original.
    \end{itemize}
    \item The input grid is kept unchanged to analyze the system’s ability to handle noisy outputs while maintaining correct inputs.

\end{enumerate}
\item\textbf{Extracting Unique Values from Grids}

Before applying noise, it is essential to identify all possible values present in both input and output grids. This ensures that modifications remain within the existing numerical structure.

 \item \textbf{Parsing and Extracting Unique Values}

Each grid consists of numerical values arranged in rows.
All unique values from both input and output grids are collected.
A set operation is used to combine these values, ensuring that any introduced noise consists only of numbers that already exist within the grids.
By restricting replacements to existing values, we prevent unrealistic alterations that could disrupt the grid’s structure.

\item\textbf{Noise Injection Strategies}

Noise is introduced by modifying numerical values at selected positions based on a predefined noise level \( n \), where \( 0 \leq n \leq 1 \). The total number of modified elements (\(\mathbb{T}\)) is calculated as:

\[
\mathbb{T} = \lfloor n \times T \rfloor
\]

where \(T\) represents the total number of elements in the grid. The selected positions are modified according to one of the three noise injection strategies described below:
\begin{itemize}
    \item \textbf{Understanding the Noise Level Parameter:}
    The noise level is a floating-point value, typically denoted as \( n \) (\( 0 \leq n \leq 1 \)). 
    It represents the fraction of total elements in the grid that will be altered.
    For example, if \( n = 0.2 \), then 20\% of the elements in the dataset will be modified.

    \item \textbf{Computing the Total Number of Elements:} 
    Given an input grid with \( m \) rows and \( n \) columns, the total number of elements (\( T \)) is calculated as:

    \[
        T = m \times n
    \]

    This represents the total count of numerical values in the input grid.

    \item \textbf{Determining the Number of Elements to be Altered:} The number of elements to be altered (\( \mathbb{T} \)) is computed as:
    \[
    \mathbb{T} = \lfloor n \times T \rfloor
    \]
    Here, \( n \) is the noise level, and \( \mathbb{T} \) is the integer count of elements to be altered.
    The floor function \( \lfloor \cdot \rfloor\) ensures that the result is rounded down to the nearest integer so that a valid number of elements are selected.
\end{itemize}
\item\textbf{Selecting Random Elements for Modification}

To ensure uniform distribution, \( \mathbb{T} \) elements are randomly selected from the set of input and output grids different from original values which we have chosen for alteration.
This is done using a random number generator that picks unique (row, column) positions within the grid.
The selection process ensures that modifications are not biased toward any specific region of the grid elements.

The noise injection process avoids excessive modifications that might distort the grid structure.
The selected elements are only replaced with valid values from the extracted unique values from grids above.
This ensures that the modifications are realistic and do not introduce completely foreign data points.







\end{enumerate}
\subsection{Experimental Setup}

In this study, we analyze how increasing the number of \(k\)-shot examples affects the model’s performance when noise is introduced in either the \textbf{input} or \textbf{output}. We systematically increase the \(k\)-shot examples while ensuring that noisy variants of a given input or output remain consistent with their corresponding pair. The experiments follow these configurations:
\newpage
\begin{itemize}
    \item \textbf{\( (1 \times k) \)-shot}:  
    \begin{itemize}
        \item If noise is added to the \textbf{input}, each noisy input grid maps to the same output grid.
        \item If noise is added to the \textbf{output}, the same input grid maps to a noisy output grid.
    \end{itemize}

    \item \textbf{\( (3 \times k) \)-shot}:  
    \begin{itemize}
        \item If noise is added to the \textbf{input}, each original input grid is transformed into three noisy input grids, all mapping to the same output grid.
        \item If noise is added to the \textbf{output}, each original input grid maps to three noisy output grids.
    \end{itemize}

    \item \textbf{\( (9 \times k) \)-shot}:  
    \begin{itemize}
        \item If noise is added to the \textbf{input}, each original input grid is transformed into nine noisy input grids, all mapping to the same output grid.
        \item If noise is added to the \textbf{output}, each original input grid maps to nine noisy output grids.
    \end{itemize}
\end{itemize}
By systematically increasing the number of noisy examples, we aim to evaluate how the model generalizes when exposed to progressively larger sets of perturbed inputs or outputs while keeping the corresponding pairs consistent.
\section{Impact of Noise and Prompt Modifications on Accuracy in ARC Tasks}

\subsection{Effect of Noise}
The model demonstrated high accuracy when provided with the original prompt containing no added noise, achieving a perfect score of \textbf{30/30} in many cases. However, the introduction of even a minimal noise level (0.05\%) led to a \textbf{sharp decline in accuracy}, often reducing it to near-zero. This drastic drop suggests that the model relies heavily on \textit{pattern memorization} rather than generalizable rule learning. The inability to maintain performance in the presence of minor perturbations highlights the model's sensitivity to noise, implying a lack of robustness in abstract reasoning.

\subsection{Effect of \(K\)-Shot Examples}
Increasing the number of \(k\)-shot examples from \textbf{1  $\rightarrow$ 3 $\rightarrow$ 9 } did not consistently improve accuracy. In several cases, accuracy declined as more examples were added. This contradicts the expectation that providing additional demonstrations enhances generalization. The observed decrease in performance may be attributed to two factors:
\begin{enumerate}
    \item \textbf{Confusion due to implicit noise handling:} Since the original prompt did not explicitly mention noise, the model was forced to infer patterns from examples. The presence of noise in training examples may have led to incorrect generalizations.
    \item \textbf{Overfitting to noisy examples:}  Instead of extracting an underlying rule, the model may have memorized noise-influenced patterns, impairing its ability to generalize to new instances.
\end{enumerate}

\subsection{Correct and Partial Matching Criteria}

In this study, we evaluate the accuracy of grid-based outputs by comparing them to target grids and analyzing statistical variations across trials. The evaluation is based on two key criteria: correct matching and partial matching.

The system processes text files containing structured grid data, where each file consists of a target grid and multiple trial grids. A parsing function extracts numerical grids from the textual representation using regular expressions. If the target grid is not found, the file is flagged as invalid.

Correct matching occurs when a trial grid exactly matches the target grid. This is determined using a cell-by-cell comparison:
\begin{equation}
  \text{Match Percentage} = \left( \frac{\text{Matching Cells}}{\text{Total Cells}} \right) \times 100
\end{equation}
A trial grid achieving \textbf{100\% match} is classified as a correct prediction, and the total number of correct predictions is recorded to assess overall accuracy.

Partial matching is evaluated when a trial grid does not achieve a full match but contains correctly predicted cells. The percentage of correctly matched cells is calculated using the same formula as correct matching. To quantify the variation in accuracy across different trials, we compute the mean match percentage and standard deviation.

The \textbf{mean match percentage} represents the average of match percentages across multiple trials:
\begin{equation}
  \mu = \frac{1}{N} \sum_{i=1}^{N} X_i
\end{equation}
where $X_i$ represents the match percentage for the $i$-th trial, and $N$ is the total number of trials.

The \textbf{standard deviation} measures the variability of match percentages:
\begin{equation}
  \sigma = \sqrt{\frac{1}{N} \sum_{i=1}^{N} (X_i - \mu)^2}
\end{equation}
A lower standard deviation indicates consistent performance, while a higher value suggests greater variability in predictions.

For each task and each noise level, we compute the output for 30 trials per task. The match percentage and statistical measures are calculated separately for each noise condition, providing insights into how different noise levels impact performance trends.

All results are stored in an output directory, and performance metrics such as mean accuracy and standard deviation are computed for each dataset, ensuring an efficient and structured evaluation of grid-based accuracy.

These findings indicate that merely increasing \(k\)-shot examples without explicitly addressing noise \textbf{does not ensure better performance} and may, in fact, lead to performance degradation.

\subsection{Role of Model Temperature}
Model temperature played a significant role in accuracy. At \textbf{temperature = 0}, where the model operates deterministically, performance was highest. However, as the temperature increased toward \textbf{1}, accuracy \textbf{progressively declined}. The drop in performance suggests that \textbf{higher randomness disrupts structured reasoning}, making it difficult for the model to follow consistent logical patterns. This is particularly important in tasks like ARC, where structured problem-solving is required rather than diverse generative outputs.

\subsection{Impact of Prompt Modifications}
Introducing explicit information about noise in the modified prompt \textbf{slightly improved accuracy}, particularly at temperature 0. By clarifying that different input grids could still correspond to the same output, the modified prompt allowed the model to \textbf{better interpret variations caused by noise}. This suggests that \textbf{precise instructions and explicit problem constraints enhance model robustness}, leading to improved generalization. However, the improvement was modest, indicating that while prompt engineering helps, it does not entirely mitigate the model’s sensitivity to noise.

\subsection{Pictorial Representation by using original prompt}
In Figure ~\ref{fig:figure2}, we present a pictorial representation of how noise is added in different settings to illustrate its effect on input and output grids
\begin{figure}
    \centering

        \centering
        
        \vspace{1em}
        \includegraphics[width=1.0\textwidth]{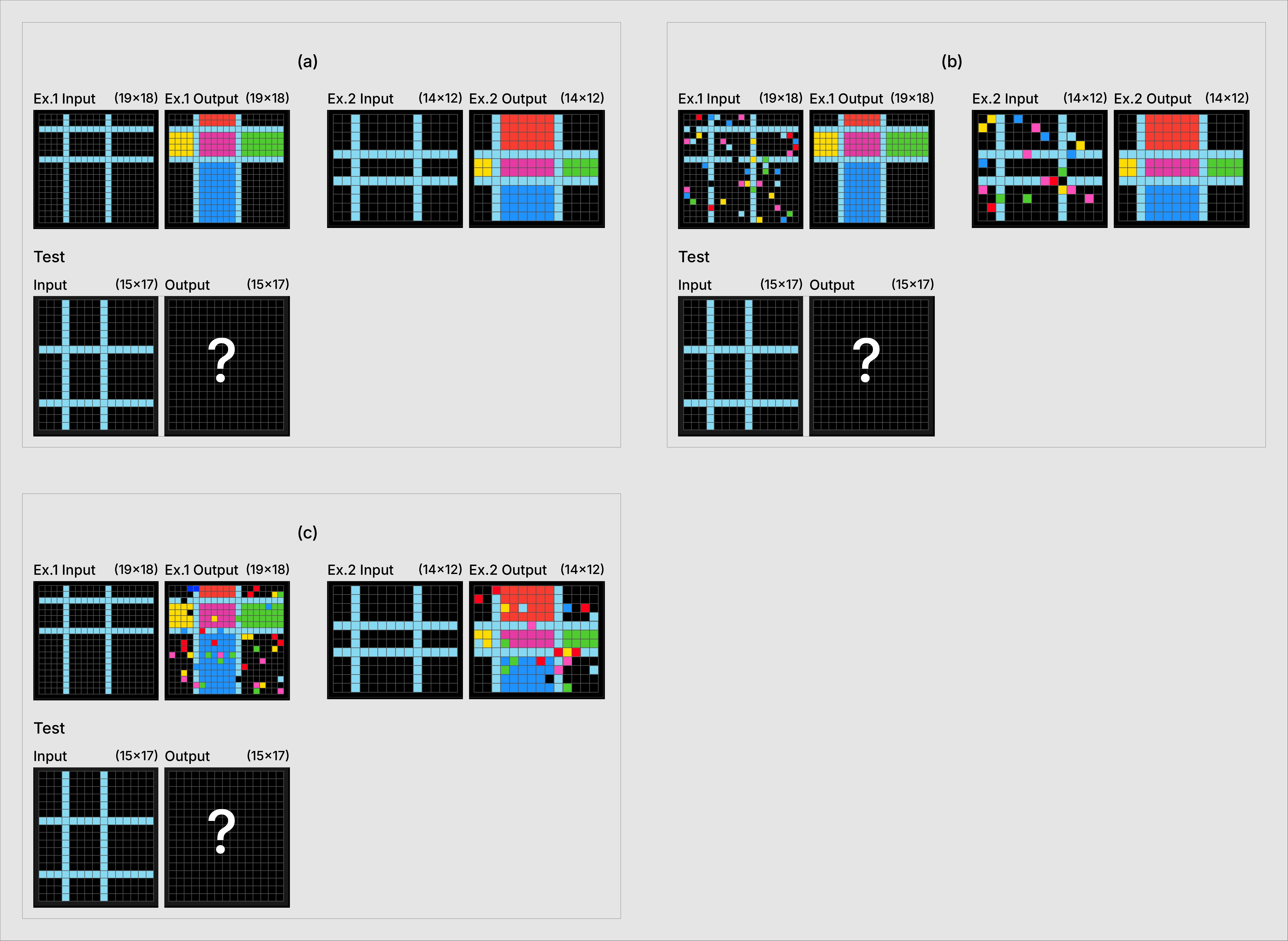}
        \caption{Illustration of the impact of noise on task id 
 \texttt{272f95fa}.  
(a) Represents the original task. (b) Represents same task with 0.125\% of noise in the input grids. (c) Represents same task with 0.125\% of noise in the output grids. The pictorial representation of new prompt is in appendix section~\ref{appendix:effect_noise}}
    \label{fig:figure2}   
\end{figure}       
\subsection{Pictorial Representation by using new prompt}
In Figure ~\ref{fig:figure2}, we present a pictorial representation of how noise is added in different settings to illustrate its effect on input and output grids
\begin{figure}
    \centering

        \centering
        
        \vspace{1em}
        \includegraphics[width=1.0\textwidth]{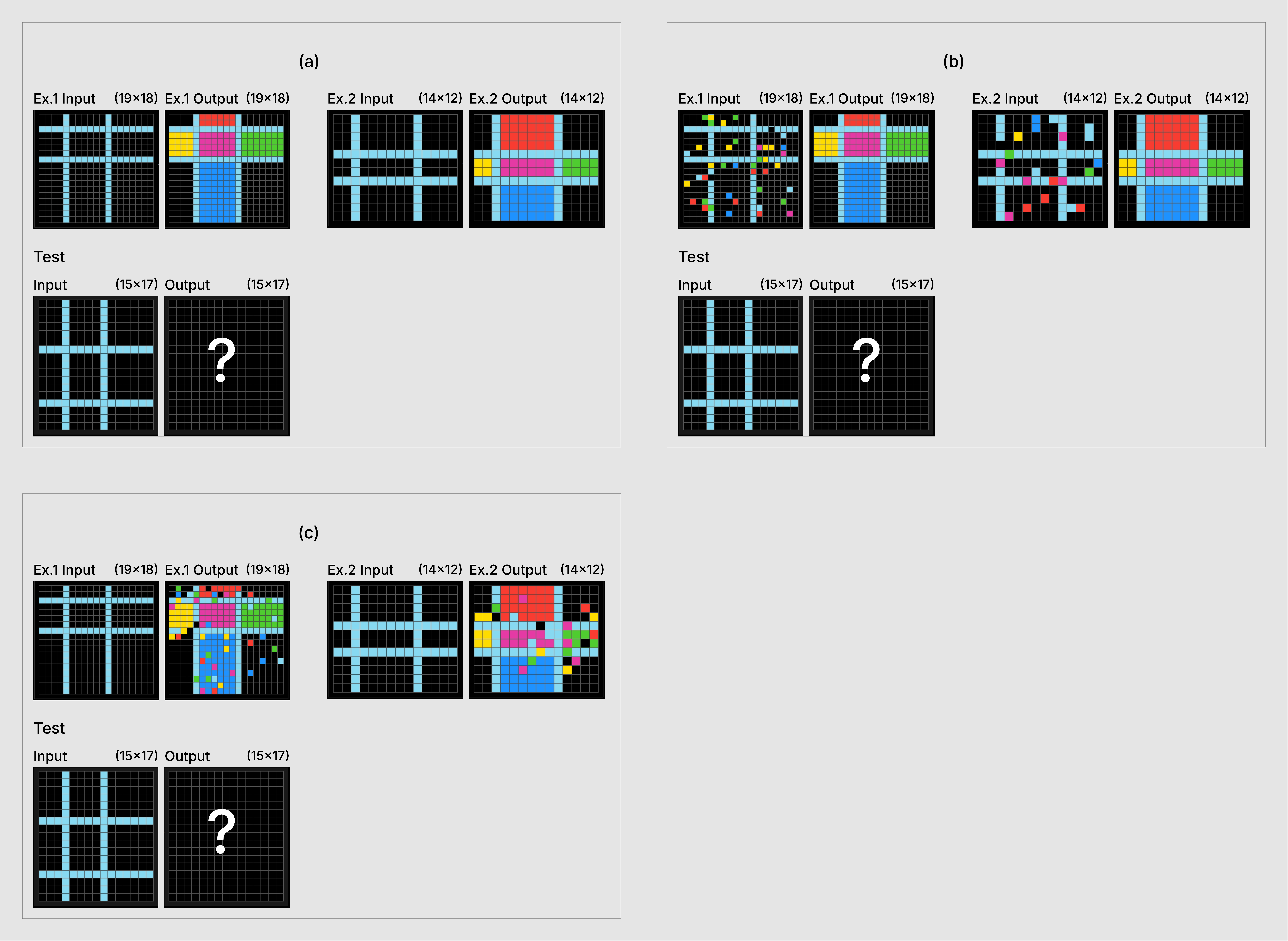}
        \caption{Illustration of the impact of noise on task id 
 \texttt{272f95fa}.  
(a) Represents the original task. (b) Represents same task with 0.125\% of noise in the input grids. (c) Represents same task with 0.125\% of noise in the output grids. The pictorial representation of new prompt is in appendix section~\ref{appendix:effect_noise}}
    \label{fig:figure2}   
\end{figure}       
        \vspace{1em} 
        

\subsection{Introducing Model Temperature for Solving ARC Tasks}
In addition to evaluating the impact of noise, we also investigated how model temperature influences performance on ARC tasks. The temperature parameter in language models controls the degree of randomness in predictions.

\textbf{Temperature Levels and Their Impact:}

\begin{itemize}
    \item \textbf{Temperature = 0 (Deterministic Mode):}  
    The model always selects the most probable response, leading to stable and accurate predictions, especially when noise is low. This setting yields the best accuracy in clean conditions but lacks adaptability to noisy inputs, as the model strictly follows learned patterns.

    \item \textbf{Temperature = 1 (Moderate Randomness):}  
    Introduces controlled randomness in predictions, allowing slight variations in responses, which may help with noisy inputs. However, it can also cause inconsistencies in structured reasoning tasks.

\end{itemize}

\section{Results}

GPT-4o initially exhibited strong problem-solving capabilities on the ARC challenge, accurately solving the following 7 tasks under noise-free conditions and a deterministic setup (temperature = 0):
\texttt{272f95fa, 539a4f51, aabf363d, bda2d7a6-a, bda2d7a6, bdad9b1f, cbded52d}.
At this configuration, the model effectively extracted and applied transformation rules from the input-output examples, indicating its ability to identify and generalize underlying patterns.
We then evaluated the model’s robustness by introducing two types of structured noise into the prompt examples noise in the input grids and noise in the output grids while also assessing the impact of increasing temperature settings.

In addition to noise, we experimented with increasing the model temperature to 1.0 (from 0.0). The temperature parameter controls the randomness in the model’s output generation higher values encourage more diverse but less deterministic responses. At a temperature of 1.0, GPT-4o began to produce significantly more incorrect predictions, even when the input-output pairs were noise-free. This indicates that successful completion of ARC tasks depends heavily on consistent, rule-based reasoning, which is hampered when the model's output space becomes more stochastic.

Taken together, these results suggest that GPT-4o's performance on ARC tasks is highly sensitive to both prompt fidelity and model determinism. Introducing noise either in the input or output examples breaks the model’s pattern recognition and generalization abilities. Similarly, increasing temperature adds unpredictability, undermining the structured reasoning necessary for solving these tasks.

Graphs in the following section visualize GPT-4o’s performance across varying temperature settings (0.0 and 1.0) and different levels of input/output noise. To systematically evaluate the impact of noise, we introduced distortions at incremental levels ranging from 0

A detailed breakdown of the noise injection process is provided in Appendix~\ref{appendix:effect_noise}, where we compare a clean prompt to a noisy one. The grid differences between the original and distorted examples highlight how even minor alterations can derail the model’s reasoning process and result in degraded performance.
\begin{enumerate} \item \textbf{Noise in Input Grids:}
Random perturbations were applied to the input grids provided in the few-shot examples. These distortions altered the visual or structural layout of the input, leading to significant confusion for the model. Since the model relies on consistent input patterns to generalize transformation rules, even slight disruptions negatively impacted its performance. In many cases, the model failed to recognize the correct transformation altogether.
\item \textbf{Noise in Output Grids:}  
Here, noise was introduced only to the output grids while keeping the input examples intact. Despite receiving correct inputs, the altered outputs misrepresented the expected transformations. This discrepancy impaired the model’s learning from examples, resulting in reduced accuracy. The model often attempted to learn incorrect rules based on the misleading output patterns.

The following section presents graphs illustrating the model’s performance across different temperature settings (0, 1) and varying noise levels: 
The impact of noise on GPT-4o’s performance was assessed by introducing structured distortions in the input and output grids of the prompt. 
We demonstrating how noise is introduced at different levels (ranging from 0\% to 0.3\%) in the input, output . A detailed illustration of this noise injection is provided in Appendix~\ref{appendix:effect_noise} , where we compare a clean prompt (without noise) to a noisy prompt (with input, output grid modifications). The exact differences in grid values highlight how small distortions can mislead the model, leading to performance degradation.
\end{enumerate}

\begin{center}
   \centering 
  \textbf{\small Task ID - 272f95fa}\\[2mm]
  \fbox{
        \includegraphics[width=1\textwidth]{272f95fa.pdf}
    }

    \captionof{figure}{The figure presents a pictorial representation of Task ID \textbf{272f95fa}, sourced from 
    \url{https://arcprize.org/play?task=272f95fa}.}
    \label{fig:figure3}

    \vspace{225mm}

    \textbf{\small (a) Model Temperature = 0} \\
    \includegraphics[width=0.95\textwidth]{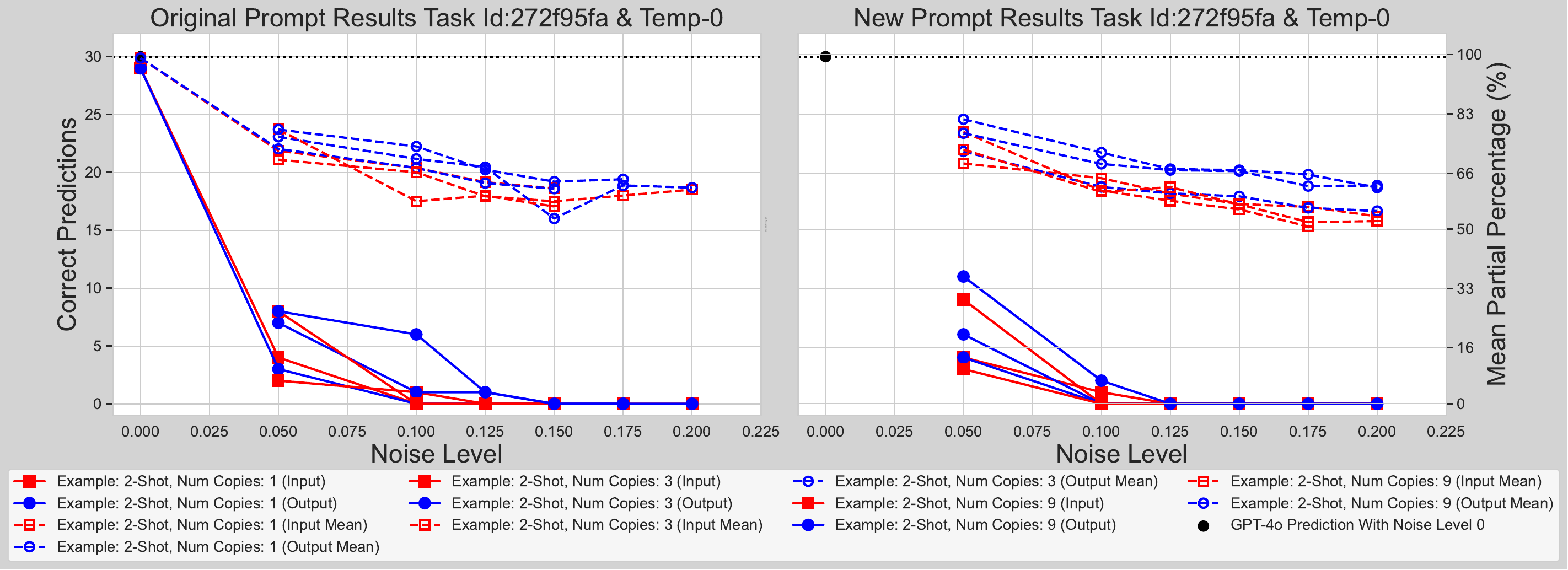}

    \vspace{4mm}

    \textbf{\small (b) Model Temperature = 1} \\
    \includegraphics[width=0.95\textwidth]{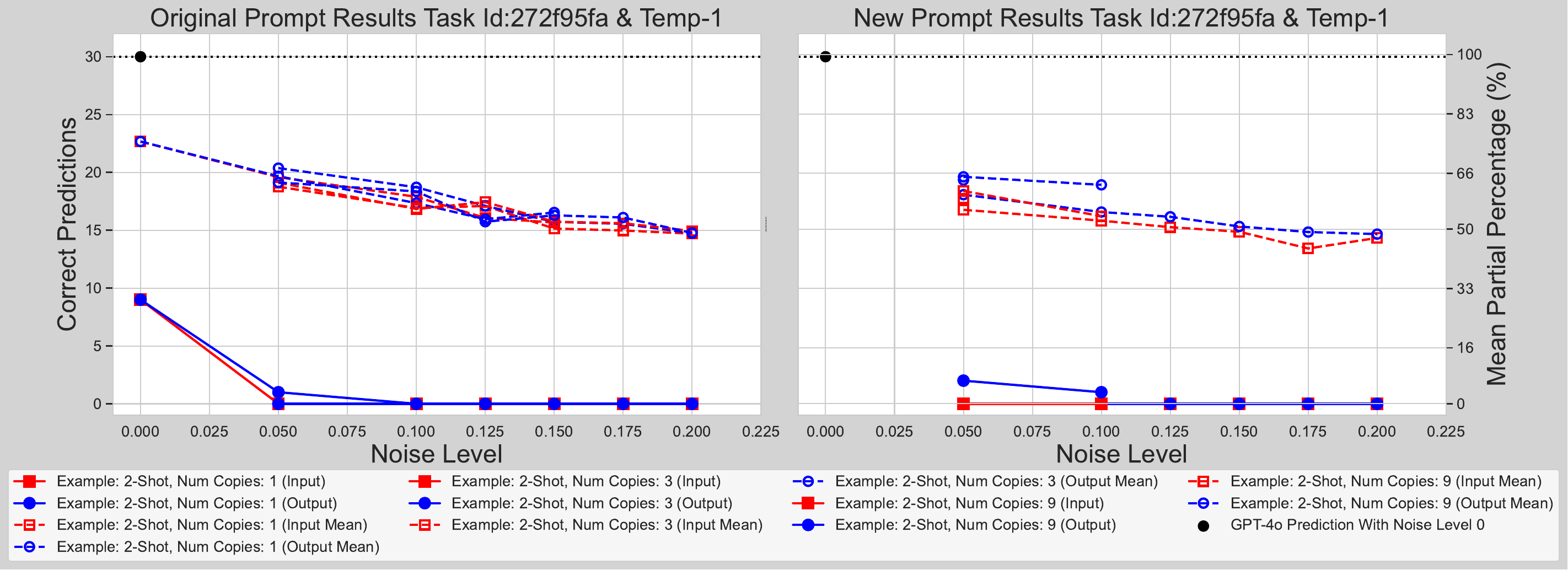}

    \captionof{figure}{These graphs illustrate the impact of noise and model temperature on GPT-4o’s ability to solve ARC tasks, specifically for Task ID \textbf{272f95fa}. The x-axis represents the noise level, the left y-axis shows correct predictions (out of 30 evaluations) and the right y-axis indicates the mean of partially correct cells, and Mean partial percentage ( out of 100\% evaluation) .\\
To evaluate the robustness of the model on Task ID: 272f95fa under varying conditions, Each subplot shows the number of correct predictions (solid lines) and mean partial correctness percentage (dotted lines) at varying noise levels for different example configurations. Subplots (a) and (b) correspond to model temperatures 0 and 1, respectively. The new prompt (right panels) demonstrates improved performance over the original prompt (left panels) at low noise levels, particularly when more examples are provided. However, at higher noise levels, both prompts perform similarly poorly. Additionally, model performance at temperature = 0 is significantly better than at temperature = 1, where variability increases and accuracy drops. Overall, the results highlight that lower noise, more examples, and lower model temperature contribute positively to model accuracy—while high noise and high temperature consistently degrade performance.
}
    \label{fig:Figure 4}
\end{center}


\begin{figure}[H]
    \centering
    \textbf{\small Task ID - 539a4f51} \\[1mm]
    \fbox{
        \includegraphics[width=0.95\textwidth]{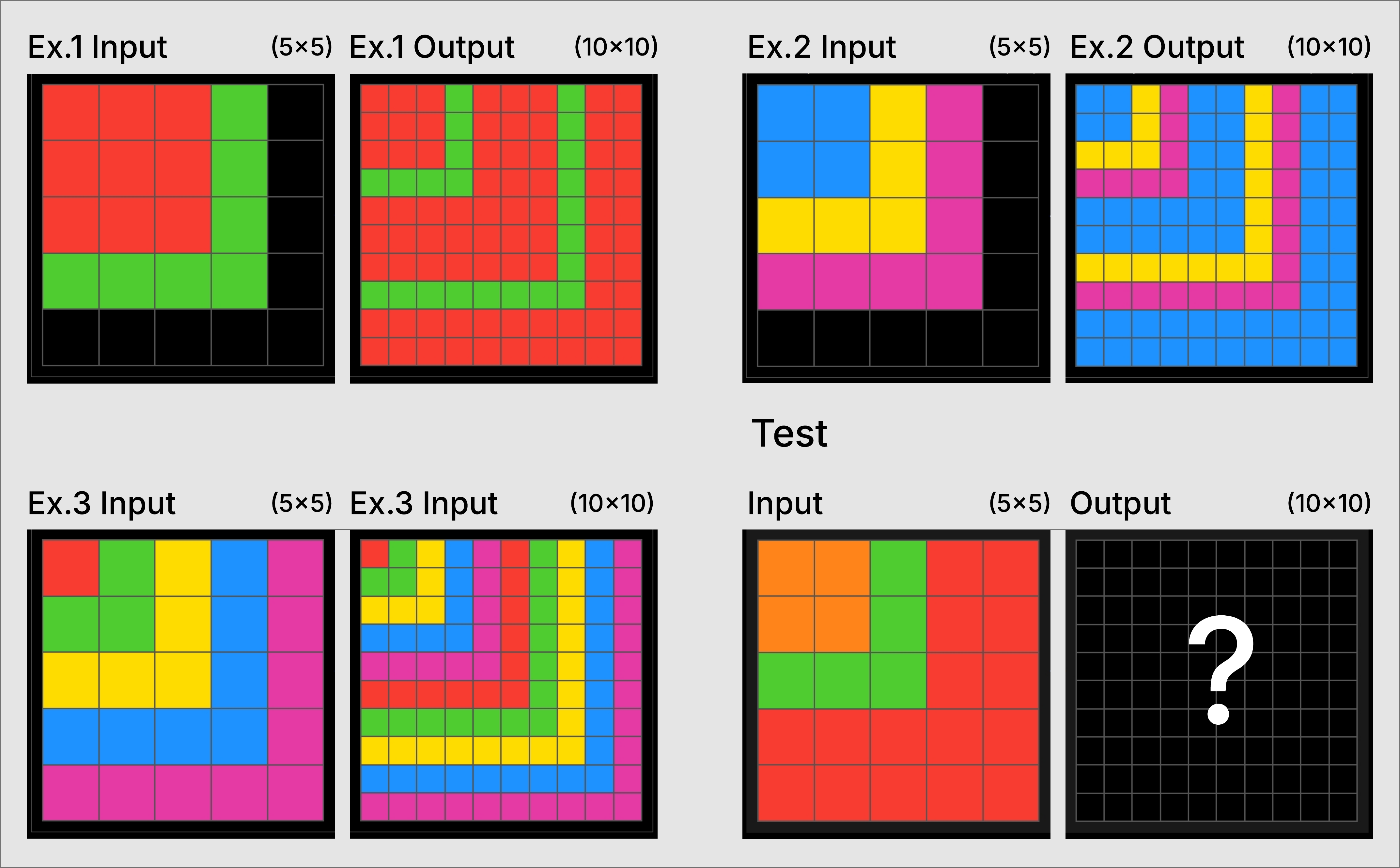}
    }
    \vspace{1mm}
    \caption{The figure presents a pictorial representation of task id \textbf{539a4f51}, sourced from 
    \href{https://arcprize.org/play?task=539a4f51}{https://arcprize.org/play?task=539a4f51}}
\end{figure}

\begin{figure}[H]
    \centering
    \textbf{\small (a) Model Temperature = 0} \\
    \includegraphics[width=0.95\textwidth]{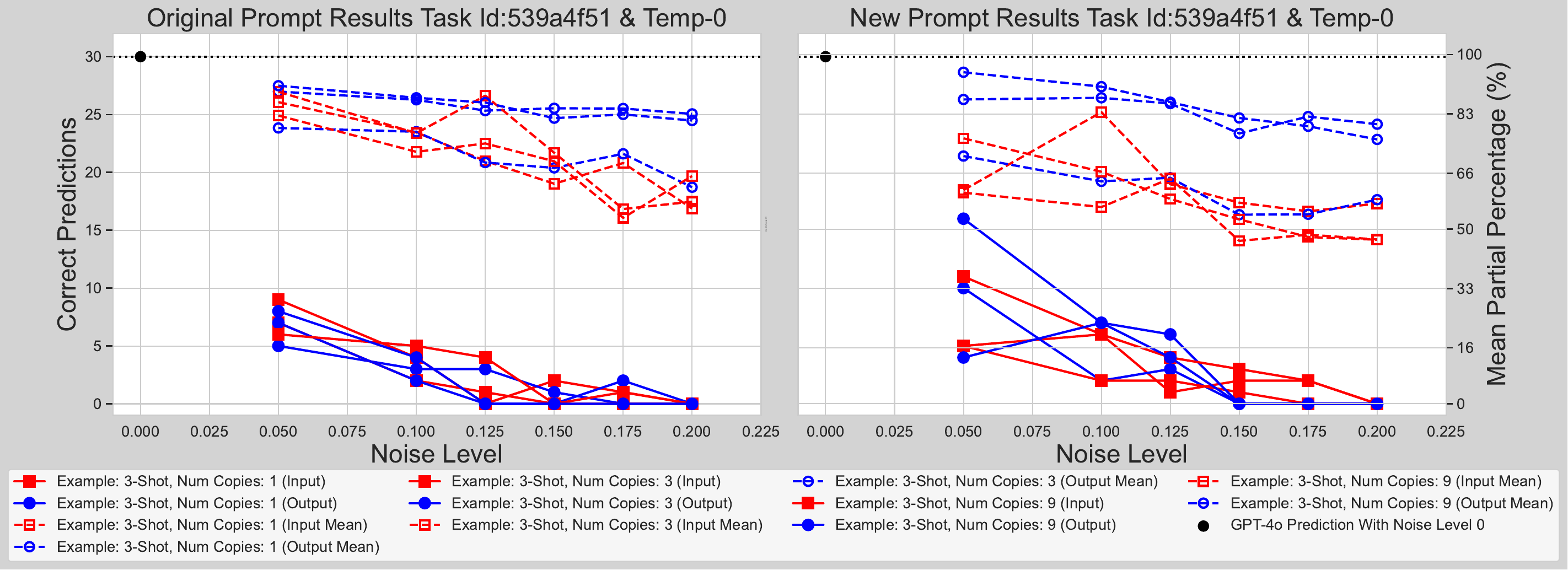}

    \vspace{3mm}

    \textbf{\small (b) Model Temperature = 1} \\
    \includegraphics[width=0.95\textwidth]{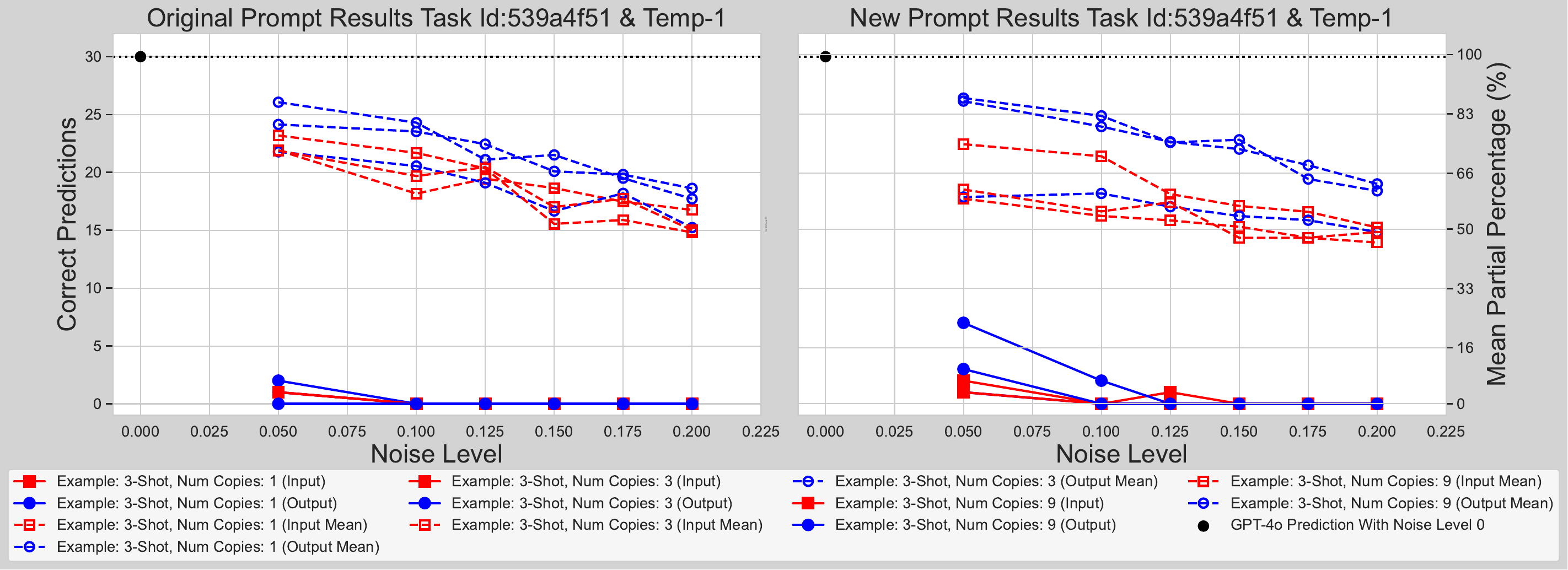} \\
    \textbf{Noise Level}

    \caption{These graphs illustrate the impact of noise and model temperature on GPT-4o’s ability to solve ARC task with task id \textbf{539a4f51}. The analysis is same as described in Figure~\ref{fig:Figure 4}, but applied to the task id 539a4f51.}
    \label{fig:539a4f51_graphs}
\end{figure}

\begin{figure}[H]
    \centering
    \textbf{\small Task ID - aabf363d} \\[1mm]
    \fbox{
        \includegraphics[width=0.95\textwidth]{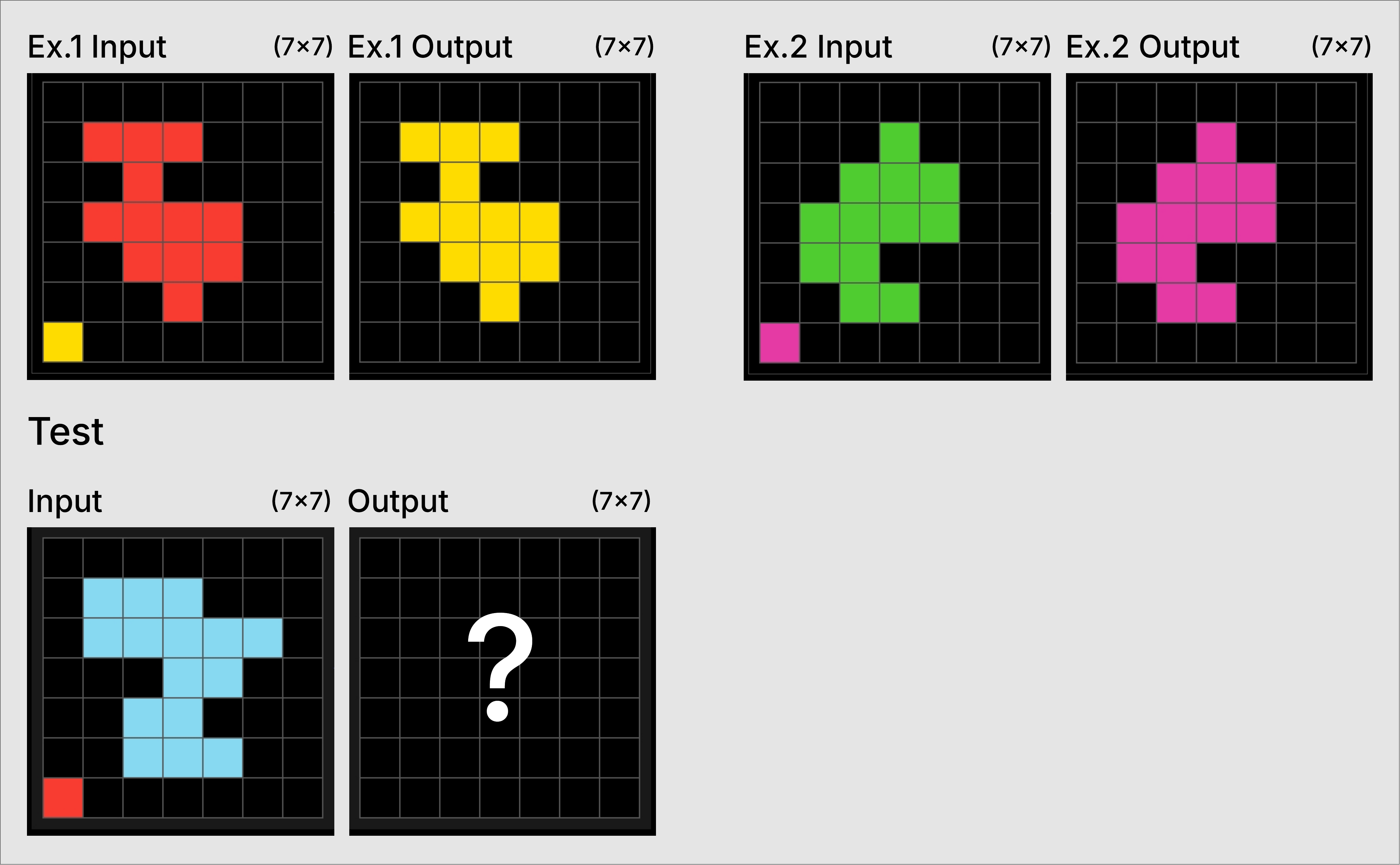}
    }
    \vspace{1mm}
    \caption{The figure presents a pictorial representation of task id \textbf{aabf363d}, sourced from 
    \href{https://arcprize.org/play?task=aabf363d}{https://arcprize.org/play?task=aabf363d}}
\end{figure}

\begin{figure}[H]
    \centering
    \textbf{\small (a) Model Temperature = 0} \\
    \includegraphics[width=0.95\textwidth]{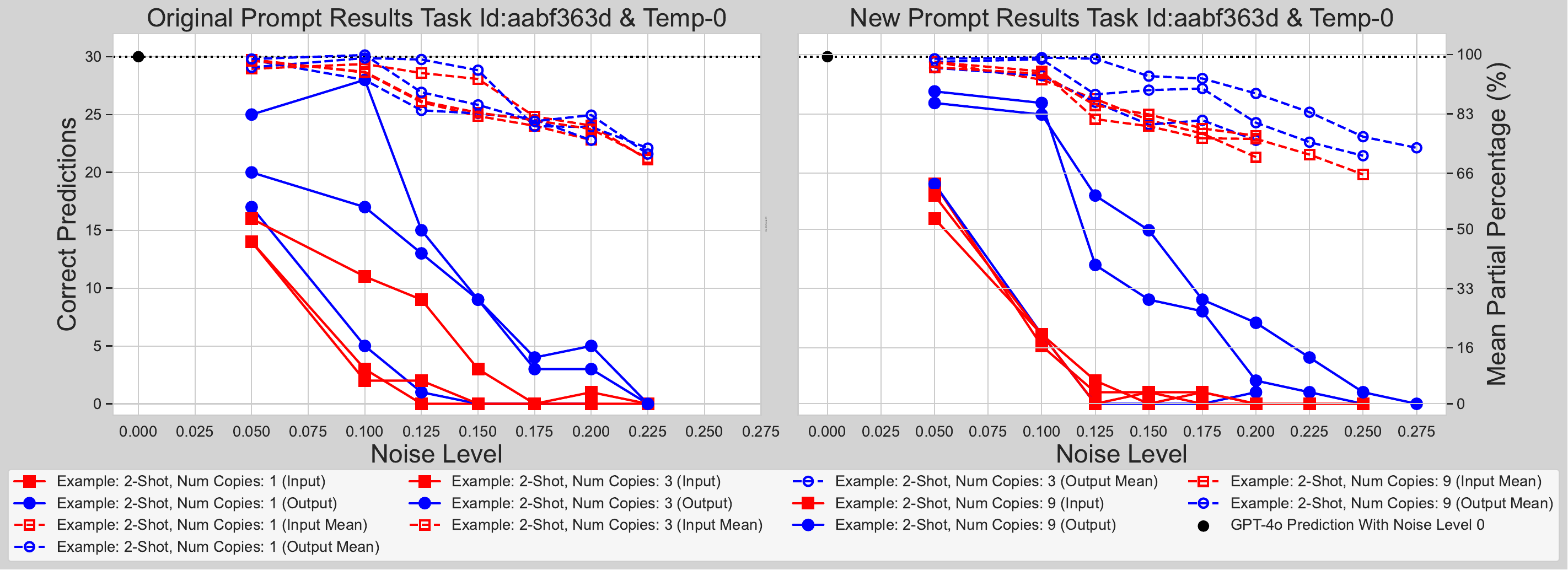}

    \vspace{3mm}

    \textbf{\small (b) Model Temperature = 1} \\
    \includegraphics[width=0.95\textwidth]{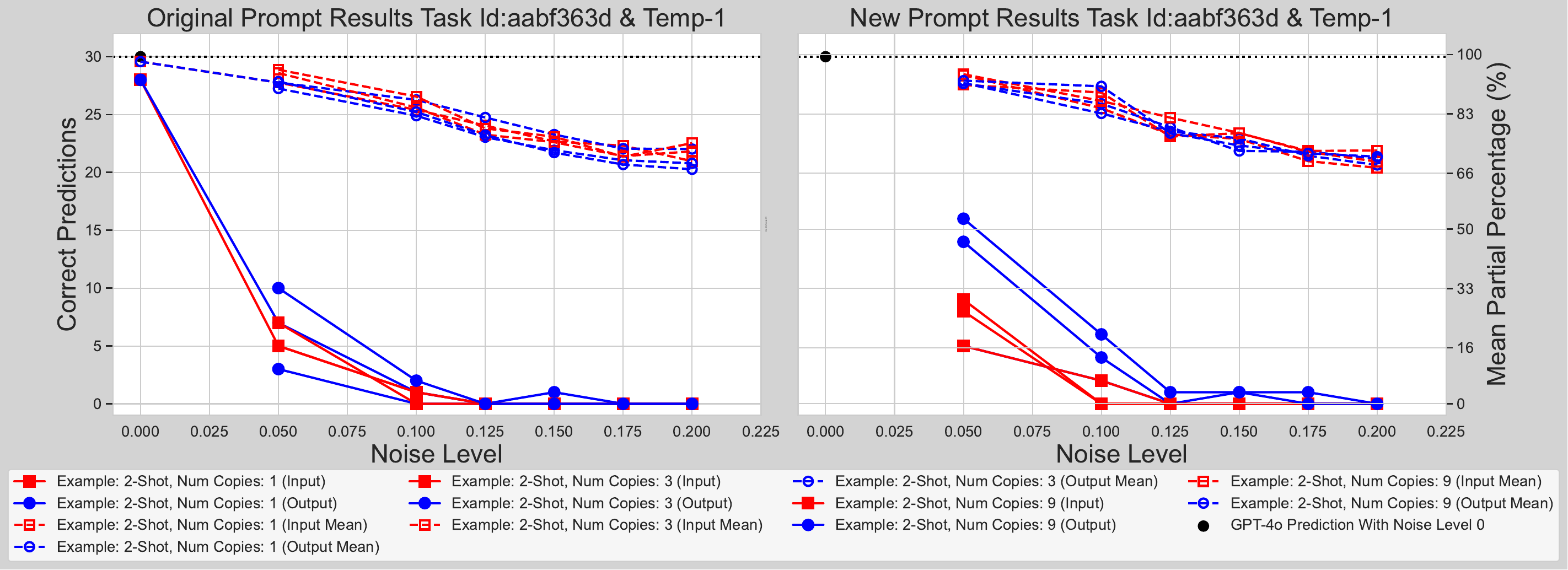} \\
    \textbf{Noise Level}

    \caption{These graphs illustrate the impact of noise and model temperature on GPT-4o’s ability to solve ARC task with task id \textbf{aabf363d}. The analysis is same as described in Figure~\ref{fig:Figure 4}, but applied to the task id aabf363d.}
    \label{fig:aabf363d_graphs}
\end{figure}

\begin{figure}[H]
    \centering
    \textbf{\small Task ID - bda2d7a6-a} \\[1mm]
    \fbox{
        \includegraphics[width=0.95\textwidth]{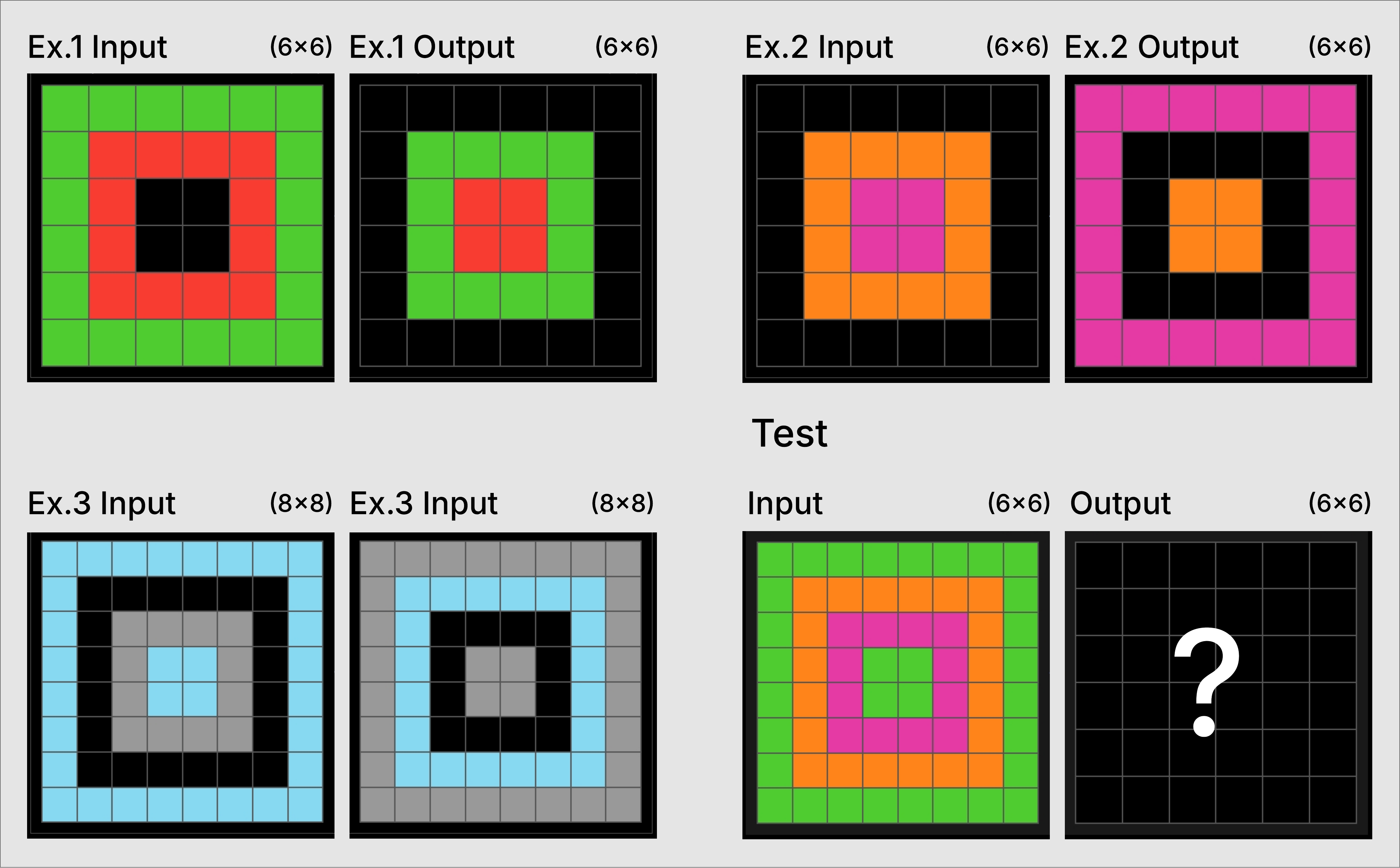}
    }
    \vspace{1mm}
    \caption{The figure presents a pictorial representation of task id \textbf{bda2d7a6-a}, sourced from 
    \href{https://arcprize.org/play?task=bda2d7a6-a}{https://arcprize.org/play?task=bda2d7a6-a}}
\end{figure}

\begin{figure}[H]
    \centering
    \textbf{\small (a) Model Temperature = 0} \\
    \includegraphics[width=0.95\textwidth]{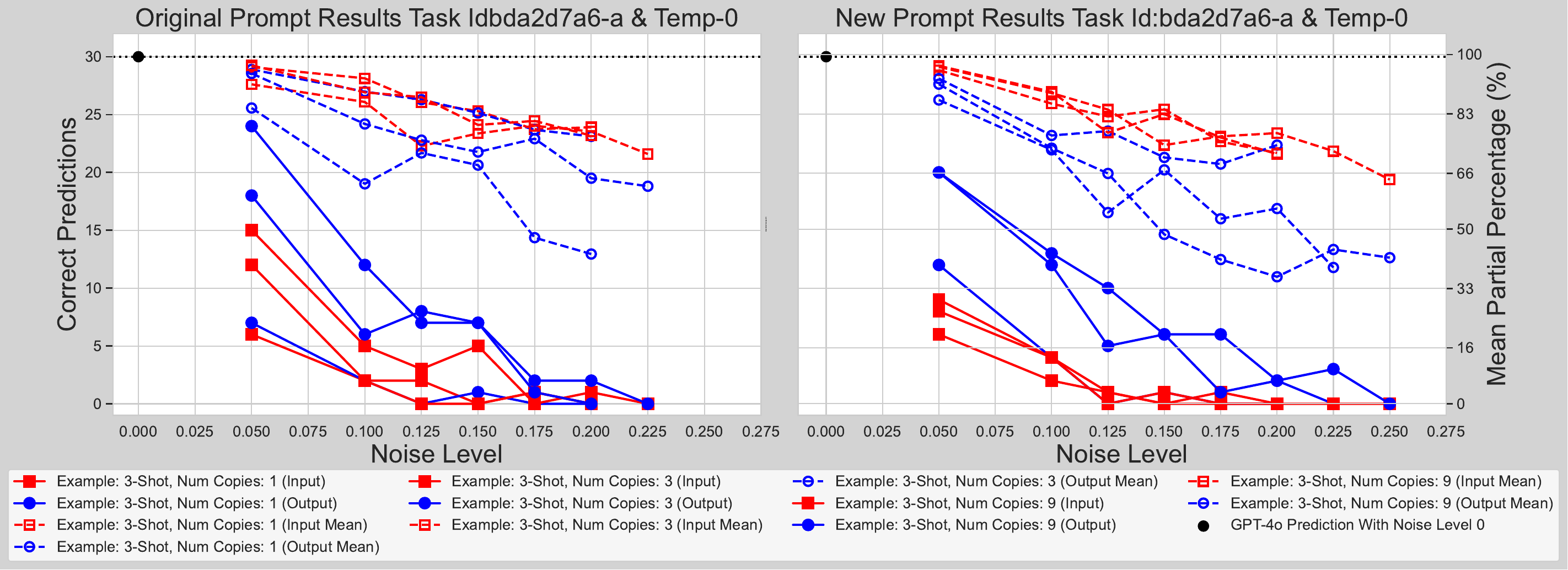}

    \vspace{3mm}

    \textbf{\small (b) Model Temperature = 1} \\
    \includegraphics[width=0.95\textwidth]{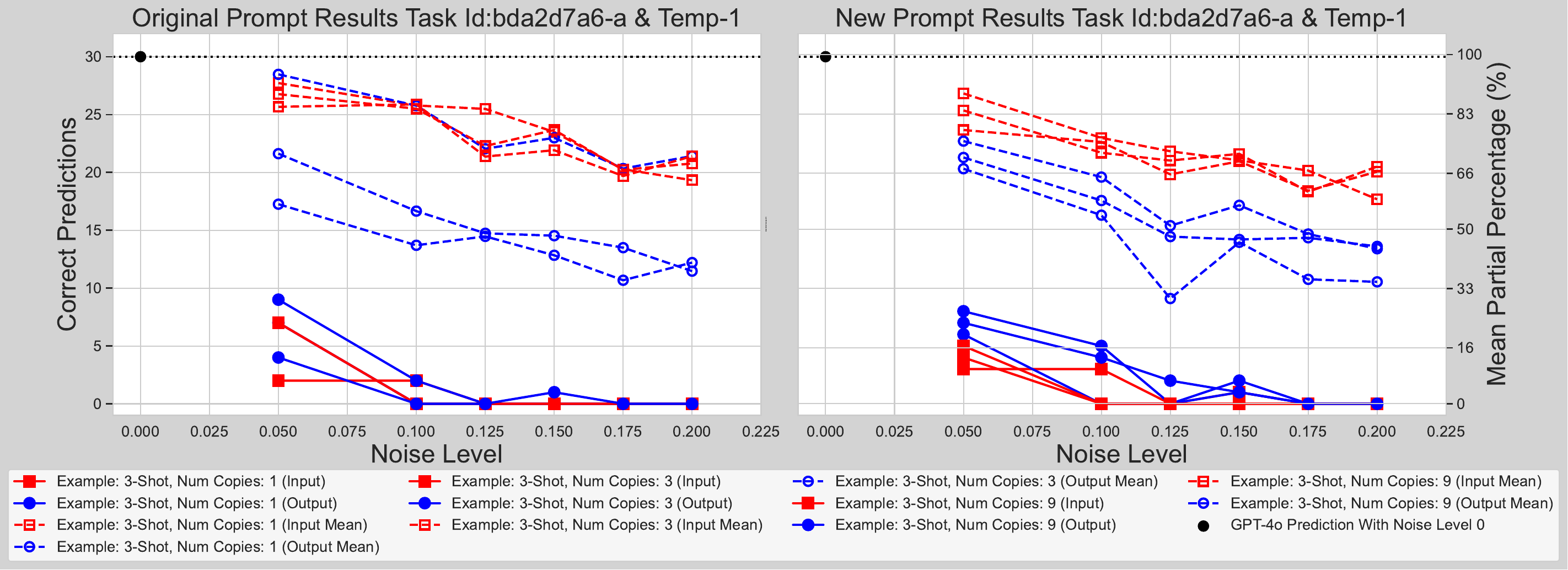} \\
    \textbf{Noise Level}

    \caption{These graphs illustrate the impact of noise and model temperature on GPT-4o’s ability to solve ARC task with task id \textbf{bda2d7a6-a}. The analysis is same as described in Figure~\ref{fig:Figure 4}, but applied to the task id bda2d7a6-a.}
    \label{fig:bda2d7a6-a_graphs}
\end{figure}

\begin{figure}[H]
    \centering
    \textbf{\small Task ID - bda2d7a6} \\[1mm]
    \fbox{
        \includegraphics[width=0.95\textwidth]{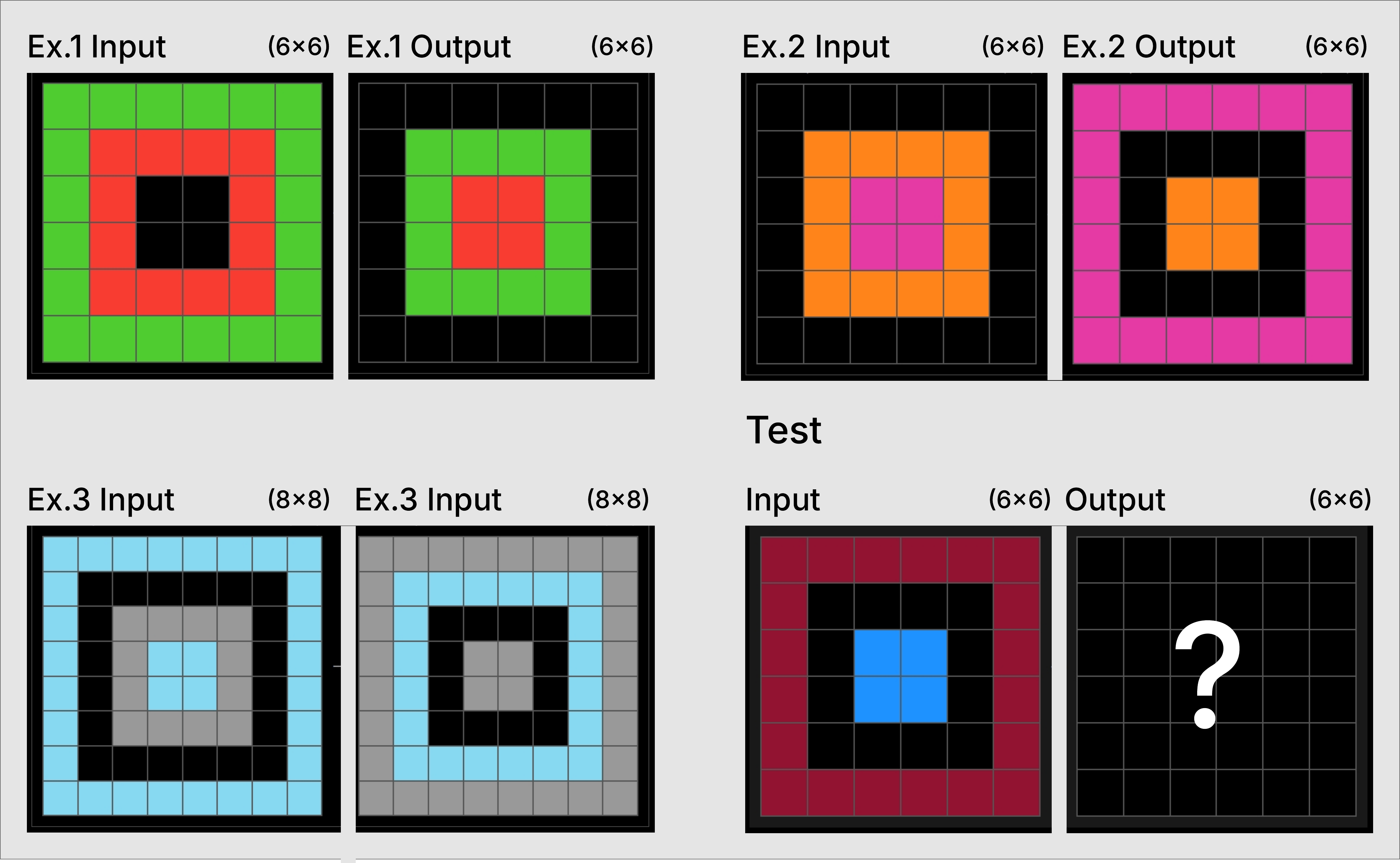}
    }
    \vspace{1mm}
    \caption{The figure presents a pictorial representation of task id \textbf{bda2d7a6}, sourced from 
    \href{https://arcprize.org/play?task=bda2d7a6}{https://arcprize.org/play?task=bda2d7a6}}
\end{figure}

\begin{figure}[H]
    \centering
    \textbf{\small (a) Model Temperature = 0} \\
    \includegraphics[width=0.95\textwidth]{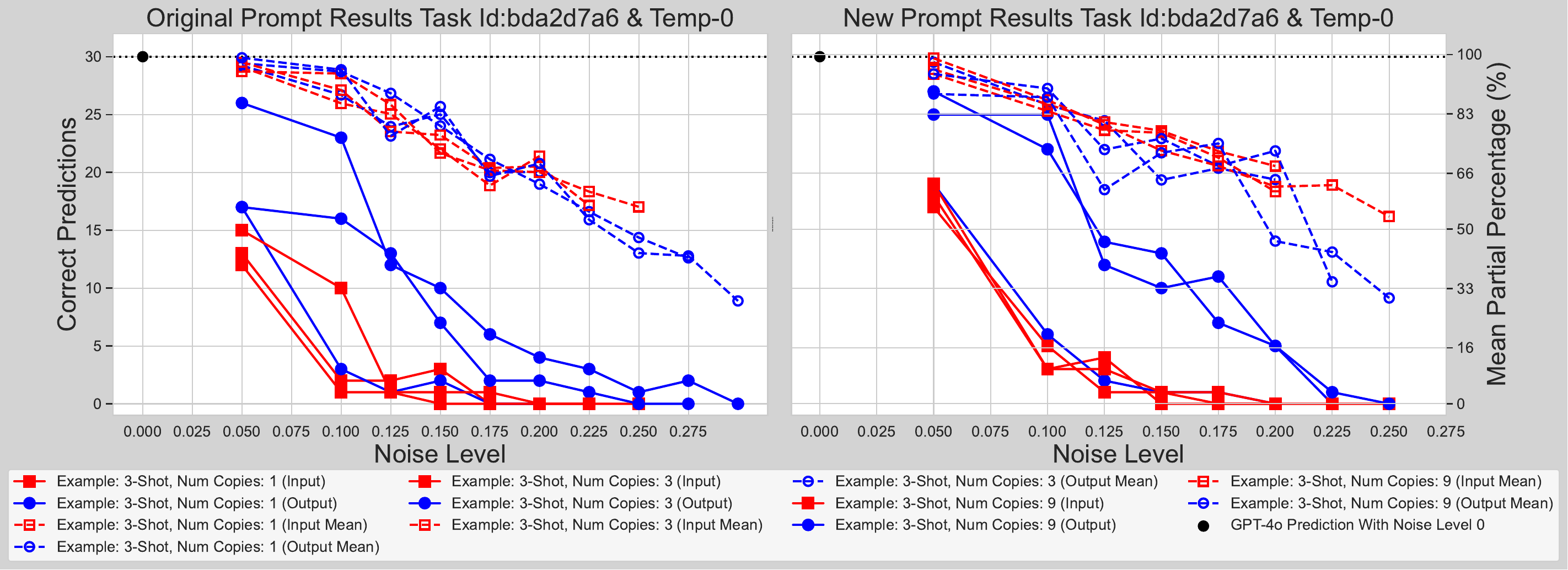}

    \vspace{3mm}

    \textbf{\small (b) Model Temperature = 1} \\
    \includegraphics[width=0.95\textwidth]{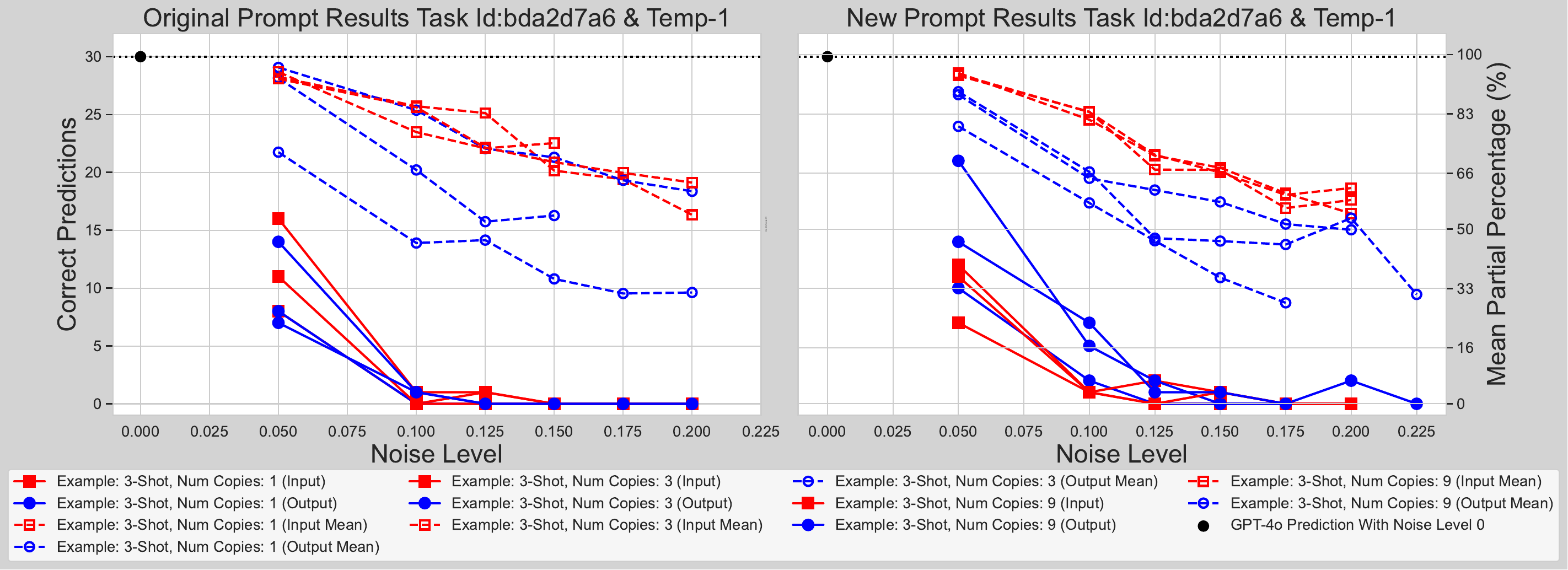} \\
    \textbf{Noise Level}

    \caption{These graphs illustrate the impact of noise and model temperature on GPT-4o’s ability to solve ARC task with task id \textbf{bda2d7a6}. The analysis is same as described in Figure~\ref{fig:Figure 4}, but applied to the task id bda2d7a6.}
    \label{fig:bda2d7a6_graphs}
\end{figure}

\begin{figure}[H]
    \centering
    \textbf{\small Task ID - bdad9b1f} \\[1mm]
    \fbox{
        \includegraphics[width=0.95\textwidth]{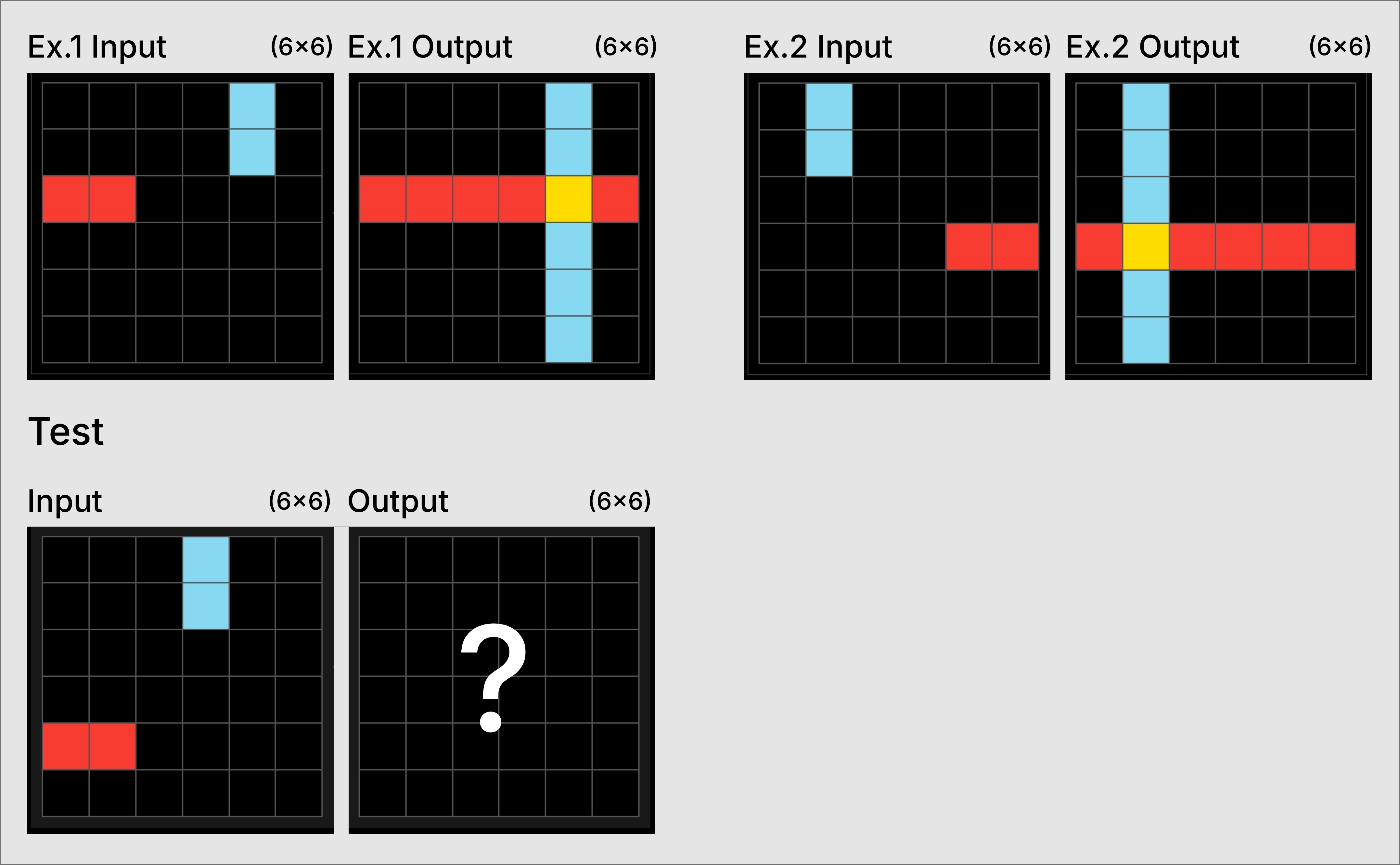}
    }
    \vspace{1mm}
    \caption{The figure presents a pictorial representation of task id \textbf{bdad9b1f}, sourced from 
    \href{https://arcprize.org/play?task=bdad9b1f}{https://arcprize.org/play?task=bdad9b1f}}
\end{figure}

\begin{figure}[H]
    \centering
    \textbf{\small (a) Model Temperature = 0} \\
    \includegraphics[width=0.95\textwidth]{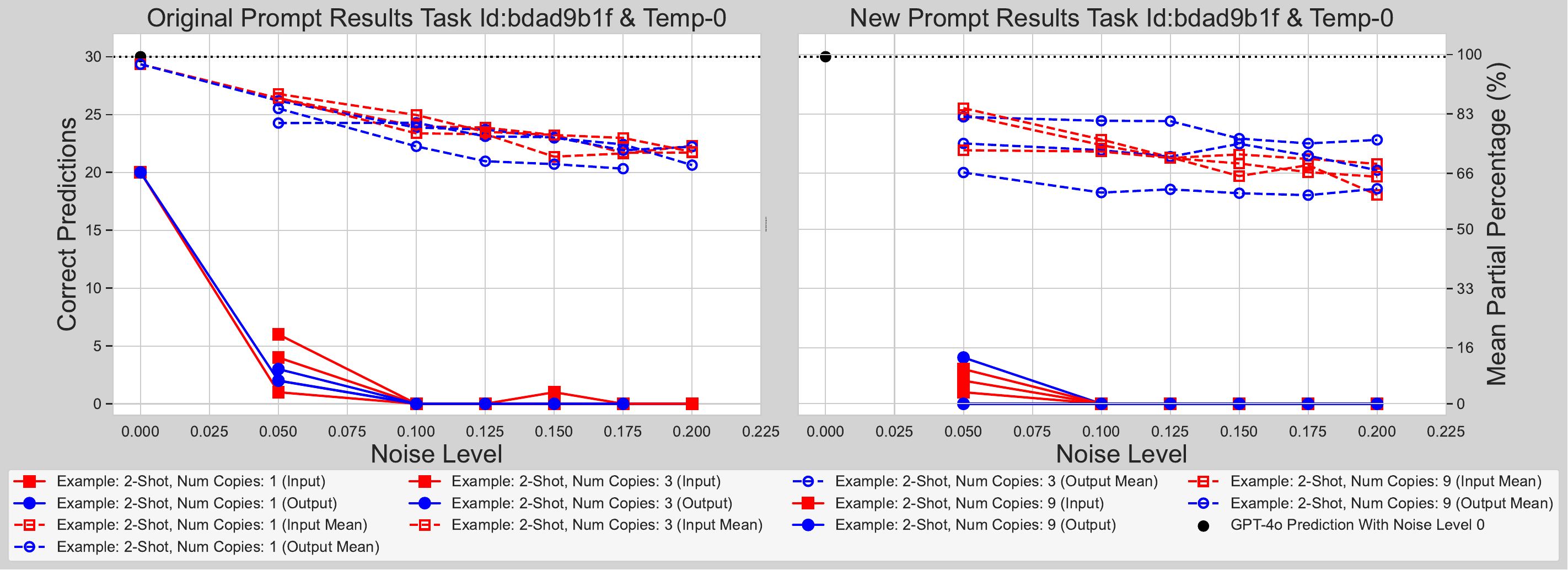}

    \vspace{3mm}

    \textbf{\small (b) Model Temperature = 1} \\
    \includegraphics[width=0.95\textwidth]{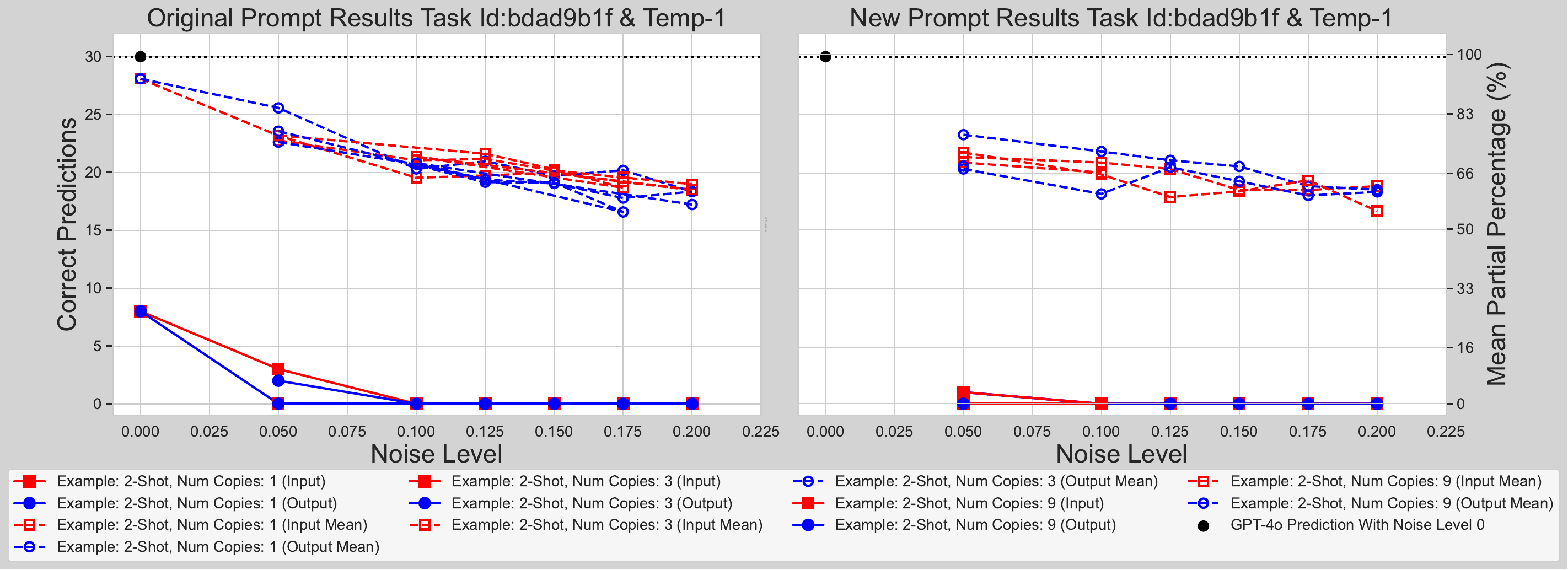} \\
    \textbf{Noise Level}

    \caption{These graphs illustrate the impact of noise and model temperature on GPT-4o’s ability to solve ARC task with task id \textbf{bdad9b1f}. The analysis is same as described in Figure~\ref{fig:Figure 4}, but applied to the task id bdad9b1f.}
    \label{fig:bdad9b1f_graphs}
\end{figure}

\begin{figure}[H]
    \centering
    \textbf{\small Task ID - cbded52d} \\[1mm]
    \fbox{
        \includegraphics[width=0.95\textwidth]{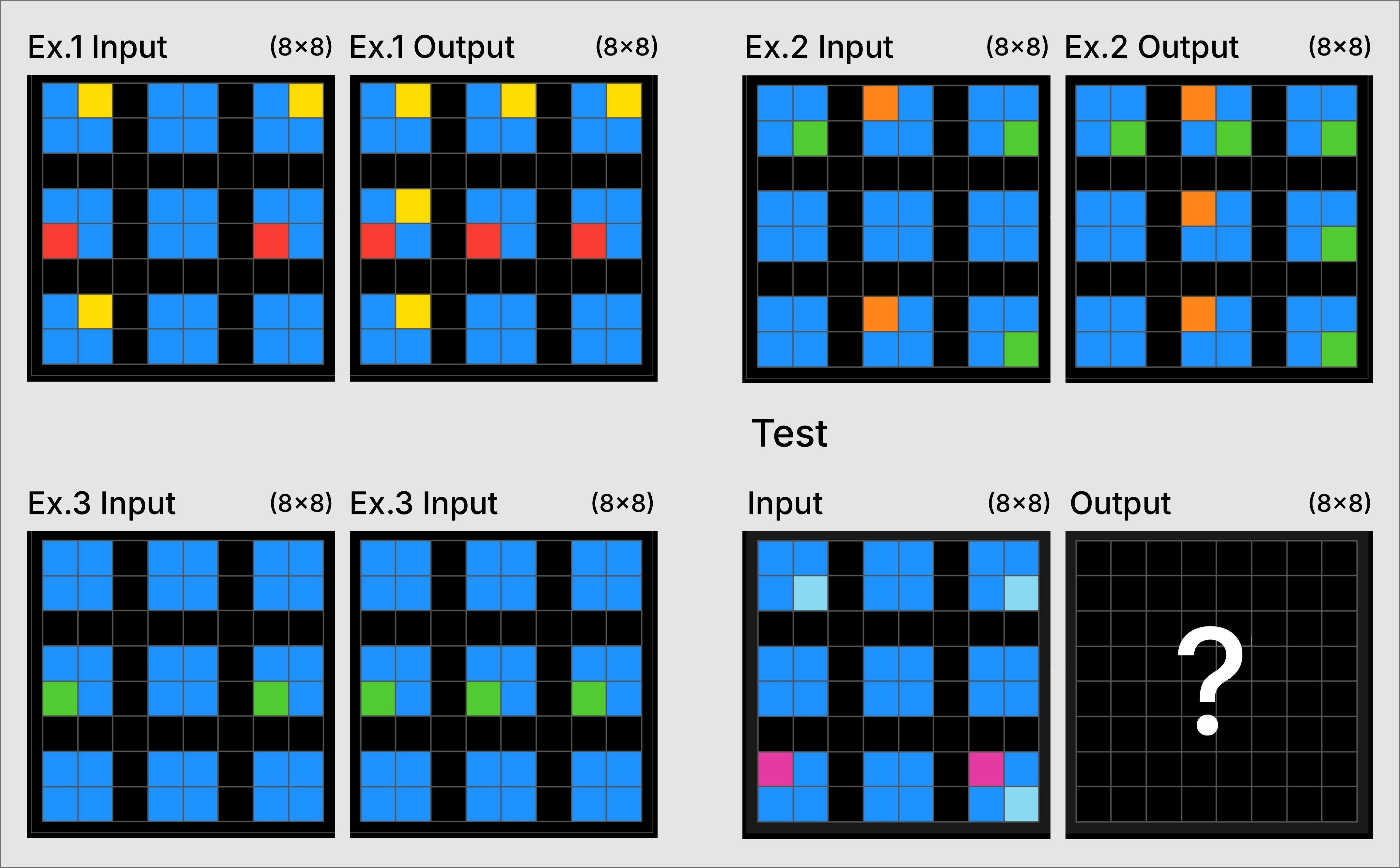}
    }
    \vspace{1mm}
    \caption{The figure presents a pictorial representation of task id \textbf{cbded52d}, sourced from 
    \href{https://arcprize.org/play?task=cbded52d}{https://arcprize.org/play?task=cbded52d}}
\end{figure}

\begin{figure}[H]
    \centering
    \textbf{\small (a) Model Temperature = 0} \\
    \includegraphics[width=0.95\textwidth]{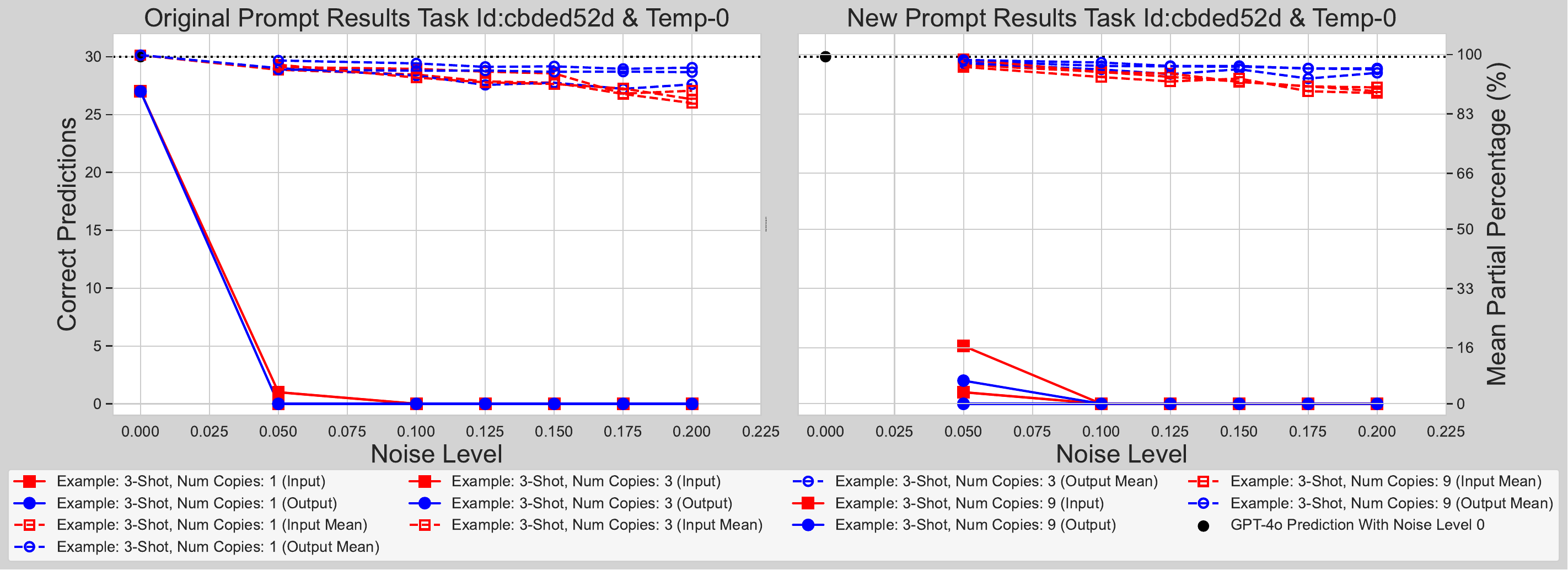}

    \vspace{3mm}

    \textbf{\small (b) Model Temperature = 1} \\
    \includegraphics[width=0.95\textwidth]{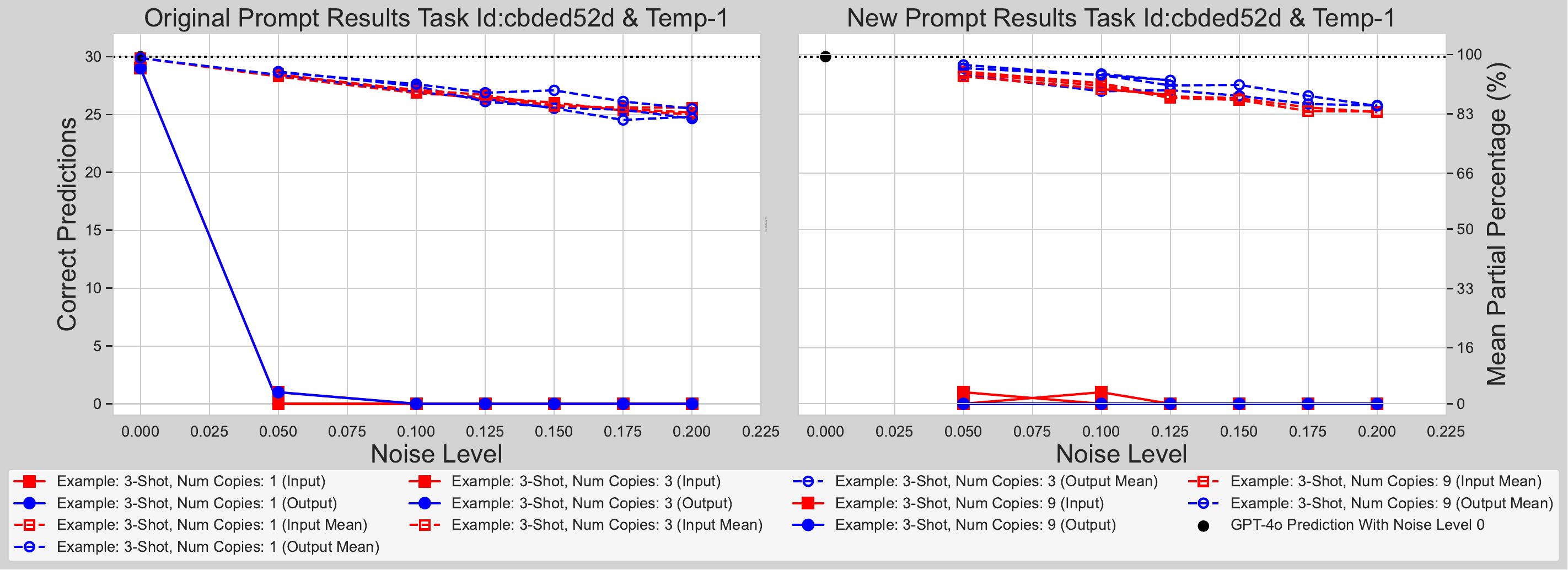} \\
    \textbf{Noise Level}

    \caption{These graphs illustrate the impact of noise and model temperature on GPT-4o’s ability to solve ARC task with task id \textbf{cbded52d}. The analysis is same as described in Figure~\ref{fig:Figure 4}, but applied to the task id cbded52d.}
    \label{fig:cbded52d_graphs}
\end{figure}

\section{Conclusion}

In this study, we evaluated the performance of AI models on Abstraction and Reasoning Corpus (ARC) tasks using both 2-shot and 3-shot learning examples. Our findings reveal that the model exhibits a high sensitivity to noise in input-output pairs, which negatively impacts its ability to generalize across diverse reasoning tasks. This observation underscores the challenge of ensuring robustness in AI-driven reasoning models when confronted with uncertain or imperfect data. To address this, we plan to extend our investigation by increasing the number of examples in a k-shot learning framework, leveraging state-of-the-art models. By systematically expanding the number of input-output pairs, we aim to determine whether providing additional examples enhances the model’s ability to recognize abstract patterns and adapt to variations more effectively. Given that real-world datasets inherently contain structured noise, our research will also focus on assessing whether controlled exposure to such noise can improve the model’s resilience in uncertain environments. Specifically, we will introduce multiple noisy outputs corresponding to a single input and multiple noisy inputs leading to a single output to examine how structured noise influences learning and generalization.

Beyond expanding the dataset, we aim to gain deeper insights into the internal workings of neural networks by analyzing how noise propagates through different activation layers. Rather than treating noise solely as an external factor affecting performance, we intend to explore its impact on internal representations and transformations within the model. By examining activation patterns, pinpointing noise-sensitive layers, and analyzing gradient flow, we aim to uncover potential vulnerabilities in the learning process and identify strategies to mitigate their effects. This investigation will help us develop more effective training methodologies that enhance the model’s robustness against structured noise and uncertainty. By integrating insights from k-shot learning, structured noise adaptation, and neural network interpretability, our future research will contribute to advancing AI models with enhanced reasoning capabilities and greater resilience to noise, ultimately improving their applicability in real-world decision-making scenarios.

This study underscores the necessity for AI models to effectively manage noise, uncertainty, and structured reasoning challenges. While some models exhibit success in solving ARC tasks, others struggle significantly, failing to complete any task. This discrepancy highlights a fundamental gap in structured reasoning capabilities between different AI architectures, particularly between models designed primarily for language-based tasks and those capable of recognizing abstract patterns. The stark contrast in performance suggests that models lacking robust pattern recognition mechanisms may struggle with tasks requiring higher-order reasoning, even if they excel in natural language processing. These findings emphasize the need for further research into AI architectures that integrate both linguistic comprehension and structured problem-solving abilities, ensuring that models can effectively handle a wider range of reasoning challenges.

By analyzing the effects of noise levels and temperature variations on model performance, we gain deeper insights into how AI models process uncertainty. The observed decline in performance under noisy conditions highlights that even the most advanced models remain vulnerable when exposed to imperfect, unstructured data. This limitation reinforces the importance of improving AI robustness, ensuring models can generalize beyond seen examples and adapt to real-world variability. Future research must prioritize building AI systems that align more closely with human cognitive flexibility, enabling them to reason effectively under uncertainty. Such advancements are especially crucial for applications in automated decision-making, problem-solving, and complex reasoning domains, where adaptability and resilience are key. By developing models that can navigate real-world complexities, we can create more reliable, efficient, and intelligent AI systems capable of robust decision-making in uncertain environments.

\section{Future Work}

To address the challenges observed in our study, we plan to extend our investigation by increasing the number of examples in a k-shot learning framework, leveraging state-of-the-art models. By systematically expanding the number of input-output pairs, we aim to determine whether providing additional examples enhances the model’s ability to recognize abstract patterns and adapt to variations more effectively.

Given that real-world datasets inherently contain structured noise, our research will also focus on assessing whether controlled exposure to such noise can improve the model’s resilience in uncertain environments. Specifically, we will introduce multiple noisy outputs corresponding to a single input and multiple noisy inputs leading to a single output to examine how structured noise influences learning and generalization.

Beyond expanding the dataset, we aim to gain deeper insights into the internal workings of neural networks by analyzing how noise propagates through different activation layers. Rather than treating noise solely as an external factor affecting performance, we intend to explore its impact on internal representations and transformations within the model. By examining activation patterns, pinpointing noise-sensitive layers, and analyzing gradient flow, we aim to uncover potential vulnerabilities in the learning process and identify strategies to mitigate their effects.

By integrating insights from k-shot learning, structured noise adaptation, and neural network interpretability, our future research will contribute to advancing AI models with enhanced reasoning capabilities and greater resilience to noise, ultimately improving their applicability in real-world decision-making scenarios.

\section{Appendix}

\appendix
\label{appendix:effect_noise}

\textbf{Pictorial Representation Prompt With Orignal Prompt} 
\begin{paracol}{2} 

\begin{tcolorbox}[colback=cyan!10!white, colframe=black,
    boxrule=0.6pt, arc=1mm, breakable, enhanced,
    title=\centering \textbf{Prompt 1}]
\scriptsize
\begin{lstlisting}
Find the common rule that maps 
an input grid to an output grid, 
given the examples below. 

Example1
Input:
0 0 0 0 8 0 0 0 0 0 0 8 0 0 0 0 0 0 0
0 0 0 0 8 0 0 0 0 0 0 8 0 0 0 0 0 0 0
8 8 8 8 8 8 8 8 8 8 8 8 8 8 8 8 8 8 8
0 0 0 0 8 0 0 0 0 0 0 8 0 0 0 0 0 0 0
0 0 0 0 8 0 0 0 0 0 0 8 0 0 0 0 0 0 0
0 0 0 0 8 0 0 0 0 0 0 8 0 0 0 0 0 0 0
0 0 0 0 8 0 0 0 0 0 0 8 0 0 0 0 0 0 0
8 8 8 8 8 8 8 8 8 8 8 8 8 8 8 8 8 8 8
0 0 0 0 8 0 0 0 0 0 0 8 0 0 0 0 0 0 0
0 0 0 0 8 0 0 0 0 0 0 8 0 0 0 0 0 0 0
0 0 0 0 8 0 0 0 0 0 0 8 0 0 0 0 0 0 0
0 0 0 0 8 0 0 0 0 0 0 8 0 0 0 0 0 0 0
0 0 0 0 8 0 0 0 0 0 0 8 0 0 0 0 0 0 0
0 0 0 0 8 0 0 0 0 0 0 8 0 0 0 0 0 0 0
0 0 0 0 8 0 0 0 0 0 0 8 0 0 0 0 0 0 0
0 0 0 0 8 0 0 0 0 0 0 8 0 0 0 0 0 0 0
0 0 0 0 8 0 0 0 0 0 0 8 0 0 0 0 0 0 0
0 0 0 0 8 0 0 0 0 0 0 8 0 0 0 0 0 0 0

Output:
0 0 0 0 8 2 2 2 2 2 2 8 0 0 0 0 0 0 0
0 0 0 0 8 2 2 2 2 2 2 8 0 0 0 0 0 0 0
8 8 8 8 8 8 8 8 8 8 8 8 8 8 8 8 8 6 8
4 4 4 4 8 6 6 6 6 6 6 8 3 3 3 3 3 3 8
4 4 4 4 8 6 6 6 6 6 6 8 3 3 3 3 3 3 3
4 4 4 4 6 6 6 6 6 6 6 8 3 3 3 3 3 3 3
4 4 4 4 8 6 6 6 0 6 6 8 3 3 3 3 3 3 3
8 8 8 8 8 8 8 8 8 8 8 8 3 8 8 8 8 8 8
0 0 0 0 8 1 1 6 1 1 1 8 0 0 0 0 0 0 0
0 0 0 0 8 1 1 1 1 1 1 1 0 0 0 0 0 0 0
0 0 1 0 8 1 1 1 1 1 1 8 0 0 0 0 0 0 0
0 0 0 0 8 1 1 1 1 1 1 8 0 0 0 0 0 0 0
0 6 0 0 8 1 1 1 1 1 1 8 1 0 0 0 0 0 0
0 0 0 0 8 1 1 1 1 1 1 8 0 0 0 0 0 0 0
0 0 0 0 8 1 1 1 1 1 1 8 0 0 0 0 0 0 0
0 0 0 0 8 6 1 1 1 1 1 8 0 0 0 0 0 0 0
0 0 6 2 8 1 1 1 1 1 1 4 0 0 0 0 0 0 0
0 0 0 0 8 1 1 6 1 1 1 8 0 0 1 3 0 0 0


Example2
Input:
0 0 0 0 8 0 0 0 0 0 0 8 0 0 0 0 0 0 0
0 0 0 0 8 0 0 0 0 0 0 8 0 0 0 0 0 0 0
8 8 8 8 8 8 8 8 8 8 8 8 8 8 8 8 8 8 8
0 0 0 0 8 0 0 0 0 0 0 8 0 0 0 0 0 0 0
0 0 0 0 8 0 0 0 0 0 0 8 0 0 0 0 0 0 0
0 0 0 0 8 0 0 0 0 0 0 8 0 0 0 0 0 0 0
0 0 0 0 8 0 0 0 0 0 0 8 0 0 0 0 0 0 0
8 8 8 8 8 8 8 8 8 8 8 8 8 8 8 8 8 8 8
0 0 0 0 8 0 0 0 0 0 0 8 0 0 0 0 0 0 0
0 0 0 0 8 0 0 0 0 0 0 8 0 0 0 0 0 0 0
0 0 0 0 8 0 0 0 0 0 0 8 0 0 0 0 0 0 0
0 0 0 0 8 0 0 0 0 0 0 8 0 0 0 0 0 0 0
0 0 0 0 8 0 0 0 0 0 0 8 0 0 0 0 0 0 0
0 0 0 0 8 0 0 0 0 0 0 8 0 0 0 0 0 0 0
0 0 0 0 8 0 0 0 0 0 0 8 0 0 0 0 0 0 0
0 0 0 0 8 0 0 0 0 0 0 8 0 0 0 0 0 0 0
0 0 0 0 8 0 0 0 0 0 0 8 0 0 0 0 0 0 0
0 0 0 0 8 0 0 0 0 0 0 8 0 0 0 0 0 0 0

Output:
0 0 0 0 8 2 2 2 2 2 2 8 0 0 0 0 0 0 0
0 0 0 0 3 3 2 2 2 6 2 8 0 1 0 0 0 0 0
8 8 8 8 8 8 8 8 8 8 8 8 8 8 8 8 8 8 8
2 4 4 4 8 6 6 6 6 6 6 8 3 3 3 3 3 3 3
4 4 4 4 8 6 6 6 6 2 6 8 3 3 3 3 3 3 3
4 4 4 4 8 6 6 6 6 6 6 8 3 3 3 3 3 3 3
4 4 4 4 8 6 6 6 6 6 6 8 3 3 1 3 3 3 3
8 8 8 8 8 8 8 8 8 8 8 8 8 8 8 8 8 2 8
0 0 0 0 8 6 1 1 1 1 1 8 0 0 0 0 0 0 0
0 0 0 0 8 1 1 1 1 1 1 8 0 0 0 0 0 0 0
0 0 0 0 8 1 1 1 1 1 1 8 0 0 0 0 0 0 0
0 0 0 0 8 1 3 1 1 1 1 8 2 0 0 6 0 0 0
0 0 0 0 8 1 1 1 1 1 1 8 0 6 0 0 0 0 0
0 0 0 0 8 1 1 1 1 1 1 8 0 0 0 0 6 0 0
0 0 0 0 8 1 1 1 1 1 1 8 0 0 0 0 0 0 0
0 0 0 0 8 1 1 1 1 1 1 8 0 0 0 1 0 0 0
0 0 0 0 8 1 1 1 1 1 1 8 0 0 0 0 0 0 0
0 0 0 0 8 1 2 1 1 6 1 8 0 0 0 0 0 0 0

\end{lstlisting}
\end{tcolorbox}

\switchcolumn 

\begin{tcolorbox}[colback=cyan!10!white, colframe=black,
    boxrule=0.6pt, arc=1mm, breakable, enhanced,
    title=\centering \textbf{Prompt 2}]
\scriptsize
\begin{lstlisting}
Find the common rule that maps an 
input grid to an output grid, given the 
examples below.
Example1:
Input:
0 0 0 0 8 0 0 0 0 0 0 8 0 0 0 0 0 0 0
0 0 0 0 8 0 0 0 0 0 0 8 0 0 0 0 0 0 0
8 8 8 8 8 8 8 8 8 8 8 8 8 8 8 8 8 8 8
0 0 0 0 8 0 0 0 0 0 0 8 0 0 0 0 0 0 0
0 0 0 0 8 0 0 0 0 0 0 8 0 0 0 0 0 0 0
0 0 0 0 8 0 0 0 0 0 0 8 0 0 0 0 0 0 0
0 0 0 0 8 0 0 0 0 0 0 8 0 0 0 0 0 0 0
8 8 8 8 8 8 8 8 8 8 8 8 8 8 8 8 8 8 8
0 0 0 0 8 0 0 0 0 0 0 8 0 0 0 0 0 0 0
0 0 0 0 8 0 0 0 0 0 0 8 0 0 0 0 0 0 0
0 0 0 0 8 0 0 0 0 0 0 8 0 0 0 0 0 0 0
0 0 0 0 8 0 0 0 0 0 0 8 0 0 0 0 0 0 0
0 0 0 0 8 0 0 0 0 0 0 8 0 0 0 0 0 0 0
0 0 0 0 8 0 0 0 0 0 0 8 0 0 0 0 0 0 0
0 0 0 0 8 0 0 0 0 0 0 8 0 0 0 0 0 0 0
0 0 0 0 8 0 0 0 0 0 0 8 0 0 0 0 0 0 0
0 0 0 0 8 0 0 0 0 0 0 8 0 0 0 0 0 0 0
0 0 0 0 8 0 0 0 0 0 0 8 0 0 0 0 0 0 0
Output:
6 0 0 0 8 2 1 2 2 2 2 8 0 0 0 0 0 0 0
0 0 0 3 2 2 2 2 1 2 2 8 0 0 0 0 0 0 0
8 8 8 8 8 4 8 8 2 8 2 8 8 8 8 8 8 8 8
4 4 4 4 8 6 6 6 6 6 6 1 3 3 3 3 3 3 4
4 4 6 4 8 6 6 6 6 6 6 8 3 2 8 3 3 3 8
4 4 4 4 8 6 6 6 6 6 6 8 3 3 3 3 3 3 3
8 2 4 6 8 6 6 6 6 6 6 8 3 3 3 3 3 3 3
0 8 8 8 8 8 8 8 8 8 8 8 8 8 8 8 8 8 8
0 0 0 0 8 1 1 1 4 1 1 8 0 0 0 0 0 0 0
0 0 0 0 8 1 8 1 6 1 2 0 0 0 0 0 0 0 3
0 0 0 0 8 1 1 1 1 1 1 8 0 0 0 0 0 0 0
0 0 0 0 1 1 1 1 1 1 1 8 0 0 0 0 1 0 0
0 0 0 0 8 1 1 1 1 1 1 8 2 4 0 0 0 0 0
0 0 0 0 8 1 1 1 1 1 1 3 0 0 0 0 0 0 4
0 0 0 0 8 1 1 1 4 8 1 0 0 0 0 0 0 1 0
3 0 0 3 8 1 1 1 1 1 1 8 0 0 0 3 0 0 0
0 0 1 0 8 1 1 1 1 1 2 8 0 0 0 1 0 0 0
0 0 0 0 8 1 1 1 1 1 1 8 0 0 0 0 2 0 4

Example2:
Input:
0 0 8 0 0 0 0 0 0 8 0 0 0 0
0 0 8 0 0 0 0 0 0 8 0 0 0 0
0 0 8 0 0 0 0 0 0 8 0 0 0 0
0 0 8 0 0 0 0 0 0 8 0 0 0 0
8 8 8 8 8 8 8 8 8 8 8 8 8 8
0 0 8 0 0 0 0 0 0 8 0 0 0 0
0 0 8 0 0 0 0 0 0 8 0 0 0 0
8 8 8 8 8 8 8 8 8 8 8 8 8 8
0 0 8 0 0 0 0 0 0 8 0 0 0 0
0 0 8 0 0 0 0 0 0 8 0 0 0 0
0 0 8 0 0 0 0 0 0 8 0 0 0 0
0 0 8 0 0 0 0 0 0 8 0 0 0 0
Output:
0 1 8 2 2 2 2 2 2 8 0 0 0 0
0 0 0 2 2 2 2 2 2 8 0 0 0 0
0 0 8 2 2 2 2 2 2 1 0 0 0 0
0 0 8 2 2 2 2 2 2 8 0 0 0 0
0 8 8 0 8 8 8 8 8 8 8 8 8 8
4 4 8 4 6 3 6 6 6 8 4 3 3 3
4 4 8 4 6 6 6 6 2 8 3 3 3 3
8 1 8 8 8 8 8 8 8 8 8 8 8 8
0 0 2 0 1 0 1 1 1 8 1 0 0 0
0 0 8 1 1 0 1 1 1 8 4 0 0 0
0 0 8 1 1 4 1 1 1 8 0 0 0 0
6 0 8 1 1 1 1 0 1 8 0 0 0 8

Below is a test input grid. 
Predict the corresponding output. 
Input:
0 0 0 8 0 0 0 0 8 0 0 0 0 0 0
0 0 0 8 0 0 0 0 8 0 0 0 0 0 0
0 0 0 8 0 0 0 0 8 0 0 0 0 0 0
0 0 0 8 0 0 0 0 8 0 0 0 0 0 0
0 0 0 8 0 0 0 0 8 0 0 0 0 0 0
0 0 0 8 0 0 0 0 8 0 0 0 0 0 0
8 8 8 8 8 8 8 8 8 8 8 8 8 8 8
0 0 0 8 0 0 0 0 8 0 0 0 0 0 0
0 0 0 8 0 0 0 0 8 0 0 0 0 0 0
0 0 0 8 0 0 0 0 8 0 0 0 0 0 0
0 0 0 8 0 0 0 0 8 0 0 0 0 0 0
0 0 0 8 0 0 0 0 8 0 0 0 0 0 0
0 0 0 8 0 0 0 0 8 0 0 0 0 0 0
8 8 8 8 8 8 8 8 8 8 8 8 8 8 8
0 0 0 8 0 0 0 0 8 0 0 0 0 0 0
0 0 0 8 0 0 0 0 8 0 0 0 0 0 0
0 0 0 8 0 0 0 0 8 0 0 0 0 0 0
\end{lstlisting}
\end{tcolorbox}

\end{paracol}

\textbf{Pictorial Representation Prompt With New Prompt} 
\begin{paracol}{2} 

\begin{tcolorbox}[colback=cyan!10!white, colframe=black,
    boxrule=0.6pt, arc=1mm, breakable, enhanced,
    title=\centering \textbf{Prompt 1}]
\scriptsize
\begin{lstlisting}

Find the common rule that maps 
an input grid to an output 
grid, given the examples below. 
Note that random noise has been 
added to the input grids, such 
that different input grids may 
map to the same output. 
In Example 1, noisy input 
grids 1 map to the same 
output grid. Similarly, 
noisy input grids 2 map 
to their respective output grid.

Example 1:
Input:
0 0 0 3 4 0 0 0 3 0 0 8 0 0 0 0 0 0 0
0 0 0 0 3 0 4 0 0 0 0 8 0 0 0 0 0 0 0
8 8 8 8 8 8 8 8 8 8 8 8 8 8 0 8 8 8 8
0 0 0 0 8 0 0 0 0 0 0 8 0 0 0 0 8 0 0
0 0 0 0 8 0 0 0 3 0 0 8 0 0 0 0 0 0 0
0 0 4 0 8 0 0 4 0 0 0 8 6 4 4 0 1 0 0
8 0 0 0 8 0 0 0 0 0 0 8 0 0 0 0 6 0 0
8 8 8 8 8 8 8 8 8 8 8 8 3 4 8 8 8 8 8
0 0 0 0 4 0 3 0 1 0 0 3 0 0 0 4 0 0 0
0 0 0 0 8 0 0 0 0 0 0 2 0 2 0 0 0 0 0
0 0 8 2 0 0 0 0 0 0 0 8 0 0 0 0 0 0 0
4 0 0 0 8 0 0 0 0 0 0 8 0 0 0 0 0 0 0
0 0 0 0 8 0 0 0 0 0 0 8 3 0 0 0 0 8 0
0 0 0 0 8 0 0 1 0 0 0 8 0 0 0 0 0 0 0
0 2 0 3 8 0 0 0 2 0 0 8 0 0 0 0 3 0 0
0 0 0 2 3 0 0 0 0 0 0 8 8 0 0 0 0 0 0
0 0 0 0 8 0 0 1 0 0 0 8 2 0 0 0 0 6 0
0 0 0 0 8 0 0 0 0 0 0 8 0 0 0 0 0 0 0

Output:
0 0 0 0 8 2 2 2 2 2 2 8 0 0 0 0 0 0 0
0 0 0 0 8 2 2 2 2 2 2 8 0 0 0 0 0 0 0
8 8 8 8 8 8 8 8 8 8 8 8 8 8 8 8 8 8 8
4 4 4 4 8 6 6 6 6 6 6 8 3 3 3 3 3 3 3
4 4 4 4 8 6 6 6 6 6 6 8 3 3 3 3 3 3 3
4 4 4 4 8 6 6 6 6 6 6 8 3 3 3 3 3 3 3
4 4 4 4 8 6 6 6 6 6 6 8 3 3 3 3 3 3 3
8 8 8 8 8 8 8 8 8 8 8 8 8 8 8 8 8 8 8
0 0 0 0 8 1 1 1 1 1 1 8 0 0 0 0 0 0 0
0 0 0 0 8 1 1 1 1 1 1 8 0 0 0 0 0 0 0
0 0 0 0 8 1 1 1 1 1 1 8 0 0 0 0 0 0 0
0 0 0 0 8 1 1 1 1 1 1 8 0 0 0 0 0 0 0
0 0 0 0 8 1 1 1 1 1 1 8 0 0 0 0 0 0 0
0 0 0 0 8 1 1 1 1 1 1 8 0 0 0 0 0 0 0
0 0 0 0 8 1 1 1 1 1 1 8 0 0 0 0 0 0 0
0 0 0 0 8 1 1 1 1 1 1 8 0 0 0 0 0 0 0
0 0 0 0 8 1 1 1 1 1 1 8 0 0 0 0 0 0 0
0 0 0 0 8 1 1 1 1 1 1 8 0 0 0 0 0 0 0

Example 2:
Input:
0 0 8 0 0 0 1 0 8 8 0 0 0 0
0 0 8 0 0 0 1 0 0 8 0 0 8 0
0 0 8 0 4 0 0 0 0 6 0 0 8 0
0 0 0 0 0 0 0 0 0 8 0 0 0 0
8 8 8 8 3 8 8 8 8 8 8 8 8 8
0 0 6 0 0 0 0 0 0 8 0 0 0 1
0 0 8 0 0 0 8 0 0 8 0 0 3 0
8 8 8 8 8 6 8 8 2 6 8 8 8 8
0 0 8 0 0 0 0 0 0 8 0 0 0 0
0 0 8 0 0 0 0 2 0 8 0 0 0 0
0 0 8 0 0 2 0 0 0 8 8 2 0 0
0 0 8 6 0 0 0 0 0 8 0 0 0 0

Output:
0 0 8 2 2 2 2 2 2 8 0 0 0 0
0 0 8 2 2 2 2 2 2 8 0 0 0 0
0 0 8 2 2 2 2 2 2 8 0 0 0 0
0 0 8 2 2 2 2 2 2 8 0 0 0 0
8 8 8 8 8 8 8 8 8 8 8 8 8 8
4 4 8 6 6 6 6 6 6 8 3 3 3 3
4 4 8 6 6 6 6 6 6 8 3 3 3 3
8 8 8 8 8 8 8 8 8 8 8 8 8 8
0 0 8 1 1 1 1 1 1 8 0 0 0 0
0 0 8 1 1 1 1 1 1 8 0 0 0 0
0 0 8 1 1 1 1 1 1 8 0 0 0 0
0 0 8 1 1 1 1 1 1 8 0 0 0 0

Below is a test input grid.
Predict the corresponding output.

Input:
0 0 0 8 0 0 0 0 8 0 0 0 0 0 0
0 0 0 8 0 0 0 0 8 0 0 0 0 0 0
0 0 0 8 0 0 0 0 8 0 0 0 0 0 0
0 0 0 8 0 0 0 0 8 0 0 0 0 0 0
0 0 0 8 0 0 0 0 8 0 0 0 0 0 0
0 0 0 8 0 0 0 0 8 0 0 0 0 0 0
8 8 8 8 8 8 8 8 8 8 8 8 8 8 8
0 0 0 8 0 0 0 0 8 0 0 0 0 0 0
0 0 0 8 0 0 0 0 8 0 0 0 0 0 0
0 0 0 8 0 0 0 0 8 0 0 0 0 0 0
0 0 0 8 0 0 0 0 8 0 0 0 0 0 0
0 0 0 8 0 0 0 0 8 0 0 0 0 0 0
0 0 0 8 0 0 0 0 8 0 0 0 0 0 0
8 8 8 8 8 8 8 8 8 8 8 8 8 8 8
0 0 0 8 0 0 0 0 8 0 0 0 0 0 0
0 0 0 8 0 0 0 0 8 0 0 0 0 0 0
0 0 0 8 0 0 0 0 8 0 0 0 0 0 0



\end{lstlisting}
\end{tcolorbox}

\switchcolumn 

\begin{tcolorbox}[colback=cyan!10!white, colframe=black,
    boxrule=0.6pt, arc=1mm, breakable, enhanced,
    title=\centering \textbf{Prompt 2}]
\scriptsize
\begin{lstlisting}
Find the common rule that maps an 
input grid to an output grid, 
given the examples below. 
Note that random noise has 
been added to the output grids, 
such that different output 
grids may map to the same 
input. In Example 1, 
noisy output grids 1 map 
to the same input grid. 
Similarly, noisy output 
grids 2 map to their 
respective input grid.

Example1
Input:
0 0 0 0 8 0 0 0 0 0 0 8 0 0 0 0 0 0 0       
0 0 0 0 8 0 0 0 0 0 0 8 0 0 0 0 0 0 0       
8 8 8 8 8 8 8 8 8 8 8 8 8 8 8 8 8 8 8       
0 0 0 0 8 0 0 0 0 0 0 8 0 0 0 0 0 0 0       
0 0 0 0 8 0 0 0 0 0 0 8 0 0 0 0 0 0 0       
0 0 0 0 8 0 0 0 0 0 0 8 0 0 0 0 0 0 0       
0 0 0 0 8 0 0 0 0 0 0 8 0 0 0 0 0 0 0       
8 8 8 8 8 8 8 8 8 8 8 8 8 8 8 8 8 8 8       
0 0 0 0 8 0 0 0 0 0 0 8 0 0 0 0 0 0 0       
0 0 0 0 8 0 0 0 0 0 0 8 0 0 0 0 0 0 0       
0 0 0 0 8 0 0 0 0 0 0 8 0 0 0 0 0 0 0       
0 0 0 0 8 0 0 0 0 0 0 8 0 0 0 0 0 0 0       
0 0 0 0 8 0 0 0 0 0 0 8 0 0 0 0 0 0 0       
0 0 0 0 8 0 0 0 0 0 0 8 0 0 0 0 0 0 0       
0 0 0 0 8 0 0 0 0 0 0 8 0 0 0 0 0 0 0       
0 0 0 0 8 0 0 0 0 0 0 8 0 0 0 0 0 0 0       
0 0 0 0 8 0 0 0 0 0 0 8 0 0 0 0 0 0 0       
0 0 0 0 8 0 0 0 0 0 0 8 0 0 0 0 0 0 0       

Output:
0 3 0 0 8 2 0 2 2 2 2 2 0 0 0 0 0 0 0       
0 1 0 8 3 2 2 1 0 6 2 8 0 2 0 0 0 0 0       
8 4 8 8 6 8 8 3 8 8 8 8 8 8 8 8 3 8 8       
6 4 4 4 8 6 6 6 6 6 6 8 3 8 3 3 3 3 3       
4 4 4 4 8 6 6 6 6 6 6 8 3 3 3 3 3 3 3       
4 4 4 4 8 6 6 6 6 6 6 8 3 3 3 3 3 8 3       
4 4 4 4 0 6 1 6 6 6 6 8 3 3 3 3 3 3 3       
8 8 4 0 8 8 8 8 8 8 8 8 8 8 8 8 8 8 8       
4 2 0 0 8 4 1 1 1 4 1 8 0 0 0 1 0 0 0       
0 0 0 0 8 1 1 1 1 1 1 8 0 2 0 0 0 0 0       
0 0 0 0 8 1 1 1 1 4 1 8 0 0 0 0 0 3 0       
0 0 0 0 8 1 3 1 1 1 1 8 0 0 0 0 0 0 0       
0 0 0 0 8 2 1 1 1 1 1 8 0 0 0 1 0 0 8       
0 0 0 0 8 1 1 6 1 1 1 8 0 0 0 0 0 0 0       
0 0 0 0 8 1 1 1 1 1 6 8 0 1 0 0 0 0 0       
0 0 0 0 3 1 3 8 1 1 1 8 0 0 0 0 0 0 0       
0 0 0 0 8 1 1 1 4 1 1 8 0 0 0 0 0 0 0       
0 0 0 0 8 6 1 2 1 1 1 8 0 0 0 0 0 0 0       
Example2
Input:
0 0 8 0 0 0 0 0 0 8 0 0 0 0
0 0 8 0 0 0 0 0 0 8 0 0 0 0
0 0 8 0 0 0 0 0 0 8 0 0 0 0
0 0 8 0 0 0 0 0 0 8 0 0 0 0
8 8 8 8 8 8 8 8 8 8 8 8 8 8
0 0 8 0 0 0 0 0 0 8 0 0 0 0
0 0 8 0 0 0 0 0 0 8 0 0 0 0
8 8 8 8 8 8 8 8 8 8 8 8 8 8
0 0 8 0 0 0 0 0 0 8 0 0 0 0
0 0 8 0 0 0 0 0 0 8 0 0 0 0
0 0 8 0 0 0 0 0 0 8 0 0 0 0
0 0 8 0 0 0 0 0 0 8 0 0 0 0

Output:
0 0 8 2 2 2 2 2 2 8 0 0 0 0
0 0 8 2 2 6 2 2 2 8 0 0 0 0
0 0 3 2 2 2 2 2 2 8 0 0 2 0
4 4 0 2 8 2 2 2 2 8 0 0 0 4
8 8 8 8 8 8 8 0 8 8 6 8 8 8
4 4 8 6 6 6 6 6 8 8 3 3 3 2
4 4 8 6 6 6 8 6 6 8 6 3 0 3
8 8 8 8 8 8 8 4 8 8 3 8 8 8
0 0 8 1 1 1 3 1 1 8 0 6 0 0
0 0 8 1 1 6 1 1 1 8 4 0 0 0
0 0 8 1 1 1 1 1 1 8 0 0 0 0
0 0 8 1 1 1 1 1 1 8 0 0 0 0

Below is a test input grid.
Predict the corresponding output.

Input:
0 0 0 8 0 0 0 0 8 0 0 0 0 0 0
0 0 0 8 0 0 0 0 8 0 0 0 0 0 0
0 0 0 8 0 0 0 0 8 0 0 0 0 0 0
0 0 0 8 0 0 0 0 8 0 0 0 0 0 0
0 0 0 8 0 0 0 0 8 0 0 0 0 0 0
0 0 0 8 0 0 0 0 8 0 0 0 0 0 0
8 8 8 8 8 8 8 8 8 8 8 8 8 8 8
0 0 0 8 0 0 0 0 8 0 0 0 0 0 0
0 0 0 8 0 0 0 0 8 0 0 0 0 0 0
0 0 0 8 0 0 0 0 8 0 0 0 0 0 0
0 0 0 8 0 0 0 0 8 0 0 0 0 0 0
0 0 0 8 0 0 0 0 8 0 0 0 0 0 0
0 0 0 8 0 0 0 0 8 0 0 0 0 0 0
8 8 8 8 8 8 8 8 8 8 8 8 8 8 8
0 0 0 8 0 0 0 0 8 0 0 0 0 0 0
0 0 0 8 0 0 0 0 8 0 0 0 0 0 0
0 0 0 8 0 0 0 0 8 0 0 0 0 0 0
\end{lstlisting}
\end{tcolorbox}

\end{paracol}

\begin{paracol}{2} 

\begin{tcolorbox}[
  colback=cyan!10!white,
  colframe=black,
  boxrule=0.6pt,
  arc=1mm,
  breakable,
  enhanced,
  title=\centering \textbf{1-Noisy Output Example With Original Prompt}
]
\scriptsize
\begin{lstlisting}
Find the common rule that maps 
an input grid to an output grid, 
given the examples below.

Example1:
Input:
0 0 0 0 8 0 0 0 0 0 0 8 0 0 0 0 0 0 0
0 0 0 0 8 0 0 0 0 0 0 8 0 0 0 0 0 0 0
8 8 8 8 8 8 8 8 8 8 8 8 8 8 8 8 8 8 6
0 0 0 0 2 0 0 0 0 0 0 8 0 0 0 0 0 0 0
0 0 0 0 8 0 0 0 0 0 0 8 0 0 0 0 0 0 0
0 8 0 0 8 2 0 0 0 0 0 8 0 0 0 0 0 4 0
0 0 0 0 8 0 0 0 2 0 0 8 8 0 0 0 0 0 0
8 8 8 8 8 8 8 8 8 8 8 8 8 8 8 8 8 8 8
0 0 6 0 8 3 0 0 0 0 0 6 0 0 0 0 0 0 0
0 0 0 0 8 6 0 0 0 0 0 8 0 0 0 0 0 0 0
0 0 0 0 8 0 0 0 0 0 0 8 0 0 0 0 0 0 0
0 0 0 0 8 0 0 3 0 0 0 8 0 0 0 0 0 3 0
0 0 0 0 8 0 0 0 0 0 0 8 0 2 0 0 0 0 0
0 0 0 0 8 0 0 0 0 0 0 8 0 0 0 0 0 0 0
0 0 0 0 8 0 0 0 0 0 0 8 0 0 0 0 0 0 0
0 0 0 0 8 0 0 0 0 0 0 8 0 0 0 0 0 0 0
0 0 0 0 8 0 0 4 0 0 0 8 1 0 0 8 0 0 0
0 0 0 0 8 0 0 0 0 0 0 8 0 0 0 0 0 0 0

Output:
0 0 0 0 8 2 2 2 2 2 2 8 0 0 0 0 0 0 0
0 0 0 0 8 2 2 2 2 2 2 8 0 0 0 0 0 0 0
8 8 8 8 8 8 8 8 8 8 8 8 8 8 8 8 8 8 8
4 4 4 4 8 6 6 6 6 6 6 8 3 3 3 3 3 3 3
4 4 4 4 8 6 6 6 6 6 6 8 3 3 3 3 3 3 3
4 4 4 4 8 6 6 6 6 6 6 8 3 3 3 3 3 3 3
4 4 4 4 8 6 6 6 6 6 6 8 3 3 3 3 3 3 3
8 8 8 8 8 8 8 8 8 8 8 8 8 8 8 8 8 8 8
0 0 0 0 8 1 1 1 1 1 1 8 0 0 0 0 0 0 0
0 0 0 0 8 1 1 1 1 1 1 8 0 0 0 0 0 0 0
0 0 0 0 8 1 1 1 1 1 1 8 0 0 0 0 0 0 0
0 0 0 0 8 1 1 1 1 1 1 8 0 0 0 0 0 0 0
0 0 0 0 8 1 1 1 1 1 1 8 0 0 0 0 0 0 0
0 0 0 0 8 1 1 1 1 1 1 8 0 0 0 0 0 0 0
0 0 0 0 8 1 1 1 1 1 1 8 0 0 0 0 0 0 0
0 0 0 0 8 1 1 1 1 1 1 8 0 0 0 0 0 0 0
0 0 0 0 8 1 1 1 1 1 1 8 0 0 0 0 0 0 0
0 0 0 0 8 1 1 1 1 1 1 8 0 0 0 0 0 0 0

Example2:
Input:
0 4 8 0 0 0 0 0 0 8 0 0 0 0
2 0 8 0 0 0 0 0 0 8 0 0 0 0
0 0 8 0 0 0 0 0 0 8 0 0 0 0
0 0 8 0 0 0 0 0 0 8 0 0 0 0
8 8 8 8 8 8 8 8 8 8 8 8 8 8
0 0 8 0 0 0 0 3 0 0 0 0 0 0
0 0 8 0 0 2 0 0 0 8 0 0 0 0
8 8 8 8 8 8 2 8 8 8 8 8 8 8
0 0 8 0 0 0 0 0 0 8 0 1 0 0
0 0 8 0 0 0 0 0 0 8 0 0 0 0
0 0 8 0 0 0 0 0 0 8 0 0 0 0
0 0 8 0 0 0 0 6 0 8 0 0 0 0

Output:
0 0 8 2 2 2 2 2 2 8 0 0 0 0
0 0 8 2 2 2 2 2 2 8 0 0 0 0
0 0 8 2 2 2 2 2 2 8 0 0 0 0
0 0 8 2 2 2 2 2 2 8 0 0 0 0
8 8 8 8 8 8 8 8 8 8 8 8 8 8
4 4 8 6 6 6 6 6 6 8 3 3 3 3
4 4 8 6 6 6 6 6 6 8 3 3 3 3
8 8 8 8 8 8 8 8 8 8 8 8 8 8
0 0 8 1 1 1 1 1 1 8 0 0 0 0
0 0 8 1 1 1 1 1 1 8 0 0 0 0
0 0 8 1 1 1 1 1 1 8 0 0 0 0
0 0 8 1 1 1 1 1 1 8 0 0 0 0

Below is a test input grid. 
Predict the corresponding output. 

Input:
0 0 0 8 0 0 0 0 8 0 0 0 0 0 0
0 0 0 8 0 0 0 0 8 0 0 0 0 0 0
0 0 0 8 0 0 0 0 8 0 0 0 0 0 0
0 0 0 8 0 0 0 0 8 0 0 0 0 0 0
0 0 0 8 0 0 0 0 8 0 0 0 0 0 0
0 0 0 8 0 0 0 0 8 0 0 0 0 0 0
8 8 8 8 8 8 8 8 8 8 8 8 8 8 8
0 0 0 8 0 0 0 0 8 0 0 0 0 0 0
0 0 0 8 0 0 0 0 8 0 0 0 0 0 0
0 0 0 8 0 0 0 0 8 0 0 0 0 0 0
0 0 0 8 0 0 0 0 8 0 0 0 0 0 0
0 0 0 8 0 0 0 0 8 0 0 0 0 0 0
0 0 0 8 0 0 0 0 8 0 0 0 0 0 0
8 8 8 8 8 8 8 8 8 8 8 8 8 8 8
0 0 0 8 0 0 0 0 8 0 0 0 0 0 0
0 0 0 8 0 0 0 0 8 0 0 0 0 0 0
0 0 0 8 0 0 0 0 8 0 0 0 0 0 0

\end{lstlisting}
\end{tcolorbox}

\switchcolumn 

\begin{tcolorbox}[
  colback=cyan!10!white,
  colframe=black,
  boxrule=0.6pt,
  arc=1mm,
  breakable,
  enhanced,
  title=\centering \textbf{1-Noisy Output Example With Original Prompt}
]
\scriptsize
\begin{lstlisting}
Find the common rule that maps 
an input grid to an output 
grid, given the examples below.

Example1:
Input:
0 0 0 0 8 0 0 0 0 0 0 8 0 0 0 0 0 0 0
0 0 0 0 8 0 0 0 0 0 0 8 0 0 0 0 0 0 0
8 8 8 8 8 8 8 8 8 8 8 8 8 8 8 8 8 8 8
0 0 0 0 8 0 0 0 0 0 0 8 0 0 0 0 0 0 0
0 0 0 0 8 0 0 0 0 0 0 8 0 0 0 0 0 0 0
0 0 0 0 8 0 0 0 0 0 0 8 0 0 0 0 0 0 0
0 0 0 0 8 0 0 0 0 0 0 8 0 0 0 0 0 0 0
8 8 8 8 8 8 8 8 8 8 8 8 8 8 8 8 8 8 8
0 0 0 0 8 0 0 0 0 0 0 8 0 0 0 0 0 0 0
0 0 0 0 8 0 0 0 0 0 0 8 0 0 0 0 0 0 0
0 0 0 0 8 0 0 0 0 0 0 8 0 0 0 0 0 0 0
0 0 0 0 8 0 0 0 0 0 0 8 0 0 0 0 0 0 0
0 0 0 0 8 0 0 0 0 0 0 8 0 0 0 0 0 0 0
0 0 0 0 8 0 0 0 0 0 0 8 0 0 0 0 0 0 0
0 0 0 0 8 0 0 0 0 0 0 8 0 0 0 0 0 0 0
0 0 0 0 8 0 0 0 0 0 0 8 0 0 0 0 0 0 0
0 0 0 0 8 0 0 0 0 0 0 8 0 0 0 0 0 0 0
0 0 0 0 8 0 0 0 0 0 0 8 0 0 0 0 0 0 0

Output:
0 0 0 0 8 2 4 2 2 2 2 8 0 0 0 0 0 0 0
0 0 0 0 8 2 2 2 8 1 2 8 0 0 0 0 0 0 0
8 8 8 8 8 8 8 8 8 8 8 8 8 8 4 8 8 8 8
4 4 4 4 8 6 1 6 6 6 6 8 3 3 3 3 3 3 3
4 6 4 4 8 6 6 6 6 6 6 8 3 3 3 3 3 3 3
4 4 4 4 8 6 6 6 6 6 6 8 3 3 6 3 3 3 3
4 4 4 4 8 6 6 6 6 6 6 8 3 3 3 3 3 3 3
8 8 8 8 8 8 8 8 8 8 8 8 8 8 8 8 8 8 8
0 0 0 0 8 1 1 1 1 1 1 8 0 0 0 0 0 0 0
0 0 0 0 8 1 1 1 1 6 1 8 0 0 0 4 0 0 0
0 0 0 0 8 1 1 0 1 1 1 8 0 0 0 0 0 0 0
0 0 0 0 8 1 1 1 1 1 1 8 0 0 0 0 0 0 0
0 0 0 0 8 1 1 1 4 1 3 8 0 0 0 0 0 0 3
0 0 0 0 8 1 1 1 6 1 1 8 0 0 0 0 0 0 0
0 0 0 0 8 1 1 1 1 1 1 8 0 0 0 0 0 0 0
0 0 8 0 8 1 1 1 1 1 1 8 0 0 0 0 0 0 0
0 0 0 0 8 1 1 1 1 1 1 8 0 0 0 0 0 0 0
0 0 0 0 8 1 1 1 1 0 1 8 8 0 0 0 0 0 0

Example2:
Input:
0 0 8 0 0 0 0 0 0 8 0 0 0 0
0 0 8 0 0 0 0 0 0 8 0 0 0 0
0 0 8 0 0 0 0 0 0 8 0 0 0 0
0 0 8 0 0 0 0 0 0 8 0 0 0 0
8 8 8 8 8 8 8 8 8 8 8 8 8 8
0 0 8 0 0 0 0 0 0 8 0 0 0 0
0 0 8 0 0 0 0 0 0 8 0 0 0 0
8 8 8 8 8 8 8 8 8 8 8 8 8 8
0 0 8 0 0 0 0 0 0 8 0 0 0 0
0 0 8 0 0 0 0 0 0 8 0 0 0 0
0 0 8 0 0 0 0 0 0 8 0 0 0 0
0 0 8 0 0 0 0 0 0 8 0 0 0 0

Output:
0 0 2 2 2 2 2 2 2 8 0 0 0 0
0 0 8 2 2 2 2 2 2 8 0 0 0 0
0 0 8 2 2 2 2 2 2 8 4 0 3 0
0 0 8 2 2 2 2 2 2 8 8 0 0 0
8 8 8 8 8 8 8 8 8 8 8 8 8 8
4 4 8 6 6 6 6 6 6 8 3 3 3 3
4 4 8 6 6 6 6 6 6 8 3 3 3 3
8 8 1 8 8 8 8 8 8 8 8 8 8 8
0 2 8 1 1 1 1 1 1 8 0 0 0 0
0 0 8 1 1 1 1 1 1 8 0 0 0 0
0 0 8 1 1 1 1 2 1 8 0 0 0 0
0 0 8 1 1 1 8 1 1 8 0 0 0 0

Below is a test input grid. 
Predict the corresponding output. 

Input:
0 0 0 8 0 0 0 0 8 0 0 0 0 0 0
0 0 0 8 0 0 0 0 8 0 0 0 0 0 0
0 0 0 8 0 0 0 0 8 0 0 0 0 0 0
0 0 0 8 0 0 0 0 8 0 0 0 0 0 0
0 0 0 8 0 0 0 0 8 0 0 0 0 0 0
0 0 0 8 0 0 0 0 8 0 0 0 0 0 0
8 8 8 8 8 8 8 8 8 8 8 8 8 8 8
0 0 0 8 0 0 0 0 8 0 0 0 0 0 0
0 0 0 8 0 0 0 0 8 0 0 0 0 0 0
0 0 0 8 0 0 0 0 8 0 0 0 0 0 0
0 0 0 8 0 0 0 0 8 0 0 0 0 0 0
0 0 0 8 0 0 0 0 8 0 0 0 0 0 0
0 0 0 8 0 0 0 0 8 0 0 0 0 0 0
8 8 8 8 8 8 8 8 8 8 8 8 8 8 8
0 0 0 8 0 0 0 0 8 0 0 0 0 0 0
0 0 0 8 0 0 0 0 8 0 0 0 0 0 0
0 0 0 8 0 0 0 0 8 0 0 0 0 0 0
\end{lstlisting}
\end{tcolorbox}

\end{paracol}
\begin{paracol}{2} 

\begin{tcolorbox}[
  colback=cyan!10!white,
  colframe=black,
  boxrule=0.6pt,
  arc=1mm,
  breakable,
  enhanced,
  title=\centering \textbf{1-Noisy Input Example With New Prompt}
]
\scriptsize
\begin{lstlisting}
Find the common rule that 
maps an input grid to an 
output grid, given the 
examples below. Note that 
random noise has been added 
to the input grids, such 
that different input grids 
may map to the same output. 
In Example 1, noisy input 
grids 1 map to the same 
output grid. Similarly,
Example 2 noisy input 
grids 2 map to their 
respective output grid

Example 1:

Input:
0 0 0 0 8 0 0 2 4 0 2 8 0 0 0 0 0 0 0
0 0 0 0 8 0 0 0 0 1 0 8 0 0 0 0 0 0 0
8 8 3 8 8 8 8 8 8 8 8 8 8 8 8 8 8 8 8
0 0 0 0 8 0 0 0 4 0 0 8 0 0 0 0 0 0 0
0 0 0 0 8 0 0 0 0 0 0 8 3 0 0 0 0 0 0
0 0 0 0 8 0 0 0 0 0 0 8 0 0 0 0 0 0 0
0 0 0 0 8 0 0 3 0 0 0 8 0 0 0 0 0 0 0
8 8 8 8 8 8 8 8 8 8 8 8 8 8 8 8 8 8 8
0 0 0 0 8 0 0 0 0 0 0 8 0 0 0 0 0 0 1
0 0 0 0 8 0 0 0 0 0 0 8 0 0 0 0 0 0 0
1 0 0 0 8 0 0 0 0 0 0 8 0 0 4 0 0 8 0
0 0 0 0 8 0 0 0 0 0 0 8 0 0 0 0 0 0 0
0 0 0 0 8 0 0 0 0 3 0 8 0 0 0 0 0 0 0
0 0 0 0 8 0 0 0 0 0 0 8 0 0 0 0 0 0 0
0 0 0 0 8 0 0 0 0 0 0 8 0 0 0 0 0 0 0
2 0 0 0 8 0 0 0 0 0 0 8 0 0 0 0 0 0 0
0 0 0 0 8 0 0 0 0 0 0 8 0 0 0 0 0 0 0
0 0 0 0 0 0 0 1 4 0 0 8 0 0 0 0 0 0 0

Output:
0 0 0 0 8 2 2 2 2 2 2 8 0 0 0 0 0 0 0
0 0 0 0 8 2 2 2 2 2 2 8 0 0 0 0 0 0 0
8 8 8 8 8 8 8 8 8 8 8 8 8 8 8 8 8 8 8
4 4 4 4 8 6 6 6 6 6 6 8 3 3 3 3 3 3 3
4 4 4 4 8 6 6 6 6 6 6 8 3 3 3 3 3 3 3
4 4 4 4 8 6 6 6 6 6 6 8 3 3 3 3 3 3 3
4 4 4 4 8 6 6 6 6 6 6 8 3 3 3 3 3 3 3
8 8 8 8 8 8 8 8 8 8 8 8 8 8 8 8 8 8 8
0 0 0 0 8 1 1 1 1 1 1 8 0 0 0 0 0 0 0
0 0 0 0 8 1 1 1 1 1 1 8 0 0 0 0 0 0 0
0 0 0 0 8 1 1 1 1 1 1 8 0 0 0 0 0 0 0
0 0 0 0 8 1 1 1 1 1 1 8 0 0 0 0 0 0 0
0 0 0 0 8 1 1 1 1 1 1 8 0 0 0 0 0 0 0
0 0 0 0 8 1 1 1 1 1 1 8 0 0 0 0 0 0 0
0 0 0 0 8 1 1 1 1 1 1 8 0 0 0 0 0 0 0
0 0 0 0 8 1 1 1 1 1 1 8 0 0 0 0 0 0 0
0 0 0 0 8 1 1 1 1 1 1 8 0 0 0 0 0 0 0
0 0 0 0 8 1 1 1 1 1 1 8 0 0 0 0 0 0 0

Example 2:

Input:
0 0 8 0 0 0 0 0 6 8 0 0 0 0
0 0 8 0 0 0 0 0 0 8 0 0 0 0
0 0 8 0 0 0 0 0 0 8 3 0 0 0
0 0 8 0 0 0 0 0 3 8 0 0 0 0
8 8 8 8 8 8 8 8 8 2 8 8 8 8
0 0 8 0 0 0 0 1 0 8 0 0 0 0
0 0 8 0 0 0 0 0 8 8 0 0 0 1
8 8 8 8 8 8 8 8 8 8 8 8 8 3
0 0 8 0 0 0 0 0 0 8 0 0 0 0
0 0 8 0 0 0 0 0 0 8 0 0 0 0
0 0 8 0 0 0 0 0 0 8 0 0 0 0
0 0 8 0 0 0 0 0 0 8 0 0 0 0

Output:
0 0 8 2 2 2 2 2 2 8 0 0 0 0
0 0 8 2 2 2 2 2 2 8 0 0 0 0
0 0 8 2 2 2 2 2 2 8 0 0 0 0
0 0 8 2 2 2 2 2 2 8 0 0 0 0
8 8 8 8 8 8 8 8 8 8 8 8 8 8
4 4 8 6 6 6 6 6 6 8 3 3 3 3
4 4 8 6 6 6 6 6 6 8 3 3 3 3
8 8 8 8 8 8 8 8 8 8 8 8 8 8
0 0 8 1 1 1 1 1 1 8 0 0 0 0
0 0 8 1 1 1 1 1 1 8 0 0 0 0
0 0 8 1 1 1 1 1 1 8 0 0 0 0
0 0 8 1 1 1 1 1 1 8 0 0 0 0

Below is a test input grid. 
Predict the corresponding output. 

Input:
0 0 0 8 0 0 0 0 8 0 0 0 0 0 0
0 0 0 8 0 0 0 0 8 0 0 0 0 0 0
0 0 0 8 0 0 0 0 8 0 0 0 0 0 0
0 0 0 8 0 0 0 0 8 0 0 0 0 0 0
0 0 0 8 0 0 0 0 8 0 0 0 0 0 0
0 0 0 8 0 0 0 0 8 0 0 0 0 0 0
8 8 8 8 8 8 8 8 8 8 8 8 8 8 8
0 0 0 8 0 0 0 0 8 0 0 0 0 0 0
0 0 0 8 0 0 0 0 8 0 0 0 0 0 0
0 0 0 8 0 0 0 0 8 0 0 0 0 0 0
0 0 0 8 0 0 0 0 8 0 0 0 0 0 0
0 0 0 8 0 0 0 0 8 0 0 0 0 0 0
0 0 0 8 0 0 0 0 8 0 0 0 0 0 0
8 8 8 8 8 8 8 8 8 8 8 8 8 8 8
0 0 0 8 0 0 0 0 8 0 0 0 0 0 0
0 0 0 8 0 0 0 0 8 0 0 0 0 0 0
0 0 0 8 0 0 0 0 8 0 0 0 0 0 0
\end{lstlisting}
\end{tcolorbox}

\switchcolumn 

\begin{tcolorbox}[
  colback=cyan!10!white,
  colframe=black,
  boxrule=0.6pt,
  arc=1mm,
  breakable,
  enhanced,
  title=\centering \textbf{1-Noisy Output Example With New Prompt}
]
\scriptsize
\begin{lstlisting}
Find the common rule that 
maps an input grid to an 
output grid, given the 
examples below. Note that 
random noise has been added 
to the output grids, such 
that different output grids 
may map to the same input. 
In Example 1, noisy output 
grids 1 map to the same 
input grid. Similarly, 
Example 2 noisy output 
grids 2  map to their 
respective input grid.

Example1
Input:
0 0 0 0 8 0 0 0 0 0 0 8 0 0 0 0 0 0 0
0 0 0 0 8 0 0 0 0 0 0 8 0 0 0 0 0 0 0
8 8 8 8 8 8 8 8 8 8 8 8 8 8 8 8 8 8 8
0 0 0 0 8 0 0 0 0 0 0 8 0 0 0 0 0 0 0
0 0 0 0 8 0 0 0 0 0 0 8 0 0 0 0 0 0 0
0 0 0 0 8 0 0 0 0 0 0 8 0 0 0 0 0 0 0
0 0 0 0 8 0 0 0 0 0 0 8 0 0 0 0 0 0 0
8 8 8 8 8 8 8 8 8 8 8 8 8 8 8 8 8 8 8
0 0 0 0 8 0 0 0 0 0 0 8 0 0 0 0 0 0 0
0 0 0 0 8 0 0 0 0 0 0 8 0 0 0 0 0 0 0
0 0 0 0 8 0 0 0 0 0 0 8 0 0 0 0 0 0 0
0 0 0 0 8 0 0 0 0 0 0 8 0 0 0 0 0 0 0
0 0 0 0 8 0 0 0 0 0 0 8 0 0 0 0 0 0 0
0 0 0 0 8 0 0 0 0 0 0 8 0 0 0 0 0 0 0
0 0 0 0 8 0 0 0 0 0 0 8 0 0 0 0 0 0 0
0 0 0 0 8 0 0 0 0 0 0 8 0 0 0 0 0 0 0
0 0 0 0 8 0 0 0 0 0 0 8 0 0 0 0 0 0 0
0 0 0 0 8 0 0 0 0 0 0 8 0 0 0 0 0 0 0

Output:
0 0 0 0 8 2 2 2 2 2 2 8 0 0 0 0 8 0 0
0 0 8 0 8 2 2 2 2 2 2 8 0 0 0 0 0 0 0
8 8 8 8 8 8 8 8 8 8 8 8 8 8 8 8 8 8 8
4 4 4 4 8 6 6 6 6 6 6 8 3 3 3 3 3 3 4
4 4 4 4 8 6 6 6 6 6 6 8 3 3 3 3 3 3 3
4 4 4 4 8 6 6 6 6 6 6 8 3 3 3 3 3 3 3
4 4 4 4 8 6 6 6 6 6 6 8 3 3 3 3 3 3 3
8 8 8 8 8 8 8 8 8 4 4 8 8 8 8 8 8 1 8
0 0 0 6 8 1 1 1 1 1 1 8 0 0 0 0 0 0 0
0 0 4 0 8 1 1 1 1 1 1 8 0 0 0 0 0 0 0
0 0 0 0 8 1 6 1 1 1 1 8 0 2 0 0 4 0 0
0 0 0 6 8 1 1 1 1 1 1 8 0 0 0 0 0 0 0
0 0 0 0 8 1 1 1 1 1 1 8 0 0 0 0 0 0 0
6 0 0 0 8 1 1 1 1 1 1 8 0 6 0 0 0 0 0
0 0 0 0 8 1 1 1 1 1 1 8 0 0 0 0 0 0 0
0 0 0 0 8 1 1 1 1 1 1 8 0 1 0 0 0 0 0
0 0 0 0 8 1 1 1 1 1 1 8 0 0 0 0 0 0 0
0 0 8 0 8 1 1 1 1 1 1 8 0 0 6 0 0 0 0
Example2
Input:
0 0 8 0 0 0 0 0 0 8 0 0 0 0
0 0 8 0 0 0 0 0 0 8 0 0 0 0
0 0 8 0 0 0 0 0 0 8 0 0 0 0
0 0 8 0 0 0 0 0 0 8 0 0 0 0
8 8 8 8 8 8 8 8 8 8 8 8 8 8
0 0 8 0 0 0 0 0 0 8 0 0 0 0
0 0 8 0 0 0 0 0 0 8 0 0 0 0
8 8 8 8 8 8 8 8 8 8 8 8 8 8
0 0 8 0 0 0 0 0 0 8 0 0 0 0
0 0 8 0 0 0 0 0 0 8 0 0 0 0
0 0 8 0 0 0 0 0 0 8 0 0 0 0
0 0 8 0 0 0 0 0 0 8 0 0 0 0

Output:
0 0 8 2 2 2 2 2 2 8 0 0 0 0
0 0 8 2 2 2 2 2 2 8 0 0 0 0
0 0 8 2 2 2 2 2 2 6 0 0 0 0
0 0 8 2 0 2 2 2 2 8 0 0 0 0
8 8 8 8 8 8 8 8 8 8 8 8 8 8
4 4 8 6 6 6 6 6 6 8 3 6 3 3
4 4 8 6 6 6 6 6 6 8 3 3 3 8
8 8 8 8 8 8 8 8 8 8 8 8 0 8
0 0 8 1 1 1 1 1 1 8 8 0 0 0
0 0 4 1 1 1 1 1 1 8 0 0 0 0
4 0 8 1 1 1 1 1 1 8 0 0 0 0
0 0 8 1 1 1 1 1 1 8 0 0 0 0
Below is a test input grid. 
Predict the corresponding output. 

Input:
0 0 0 8 0 0 0 0 8 0 0 0 0 0 0
0 0 0 8 0 0 0 0 8 0 0 0 0 0 0
0 0 0 8 0 0 0 0 8 0 0 0 0 0 0
0 0 0 8 0 0 0 0 8 0 0 0 0 0 0
0 0 0 8 0 0 0 0 8 0 0 0 0 0 0
0 0 0 8 0 0 0 0 8 0 0 0 0 0 0
8 8 8 8 8 8 8 8 8 8 8 8 8 8 8
0 0 0 8 0 0 0 0 8 0 0 0 0 0 0
0 0 0 8 0 0 0 0 8 0 0 0 0 0 0
0 0 0 8 0 0 0 0 8 0 0 0 0 0 0
0 0 0 8 0 0 0 0 8 0 0 0 0 0 0
0 0 0 8 0 0 0 0 8 0 0 0 0 0 0
0 0 0 8 0 0 0 0 8 0 0 0 0 0 0
8 8 8 8 8 8 8 8 8 8 8 8 8 8 8
0 0 0 8 0 0 0 0 8 0 0 0 0 0 0
0 0 0 8 0 0 0 0 8 0 0 0 0 0 0
0 0 0 8 0 0 0 0 8 0 0 0 0 0 0

\end{lstlisting}
\end{tcolorbox}

\end{paracol}
\begin{paracol}{2} 

\begin{tcolorbox}[
  colback=cyan!10!white,
  colframe=black,
  boxrule=0.6pt,
  arc=1mm,
  breakable,
  enhanced,
  title=\centering \textbf{3-Noisy Input Example With Original Prompt}
]
\scriptsize
\begin{lstlisting}
Find the common rule that maps 
an input grid to an output grid, 
given the examples below.

Example 1:

Input:
0 0 0 0 8 0 0 0 0 0 0 8 0 0 4 0 0 8 0
0 0 0 0 8 0 0 0 0 0 0 8 0 0 0 0 1 0 0
8 8 8 8 8 0 8 8 8 8 8 8 8 8 8 8 8 8 8
0 0 0 0 8 0 0 0 0 0 0 8 0 8 0 0 0 0 0
0 0 0 1 8 0 0 0 0 0 0 8 0 0 0 0 0 0 0
0 0 0 0 8 0 0 0 0 0 0 8 0 0 0 2 0 0 0
0 0 0 0 4 0 0 0 0 0 0 8 0 0 0 0 1 0 0
8 8 8 8 8 8 8 8 8 4 8 8 8 8 8 8 8 8 8
0 0 0 0 8 0 0 6 0 0 0 8 0 0 0 0 0 0 0
0 0 0 0 8 0 0 0 0 0 0 8 0 0 0 0 0 0 2
0 0 0 8 8 0 0 0 0 0 0 8 0 0 0 0 0 0 0
0 0 0 0 8 0 0 0 0 0 6 8 0 0 0 0 0 0 0
2 0 0 0 8 0 0 0 0 0 0 8 0 0 0 0 0 0 0
0 0 0 0 8 0 0 0 0 0 0 8 0 0 0 0 0 0 0
0 0 0 0 8 0 0 0 0 0 0 8 0 0 0 0 0 0 0
0 0 0 0 8 0 0 0 0 0 0 8 0 0 0 0 0 0 0
0 0 0 0 8 0 0 0 0 0 0 8 0 0 0 0 0 0 0
0 0 0 0 8 8 0 0 0 0 0 8 0 0 0 0 6 0 0

Output:
0 0 0 0 8 2 2 2 2 2 2 8 0 0 0 0 0 0 0
0 0 0 0 8 2 2 2 2 2 2 8 0 0 0 0 0 0 0
8 8 8 8 8 8 8 8 8 8 8 8 8 8 8 8 8 8 8
4 4 4 4 8 6 6 6 6 6 6 8 3 3 3 3 3 3 3
4 4 4 4 8 6 6 6 6 6 6 8 3 3 3 3 3 3 3
4 4 4 4 8 6 6 6 6 6 6 8 3 3 3 3 3 3 3
4 4 4 4 8 6 6 6 6 6 6 8 3 3 3 3 3 3 3
8 8 8 8 8 8 8 8 8 8 8 8 8 8 8 8 8 8 8
0 0 0 0 8 1 1 1 1 1 1 8 0 0 0 0 0 0 0
0 0 0 0 8 1 1 1 1 1 1 8 0 0 0 0 0 0 0
0 0 0 0 8 1 1 1 1 1 1 8 0 0 0 0 0 0 0
0 0 0 0 8 1 1 1 1 1 1 8 0 0 0 0 0 0 0
0 0 0 0 8 1 1 1 1 1 1 8 0 0 0 0 0 0 0
0 0 0 0 8 1 1 1 1 1 1 8 0 0 0 0 0 0 0
0 0 0 0 8 1 1 1 1 1 1 8 0 0 0 0 0 0 0
0 0 0 0 8 1 1 1 1 1 1 8 0 0 0 0 0 0 0
0 0 0 0 8 1 1 1 1 1 1 8 0 0 0 0 0 0 0
0 0 0 0 8 1 1 1 1 1 1 8 0 0 0 0 0 0 0


Example 2:

Input:
1 0 0 0 8 0 0 0 0 0 0 8 0 6 0 0 0 0 0
0 0 0 0 1 0 0 0 0 0 0 8 0 0 0 0 0 0 0
8 8 8 8 8 8 8 8 8 8 8 8 8 8 8 8 8 8 8
0 0 0 0 8 0 2 0 0 0 0 8 0 0 0 0 0 2 0
0 0 0 1 8 0 0 0 0 0 0 8 0 0 0 0 4 0 0
0 0 0 0 8 0 0 0 0 0 0 8 0 0 0 0 0 0 0
0 0 0 0 8 0 0 0 0 0 0 8 3 0 0 0 0 0 0
8 8 8 8 8 0 8 8 8 8 8 8 8 8 8 8 8 8 8
0 0 0 0 8 0 0 0 0 0 0 8 0 0 1 0 0 0 0
0 0 0 0 8 0 0 0 0 0 0 8 0 0 0 0 0 0 0
0 0 0 0 8 0 0 0 0 0 0 8 0 0 6 0 0 0 0
0 0 0 0 8 0 0 0 6 0 0 8 0 0 0 0 0 0 0
0 0 0 0 8 0 0 0 0 0 0 8 0 0 0 0 0 0 0
0 0 0 0 8 0 0 0 0 0 0 8 0 0 0 0 0 0 8
3 0 0 0 8 0 0 0 0 0 0 8 0 0 0 0 0 0 0
0 0 0 0 8 0 0 0 0 0 0 8 0 0 0 0 0 0 0
0 0 0 0 8 0 0 0 0 6 0 8 0 0 0 0 0 0 0
0 1 0 0 8 0 0 0 0 0 8 8 0 0 0 0 0 0 0

Output:
0 0 0 0 8 2 2 2 2 2 2 8 0 0 0 0 0 0 0
0 0 0 0 8 2 2 2 2 2 2 8 0 0 0 0 0 0 0
8 8 8 8 8 8 8 8 8 8 8 8 8 8 8 8 8 8 8
4 4 4 4 8 6 6 6 6 6 6 8 3 3 3 3 3 3 3
4 4 4 4 8 6 6 6 6 6 6 8 3 3 3 3 3 3 3
4 4 4 4 8 6 6 6 6 6 6 8 3 3 3 3 3 3 3
4 4 4 4 8 6 6 6 6 6 6 8 3 3 3 3 3 3 3
8 8 8 8 8 8 8 8 8 8 8 8 8 8 8 8 8 8 8
0 0 0 0 8 1 1 1 1 1 1 8 0 0 0 0 0 0 0
0 0 0 0 8 1 1 1 1 1 1 8 0 0 0 0 0 0 0
0 0 0 0 8 1 1 1 1 1 1 8 0 0 0 0 0 0 0
0 0 0 0 8 1 1 1 1 1 1 8 0 0 0 0 0 0 0
0 0 0 0 8 1 1 1 1 1 1 8 0 0 0 0 0 0 0
0 0 0 0 8 1 1 1 1 1 1 8 0 0 0 0 0 0 0
0 0 0 0 8 1 1 1 1 1 1 8 0 0 0 0 0 0 0
0 0 0 0 8 1 1 1 1 1 1 8 0 0 0 0 0 0 0
0 0 0 0 8 1 1 1 1 1 1 8 0 0 0 0 0 0 0
0 0 0 0 8 1 1 1 1 1 1 8 0 0 0 0 0 0 0


Example 3:

Input:
0 0 4 0 8 0 0 0 0 0 0 8 0 0 0 0 0 0 0
0 0 0 0 8 0 0 0 0 0 0 8 0 0 0 0 0 4 0
8 8 8 8 8 8 8 8 8 8 8 8 8 8 8 8 8 8 8
0 0 0 0 8 0 0 0 0 0 0 8 0 0 0 0 2 0 4
0 0 0 0 8 0 0 0 0 0 0 8 0 8 0 0 0 0 0
0 0 0 0 8 0 0 0 4 0 0 8 3 0 0 0 0 0 0
0 0 0 0 8 0 0 2 0 6 0 8 0 0 0 0 0 0 0
8 8 8 8 8 8 8 8 8 8 8 8 8 8 8 8 8 8 8
0 0 0 0 8 0 0 0 0 0 0 8 0 0 0 0 0 4 0
0 0 0 0 8 0 0 0 0 0 4 8 0 0 0 0 0 0 0
0 0 0 0 8 0 0 0 0 0 0 8 0 0 0 0 0 0 0
0 8 0 2 8 0 0 0 0 0 0 8 0 1 0 0 0 0 0
0 0 0 0 8 0 0 0 0 0 0 8 4 0 0 0 0 0 0
0 0 0 0 3 0 0 0 0 0 0 8 0 0 0 0 0 0 0
0 0 0 0 8 0 0 0 0 0 0 8 0 0 0 0 0 0 0
0 0 6 0 8 0 0 0 0 0 0 8 0 0 0 0 0 0 0
0 0 0 0 8 0 0 0 0 0 0 8 0 0 0 0 0 0 0
0 0 0 0 8 0 0 0 0 0 0 8 0 0 0 0 0 0 0

Output:
0 0 0 0 8 2 2 2 2 2 2 8 0 0 0 0 0 0 0
0 0 0 0 8 2 2 2 2 2 2 8 0 0 0 0 0 0 0
8 8 8 8 8 8 8 8 8 8 8 8 8 8 8 8 8 8 8
4 4 4 4 8 6 6 6 6 6 6 8 3 3 3 3 3 3 3
4 4 4 4 8 6 6 6 6 6 6 8 3 3 3 3 3 3 3
4 4 4 4 8 6 6 6 6 6 6 8 3 3 3 3 3 3 3
4 4 4 4 8 6 6 6 6 6 6 8 3 3 3 3 3 3 3
8 8 8 8 8 8 8 8 8 8 8 8 8 8 8 8 8 8 8
0 0 0 0 8 1 1 1 1 1 1 8 0 0 0 0 0 0 0
0 0 0 0 8 1 1 1 1 1 1 8 0 0 0 0 0 0 0
0 0 0 0 8 1 1 1 1 1 1 8 0 0 0 0 0 0 0
0 0 0 0 8 1 1 1 1 1 1 8 0 0 0 0 0 0 0
0 0 0 0 8 1 1 1 1 1 1 8 0 0 0 0 0 0 0
0 0 0 0 8 1 1 1 1 1 1 8 0 0 0 0 0 0 0
0 0 0 0 8 1 1 1 1 1 1 8 0 0 0 0 0 0 0
0 0 0 0 8 1 1 1 1 1 1 8 0 0 0 0 0 0 0
0 0 0 0 8 1 1 1 1 1 1 8 0 0 0 0 0 0 0
0 0 0 0 8 1 1 1 1 1 1 8 0 0 0 0 0 0 0


Example 4:

Input:
0 0 8 0 0 0 0 0 0 8 0 0 0 0
0 0 8 0 0 0 0 0 0 8 0 0 0 0
0 0 8 0 0 0 0 0 0 8 0 0 0 0
0 0 8 0 0 0 0 0 0 3 0 0 0 4
8 8 8 8 8 8 8 8 8 8 8 6 8 8
0 0 8 0 0 0 0 0 0 8 0 6 0 0
0 0 8 0 0 0 0 0 0 8 3 0 0 0
8 8 8 8 8 8 8 1 3 8 8 8 8 8
0 0 8 0 0 0 0 0 0 8 0 0 0 0
0 0 8 0 0 0 0 0 0 8 0 0 0 0
0 0 8 0 0 0 0 0 0 8 0 0 0 0
0 0 8 0 0 0 0 8 0 8 0 0 0 0

Output:
0 0 8 2 2 2 2 2 2 8 0 0 0 0
0 0 8 2 2 2 2 2 2 8 0 0 0 0
0 0 8 2 2 2 2 2 2 8 0 0 0 0
0 0 8 2 2 2 2 2 2 8 0 0 0 0
8 8 8 8 8 8 8 8 8 8 8 8 8 8
4 4 8 6 6 6 6 6 6 8 3 3 3 3
4 4 8 6 6 6 6 6 6 8 3 3 3 3
8 8 8 8 8 8 8 8 8 8 8 8 8 8
0 0 8 1 1 1 1 1 1 8 0 0 0 0
0 0 8 1 1 1 1 1 1 8 0 0 0 0
0 0 8 1 1 1 1 1 1 8 0 0 0 0
0 0 8 1 1 1 1 1 1 8 0 0 0 0


Example 5:

Input:
0 0 8 0 0 0 0 0 0 8 0 8 0 0
0 0 8 0 0 3 0 0 0 8 0 0 0 0
0 0 8 0 0 0 0 0 0 8 0 0 0 0
2 0 8 0 0 0 0 0 0 8 3 0 0 0
8 8 8 8 8 8 8 8 8 8 8 8 8 8
0 0 8 0 0 0 0 0 0 8 0 0 0 0
0 0 8 0 0 0 0 0 0 8 0 0 0 0
8 8 8 8 8 8 8 8 8 8 8 8 8 8
0 0 8 0 0 0 0 0 0 8 3 0 0 0
0 0 8 0 0 0 0 0 0 8 4 0 0 0
0 0 8 4 0 0 0 0 0 8 0 0 0 0
0 0 8 0 0 0 0 0 1 8 0 0 0 0

Output:
0 0 8 2 2 2 2 2 2 8 0 0 0 0
0 0 8 2 2 2 2 2 2 8 0 0 0 0
0 0 8 2 2 2 2 2 2 8 0 0 0 0
0 0 8 2 2 2 2 2 2 8 0 0 0 0
8 8 8 8 8 8 8 8 8 8 8 8 8 8
4 4 8 6 6 6 6 6 6 8 3 3 3 3
4 4 8 6 6 6 6 6 6 8 3 3 3 3
8 8 8 8 8 8 8 8 8 8 8 8 8 8
0 0 8 1 1 1 1 1 1 8 0 0 0 0
0 0 8 1 1 1 1 1 1 8 0 0 0 0
0 0 8 1 1 1 1 1 1 8 0 0 0 0
0 0 8 1 1 1 1 1 1 8 0 0 0 0


Example 6:

Input:
0 0 8 0 0 0 0 0 0 8 0 0 0 0
0 0 8 0 0 0 0 0 1 8 0 0 0 0
0 0 8 0 0 0 0 0 0 8 0 0 0 0
0 0 8 0 0 0 4 0 0 0 0 0 0 0
8 8 8 8 8 8 8 8 8 8 8 8 8 8
0 0 8 0 0 0 0 0 0 8 0 0 0 0
0 0 8 0 0 0 0 0 1 8 0 0 2 0
8 8 8 8 8 8 8 8 8 0 8 8 8 8
0 0 8 0 0 0 0 0 0 8 0 0 0 0
0 0 8 0 0 0 0 0 0 8 0 8 0 0
0 0 8 0 0 0 0 0 0 8 0 0 0 0
0 0 8 0 0 0 0 0 3 8 0 0 0 0

Output:
0 0 8 2 2 2 2 2 2 8 0 0 0 0
0 0 8 2 2 2 2 2 2 8 0 0 0 0
0 0 8 2 2 2 2 2 2 8 0 0 0 0
0 0 8 2 2 2 2 2 2 8 0 0 0 0
8 8 8 8 8 8 8 8 8 8 8 8 8 8
4 4 8 6 6 6 6 6 6 8 3 3 3 3
4 4 8 6 6 6 6 6 6 8 3 3 3 3
8 8 8 8 8 8 8 8 8 8 8 8 8 8
0 0 8 1 1 1 1 1 1 8 0 0 0 0
0 0 8 1 1 1 1 1 1 8 0 0 0 0
0 0 8 1 1 1 1 1 1 8 0 0 0 0
0 0 8 1 1 1 1 1 1 8 0 0 0 0



Below is a test input grid. 
Predict the corresponding output. 


Input:
0 0 0 8 0 0 0 0 8 0 0 0 0 0 0
0 0 0 8 0 0 0 0 8 0 0 0 0 0 0
0 0 0 8 0 0 0 0 8 0 0 0 0 0 0
0 0 0 8 0 0 0 0 8 0 0 0 0 0 0
0 0 0 8 0 0 0 0 8 0 0 0 0 0 0
0 0 0 8 0 0 0 0 8 0 0 0 0 0 0
8 8 8 8 8 8 8 8 8 8 8 8 8 8 8
0 0 0 8 0 0 0 0 8 0 0 0 0 0 0
0 0 0 8 0 0 0 0 8 0 0 0 0 0 0
0 0 0 8 0 0 0 0 8 0 0 0 0 0 0
0 0 0 8 0 0 0 0 8 0 0 0 0 0 0
0 0 0 8 0 0 0 0 8 0 0 0 0 0 0
0 0 0 8 0 0 0 0 8 0 0 0 0 0 0
8 8 8 8 8 8 8 8 8 8 8 8 8 8 8
0 0 0 8 0 0 0 0 8 0 0 0 0 0 0
0 0 0 8 0 0 0 0 8 0 0 0 0 0 0
0 0 0 8 0 0 0 0 8 0 0 0 0 0 0

\end{lstlisting}
\end{tcolorbox}

\switchcolumn 

\begin{tcolorbox}[
  colback=cyan!10!white,
  colframe=black,
  boxrule=0.6pt,
  arc=1mm,
  breakable,
  enhanced,
  title=\centering \textbf{3-Noisy output Example With Original Prompt}
]
\scriptsize
\begin{lstlisting}
Find the common rule that maps 
an input grid to an output grid, 
given the examples below.

Example1
Input:
0 0 0 0 8 0 0 0 0 0 0 8 0 0 0 0 0 0 0
0 0 0 0 8 0 0 0 0 0 0 8 0 0 0 0 0 0 0
8 8 8 8 8 8 8 8 8 8 8 8 8 8 8 8 8 8 8
0 0 0 0 8 0 0 0 0 0 0 8 0 0 0 0 0 0 0
0 0 0 0 8 0 0 0 0 0 0 8 0 0 0 0 0 0 0
0 0 0 0 8 0 0 0 0 0 0 8 0 0 0 0 0 0 0
0 0 0 0 8 0 0 0 0 0 0 8 0 0 0 0 0 0 0
8 8 8 8 8 8 8 8 8 8 8 8 8 8 8 8 8 8 8
0 0 0 0 8 0 0 0 0 0 0 8 0 0 0 0 0 0 0
0 0 0 0 8 0 0 0 0 0 0 8 0 0 0 0 0 0 0
0 0 0 0 8 0 0 0 0 0 0 8 0 0 0 0 0 0 0
0 0 0 0 8 0 0 0 0 0 0 8 0 0 0 0 0 0 0
0 0 0 0 8 0 0 0 0 0 0 8 0 0 0 0 0 0 0
0 0 0 0 8 0 0 0 0 0 0 8 0 0 0 0 0 0 0
0 0 0 0 8 0 0 0 0 0 0 8 0 0 0 0 0 0 0
0 0 0 0 8 0 0 0 0 0 0 8 0 0 0 0 0 0 0
0 0 0 0 8 0 0 0 0 0 0 8 0 0 0 0 0 0 0
0 0 0 0 8 0 0 0 0 0 0 8 0 0 0 0 0 0 0

Output:
0 0 0 0 8 2 2 2 2 2 2 8 0 0 0 0 0 0 0
0 0 3 0 8 2 2 2 2 2 2 8 0 0 0 1 0 0 0
8 8 8 8 8 8 8 8 8 8 8 8 8 8 8 8 8 8 8
4 4 4 4 8 6 6 6 6 6 6 8 3 3 1 3 3 3 3
4 4 4 4 8 6 6 6 6 6 4 8 3 3 3 3 3 3 2
4 4 4 4 8 6 6 6 6 6 6 8 3 3 3 3 3 3 3
4 4 4 4 8 6 4 3 6 4 6 8 3 3 3 3 3 3 3
8 8 8 0 8 8 8 8 8 8 8 8 8 8 0 8 8 8 8
0 0 0 0 8 1 1 1 1 1 1 8 0 6 0 0 0 0 0
0 0 0 0 8 1 1 1 1 1 1 8 0 0 0 0 0 0 0
0 0 0 0 8 1 1 1 1 1 4 8 0 0 0 0 0 0 0
0 0 0 0 8 1 1 1 1 1 1 8 0 0 0 0 4 0 0
0 0 0 0 8 1 1 1 1 1 1 8 0 0 0 0 0 0 0
0 0 0 1 8 1 1 1 1 1 1 8 0 0 0 0 0 0 0
0 0 0 0 8 1 1 1 1 1 1 8 0 0 0 6 0 0 0
0 0 0 0 8 1 1 1 1 1 1 8 0 0 0 0 0 0 0
0 0 0 0 8 1 1 1 1 1 1 8 0 0 0 0 0 0 0
0 0 0 0 8 1 1 1 1 1 1 3 0 1 0 0 0 0 0


Example2
Input:
0 0 0 0 8 0 0 0 0 0 0 8 0 0 0 0 0 0 0
0 0 0 0 8 0 0 0 0 0 0 8 0 0 0 0 0 0 0
8 8 8 8 8 8 8 8 8 8 8 8 8 8 8 8 8 8 8
0 0 0 0 8 0 0 0 0 0 0 8 0 0 0 0 0 0 0
0 0 0 0 8 0 0 0 0 0 0 8 0 0 0 0 0 0 0
0 0 0 0 8 0 0 0 0 0 0 8 0 0 0 0 0 0 0
0 0 0 0 8 0 0 0 0 0 0 8 0 0 0 0 0 0 0
8 8 8 8 8 8 8 8 8 8 8 8 8 8 8 8 8 8 8
0 0 0 0 8 0 0 0 0 0 0 8 0 0 0 0 0 0 0
0 0 0 0 8 0 0 0 0 0 0 8 0 0 0 0 0 0 0
0 0 0 0 8 0 0 0 0 0 0 8 0 0 0 0 0 0 0
0 0 0 0 8 0 0 0 0 0 0 8 0 0 0 0 0 0 0
0 0 0 0 8 0 0 0 0 0 0 8 0 0 0 0 0 0 0
0 0 0 0 8 0 0 0 0 0 0 8 0 0 0 0 0 0 0
0 0 0 0 8 0 0 0 0 0 0 8 0 0 0 0 0 0 0
0 0 0 0 8 0 0 0 0 0 0 8 0 0 0 0 0 0 0
0 0 0 0 8 0 0 0 0 0 0 8 0 0 0 0 0 0 0
0 0 0 0 8 0 0 0 0 0 0 8 0 0 0 0 0 0 0

Output:
0 4 0 0 8 2 2 2 2 2 2 8 0 0 0 0 0 0 0
0 0 0 0 8 2 2 2 2 2 3 8 0 0 0 0 0 0 0
4 8 8 8 0 8 8 8 8 2 8 8 8 8 8 8 8 4 8
4 4 4 4 8 6 6 6 6 6 6 8 3 3 3 3 3 3 3
4 4 4 4 8 6 6 6 6 6 2 8 3 3 1 3 3 3 3
4 4 4 4 8 6 6 6 6 6 6 2 3 3 3 3 3 3 3
4 4 4 4 8 6 6 6 6 6 6 8 3 3 3 3 3 3 3
8 8 8 8 8 8 8 8 8 8 8 8 8 8 8 8 8 8 8
0 0 0 0 8 1 2 1 1 1 1 8 0 0 0 0 0 0 0
0 0 0 0 8 1 1 1 1 1 1 8 0 0 1 6 0 0 0
0 0 0 0 8 1 1 1 1 1 1 6 0 0 0 0 0 0 0
0 0 0 0 8 1 1 1 1 1 1 8 0 0 0 0 0 0 0
0 0 0 0 8 1 4 1 1 1 1 8 0 0 0 0 0 6 0
0 0 0 0 8 1 1 1 1 1 1 8 0 0 0 0 0 0 0
0 0 0 0 8 1 1 1 1 1 1 8 0 0 0 0 0 0 0
0 0 0 0 8 1 1 1 1 1 1 8 0 0 0 0 0 0 6
0 0 0 0 8 1 1 1 1 1 1 8 0 0 0 0 0 0 0
0 0 0 0 8 1 0 1 1 1 1 8 0 0 0 0 0 0 0


Example3
Input:
0 0 0 0 8 0 0 0 0 0 0 8 0 0 0 0 0 0 0
0 0 0 0 8 0 0 0 0 0 0 8 0 0 0 0 0 0 0
8 8 8 8 8 8 8 8 8 8 8 8 8 8 8 8 8 8 8
0 0 0 0 8 0 0 0 0 0 0 8 0 0 0 0 0 0 0
0 0 0 0 8 0 0 0 0 0 0 8 0 0 0 0 0 0 0
0 0 0 0 8 0 0 0 0 0 0 8 0 0 0 0 0 0 0
0 0 0 0 8 0 0 0 0 0 0 8 0 0 0 0 0 0 0
8 8 8 8 8 8 8 8 8 8 8 8 8 8 8 8 8 8 8
0 0 0 0 8 0 0 0 0 0 0 8 0 0 0 0 0 0 0
0 0 0 0 8 0 0 0 0 0 0 8 0 0 0 0 0 0 0
0 0 0 0 8 0 0 0 0 0 0 8 0 0 0 0 0 0 0
0 0 0 0 8 0 0 0 0 0 0 8 0 0 0 0 0 0 0
0 0 0 0 8 0 0 0 0 0 0 8 0 0 0 0 0 0 0
0 0 0 0 8 0 0 0 0 0 0 8 0 0 0 0 0 0 0
0 0 0 0 8 0 0 0 0 0 0 8 0 0 0 0 0 0 0
0 0 0 0 8 0 0 0 0 0 0 8 0 0 0 0 0 0 0
0 0 0 0 8 0 0 0 0 0 0 8 0 0 0 0 0 0 0
0 0 0 0 8 0 0 0 0 0 0 8 0 0 0 0 0 0 0

Output:
0 0 0 0 8 2 2 2 2 2 2 8 0 0 0 0 0 0 0
0 0 0 0 8 2 2 2 2 2 2 8 0 0 0 0 0 2 0
8 8 2 8 8 8 8 8 8 8 8 8 8 4 8 8 8 8 8
4 4 4 4 8 6 6 6 6 6 6 8 3 3 3 3 3 3 3
4 4 4 4 8 6 6 6 6 6 6 8 3 3 3 1 3 3 3
4 4 4 4 8 6 6 6 6 6 6 8 3 3 3 3 3 3 1
4 4 4 4 8 6 6 6 6 6 6 8 3 3 3 3 3 3 3
8 8 8 8 8 8 8 4 8 8 8 8 8 8 8 8 8 8 8
0 1 0 0 8 1 1 1 1 1 1 8 0 0 0 0 0 0 0
0 0 0 6 8 1 8 1 1 1 1 8 0 0 0 0 0 0 0
0 0 0 0 8 1 1 1 1 1 1 8 0 0 0 0 0 0 0
0 0 0 0 8 1 2 1 0 1 1 1 0 0 6 0 0 0 0
0 0 0 0 8 1 1 1 1 1 1 8 0 0 0 0 0 0 0
0 0 0 0 8 1 1 1 1 1 1 8 0 0 3 0 0 0 0
0 0 0 0 8 1 1 1 1 1 1 2 0 0 0 0 0 0 0
0 0 0 0 8 1 1 1 1 1 2 8 0 0 0 0 0 0 0
0 0 0 0 8 1 1 1 1 1 1 8 0 0 0 0 0 0 0
0 0 0 0 2 1 1 1 1 1 1 8 0 0 0 0 0 0 0
Example4
Input:
0 0 8 0 0 0 0 0 0 8 0 0 0 0
0 0 8 0 0 0 0 0 0 8 0 0 0 0
0 0 8 0 0 0 0 0 0 8 0 0 0 0
0 0 8 0 0 0 0 0 0 8 0 0 0 0
8 8 8 8 8 8 8 8 8 8 8 8 8 8
0 0 8 0 0 0 0 0 0 8 0 0 0 0
0 0 8 0 0 0 0 0 0 8 0 0 0 0
8 8 8 8 8 8 8 8 8 8 8 8 8 8
0 0 8 0 0 0 0 0 0 8 0 0 0 0
0 0 8 0 0 0 0 0 0 8 0 0 0 0
0 0 8 0 0 0 0 0 0 8 0 0 0 0
0 0 8 0 0 0 0 0 0 8 0 0 0 0

Output:
0 0 8 2 2 2 2 2 2 8 0 0 0 0
0 0 8 2 1 2 2 2 2 8 0 0 0 0
0 0 8 2 2 2 2 2 2 8 0 0 0 0
0 0 8 2 2 2 2 2 2 8 0 6 0 0
8 8 8 8 8 8 8 8 8 8 8 8 8 8
4 4 8 6 6 6 6 6 6 8 3 3 3 3
1 4 8 6 6 8 6 6 6 8 3 3 3 3
8 8 8 8 8 0 8 8 8 8 8 8 8 8
0 0 8 1 1 1 1 1 1 8 0 0 0 0
0 0 8 1 1 1 1 1 1 8 0 0 0 0
0 0 8 1 1 1 1 1 1 0 0 0 0 8
0 0 8 1 2 1 1 1 1 8 0 0 0 0


Example5
Input:
0 0 8 0 0 0 0 0 0 8 0 0 0 0
0 0 8 0 0 0 0 0 0 8 0 0 0 0
0 0 8 0 0 0 0 0 0 8 0 0 0 0
0 0 8 0 0 0 0 0 0 8 0 0 0 0
8 8 8 8 8 8 8 8 8 8 8 8 8 8
0 0 8 0 0 0 0 0 0 8 0 0 0 0
0 0 8 0 0 0 0 0 0 8 0 0 0 0
8 8 8 8 8 8 8 8 8 8 8 8 8 8
0 0 8 0 0 0 0 0 0 8 0 0 0 0
0 0 8 0 0 0 0 0 0 8 0 0 0 0
0 0 8 0 0 0 0 0 0 8 0 0 0 0
0 0 8 0 0 0 0 0 0 8 0 0 0 0

Output:
0 0 8 2 2 2 2 8 2 8 0 0 0 0
0 0 8 2 2 2 8 2 2 8 0 0 0 0
0 0 8 2 2 2 2 2 2 8 0 0 0 0
0 0 8 2 2 2 2 2 2 8 0 0 0 0
8 8 8 8 8 8 8 8 8 8 8 8 8 8
4 4 8 6 6 6 6 6 0 8 3 3 3 3
4 4 8 6 6 6 6 6 6 8 3 3 3 3
8 8 8 8 8 8 8 8 8 8 8 3 8 8
0 0 8 1 1 1 1 1 1 6 6 0 0 0
0 3 8 1 1 1 1 1 1 8 0 0 0 2
0 0 8 1 1 1 1 1 1 8 0 0 0 0
0 0 8 1 1 1 1 1 1 8 0 0 0 0


Example6
Input:
0 0 8 0 0 0 0 0 0 8 0 0 0 0
0 0 8 0 0 0 0 0 0 8 0 0 0 0
0 0 8 0 0 0 0 0 0 8 0 0 0 0
0 0 8 0 0 0 0 0 0 8 0 0 0 0
8 8 8 8 8 8 8 8 8 8 8 8 8 8
0 0 8 0 0 0 0 0 0 8 0 0 0 0
0 0 8 0 0 0 0 0 0 8 0 0 0 0
8 8 8 8 8 8 8 8 8 8 8 8 8 8
0 0 8 0 0 0 0 0 0 8 0 0 0 0
0 0 8 0 0 0 0 0 0 8 0 0 0 0
0 0 8 0 0 0 0 0 0 8 0 0 0 0
0 0 8 0 0 0 0 0 0 8 0 0 0 0

Output:
0 0 8 2 2 2 2 2 2 8 0 0 0 0
0 0 8 2 2 2 2 2 2 8 0 0 0 0
0 0 8 2 2 2 2 2 8 8 0 0 0 0
0 0 8 2 2 2 2 2 2 8 0 0 0 0
8 8 8 8 8 8 8 3 8 8 8 8 8 8
4 4 8 6 6 6 6 6 6 8 3 3 3 3
4 4 8 6 6 6 6 6 6 8 0 3 3 3
8 8 8 8 8 8 8 8 8 8 8 8 8 8
0 0 8 1 1 1 1 1 2 4 0 0 0 0
0 0 8 1 1 1 1 1 1 8 0 0 8 8
0 0 8 1 6 1 1 1 1 8 0 0 0 0
0 0 8 1 1 1 1 1 1 8 0 0 0 0
Below is a test input grid. 
Predict the corresponding output. 

Input:
0 0 0 8 0 0 0 0 8 0 0 0 0 0 0
0 0 0 8 0 0 0 0 8 0 0 0 0 0 0
0 0 0 8 0 0 0 0 8 0 0 0 0 0 0
0 0 0 8 0 0 0 0 8 0 0 0 0 0 0
0 0 0 8 0 0 0 0 8 0 0 0 0 0 0
0 0 0 8 0 0 0 0 8 0 0 0 0 0 0
8 8 8 8 8 8 8 8 8 8 8 8 8 8 8
0 0 0 8 0 0 0 0 8 0 0 0 0 0 0
0 0 0 8 0 0 0 0 8 0 0 0 0 0 0
0 0 0 8 0 0 0 0 8 0 0 0 0 0 0
0 0 0 8 0 0 0 0 8 0 0 0 0 0 0
0 0 0 8 0 0 0 0 8 0 0 0 0 0 0
0 0 0 8 0 0 0 0 8 0 0 0 0 0 0
8 8 8 8 8 8 8 8 8 8 8 8 8 8 8
0 0 0 8 0 0 0 0 8 0 0 0 0 0 0
0 0 0 8 0 0 0 0 8 0 0 0 0 0 0
0 0 0 8 0 0 0 0 8 0 0 0 0 0 0

\end{lstlisting}
\end{tcolorbox}

\end{paracol}
\begin{paracol}{2} 

\begin{tcolorbox}[
  colback=cyan!10!white,
  colframe=black,
  boxrule=0.6pt,
  arc=1mm,
  breakable,
  enhanced,
  title=\centering \textbf{3-Noisy Input Example With New Prompt}
]
\scriptsize
\begin{lstlisting}
Find the common rule that maps 
an input grid to an output grid, 
given the examples below. 
Note that random noise has 
been added to the input grids, 
meaning that different noisy 
input grids may map to the same 
output. In Examples 1 to 3, 
noisy input grids map to the 
same output grid. In Examples 4
to 6, noisy input grids map to 
their respective output grids.

Example 1:

Input:
0 0 0 0 8 0 0 8 0 0 0 8 0 0 0 0 0 0 0
0 0 0 0 8 0 0 0 0 0 0 8 2 3 0 0 0 0 0
8 8 8 8 8 8 8 8 8 8 8 8 8 8 8 8 8 8 8
0 0 0 0 8 0 0 0 0 0 0 8 0 0 0 0 0 0 0
0 0 0 0 8 0 0 0 0 0 0 8 0 0 0 0 0 0 1
0 0 0 0 8 0 0 0 0 0 0 8 0 0 0 0 0 0 0
0 0 0 0 8 0 0 0 0 0 0 8 0 0 0 0 0 0 0
8 8 8 1 8 8 8 8 8 8 8 8 8 8 8 8 8 8 8
0 0 0 0 8 0 0 0 0 0 0 8 0 0 0 0 0 0 0
0 0 0 0 8 0 0 0 0 0 0 8 0 0 0 0 0 0 0
0 0 0 0 8 0 0 0 0 0 0 8 0 0 0 0 0 0 0
0 0 0 0 8 0 0 8 0 0 0 8 6 0 0 0 0 0 0
0 0 0 0 8 0 0 0 0 0 0 8 0 0 0 6 0 0 0
0 0 0 0 8 0 0 0 0 4 0 8 0 0 0 0 0 0 0
0 0 0 0 8 0 0 0 0 6 0 8 2 0 0 0 0 0 0
0 0 0 0 8 0 0 0 0 0 0 8 0 0 0 0 0 0 0
0 0 0 0 8 0 0 0 0 0 2 0 0 0 0 0 0 0 0
0 0 0 0 3 0 1 6 0 0 4 8 0 0 0 0 0 0 0

Output:
0 0 0 0 8 2 2 2 2 2 2 8 0 0 0 0 0 0 0
0 0 0 0 8 2 2 2 2 2 2 8 0 0 0 0 0 0 0
8 8 8 8 8 8 8 8 8 8 8 8 8 8 8 8 8 8 8
4 4 4 4 8 6 6 6 6 6 6 8 3 3 3 3 3 3 3
4 4 4 4 8 6 6 6 6 6 6 8 3 3 3 3 3 3 3
4 4 4 4 8 6 6 6 6 6 6 8 3 3 3 3 3 3 3
4 4 4 4 8 6 6 6 6 6 6 8 3 3 3 3 3 3 3
8 8 8 8 8 8 8 8 8 8 8 8 8 8 8 8 8 8 8
0 0 0 0 8 1 1 1 1 1 1 8 0 0 0 0 0 0 0
0 0 0 0 8 1 1 1 1 1 1 8 0 0 0 0 0 0 0
0 0 0 0 8 1 1 1 1 1 1 8 0 0 0 0 0 0 0
0 0 0 0 8 1 1 1 1 1 1 8 0 0 0 0 0 0 0
0 0 0 0 8 1 1 1 1 1 1 8 0 0 0 0 0 0 0
0 0 0 0 8 1 1 1 1 1 1 8 0 0 0 0 0 0 0
0 0 0 0 8 1 1 1 1 1 1 8 0 0 0 0 0 0 0
0 0 0 0 8 1 1 1 1 1 1 8 0 0 0 0 0 0 0
0 0 0 0 8 1 1 1 1 1 1 8 0 0 0 0 0 0 0
0 0 0 0 8 1 1 1 1 1 1 8 0 0 0 0 0 0 0


Example 2:

Input:
0 0 0 0 8 0 0 0 0 0 0 8 0 0 0 0 0 0 0
0 2 0 0 8 0 0 0 0 0 0 8 0 0 0 0 0 0 0
8 8 8 8 8 8 1 8 8 8 8 8 8 8 8 8 8 8 8
0 0 0 0 8 0 0 0 0 0 0 8 0 0 0 0 0 0 0
0 0 0 8 8 0 0 0 0 0 0 8 0 0 0 0 0 0 0
0 0 0 0 3 0 0 0 3 0 0 8 3 0 0 0 0 0 0
0 0 0 0 8 0 0 0 0 0 0 8 0 0 0 1 0 0 0
8 8 8 8 8 8 8 8 8 8 8 8 8 3 8 8 8 8 8
0 0 0 0 8 0 0 0 0 0 0 8 0 0 0 0 0 0 0
0 0 0 0 1 6 0 0 0 0 0 8 0 3 0 0 8 0 0
0 0 0 0 8 0 0 0 0 0 0 8 0 0 0 0 0 0 0
0 0 0 0 4 0 0 0 0 0 0 8 0 0 0 0 1 0 0
0 0 0 0 8 0 0 0 1 0 0 8 0 0 0 0 0 0 0
0 0 0 0 8 0 0 0 0 0 0 8 0 0 0 0 0 0 0
0 0 0 0 8 0 0 0 0 0 0 8 0 0 0 0 0 0 0
0 0 0 0 8 0 0 0 0 0 0 8 0 0 0 0 0 0 0
0 0 0 0 8 0 0 0 0 0 0 8 0 0 0 0 8 0 0
1 0 0 0 8 0 0 0 0 0 0 8 0 0 0 0 0 0 0

Output:
0 0 0 0 8 2 2 2 2 2 2 8 0 0 0 0 0 0 0
0 0 0 0 8 2 2 2 2 2 2 8 0 0 0 0 0 0 0
8 8 8 8 8 8 8 8 8 8 8 8 8 8 8 8 8 8 8
4 4 4 4 8 6 6 6 6 6 6 8 3 3 3 3 3 3 3
4 4 4 4 8 6 6 6 6 6 6 8 3 3 3 3 3 3 3
4 4 4 4 8 6 6 6 6 6 6 8 3 3 3 3 3 3 3
4 4 4 4 8 6 6 6 6 6 6 8 3 3 3 3 3 3 3
8 8 8 8 8 8 8 8 8 8 8 8 8 8 8 8 8 8 8
0 0 0 0 8 1 1 1 1 1 1 8 0 0 0 0 0 0 0
0 0 0 0 8 1 1 1 1 1 1 8 0 0 0 0 0 0 0
0 0 0 0 8 1 1 1 1 1 1 8 0 0 0 0 0 0 0
0 0 0 0 8 1 1 1 1 1 1 8 0 0 0 0 0 0 0
0 0 0 0 8 1 1 1 1 1 1 8 0 0 0 0 0 0 0
0 0 0 0 8 1 1 1 1 1 1 8 0 0 0 0 0 0 0
0 0 0 0 8 1 1 1 1 1 1 8 0 0 0 0 0 0 0
0 0 0 0 8 1 1 1 1 1 1 8 0 0 0 0 0 0 0
0 0 0 0 8 1 1 1 1 1 1 8 0 0 0 0 0 0 0
0 0 0 0 8 1 1 1 1 1 1 8 0 0 0 0 0 0 0


Example 3:

Input:
0 0 0 0 8 0 0 0 0 0 0 8 0 0 2 0 0 0 0
0 0 0 0 8 0 0 0 0 0 0 8 0 0 0 0 0 0 0
8 8 8 8 8 8 8 8 8 8 8 8 8 8 8 3 8 8 0
0 0 0 0 8 0 0 0 0 0 0 8 0 0 0 0 1 0 0
0 0 0 0 8 0 8 0 0 0 0 8 0 0 0 0 2 0 0
0 0 0 0 8 0 0 0 8 1 0 8 0 8 0 0 0 0 0
0 0 0 0 8 0 0 0 0 0 3 8 0 0 0 0 6 0 0
8 8 8 8 8 8 8 8 8 8 8 8 8 8 8 8 8 8 8
0 0 0 0 8 0 0 0 0 0 0 8 0 0 0 0 0 0 0
0 0 0 0 8 0 0 0 0 0 0 8 0 0 0 0 0 0 0
0 0 0 0 8 0 0 0 0 0 0 8 0 0 0 0 0 0 0
0 0 0 0 8 4 0 0 0 0 0 8 8 0 0 0 0 0 0
0 0 0 0 8 0 0 0 0 0 4 8 0 0 0 0 0 0 0
0 0 0 0 8 0 0 0 0 0 0 8 0 0 0 0 0 0 0
0 0 0 0 8 0 0 0 0 0 2 8 0 0 0 3 0 0 0
0 0 0 0 8 0 0 0 0 0 0 8 0 0 0 0 0 0 0
0 0 0 0 8 0 8 0 0 0 0 8 0 0 0 0 0 0 0
0 0 0 0 8 0 0 0 0 0 0 8 0 0 0 0 0 0 0

Output:
0 0 0 0 8 2 2 2 2 2 2 8 0 0 0 0 0 0 0
0 0 0 0 8 2 2 2 2 2 2 8 0 0 0 0 0 0 0
8 8 8 8 8 8 8 8 8 8 8 8 8 8 8 8 8 8 8
4 4 4 4 8 6 6 6 6 6 6 8 3 3 3 3 3 3 3
4 4 4 4 8 6 6 6 6 6 6 8 3 3 3 3 3 3 3
4 4 4 4 8 6 6 6 6 6 6 8 3 3 3 3 3 3 3
4 4 4 4 8 6 6 6 6 6 6 8 3 3 3 3 3 3 3
8 8 8 8 8 8 8 8 8 8 8 8 8 8 8 8 8 8 8
0 0 0 0 8 1 1 1 1 1 1 8 0 0 0 0 0 0 0
0 0 0 0 8 1 1 1 1 1 1 8 0 0 0 0 0 0 0
0 0 0 0 8 1 1 1 1 1 1 8 0 0 0 0 0 0 0
0 0 0 0 8 1 1 1 1 1 1 8 0 0 0 0 0 0 0
0 0 0 0 8 1 1 1 1 1 1 8 0 0 0 0 0 0 0
0 0 0 0 8 1 1 1 1 1 1 8 0 0 0 0 0 0 0
0 0 0 0 8 1 1 1 1 1 1 8 0 0 0 0 0 0 0
0 0 0 0 8 1 1 1 1 1 1 8 0 0 0 0 0 0 0
0 0 0 0 8 1 1 1 1 1 1 8 0 0 0 0 0 0 0
0 0 0 0 8 1 1 1 1 1 1 8 0 0 0 0 0 0 0


Example 4:

Input:
0 0 8 0 0 0 0 0 0 8 0 0 0 2
0 0 8 0 0 0 0 0 0 8 0 0 0 0
0 2 8 0 0 0 0 0 0 8 0 0 0 0
8 0 8 4 0 0 0 0 0 8 0 0 0 0
8 8 8 8 8 8 8 8 8 8 8 8 8 8
0 0 4 0 0 0 0 0 0 8 0 0 0 0
0 0 8 0 0 0 0 0 0 8 0 0 0 0
8 8 8 8 8 8 8 8 8 8 8 8 8 1
0 0 8 0 0 0 0 0 0 8 0 0 0 0
0 0 8 0 0 0 0 0 0 8 0 0 0 0
0 8 8 0 4 0 0 0 0 8 0 0 0 0
0 0 8 0 0 0 0 0 0 8 0 0 0 0

Output:
0 0 8 2 2 2 2 2 2 8 0 0 0 0
0 0 8 2 2 2 2 2 2 8 0 0 0 0
0 0 8 2 2 2 2 2 2 8 0 0 0 0
0 0 8 2 2 2 2 2 2 8 0 0 0 0
8 8 8 8 8 8 8 8 8 8 8 8 8 8
4 4 8 6 6 6 6 6 6 8 3 3 3 3
4 4 8 6 6 6 6 6 6 8 3 3 3 3
8 8 8 8 8 8 8 8 8 8 8 8 8 8
0 0 8 1 1 1 1 1 1 8 0 0 0 0
0 0 8 1 1 1 1 1 1 8 0 0 0 0
0 0 8 1 1 1 1 1 1 8 0 0 0 0
0 0 8 1 1 1 1 1 1 8 0 0 0 0


Example 5:

Input:
0 0 8 0 0 0 0 0 0 8 0 0 2 0
0 0 8 0 0 0 0 0 0 3 0 0 0 0
0 0 8 0 0 0 0 0 0 8 0 0 0 0
0 0 8 0 0 2 0 0 0 8 0 0 0 0
8 8 8 8 8 8 8 8 8 8 8 8 8 8
0 0 8 0 0 0 0 0 4 8 1 0 0 0
0 0 8 0 0 0 0 0 0 8 0 0 0 0
8 8 8 8 8 8 8 8 8 8 8 2 8 8
0 0 8 0 0 0 0 0 0 8 0 0 0 0
0 0 8 0 0 0 0 0 0 8 0 0 0 0
0 0 8 0 0 0 0 0 0 8 4 0 0 0
0 0 0 0 0 0 0 0 0 8 0 0 0 0

Output:
0 0 8 2 2 2 2 2 2 8 0 0 0 0
0 0 8 2 2 2 2 2 2 8 0 0 0 0
0 0 8 2 2 2 2 2 2 8 0 0 0 0
0 0 8 2 2 2 2 2 2 8 0 0 0 0
8 8 8 8 8 8 8 8 8 8 8 8 8 8
4 4 8 6 6 6 6 6 6 8 3 3 3 3
4 4 8 6 6 6 6 6 6 8 3 3 3 3
8 8 8 8 8 8 8 8 8 8 8 8 8 8
0 0 8 1 1 1 1 1 1 8 0 0 0 0
0 0 8 1 1 1 1 1 1 8 0 0 0 0
0 0 8 1 1 1 1 1 1 8 0 0 0 0
0 0 8 1 1 1 1 1 1 8 0 0 0 0


Example 6:

Input:
0 0 8 0 0 0 0 0 0 8 0 0 0 0
0 0 8 0 2 0 0 0 0 8 0 0 0 0
0 0 8 0 0 0 0 0 0 8 0 0 0 0
0 0 8 0 0 0 0 0 0 8 0 0 0 0
8 8 8 8 8 8 8 8 8 8 8 8 8 8
0 0 8 0 0 0 0 0 0 8 0 0 0 2
0 0 8 0 0 0 0 0 0 3 0 0 0 0
8 8 8 8 8 8 8 1 8 8 8 8 8 8
0 0 8 0 0 0 6 0 4 0 0 0 0 0
0 0 8 0 0 0 0 0 0 8 0 0 0 0
0 0 8 0 0 0 0 0 0 8 0 0 0 0
0 0 8 0 0 0 0 0 0 8 0 3 0 0

Output:
0 0 8 2 2 2 2 2 2 8 0 0 0 0
0 0 8 2 2 2 2 2 2 8 0 0 0 0
0 0 8 2 2 2 2 2 2 8 0 0 0 0
0 0 8 2 2 2 2 2 2 8 0 0 0 0
8 8 8 8 8 8 8 8 8 8 8 8 8 8
4 4 8 6 6 6 6 6 6 8 3 3 3 3
4 4 8 6 6 6 6 6 6 8 3 3 3 3
8 8 8 8 8 8 8 8 8 8 8 8 8 8
0 0 8 1 1 1 1 1 1 8 0 0 0 0
0 0 8 1 1 1 1 1 1 8 0 0 0 0
0 0 8 1 1 1 1 1 1 8 0 0 0 0
0 0 8 1 1 1 1 1 1 8 0 0 0 0

Below is a test input grid. 
Predict the corresponding output. 

Input:
0 0 0 8 0 0 0 0 8 0 0 0 0 0 0
0 0 0 8 0 0 0 0 8 0 0 0 0 0 0
0 0 0 8 0 0 0 0 8 0 0 0 0 0 0
0 0 0 8 0 0 0 0 8 0 0 0 0 0 0
0 0 0 8 0 0 0 0 8 0 0 0 0 0 0
0 0 0 8 0 0 0 0 8 0 0 0 0 0 0
8 8 8 8 8 8 8 8 8 8 8 8 8 8 8
0 0 0 8 0 0 0 0 8 0 0 0 0 0 0
0 0 0 8 0 0 0 0 8 0 0 0 0 0 0
0 0 0 8 0 0 0 0 8 0 0 0 0 0 0
0 0 0 8 0 0 0 0 8 0 0 0 0 0 0
0 0 0 8 0 0 0 0 8 0 0 0 0 0 0
0 0 0 8 0 0 0 0 8 0 0 0 0 0 0
8 8 8 8 8 8 8 8 8 8 8 8 8 8 8
0 0 0 8 0 0 0 0 8 0 0 0 0 0 0
0 0 0 8 0 0 0 0 8 0 0 0 0 0 0
0 0 0 8 0 0 0 0 8 0 0 0 0 0 0

\end{lstlisting}
\end{tcolorbox}

\switchcolumn 

\begin{tcolorbox}[
  colback=cyan!10!white,
  colframe=black,
  boxrule=0.6pt,
  arc=1mm,
  breakable,
  enhanced,
  title=\centering \textbf{3-Noisy output Example With New Prompt}
]
\scriptsize
\begin{lstlisting}
Find the common rule that maps 
an input grid to an output grid, 
given the examples below. 
Note that random noise has 
been added to the output grids, 
meaning that different noisy 
output grids may map to the same input. 
In Examples 1 to 3, 
each example contains a 
noisy output grid that maps 
to the same input grid.   
In Examples 4 to 6, each 
example contains a noisy 
output grid that maps to 
its respective input grid.

Example1
Input:
0 0 0 0 8 0 0 0 0 0 0 8 0 0 0 0 0 0 0
0 0 0 0 8 0 0 0 0 0 0 8 0 0 0 0 0 0 0
8 8 8 8 8 8 8 8 8 8 8 8 8 8 8 8 8 8 8
0 0 0 0 8 0 0 0 0 0 0 8 0 0 0 0 0 0 0
0 0 0 0 8 0 0 0 0 0 0 8 0 0 0 0 0 0 0
0 0 0 0 8 0 0 0 0 0 0 8 0 0 0 0 0 0 0
0 0 0 0 8 0 0 0 0 0 0 8 0 0 0 0 0 0 0
8 8 8 8 8 8 8 8 8 8 8 8 8 8 8 8 8 8 8
0 0 0 0 8 0 0 0 0 0 0 8 0 0 0 0 0 0 0
0 0 0 0 8 0 0 0 0 0 0 8 0 0 0 0 0 0 0
0 0 0 0 8 0 0 0 0 0 0 8 0 0 0 0 0 0 0
0 0 0 0 8 0 0 0 0 0 0 8 0 0 0 0 0 0 0
0 0 0 0 8 0 0 0 0 0 0 8 0 0 0 0 0 0 0
0 0 0 0 8 0 0 0 0 0 0 8 0 0 0 0 0 0 0
0 0 0 0 8 0 0 0 0 0 0 8 0 0 0 0 0 0 0
0 0 0 0 8 0 0 0 0 0 0 8 0 0 0 0 0 0 0
0 0 0 0 8 0 0 0 0 0 0 8 0 0 0 0 0 0 0
0 0 0 0 8 0 0 0 0 0 0 8 0 0 0 0 0 0 0

Output:
0 0 0 0 8 2 2 2 2 2 2 8 0 0 0 0 0 0 6
0 0 0 0 8 2 2 3 2 2 2 8 0 0 0 0 0 0 0
8 8 8 8 8 8 8 8 8 8 8 8 8 8 2 1 8 8 8
4 4 4 4 8 6 6 4 6 6 6 8 3 3 3 3 3 3 3
4 4 4 4 8 6 6 6 1 6 6 8 3 3 3 3 3 3 3
4 4 4 4 8 6 6 6 6 6 6 8 3 3 3 3 3 3 3
4 4 4 8 8 6 6 6 6 6 6 8 3 3 4 3 2 3 3
8 8 8 8 8 8 8 1 8 8 8 8 8 8 8 8 8 8 8
0 0 0 0 8 1 1 0 1 1 1 4 0 0 0 0 0 0 0
0 0 0 0 8 1 1 1 1 1 1 8 0 0 0 0 0 0 0
0 0 0 0 8 6 1 1 1 1 1 8 2 0 0 0 0 0 0
0 0 0 0 8 1 1 1 1 1 1 8 0 0 0 0 0 0 0
0 0 0 0 8 1 1 1 1 1 1 8 0 0 0 0 0 0 0
0 0 0 0 8 1 1 1 1 1 1 8 0 0 0 0 0 0 0
0 0 0 0 8 1 1 1 1 1 3 8 0 0 0 0 0 0 0
0 0 0 0 8 1 1 1 1 1 1 8 0 0 0 0 0 0 0
0 0 0 0 8 1 1 1 1 1 1 8 0 0 0 0 0 0 0
0 0 0 1 8 1 1 1 2 1 1 8 0 0 0 0 0 0 0


Example2
Input:
0 0 0 0 8 0 0 0 0 0 0 8 0 0 0 0 0 0 0
0 0 0 0 8 0 0 0 0 0 0 8 0 0 0 0 0 0 0
8 8 8 8 8 8 8 8 8 8 8 8 8 8 8 8 8 8 8
0 0 0 0 8 0 0 0 0 0 0 8 0 0 0 0 0 0 0
0 0 0 0 8 0 0 0 0 0 0 8 0 0 0 0 0 0 0
0 0 0 0 8 0 0 0 0 0 0 8 0 0 0 0 0 0 0
0 0 0 0 8 0 0 0 0 0 0 8 0 0 0 0 0 0 0
8 8 8 8 8 8 8 8 8 8 8 8 8 8 8 8 8 8 8
0 0 0 0 8 0 0 0 0 0 0 8 0 0 0 0 0 0 0
0 0 0 0 8 0 0 0 0 0 0 8 0 0 0 0 0 0 0
0 0 0 0 8 0 0 0 0 0 0 8 0 0 0 0 0 0 0
0 0 0 0 8 0 0 0 0 0 0 8 0 0 0 0 0 0 0
0 0 0 0 8 0 0 0 0 0 0 8 0 0 0 0 0 0 0
0 0 0 0 8 0 0 0 0 0 0 8 0 0 0 0 0 0 0
0 0 0 0 8 0 0 0 0 0 0 8 0 0 0 0 0 0 0
0 0 0 0 8 0 0 0 0 0 0 8 0 0 0 0 0 0 0
0 0 0 0 8 0 0 0 0 0 0 8 0 0 0 0 0 0 0
0 0 0 0 8 0 0 0 0 0 0 8 0 0 0 0 0 0 0

Output:
0 0 0 0 8 2 2 2 2 2 2 8 0 0 0 0 0 0 0
0 0 0 0 8 2 2 2 2 3 2 8 0 0 0 0 0 0 0
8 8 8 8 8 8 8 8 8 8 8 8 8 8 8 8 8 8 8
4 4 4 4 8 6 6 6 6 6 1 8 3 3 3 3 3 3 3
4 4 4 4 8 6 6 6 6 2 6 8 3 8 3 3 3 3 3
4 4 4 4 8 6 6 3 6 6 6 8 3 3 3 3 3 3 3
4 4 4 4 8 6 6 6 6 8 6 8 3 4 3 3 3 3 3
8 8 8 8 8 8 8 8 8 8 8 8 8 8 8 8 8 8 8
0 0 4 0 8 1 1 1 1 1 1 8 0 0 0 0 0 0 0
0 0 0 0 8 1 1 1 1 1 1 8 0 0 0 0 0 0 0
0 0 0 0 8 1 1 1 1 1 1 8 0 0 0 0 0 0 0
0 0 0 0 8 1 1 1 1 1 1 2 0 0 0 0 0 0 0
0 0 0 0 4 1 4 8 1 1 1 8 0 0 0 0 0 0 0
0 0 0 0 8 1 1 1 1 1 6 8 0 0 0 0 0 0 2
0 0 0 0 8 1 1 1 1 1 1 8 0 0 0 0 0 0 0
0 0 0 0 8 1 1 1 1 1 1 8 8 0 0 0 0 0 0
0 0 0 6 8 1 1 1 1 1 4 8 0 0 0 0 0 0 0
0 0 0 0 8 1 1 1 1 1 1 8 0 0 0 0 0 0 0


Example3
Input:
0 0 0 0 8 0 0 0 0 0 0 8 0 0 0 0 0 0 0
0 0 0 0 8 0 0 0 0 0 0 8 0 0 0 0 0 0 0
8 8 8 8 8 8 8 8 8 8 8 8 8 8 8 8 8 8 8
0 0 0 0 8 0 0 0 0 0 0 8 0 0 0 0 0 0 0
0 0 0 0 8 0 0 0 0 0 0 8 0 0 0 0 0 0 0
0 0 0 0 8 0 0 0 0 0 0 8 0 0 0 0 0 0 0
0 0 0 0 8 0 0 0 0 0 0 8 0 0 0 0 0 0 0
8 8 8 8 8 8 8 8 8 8 8 8 8 8 8 8 8 8 8
0 0 0 0 8 0 0 0 0 0 0 8 0 0 0 0 0 0 0
0 0 0 0 8 0 0 0 0 0 0 8 0 0 0 0 0 0 0
0 0 0 0 8 0 0 0 0 0 0 8 0 0 0 0 0 0 0
0 0 0 0 8 0 0 0 0 0 0 8 0 0 0 0 0 0 0
0 0 0 0 8 0 0 0 0 0 0 8 0 0 0 0 0 0 0
0 0 0 0 8 0 0 0 0 0 0 8 0 0 0 0 0 0 0
0 0 0 0 8 0 0 0 0 0 0 8 0 0 0 0 0 0 0
0 0 0 0 8 0 0 0 0 0 0 8 0 0 0 0 0 0 0
0 0 0 0 8 0 0 0 0 0 0 8 0 0 0 0 0 0 0
0 0 0 0 8 0 0 0 0 0 0 8 0 0 0 0 0 0 0

Output:
0 0 0 0 8 2 2 2 2 2 2 8 0 0 0 2 0 0 0
0 3 0 0 8 2 3 2 1 2 2 8 2 0 0 0 0 0 0
8 8 8 8 8 8 8 8 8 8 8 8 8 8 8 8 8 8 8
4 4 4 4 8 6 6 6 6 6 6 4 3 3 3 3 3 3 3
4 4 4 4 8 6 6 4 6 8 6 8 3 3 3 3 3 3 3
4 4 4 4 8 6 6 6 6 6 6 8 3 3 3 3 3 3 0
4 4 4 4 8 6 6 6 6 6 6 8 3 3 3 3 3 3 3
8 8 8 8 8 8 8 8 8 8 8 8 8 8 8 8 8 8 8
0 0 0 6 8 1 1 1 1 1 1 8 0 0 0 0 0 0 0
0 0 0 0 8 1 1 1 1 1 1 8 0 0 0 0 0 0 6
0 0 0 0 8 1 4 1 1 6 1 8 0 0 0 0 0 0 0
0 0 0 0 8 3 1 1 1 1 1 8 0 0 0 0 0 0 0
0 0 0 0 8 1 1 3 1 1 1 8 0 0 0 0 0 0 0
0 0 0 0 8 1 1 1 1 1 1 1 0 0 0 0 0 0 0
0 0 0 0 8 1 1 1 1 1 1 8 0 0 0 0 0 0 0
0 0 0 0 8 1 1 1 1 1 1 8 0 0 0 4 0 0 0
0 0 0 0 8 1 1 1 1 1 1 8 0 0 0 0 0 0 0
0 0 0 0 8 1 1 1 1 1 1 8 0 0 0 0 0 0 0
Example4
Input:
0 0 8 0 0 0 0 0 0 8 0 0 0 0
0 0 8 0 0 0 0 0 0 8 0 0 0 0
0 0 8 0 0 0 0 0 0 8 0 0 0 0
0 0 8 0 0 0 0 0 0 8 0 0 0 0
8 8 8 8 8 8 8 8 8 8 8 8 8 8
0 0 8 0 0 0 0 0 0 8 0 0 0 0
0 0 8 0 0 0 0 0 0 8 0 0 0 0
8 8 8 8 8 8 8 8 8 8 8 8 8 8
0 0 8 0 0 0 0 0 0 8 0 0 0 0
0 0 8 0 0 0 0 0 0 8 0 0 0 0
0 0 8 0 0 0 0 0 0 8 0 0 0 0
0 0 8 0 0 0 0 0 0 8 0 0 0 0

Output:
0 6 8 2 2 2 2 2 2 8 0 0 0 0
0 0 8 2 0 2 2 2 2 8 0 0 0 0
0 0 8 2 2 2 2 2 2 8 0 0 0 0
0 0 8 2 2 2 2 2 2 8 0 0 0 0
8 8 8 8 8 8 8 8 8 8 8 8 8 8
4 4 8 6 6 6 6 6 6 0 3 3 3 3
4 4 8 6 6 6 6 6 6 8 3 3 3 4
8 8 8 8 8 8 8 8 8 8 8 8 8 8
0 0 8 1 1 1 8 1 1 8 0 0 0 0
0 0 8 1 3 1 2 1 1 8 0 0 0 0
0 0 8 1 1 1 1 1 1 8 0 0 0 0
0 0 8 1 1 1 1 1 1 8 4 0 0 0


Example5
Input:
0 0 8 0 0 0 0 0 0 8 0 0 0 0
0 0 8 0 0 0 0 0 0 8 0 0 0 0
0 0 8 0 0 0 0 0 0 8 0 0 0 0
0 0 8 0 0 0 0 0 0 8 0 0 0 0
8 8 8 8 8 8 8 8 8 8 8 8 8 8
0 0 8 0 0 0 0 0 0 8 0 0 0 0
0 0 8 0 0 0 0 0 0 8 0 0 0 0
8 8 8 8 8 8 8 8 8 8 8 8 8 8
0 0 8 0 0 0 0 0 0 8 0 0 0 0
0 0 8 0 0 0 0 0 0 8 0 0 0 0
0 0 8 0 0 0 0 0 0 8 0 0 0 0
0 0 8 0 0 0 0 0 0 8 0 0 0 0

Output:
0 0 8 2 2 2 2 2 2 8 0 0 0 0
0 0 8 2 2 2 2 2 2 8 2 0 0 0
0 0 8 2 2 2 2 2 2 8 0 0 0 0
0 0 8 2 2 2 2 2 2 8 0 0 0 8
8 8 1 8 8 8 8 8 8 8 8 8 8 8
4 4 8 6 6 6 6 6 6 8 3 3 3 3
4 4 8 6 6 6 6 6 6 8 3 3 3 3
8 8 8 8 8 8 8 1 8 8 8 8 8 8
2 0 8 1 1 1 1 1 1 8 0 0 0 0
0 0 8 1 1 1 1 1 2 8 0 0 0 0
0 0 8 1 1 1 1 1 1 8 0 0 8 0
0 8 8 1 1 1 1 1 1 8 0 0 0 0


Example6
Input:
0 0 8 0 0 0 0 0 0 8 0 0 0 0
0 0 8 0 0 0 0 0 0 8 0 0 0 0
0 0 8 0 0 0 0 0 0 8 0 0 0 0
0 0 8 0 0 0 0 0 0 8 0 0 0 0
8 8 8 8 8 8 8 8 8 8 8 8 8 8
0 0 8 0 0 0 0 0 0 8 0 0 0 0
0 0 8 0 0 0 0 0 0 8 0 0 0 0
8 8 8 8 8 8 8 8 8 8 8 8 8 8
0 0 8 0 0 0 0 0 0 8 0 0 0 0
0 0 8 0 0 0 0 0 0 8 0 0 0 0
0 0 8 0 0 0 0 0 0 8 0 0 0 0
0 0 8 0 0 0 0 0 0 8 0 0 0 0

Output:
0 0 8 2 2 2 2 2 2 8 0 0 3 0
0 0 8 2 2 2 2 2 2 1 0 0 0 0
0 0 8 2 2 2 2 2 3 8 0 1 0 0
0 0 8 2 2 2 2 2 2 8 0 0 0 0
8 8 8 8 8 8 8 8 8 8 8 8 8 8
4 4 8 6 6 6 6 6 6 8 3 3 3 0
8 4 8 6 6 6 6 6 6 8 3 3 3 3
8 8 8 8 2 8 8 8 8 8 0 8 8 8
0 0 8 1 1 1 1 1 1 8 0 0 0 0
0 0 8 1 1 1 1 1 1 8 0 0 0 0
0 0 8 1 1 1 1 1 1 8 0 0 0 0
0 0 8 1 1 1 1 1 1 8 0 0 0 0

Below is a test input grid. 
Predict the corresponding output. 

Input:
0 0 0 8 0 0 0 0 8 0 0 0 0 0 0
0 0 0 8 0 0 0 0 8 0 0 0 0 0 0
0 0 0 8 0 0 0 0 8 0 0 0 0 0 0
0 0 0 8 0 0 0 0 8 0 0 0 0 0 0
0 0 0 8 0 0 0 0 8 0 0 0 0 0 0
0 0 0 8 0 0 0 0 8 0 0 0 0 0 0
8 8 8 8 8 8 8 8 8 8 8 8 8 8 8
0 0 0 8 0 0 0 0 8 0 0 0 0 0 0
0 0 0 8 0 0 0 0 8 0 0 0 0 0 0
0 0 0 8 0 0 0 0 8 0 0 0 0 0 0
0 0 0 8 0 0 0 0 8 0 0 0 0 0 0
0 0 0 8 0 0 0 0 8 0 0 0 0 0 0
0 0 0 8 0 0 0 0 8 0 0 0 0 0 0
8 8 8 8 8 8 8 8 8 8 8 8 8 8 8
0 0 0 8 0 0 0 0 8 0 0 0 0 0 0
0 0 0 8 0 0 0 0 8 0 0 0 0 0 0
0 0 0 8 0 0 0 0 8 0 0 0 0 0 0
\end{lstlisting}
\end{tcolorbox}

\end{paracol}
\begin{paracol}{2} 

\begin{tcolorbox}[
  colback=cyan!10!white,
  colframe=black,
  boxrule=0.6pt,
  arc=1mm,
  breakable,
  enhanced,
  title=\centering \textbf{9-Noisy Input Example With Original Prompt}
]
\scriptsize
\begin{lstlisting}
Find the common rule that maps 
an input grid to an output grid, 
given the examples below.

Example1
Input:
0 0 0 0 8 0 0 0 0 0 0 8 0 0 0 0 0 0 0
0 0 0 0 8 0 0 0 0 0 0 8 0 0 0 0 0 0 0
8 8 8 8 8 8 8 8 8 8 8 8 8 8 8 8 8 8 8
0 0 0 0 8 0 0 0 0 0 0 8 0 0 0 0 0 0 0
0 0 0 0 8 0 0 0 0 0 0 8 0 0 0 0 0 0 0
0 0 0 0 8 0 0 0 0 0 0 8 0 0 0 0 0 0 0
0 0 0 0 8 0 0 0 0 0 0 8 0 0 0 0 0 0 0
8 8 8 8 8 8 8 8 8 8 8 8 8 8 8 8 8 8 8
0 0 0 0 8 0 0 0 0 0 0 8 0 0 0 0 0 0 0
0 0 0 0 8 0 0 0 0 0 0 8 0 0 0 0 0 0 0
0 0 0 0 8 0 0 0 0 0 0 8 0 0 0 0 0 0 0
0 0 0 0 8 0 0 0 0 0 0 8 0 0 0 0 0 0 0
0 0 0 0 8 0 0 0 0 0 0 8 0 0 0 0 0 0 0
0 0 0 0 8 0 0 0 0 0 0 8 0 0 0 0 0 0 0
0 0 0 0 8 0 0 0 0 0 0 8 0 0 0 0 0 0 0
0 0 0 0 8 0 0 0 0 0 0 8 0 0 0 0 0 0 0
0 0 0 0 8 0 0 0 0 0 0 8 0 0 0 0 0 0 0
0 0 0 0 8 0 0 0 0 0 0 8 0 0 0 0 0 0 0

Output:
0 0 0 0 8 2 3 2 2 2 2 8 0 0 0 0 0 0 0
0 0 0 0 8 2 2 2 2 2 2 8 0 0 0 0 0 0 0
8 8 8 8 8 8 8 8 8 8 8 8 8 8 2 8 2 8 8
4 4 4 4 8 6 6 6 6 6 6 8 3 3 3 3 3 3 0
4 0 4 4 8 6 6 6 6 6 6 8 3 3 3 3 3 3 3
4 4 4 4 8 0 6 6 6 6 6 8 3 3 3 3 3 3 3
4 4 4 4 8 6 6 6 6 6 6 8 3 3 3 3 3 3 3
8 8 8 8 8 8 8 8 8 8 8 8 8 8 8 8 4 8 8
0 0 0 0 8 1 1 1 6 1 1 8 0 0 0 0 0 0 0
1 0 0 0 8 1 1 1 1 1 1 8 0 0 0 1 0 0 0
0 0 0 0 8 1 1 1 1 6 1 8 0 0 0 0 0 0 0
0 0 0 0 8 1 1 1 1 1 1 8 0 0 0 0 0 0 0
0 0 0 0 8 1 1 1 1 1 1 8 2 0 0 0 0 0 0
0 0 0 0 8 1 1 1 1 1 1 8 0 0 0 0 0 0 0
0 0 0 0 8 1 1 8 1 1 6 8 0 0 0 0 0 0 3
0 0 0 0 8 1 1 1 1 1 1 8 0 0 0 0 0 0 2
0 0 0 0 8 1 1 1 1 1 1 8 0 0 2 0 0 0 0
0 0 0 0 8 1 1 1 1 1 1 8 0 0 0 0 0 0 0


Example2
Input:
0 0 0 0 8 0 0 0 0 0 0 8 0 0 0 0 0 0 0
0 0 0 0 8 0 0 0 0 0 0 8 0 0 0 0 0 0 0
8 8 8 8 8 8 8 8 8 8 8 8 8 8 8 8 8 8 8
0 0 0 0 8 0 0 0 0 0 0 8 0 0 0 0 0 0 0
0 0 0 0 8 0 0 0 0 0 0 8 0 0 0 0 0 0 0
0 0 0 0 8 0 0 0 0 0 0 8 0 0 0 0 0 0 0
0 0 0 0 8 0 0 0 0 0 0 8 0 0 0 0 0 0 0
8 8 8 8 8 8 8 8 8 8 8 8 8 8 8 8 8 8 8
0 0 0 0 8 0 0 0 0 0 0 8 0 0 0 0 0 0 0
0 0 0 0 8 0 0 0 0 0 0 8 0 0 0 0 0 0 0
0 0 0 0 8 0 0 0 0 0 0 8 0 0 0 0 0 0 0
0 0 0 0 8 0 0 0 0 0 0 8 0 0 0 0 0 0 0
0 0 0 0 8 0 0 0 0 0 0 8 0 0 0 0 0 0 0
0 0 0 0 8 0 0 0 0 0 0 8 0 0 0 0 0 0 0
0 0 0 0 8 0 0 0 0 0 0 8 0 0 0 0 0 0 0
0 0 0 0 8 0 0 0 0 0 0 8 0 0 0 0 0 0 0
0 0 0 0 8 0 0 0 0 0 0 8 0 0 0 0 0 0 0
0 0 0 0 8 0 0 0 0 0 0 8 0 0 0 0 0 0 0

Output:
0 1 0 0 8 2 2 2 2 2 2 8 0 0 3 4 0 0 0
0 0 0 0 8 2 2 2 2 2 3 8 0 0 0 0 0 0 0
8 8 8 8 8 8 8 8 8 8 8 8 8 8 8 8 8 8 8
4 4 4 4 8 6 6 6 6 6 6 8 3 3 3 3 3 3 3
4 4 4 4 8 6 6 6 6 6 6 8 3 3 3 3 3 3 3
4 4 4 4 8 6 6 6 6 6 6 8 3 3 3 3 3 3 3
4 4 4 4 8 6 6 6 1 6 6 8 3 3 3 3 3 3 3
8 8 8 8 8 8 8 8 8 8 8 8 1 8 8 8 8 8 8
0 0 0 0 8 1 1 1 1 1 1 8 0 0 0 0 0 0 0
0 0 0 6 8 1 1 1 2 1 4 8 8 0 0 0 0 0 0
0 0 4 0 8 1 1 1 1 1 1 8 0 0 0 0 0 0 0
0 0 0 0 8 1 1 1 1 1 1 8 0 0 0 0 0 0 0
0 0 0 0 8 1 1 1 1 1 1 8 0 3 0 0 0 0 0
0 3 0 0 8 1 1 1 1 1 8 8 0 0 0 0 0 0 0
0 0 0 0 8 3 1 1 1 1 1 8 0 0 0 0 0 0 0
0 0 0 0 8 1 1 1 1 1 1 8 0 0 0 0 0 0 0
0 0 0 0 8 1 1 1 1 1 1 8 0 0 0 0 0 0 0
0 0 0 0 8 1 1 1 1 1 1 4 0 0 0 0 0 6 0


Example3
Input:
0 0 0 0 8 0 0 0 0 0 0 8 0 0 0 0 0 0 0
0 0 0 0 8 0 0 0 0 0 0 8 0 0 0 0 0 0 0
8 8 8 8 8 8 8 8 8 8 8 8 8 8 8 8 8 8 8
0 0 0 0 8 0 0 0 0 0 0 8 0 0 0 0 0 0 0
0 0 0 0 8 0 0 0 0 0 0 8 0 0 0 0 0 0 0
0 0 0 0 8 0 0 0 0 0 0 8 0 0 0 0 0 0 0
0 0 0 0 8 0 0 0 0 0 0 8 0 0 0 0 0 0 0
8 8 8 8 8 8 8 8 8 8 8 8 8 8 8 8 8 8 8
0 0 0 0 8 0 0 0 0 0 0 8 0 0 0 0 0 0 0
0 0 0 0 8 0 0 0 0 0 0 8 0 0 0 0 0 0 0
0 0 0 0 8 0 0 0 0 0 0 8 0 0 0 0 0 0 0
0 0 0 0 8 0 0 0 0 0 0 8 0 0 0 0 0 0 0
0 0 0 0 8 0 0 0 0 0 0 8 0 0 0 0 0 0 0
0 0 0 0 8 0 0 0 0 0 0 8 0 0 0 0 0 0 0
0 0 0 0 8 0 0 0 0 0 0 8 0 0 0 0 0 0 0
0 0 0 0 8 0 0 0 0 0 0 8 0 0 0 0 0 0 0
0 0 0 0 8 0 0 0 0 0 0 8 0 0 0 0 0 0 0
0 0 0 0 8 0 0 0 0 0 0 8 0 0 0 0 0 0 0

Output:
0 0 0 0 8 2 2 2 2 2 2 8 0 0 0 0 0 0 0
0 0 0 0 8 2 2 2 2 2 2 8 0 0 0 0 0 0 0
8 8 8 8 8 8 8 8 8 8 3 8 8 8 8 8 8 8 8
4 4 4 4 8 6 6 6 6 6 6 8 3 3 3 3 3 3 3
4 2 4 4 8 6 6 3 6 6 6 8 3 3 3 3 3 3 3
4 4 4 4 8 6 6 6 6 6 6 8 3 3 3 3 3 2 3
4 4 4 3 8 6 6 6 6 6 6 8 3 3 3 3 3 3 3
8 8 8 8 8 8 8 8 8 8 8 4 8 8 8 8 8 8 8
0 0 0 0 8 1 1 1 1 1 1 8 0 0 0 0 0 0 0
4 0 0 0 2 1 1 1 1 1 1 8 0 0 0 0 0 0 0
0 0 0 0 8 1 1 1 1 1 1 8 0 0 0 0 0 0 0
0 0 0 4 8 1 1 1 1 1 1 8 0 0 0 0 0 0 0
4 0 0 0 8 1 1 1 1 1 1 8 0 0 0 0 0 4 0
3 3 0 0 8 1 1 1 1 1 1 8 0 6 0 0 0 0 0
0 2 0 0 8 1 1 1 1 1 1 8 0 0 0 0 0 0 0
0 0 0 0 8 1 1 1 1 1 1 8 0 0 0 0 0 0 0
0 0 0 0 8 1 1 1 1 8 1 8 0 0 0 0 0 0 0
0 0 0 0 8 1 1 1 1 1 1 8 0 0 0 0 0 0 2


Example4
Input:
0 0 0 0 8 0 0 0 0 0 0 8 0 0 0 0 0 0 0
0 0 0 0 8 0 0 0 0 0 0 8 0 0 0 0 0 0 0
8 8 8 8 8 8 8 8 8 8 8 8 8 8 8 8 8 8 8
0 0 0 0 8 0 0 0 0 0 0 8 0 0 0 0 0 0 0
0 0 0 0 8 0 0 0 0 0 0 8 0 0 0 0 0 0 0
0 0 0 0 8 0 0 0 0 0 0 8 0 0 0 0 0 0 0
0 0 0 0 8 0 0 0 0 0 0 8 0 0 0 0 0 0 0
8 8 8 8 8 8 8 8 8 8 8 8 8 8 8 8 8 8 8
0 0 0 0 8 0 0 0 0 0 0 8 0 0 0 0 0 0 0
0 0 0 0 8 0 0 0 0 0 0 8 0 0 0 0 0 0 0
0 0 0 0 8 0 0 0 0 0 0 8 0 0 0 0 0 0 0
0 0 0 0 8 0 0 0 0 0 0 8 0 0 0 0 0 0 0
0 0 0 0 8 0 0 0 0 0 0 8 0 0 0 0 0 0 0
0 0 0 0 8 0 0 0 0 0 0 8 0 0 0 0 0 0 0
0 0 0 0 8 0 0 0 0 0 0 8 0 0 0 0 0 0 0
0 0 0 0 8 0 0 0 0 0 0 8 0 0 0 0 0 0 0
0 0 0 0 8 0 0 0 0 0 0 8 0 0 0 0 0 0 0
0 0 0 0 8 0 0 0 0 0 0 8 0 0 0 0 0 0 0

Output:
0 0 0 0 3 2 2 2 2 2 8 8 0 0 0 0 0 0 0
0 0 0 0 8 2 2 2 2 2 2 8 0 0 0 0 0 0 0
8 8 8 8 8 8 8 8 8 8 8 8 8 8 8 8 8 8 8
4 4 4 4 8 6 6 4 6 6 6 8 3 3 3 3 3 3 3
4 4 4 4 8 6 6 6 6 6 6 8 3 3 3 3 4 3 3
4 4 4 4 8 6 6 6 6 6 6 8 3 3 3 3 3 3 3
4 4 4 3 8 6 6 6 6 3 6 8 3 3 3 3 3 3 3
8 8 8 8 8 8 8 8 8 8 8 8 8 8 8 8 8 8 8
0 4 0 0 8 1 1 1 1 1 1 8 1 0 0 0 0 0 0
0 0 0 0 8 1 1 1 1 1 1 8 0 0 0 0 0 0 0
0 0 0 0 8 1 1 1 1 1 2 8 0 0 0 0 0 0 0
0 0 0 0 8 1 2 4 1 1 1 2 0 0 0 0 0 0 0
0 0 0 0 8 1 1 1 1 1 1 8 0 0 0 0 0 0 0
0 0 0 0 8 1 6 1 1 1 1 8 0 0 0 0 8 4 2
0 0 0 0 8 1 1 1 1 1 1 8 0 0 0 0 0 1 0
0 0 0 0 8 1 1 1 1 1 1 8 0 0 0 0 0 0 0
0 0 0 0 8 1 1 1 1 1 1 8 0 0 0 0 0 0 0
0 0 0 0 8 1 1 1 1 1 1 8 0 0 0 0 0 0 0


Example5
Input:
0 0 0 0 8 0 0 0 0 0 0 8 0 0 0 0 0 0 0
0 0 0 0 8 0 0 0 0 0 0 8 0 0 0 0 0 0 0
8 8 8 8 8 8 8 8 8 8 8 8 8 8 8 8 8 8 8
0 0 0 0 8 0 0 0 0 0 0 8 0 0 0 0 0 0 0
0 0 0 0 8 0 0 0 0 0 0 8 0 0 0 0 0 0 0
0 0 0 0 8 0 0 0 0 0 0 8 0 0 0 0 0 0 0
0 0 0 0 8 0 0 0 0 0 0 8 0 0 0 0 0 0 0
8 8 8 8 8 8 8 8 8 8 8 8 8 8 8 8 8 8 8
0 0 0 0 8 0 0 0 0 0 0 8 0 0 0 0 0 0 0
0 0 0 0 8 0 0 0 0 0 0 8 0 0 0 0 0 0 0
0 0 0 0 8 0 0 0 0 0 0 8 0 0 0 0 0 0 0
0 0 0 0 8 0 0 0 0 0 0 8 0 0 0 0 0 0 0
0 0 0 0 8 0 0 0 0 0 0 8 0 0 0 0 0 0 0
0 0 0 0 8 0 0 0 0 0 0 8 0 0 0 0 0 0 0
0 0 0 0 8 0 0 0 0 0 0 8 0 0 0 0 0 0 0
0 0 0 0 8 0 0 0 0 0 0 8 0 0 0 0 0 0 0
0 0 0 0 8 0 0 0 0 0 0 8 0 0 0 0 0 0 0
0 0 0 0 8 0 0 0 0 0 0 8 0 0 0 0 0 0 0

Output:
0 0 0 0 8 2 2 2 2 2 2 8 6 0 0 0 0 0 0
0 0 0 8 8 2 2 2 2 2 2 8 0 0 0 0 0 0 0
8 8 8 8 8 8 8 8 8 8 8 8 8 8 8 8 8 8 8
4 4 4 4 8 6 6 6 6 0 6 8 3 3 3 3 3 3 3
4 4 4 4 8 6 1 4 6 6 6 8 3 3 3 3 3 3 3
4 4 4 4 8 4 6 6 6 6 6 8 3 3 3 1 3 3 3
4 4 4 3 3 6 6 6 6 6 6 8 3 3 3 3 3 3 3
8 8 8 8 8 8 8 8 8 8 8 8 8 8 8 8 8 8 8
0 0 0 0 8 1 1 1 1 1 1 8 0 0 0 0 0 0 0
0 3 0 0 8 1 1 1 1 1 1 8 3 0 0 0 0 0 0
0 0 0 0 8 1 1 1 2 1 1 8 0 0 0 0 0 0 0
0 0 0 0 3 1 1 1 1 1 1 8 0 0 0 0 0 0 0
0 0 0 0 8 1 1 1 1 1 1 8 0 0 0 0 0 0 0
0 0 0 0 2 1 1 1 1 1 1 8 0 0 0 0 0 0 0
0 0 0 0 8 1 1 1 1 1 1 8 0 0 0 0 0 0 0
0 0 0 0 8 1 1 1 1 1 1 8 0 0 0 0 0 0 0
0 0 0 1 2 1 1 1 1 1 1 8 0 0 0 0 0 0 0
0 0 0 0 8 1 1 1 1 1 1 8 0 0 0 0 0 1 0


Example6
Input:
0 0 0 0 8 0 0 0 0 0 0 8 0 0 0 0 0 0 0
0 0 0 0 8 0 0 0 0 0 0 8 0 0 0 0 0 0 0
8 8 8 8 8 8 8 8 8 8 8 8 8 8 8 8 8 8 8
0 0 0 0 8 0 0 0 0 0 0 8 0 0 0 0 0 0 0
0 0 0 0 8 0 0 0 0 0 0 8 0 0 0 0 0 0 0
0 0 0 0 8 0 0 0 0 0 0 8 0 0 0 0 0 0 0
0 0 0 0 8 0 0 0 0 0 0 8 0 0 0 0 0 0 0
8 8 8 8 8 8 8 8 8 8 8 8 8 8 8 8 8 8 8
0 0 0 0 8 0 0 0 0 0 0 8 0 0 0 0 0 0 0
0 0 0 0 8 0 0 0 0 0 0 8 0 0 0 0 0 0 0
0 0 0 0 8 0 0 0 0 0 0 8 0 0 0 0 0 0 0
0 0 0 0 8 0 0 0 0 0 0 8 0 0 0 0 0 0 0
0 0 0 0 8 0 0 0 0 0 0 8 0 0 0 0 0 0 0
0 0 0 0 8 0 0 0 0 0 0 8 0 0 0 0 0 0 0
0 0 0 0 8 0 0 0 0 0 0 8 0 0 0 0 0 0 0
0 0 0 0 8 0 0 0 0 0 0 8 0 0 0 0 0 0 0
0 0 0 0 8 0 0 0 0 0 0 8 0 0 0 0 0 0 0
0 0 0 0 8 0 0 0 0 0 0 8 0 0 0 0 0 0 0

Output:
0 0 0 0 8 2 2 2 2 2 2 8 0 0 0 0 0 0 4
0 0 4 0 8 2 2 2 2 2 2 8 3 0 0 0 0 0 0
8 8 8 8 8 8 8 8 8 8 8 8 8 8 8 8 8 8 8
4 4 4 4 8 6 6 6 6 6 6 8 0 3 4 3 3 3 3
4 4 4 0 8 6 6 6 6 6 6 8 3 3 3 3 3 3 3
4 4 4 4 8 6 6 6 6 6 6 8 3 3 3 3 3 3 3
8 4 4 4 8 6 3 1 6 6 6 8 3 3 3 3 3 3 3
8 8 8 8 8 0 8 8 8 8 8 8 8 8 8 8 8 8 8
0 0 0 0 8 1 1 1 1 1 1 8 0 0 0 0 0 0 0
0 0 0 0 8 1 1 1 1 1 1 8 0 0 0 0 0 0 0
0 0 0 0 8 1 1 1 1 1 1 8 0 0 0 0 0 0 0
0 0 0 0 8 1 1 1 1 1 1 8 0 0 0 0 0 0 0
0 0 0 0 0 1 1 1 1 1 1 8 0 0 0 0 0 0 0
0 0 0 0 8 1 1 1 1 1 6 8 0 0 0 0 0 0 0
0 0 0 0 8 1 1 1 1 1 1 8 0 0 0 8 0 0 0
0 0 0 0 8 1 1 1 1 1 1 8 4 0 0 0 0 0 0
0 0 0 0 8 1 1 1 1 1 1 8 0 0 0 8 0 0 0
0 0 0 0 8 1 1 8 1 1 1 8 0 0 0 0 1 0 0


Example7
Input:
0 0 0 0 8 0 0 0 0 0 0 8 0 0 0 0 0 0 0
0 0 0 0 8 0 0 0 0 0 0 8 0 0 0 0 0 0 0
8 8 8 8 8 8 8 8 8 8 8 8 8 8 8 8 8 8 8
0 0 0 0 8 0 0 0 0 0 0 8 0 0 0 0 0 0 0
0 0 0 0 8 0 0 0 0 0 0 8 0 0 0 0 0 0 0
0 0 0 0 8 0 0 0 0 0 0 8 0 0 0 0 0 0 0
0 0 0 0 8 0 0 0 0 0 0 8 0 0 0 0 0 0 0
8 8 8 8 8 8 8 8 8 8 8 8 8 8 8 8 8 8 8
0 0 0 0 8 0 0 0 0 0 0 8 0 0 0 0 0 0 0
0 0 0 0 8 0 0 0 0 0 0 8 0 0 0 0 0 0 0
0 0 0 0 8 0 0 0 0 0 0 8 0 0 0 0 0 0 0
0 0 0 0 8 0 0 0 0 0 0 8 0 0 0 0 0 0 0
0 0 0 0 8 0 0 0 0 0 0 8 0 0 0 0 0 0 0
0 0 0 0 8 0 0 0 0 0 0 8 0 0 0 0 0 0 0
0 0 0 0 8 0 0 0 0 0 0 8 0 0 0 0 0 0 0
0 0 0 0 8 0 0 0 0 0 0 8 0 0 0 0 0 0 0
0 0 0 0 8 0 0 0 0 0 0 8 0 0 0 0 0 0 0
0 0 0 0 8 0 0 0 0 0 0 8 0 0 0 0 0 0 0

Output:
0 0 0 0 8 2 2 2 2 2 2 8 0 0 0 4 0 0 0
0 0 0 0 0 2 2 2 2 2 2 8 0 0 0 0 0 0 0
2 8 8 8 8 8 8 4 8 8 1 8 8 8 8 8 8 8 8
4 3 4 0 8 6 6 6 6 6 6 8 3 3 3 3 3 3 3
4 4 4 4 8 6 6 6 6 6 6 8 3 3 4 3 3 3 3
4 4 4 4 8 6 6 6 6 6 6 8 3 3 3 3 3 3 3
4 4 4 4 8 6 6 6 6 6 6 8 3 3 3 3 3 3 3
8 8 8 8 8 8 8 8 8 8 0 8 8 8 8 8 8 8 8
0 0 0 0 8 1 1 1 1 1 2 8 0 0 0 3 0 0 0
0 0 0 0 8 1 1 1 1 1 1 8 0 0 0 0 0 0 0
0 0 0 0 8 1 1 1 1 1 1 8 0 0 0 0 0 0 0
0 0 0 0 8 1 1 1 1 1 1 8 4 0 0 1 0 0 0
0 0 0 0 8 1 1 1 1 1 1 8 0 0 0 0 0 0 0
1 0 0 0 8 1 1 1 1 1 1 8 0 0 0 0 0 0 0
0 0 0 0 8 1 1 1 1 1 1 8 0 0 0 0 0 0 0
0 0 0 0 8 1 1 1 1 1 2 8 0 0 0 0 0 0 0
0 0 0 0 4 1 1 1 1 1 1 8 0 0 0 0 0 3 0
0 0 0 0 8 1 1 1 1 1 1 8 0 0 0 0 0 0 0


Example8
Input:
0 0 0 0 8 0 0 0 0 0 0 8 0 0 0 0 0 0 0
0 0 0 0 8 0 0 0 0 0 0 8 0 0 0 0 0 0 0
8 8 8 8 8 8 8 8 8 8 8 8 8 8 8 8 8 8 8
0 0 0 0 8 0 0 0 0 0 0 8 0 0 0 0 0 0 0
0 0 0 0 8 0 0 0 0 0 0 8 0 0 0 0 0 0 0
0 0 0 0 8 0 0 0 0 0 0 8 0 0 0 0 0 0 0
0 0 0 0 8 0 0 0 0 0 0 8 0 0 0 0 0 0 0
8 8 8 8 8 8 8 8 8 8 8 8 8 8 8 8 8 8 8
0 0 0 0 8 0 0 0 0 0 0 8 0 0 0 0 0 0 0
0 0 0 0 8 0 0 0 0 0 0 8 0 0 0 0 0 0 0
0 0 0 0 8 0 0 0 0 0 0 8 0 0 0 0 0 0 0
0 0 0 0 8 0 0 0 0 0 0 8 0 0 0 0 0 0 0
0 0 0 0 8 0 0 0 0 0 0 8 0 0 0 0 0 0 0
0 0 0 0 8 0 0 0 0 0 0 8 0 0 0 0 0 0 0
0 0 0 0 8 0 0 0 0 0 0 8 0 0 0 0 0 0 0
0 0 0 0 8 0 0 0 0 0 0 8 0 0 0 0 0 0 0
0 0 0 0 8 0 0 0 0 0 0 8 0 0 0 0 0 0 0
0 0 0 0 8 0 0 0 0 0 0 8 0 0 0 0 0 0 0

Output:
0 0 0 0 8 2 2 2 2 2 2 8 0 0 0 0 0 0 2
0 0 0 0 0 2 2 2 2 2 2 8 6 0 0 0 0 0 0
8 8 8 8 8 8 8 8 8 8 8 8 8 8 8 8 8 8 8
4 4 3 4 8 6 6 6 6 6 6 8 3 3 3 3 3 3 3
4 4 4 4 8 6 6 6 6 6 6 8 3 3 3 4 3 3 3
4 4 1 4 8 6 6 6 6 8 6 8 3 3 3 3 3 3 3
4 4 4 4 8 6 8 6 6 6 6 8 3 3 3 3 3 3 3
8 8 8 8 8 8 8 8 8 8 2 8 8 8 8 8 8 8 8
0 0 0 0 8 1 1 3 1 1 1 8 0 0 0 0 0 0 0
0 0 0 0 8 1 1 1 1 1 1 8 0 0 0 0 0 0 0
0 0 0 0 8 1 1 1 1 1 1 8 0 0 0 0 0 3 0
0 0 0 0 8 1 1 1 1 1 1 8 0 0 0 0 0 0 0
0 0 0 0 8 1 1 1 1 1 1 8 0 0 0 2 0 0 0
0 0 0 0 8 1 1 1 1 1 1 8 0 2 0 0 0 0 0
0 0 0 0 8 1 1 1 1 1 1 8 0 0 0 0 0 0 0
0 0 0 0 8 1 1 1 1 1 1 8 3 0 0 0 0 0 0
0 0 0 0 8 1 1 1 1 1 1 8 0 1 0 0 0 0 3
0 0 0 0 8 1 0 1 1 1 1 8 0 0 0 0 0 0 0


Example9
Input:
0 0 0 0 8 0 0 0 0 0 0 8 0 0 0 0 0 0 0
0 0 0 0 8 0 0 0 0 0 0 8 0 0 0 0 0 0 0
8 8 8 8 8 8 8 8 8 8 8 8 8 8 8 8 8 8 8
0 0 0 0 8 0 0 0 0 0 0 8 0 0 0 0 0 0 0
0 0 0 0 8 0 0 0 0 0 0 8 0 0 0 0 0 0 0
0 0 0 0 8 0 0 0 0 0 0 8 0 0 0 0 0 0 0
0 0 0 0 8 0 0 0 0 0 0 8 0 0 0 0 0 0 0
8 8 8 8 8 8 8 8 8 8 8 8 8 8 8 8 8 8 8
0 0 0 0 8 0 0 0 0 0 0 8 0 0 0 0 0 0 0
0 0 0 0 8 0 0 0 0 0 0 8 0 0 0 0 0 0 0
0 0 0 0 8 0 0 0 0 0 0 8 0 0 0 0 0 0 0
0 0 0 0 8 0 0 0 0 0 0 8 0 0 0 0 0 0 0
0 0 0 0 8 0 0 0 0 0 0 8 0 0 0 0 0 0 0
0 0 0 0 8 0 0 0 0 0 0 8 0 0 0 0 0 0 0
0 0 0 0 8 0 0 0 0 0 0 8 0 0 0 0 0 0 0
0 0 0 0 8 0 0 0 0 0 0 8 0 0 0 0 0 0 0
0 0 0 0 8 0 0 0 0 0 0 8 0 0 0 0 0 0 0
0 0 0 0 8 0 0 0 0 0 0 8 0 0 0 0 0 0 0

Output:
0 0 0 0 8 2 2 2 2 2 2 8 0 0 0 0 0 0 0
3 0 0 0 8 2 1 2 2 2 2 8 0 0 0 0 0 0 0
8 8 8 8 8 8 8 8 8 8 8 8 8 8 8 8 0 8 8
4 4 4 4 8 6 6 6 6 6 2 8 3 3 3 6 3 3 3
4 4 4 4 8 6 6 6 6 6 6 8 3 3 3 3 3 3 3
4 4 4 4 8 6 6 6 6 6 6 8 3 3 3 3 3 3 3
4 4 4 4 8 6 6 6 6 6 6 8 3 3 3 3 3 3 3
8 8 8 8 8 4 8 8 8 8 8 8 8 8 8 8 8 3 8
0 0 0 0 8 1 1 8 1 1 1 8 0 0 0 0 0 0 0
0 0 0 4 8 1 1 1 8 1 1 8 0 0 0 0 0 0 0
0 0 0 0 8 1 1 1 1 1 1 8 0 0 0 0 0 0 0
0 0 0 0 8 1 1 1 1 1 1 0 0 0 4 0 0 0 0
0 0 0 0 8 1 1 1 1 1 1 8 0 0 0 0 0 0 0
0 0 6 0 8 1 1 1 1 1 1 8 0 0 0 0 0 0 0
0 0 0 0 8 1 1 1 1 1 2 8 0 0 0 0 0 0 0
0 0 0 0 8 1 1 1 1 1 1 8 0 0 0 0 8 0 4
0 0 0 0 8 1 1 1 1 1 1 0 0 0 0 0 0 0 0
0 0 0 0 8 1 1 1 1 1 1 8 0 0 0 0 0 0 0
Example10
Input:
0 0 8 0 0 0 0 0 0 8 0 0 0 0
0 0 8 0 0 0 0 0 0 8 0 0 0 0
0 0 8 0 0 0 0 0 0 8 0 0 0 0
0 0 8 0 0 0 0 0 0 8 0 0 0 0
8 8 8 8 8 8 8 8 8 8 8 8 8 8
0 0 8 0 0 0 0 0 0 8 0 0 0 0
0 0 8 0 0 0 0 0 0 8 0 0 0 0
8 8 8 8 8 8 8 8 8 8 8 8 8 8
0 0 8 0 0 0 0 0 0 8 0 0 0 0
0 0 8 0 0 0 0 0 0 8 0 0 0 0
0 0 8 0 0 0 0 0 0 8 0 0 0 0
0 0 8 0 0 0 0 0 0 8 0 0 0 0

Output:
0 0 8 2 2 2 2 2 2 8 0 0 0 0
0 0 8 2 2 2 2 2 2 8 0 0 0 0
0 0 8 2 2 2 2 2 2 8 0 0 0 0
0 0 8 2 2 2 2 2 2 8 0 0 0 0
8 8 8 8 8 8 8 8 8 8 8 3 8 8
4 4 8 6 6 1 6 6 6 8 3 3 3 3
4 4 6 6 6 6 6 6 6 8 3 3 3 3
8 8 8 8 8 8 8 8 8 8 8 8 8 8
0 0 8 1 8 1 1 3 1 8 0 0 0 0
0 0 8 1 1 1 1 1 1 8 0 0 0 0
0 0 8 1 1 1 1 1 1 8 0 0 0 0
0 1 8 3 1 1 2 1 1 8 0 0 0 0


Example11
Input:
0 0 8 0 0 0 0 0 0 8 0 0 0 0
0 0 8 0 0 0 0 0 0 8 0 0 0 0
0 0 8 0 0 0 0 0 0 8 0 0 0 0
0 0 8 0 0 0 0 0 0 8 0 0 0 0
8 8 8 8 8 8 8 8 8 8 8 8 8 8
0 0 8 0 0 0 0 0 0 8 0 0 0 0
0 0 8 0 0 0 0 0 0 8 0 0 0 0
8 8 8 8 8 8 8 8 8 8 8 8 8 8
0 0 8 0 0 0 0 0 0 8 0 0 0 0
0 0 8 0 0 0 0 0 0 8 0 0 0 0
0 0 8 0 0 0 0 0 0 8 0 0 0 0
0 0 8 0 0 0 0 0 0 8 0 0 0 0

Output:
0 0 8 2 2 2 2 2 2 8 0 0 0 4
0 0 8 2 2 2 2 2 2 8 0 0 0 0
0 0 8 1 2 2 2 2 2 8 0 0 0 0
0 0 8 2 2 2 2 2 8 8 0 0 0 0
8 8 8 8 8 8 8 8 8 8 8 8 8 8
4 4 8 6 6 6 6 6 6 8 3 3 3 3
4 4 8 6 6 6 6 6 6 8 3 3 3 3
8 8 8 8 8 1 8 8 8 8 8 8 8 8
0 0 8 1 1 1 1 1 1 8 0 0 0 0
0 0 4 1 0 1 1 1 1 8 0 0 0 0
0 0 8 1 1 1 1 1 1 8 0 8 0 0
0 0 8 1 1 1 1 1 1 8 0 0 0 3


Example12
Input:
0 0 8 0 0 0 0 0 0 8 0 0 0 0
0 0 8 0 0 0 0 0 0 8 0 0 0 0
0 0 8 0 0 0 0 0 0 8 0 0 0 0
0 0 8 0 0 0 0 0 0 8 0 0 0 0
8 8 8 8 8 8 8 8 8 8 8 8 8 8
0 0 8 0 0 0 0 0 0 8 0 0 0 0
0 0 8 0 0 0 0 0 0 8 0 0 0 0
8 8 8 8 8 8 8 8 8 8 8 8 8 8
0 0 8 0 0 0 0 0 0 8 0 0 0 0
0 0 8 0 0 0 0 0 0 8 0 0 0 0
0 0 8 0 0 0 0 0 0 8 0 0 0 0
0 0 8 0 0 0 0 0 0 8 0 0 0 0

Output:
0 0 3 2 2 2 2 2 2 8 0 0 0 0
0 0 8 2 2 2 2 2 2 8 0 0 0 0
0 0 8 2 6 2 2 2 2 8 0 0 0 0
0 0 8 2 2 2 2 2 2 8 0 0 0 0
8 8 8 8 8 8 8 8 8 8 8 8 8 8
4 4 8 6 6 6 6 6 6 8 3 3 3 3
4 4 8 6 6 6 6 6 6 8 3 3 4 3
8 8 8 8 8 8 8 8 2 8 8 8 8 4
0 6 8 1 1 1 1 1 1 1 0 0 0 0
0 0 8 1 1 1 1 1 1 8 0 0 0 0
0 0 0 1 1 1 1 1 1 8 0 0 0 0
0 0 8 1 1 1 1 1 1 8 0 0 0 0


Example13
Input:
0 0 8 0 0 0 0 0 0 8 0 0 0 0
0 0 8 0 0 0 0 0 0 8 0 0 0 0
0 0 8 0 0 0 0 0 0 8 0 0 0 0
0 0 8 0 0 0 0 0 0 8 0 0 0 0
8 8 8 8 8 8 8 8 8 8 8 8 8 8
0 0 8 0 0 0 0 0 0 8 0 0 0 0
0 0 8 0 0 0 0 0 0 8 0 0 0 0
8 8 8 8 8 8 8 8 8 8 8 8 8 8
0 0 8 0 0 0 0 0 0 8 0 0 0 0
0 0 8 0 0 0 0 0 0 8 0 0 0 0
0 0 8 0 0 0 0 0 0 8 0 0 0 0
0 0 8 0 0 0 0 0 0 8 0 0 0 0

Output:
0 0 8 2 2 2 2 2 2 8 0 0 0 0
0 2 8 2 2 8 2 2 2 8 0 0 0 2
0 0 8 2 2 2 2 2 2 8 0 0 0 0
0 0 8 2 2 2 2 2 2 8 0 0 0 0
8 8 8 8 8 8 8 8 8 8 8 8 8 8
4 4 8 6 6 6 6 6 6 8 3 3 3 3
4 4 8 2 6 6 3 6 6 8 3 3 3 3
8 8 8 8 8 8 8 0 8 8 8 8 8 8
0 0 8 1 1 1 1 1 1 8 0 0 1 0
0 0 8 1 1 1 1 1 1 8 0 0 0 0
0 0 8 1 1 1 1 1 1 8 0 0 0 0
0 0 8 1 6 1 1 1 1 8 0 0 0 0


Example14
Input:
0 0 8 0 0 0 0 0 0 8 0 0 0 0
0 0 8 0 0 0 0 0 0 8 0 0 0 0
0 0 8 0 0 0 0 0 0 8 0 0 0 0
0 0 8 0 0 0 0 0 0 8 0 0 0 0
8 8 8 8 8 8 8 8 8 8 8 8 8 8
0 0 8 0 0 0 0 0 0 8 0 0 0 0
0 0 8 0 0 0 0 0 0 8 0 0 0 0
8 8 8 8 8 8 8 8 8 8 8 8 8 8
0 0 8 0 0 0 0 0 0 8 0 0 0 0
0 0 8 0 0 0 0 0 0 8 0 0 0 0
0 0 8 0 0 0 0 0 0 8 0 0 0 0
0 0 8 0 0 0 0 0 0 8 0 0 0 0

Output:
0 0 8 2 2 2 2 2 2 8 0 0 0 0
0 0 8 2 2 2 2 2 2 8 0 0 0 0
0 0 8 2 2 2 2 2 2 8 0 0 0 0
0 0 8 2 2 2 2 2 2 8 0 0 0 0
8 8 8 8 8 8 1 8 8 8 8 8 8 8
0 4 8 6 6 6 6 6 6 8 3 3 3 3
4 4 8 6 6 6 6 6 6 8 3 3 3 3
8 8 8 8 8 8 8 8 8 8 8 4 8 8
0 0 8 1 1 1 1 1 1 8 0 0 2 0
6 0 8 3 1 1 1 1 1 4 0 0 0 0
0 0 8 1 1 1 1 1 1 8 0 0 0 0
0 0 8 1 1 1 1 1 1 8 0 0 0 2


Example15
Input:
0 0 8 0 0 0 0 0 0 8 0 0 0 0
0 0 8 0 0 0 0 0 0 8 0 0 0 0
0 0 8 0 0 0 0 0 0 8 0 0 0 0
0 0 8 0 0 0 0 0 0 8 0 0 0 0
8 8 8 8 8 8 8 8 8 8 8 8 8 8
0 0 8 0 0 0 0 0 0 8 0 0 0 0
0 0 8 0 0 0 0 0 0 8 0 0 0 0
8 8 8 8 8 8 8 8 8 8 8 8 8 8
0 0 8 0 0 0 0 0 0 8 0 0 0 0
0 0 8 0 0 0 0 0 0 8 0 0 0 0
0 0 8 0 0 0 0 0 0 8 0 0 0 0
0 0 8 0 0 0 0 0 0 8 0 0 0 0

Output:
0 0 2 4 2 2 2 8 2 8 0 0 0 0
0 0 8 2 2 2 2 2 2 8 0 0 0 3
0 0 8 2 2 2 2 2 2 8 0 0 0 0
0 0 8 2 2 2 2 2 2 8 0 0 0 0
8 8 8 8 8 8 8 8 8 8 6 8 8 8
4 4 8 6 6 6 6 6 6 8 3 3 3 3
4 4 8 6 6 6 6 6 6 8 3 3 3 3
8 8 8 8 8 8 8 8 8 8 8 8 8 8
0 8 8 1 1 1 1 1 1 8 0 0 0 0
0 0 8 1 1 1 1 1 1 8 0 0 0 0
0 0 8 1 1 1 1 1 1 8 6 0 0 0
0 0 8 1 1 1 1 4 1 8 0 0 0 0


Example16
Input:
0 0 8 0 0 0 0 0 0 8 0 0 0 0
0 0 8 0 0 0 0 0 0 8 0 0 0 0
0 0 8 0 0 0 0 0 0 8 0 0 0 0
0 0 8 0 0 0 0 0 0 8 0 0 0 0
8 8 8 8 8 8 8 8 8 8 8 8 8 8
0 0 8 0 0 0 0 0 0 8 0 0 0 0
0 0 8 0 0 0 0 0 0 8 0 0 0 0
8 8 8 8 8 8 8 8 8 8 8 8 8 8
0 0 8 0 0 0 0 0 0 8 0 0 0 0
0 0 8 0 0 0 0 0 0 8 0 0 0 0
0 0 8 0 0 0 0 0 0 8 0 0 0 0
0 0 8 0 0 0 0 0 0 8 0 0 0 0

Output:
0 0 8 2 2 2 2 2 2 8 0 0 0 0
0 0 2 2 2 2 2 2 2 8 0 8 0 0
0 0 8 2 2 2 2 2 2 8 0 0 0 0
0 0 8 2 2 2 2 2 2 8 0 0 0 0
8 8 8 8 8 8 8 8 1 8 8 8 8 8
4 4 8 6 6 6 2 6 6 8 0 3 3 3
4 4 8 6 6 6 6 6 6 8 3 3 3 3
8 8 8 4 8 8 8 8 8 8 8 8 8 8
0 0 8 1 1 1 1 1 1 8 0 0 0 0
0 0 8 1 1 1 1 1 1 8 0 0 0 0
0 0 8 1 1 1 0 1 1 8 0 0 0 0
0 0 8 1 1 1 0 1 1 8 0 0 0 0


Example17
Input:
0 0 8 0 0 0 0 0 0 8 0 0 0 0
0 0 8 0 0 0 0 0 0 8 0 0 0 0
0 0 8 0 0 0 0 0 0 8 0 0 0 0
0 0 8 0 0 0 0 0 0 8 0 0 0 0
8 8 8 8 8 8 8 8 8 8 8 8 8 8
0 0 8 0 0 0 0 0 0 8 0 0 0 0
0 0 8 0 0 0 0 0 0 8 0 0 0 0
8 8 8 8 8 8 8 8 8 8 8 8 8 8
0 0 8 0 0 0 0 0 0 8 0 0 0 0
0 0 8 0 0 0 0 0 0 8 0 0 0 0
0 0 8 0 0 0 0 0 0 8 0 0 0 0
0 0 8 0 0 0 0 0 0 8 0 0 0 0

Output:
0 0 8 2 2 2 2 6 2 8 0 3 0 0
0 0 8 2 2 2 3 2 2 8 0 0 0 0
0 0 8 2 2 2 2 2 2 8 0 0 0 0
0 0 8 1 2 2 2 2 2 8 0 0 0 0
8 8 8 8 8 8 8 8 4 8 8 8 8 8
4 4 8 6 6 6 6 6 6 8 4 3 3 3
4 4 8 6 6 6 6 6 6 8 3 0 3 3
8 8 8 8 8 8 8 8 8 8 8 8 8 8
0 0 8 1 1 1 1 1 1 8 0 0 0 0
0 0 8 1 1 1 1 1 1 8 0 0 0 0
0 0 8 1 1 1 1 1 1 8 1 0 0 0
0 0 8 1 1 1 1 1 1 8 0 0 0 0


Example18
Input:
0 0 8 0 0 0 0 0 0 8 0 0 0 0
0 0 8 0 0 0 0 0 0 8 0 0 0 0
0 0 8 0 0 0 0 0 0 8 0 0 0 0
0 0 8 0 0 0 0 0 0 8 0 0 0 0
8 8 8 8 8 8 8 8 8 8 8 8 8 8
0 0 8 0 0 0 0 0 0 8 0 0 0 0
0 0 8 0 0 0 0 0 0 8 0 0 0 0
8 8 8 8 8 8 8 8 8 8 8 8 8 8
0 0 8 0 0 0 0 0 0 8 0 0 0 0
0 0 8 0 0 0 0 0 0 8 0 0 0 0
0 0 8 0 0 0 0 0 0 8 0 0 0 0
0 0 8 0 0 0 0 0 0 8 0 0 0 0

Output:
0 0 8 2 2 2 2 2 2 8 0 0 0 0
0 0 8 2 2 2 2 2 2 8 0 0 0 0
0 0 8 1 2 2 2 2 2 8 0 3 0 0
0 0 8 2 2 2 2 2 2 8 0 0 0 0
8 8 8 8 8 8 8 8 8 8 8 8 8 8
4 4 8 6 6 6 6 6 6 8 3 3 3 3
4 4 8 6 6 6 6 6 6 8 3 3 3 3
4 8 8 8 8 8 8 8 8 0 1 8 8 8
0 0 8 1 6 1 1 1 1 8 0 0 0 0
0 0 8 1 1 1 1 1 1 8 0 0 0 0
0 0 8 1 1 1 1 1 1 1 0 0 0 0
0 0 8 1 1 1 1 8 1 8 0 0 0 0

Below is a test input grid. 
Predict the corresponding output. 

Input:
0 0 0 8 0 0 0 0 8 0 0 0 0 0 0
0 0 0 8 0 0 0 0 8 0 0 0 0 0 0
0 0 0 8 0 0 0 0 8 0 0 0 0 0 0
0 0 0 8 0 0 0 0 8 0 0 0 0 0 0
0 0 0 8 0 0 0 0 8 0 0 0 0 0 0
0 0 0 8 0 0 0 0 8 0 0 0 0 0 0
8 8 8 8 8 8 8 8 8 8 8 8 8 8 8
0 0 0 8 0 0 0 0 8 0 0 0 0 0 0
0 0 0 8 0 0 0 0 8 0 0 0 0 0 0
0 0 0 8 0 0 0 0 8 0 0 0 0 0 0
0 0 0 8 0 0 0 0 8 0 0 0 0 0 0
0 0 0 8 0 0 0 0 8 0 0 0 0 0 0
0 0 0 8 0 0 0 0 8 0 0 0 0 0 0
8 8 8 8 8 8 8 8 8 8 8 8 8 8 8
0 0 0 8 0 0 0 0 8 0 0 0 0 0 0
0 0 0 8 0 0 0 0 8 0 0 0 0 0 0
0 0 0 8 0 0 0 0 8 0 0 0 0 0 0
\end{lstlisting}
\end{tcolorbox}

\switchcolumn 

\begin{tcolorbox}[
  colback=cyan!10!white,
  colframe=black,
  boxrule=0.6pt,
  arc=1mm,
  breakable,
  enhanced,
  title=\centering \textbf{9-Noisy Input Example With Original Prompt}
]
\scriptsize
\begin{lstlisting}
Find the common rule that maps 
an input grid to an output grid, 
given the examples below.

Example1
Input:
0 0 0 0 8 0 0 0 0 0 0 8 0 0 0 0 0 0 0
0 0 0 0 8 0 0 0 0 0 0 8 0 0 0 0 0 0 0
8 8 8 8 8 8 8 8 8 8 8 8 8 8 8 8 8 8 8
0 0 0 0 8 0 0 0 0 0 0 8 0 0 0 0 0 0 0
0 0 0 0 8 0 0 0 0 0 0 8 0 0 0 0 0 0 0
0 0 0 0 8 0 0 0 0 0 0 8 0 0 0 0 0 0 0
0 0 0 0 8 0 0 0 0 0 0 8 0 0 0 0 0 0 0
8 8 8 8 8 8 8 8 8 8 8 8 8 8 8 8 8 8 8
0 0 0 0 8 0 0 0 0 0 0 8 0 0 0 0 0 0 0
0 0 0 0 8 0 0 0 0 0 0 8 0 0 0 0 0 0 0
0 0 0 0 8 0 0 0 0 0 0 8 0 0 0 0 0 0 0
0 0 0 0 8 0 0 0 0 0 0 8 0 0 0 0 0 0 0
0 0 0 0 8 0 0 0 0 0 0 8 0 0 0 0 0 0 0
0 0 0 0 8 0 0 0 0 0 0 8 0 0 0 0 0 0 0
0 0 0 0 8 0 0 0 0 0 0 8 0 0 0 0 0 0 0
0 0 0 0 8 0 0 0 0 0 0 8 0 0 0 0 0 0 0
0 0 0 0 8 0 0 0 0 0 0 8 0 0 0 0 0 0 0
0 0 0 0 8 0 0 0 0 0 0 8 0 0 0 0 0 0 0

Output:
0 0 0 0 8 2 3 2 2 2 2 8 0 0 0 0 0 0 0
0 0 0 0 8 2 2 2 2 2 2 8 0 0 0 0 0 0 0
8 8 8 8 8 8 8 8 8 8 8 8 8 8 2 8 2 8 8
4 4 4 4 8 6 6 6 6 6 6 8 3 3 3 3 3 3 0
4 0 4 4 8 6 6 6 6 6 6 8 3 3 3 3 3 3 3
4 4 4 4 8 0 6 6 6 6 6 8 3 3 3 3 3 3 3
4 4 4 4 8 6 6 6 6 6 6 8 3 3 3 3 3 3 3
8 8 8 8 8 8 8 8 8 8 8 8 8 8 8 8 4 8 8
0 0 0 0 8 1 1 1 6 1 1 8 0 0 0 0 0 0 0
1 0 0 0 8 1 1 1 1 1 1 8 0 0 0 1 0 0 0
0 0 0 0 8 1 1 1 1 6 1 8 0 0 0 0 0 0 0
0 0 0 0 8 1 1 1 1 1 1 8 0 0 0 0 0 0 0
0 0 0 0 8 1 1 1 1 1 1 8 2 0 0 0 0 0 0
0 0 0 0 8 1 1 1 1 1 1 8 0 0 0 0 0 0 0
0 0 0 0 8 1 1 8 1 1 6 8 0 0 0 0 0 0 3
0 0 0 0 8 1 1 1 1 1 1 8 0 0 0 0 0 0 2
0 0 0 0 8 1 1 1 1 1 1 8 0 0 2 0 0 0 0
0 0 0 0 8 1 1 1 1 1 1 8 0 0 0 0 0 0 0


Example2
Input:
0 0 0 0 8 0 0 0 0 0 0 8 0 0 0 0 0 0 0
0 0 0 0 8 0 0 0 0 0 0 8 0 0 0 0 0 0 0
8 8 8 8 8 8 8 8 8 8 8 8 8 8 8 8 8 8 8
0 0 0 0 8 0 0 0 0 0 0 8 0 0 0 0 0 0 0
0 0 0 0 8 0 0 0 0 0 0 8 0 0 0 0 0 0 0
0 0 0 0 8 0 0 0 0 0 0 8 0 0 0 0 0 0 0
0 0 0 0 8 0 0 0 0 0 0 8 0 0 0 0 0 0 0
8 8 8 8 8 8 8 8 8 8 8 8 8 8 8 8 8 8 8
0 0 0 0 8 0 0 0 0 0 0 8 0 0 0 0 0 0 0
0 0 0 0 8 0 0 0 0 0 0 8 0 0 0 0 0 0 0
0 0 0 0 8 0 0 0 0 0 0 8 0 0 0 0 0 0 0
0 0 0 0 8 0 0 0 0 0 0 8 0 0 0 0 0 0 0
0 0 0 0 8 0 0 0 0 0 0 8 0 0 0 0 0 0 0
0 0 0 0 8 0 0 0 0 0 0 8 0 0 0 0 0 0 0
0 0 0 0 8 0 0 0 0 0 0 8 0 0 0 0 0 0 0
0 0 0 0 8 0 0 0 0 0 0 8 0 0 0 0 0 0 0
0 0 0 0 8 0 0 0 0 0 0 8 0 0 0 0 0 0 0
0 0 0 0 8 0 0 0 0 0 0 8 0 0 0 0 0 0 0

Output:
0 1 0 0 8 2 2 2 2 2 2 8 0 0 3 4 0 0 0
0 0 0 0 8 2 2 2 2 2 3 8 0 0 0 0 0 0 0
8 8 8 8 8 8 8 8 8 8 8 8 8 8 8 8 8 8 8
4 4 4 4 8 6 6 6 6 6 6 8 3 3 3 3 3 3 3
4 4 4 4 8 6 6 6 6 6 6 8 3 3 3 3 3 3 3
4 4 4 4 8 6 6 6 6 6 6 8 3 3 3 3 3 3 3
4 4 4 4 8 6 6 6 1 6 6 8 3 3 3 3 3 3 3
8 8 8 8 8 8 8 8 8 8 8 8 1 8 8 8 8 8 8
0 0 0 0 8 1 1 1 1 1 1 8 0 0 0 0 0 0 0
0 0 0 6 8 1 1 1 2 1 4 8 8 0 0 0 0 0 0
0 0 4 0 8 1 1 1 1 1 1 8 0 0 0 0 0 0 0
0 0 0 0 8 1 1 1 1 1 1 8 0 0 0 0 0 0 0
0 0 0 0 8 1 1 1 1 1 1 8 0 3 0 0 0 0 0
0 3 0 0 8 1 1 1 1 1 8 8 0 0 0 0 0 0 0
0 0 0 0 8 3 1 1 1 1 1 8 0 0 0 0 0 0 0
0 0 0 0 8 1 1 1 1 1 1 8 0 0 0 0 0 0 0
0 0 0 0 8 1 1 1 1 1 1 8 0 0 0 0 0 0 0
0 0 0 0 8 1 1 1 1 1 1 4 0 0 0 0 0 6 0


Example3
Input:
0 0 0 0 8 0 0 0 0 0 0 8 0 0 0 0 0 0 0
0 0 0 0 8 0 0 0 0 0 0 8 0 0 0 0 0 0 0
8 8 8 8 8 8 8 8 8 8 8 8 8 8 8 8 8 8 8
0 0 0 0 8 0 0 0 0 0 0 8 0 0 0 0 0 0 0
0 0 0 0 8 0 0 0 0 0 0 8 0 0 0 0 0 0 0
0 0 0 0 8 0 0 0 0 0 0 8 0 0 0 0 0 0 0
0 0 0 0 8 0 0 0 0 0 0 8 0 0 0 0 0 0 0
8 8 8 8 8 8 8 8 8 8 8 8 8 8 8 8 8 8 8
0 0 0 0 8 0 0 0 0 0 0 8 0 0 0 0 0 0 0
0 0 0 0 8 0 0 0 0 0 0 8 0 0 0 0 0 0 0
0 0 0 0 8 0 0 0 0 0 0 8 0 0 0 0 0 0 0
0 0 0 0 8 0 0 0 0 0 0 8 0 0 0 0 0 0 0
0 0 0 0 8 0 0 0 0 0 0 8 0 0 0 0 0 0 0
0 0 0 0 8 0 0 0 0 0 0 8 0 0 0 0 0 0 0
0 0 0 0 8 0 0 0 0 0 0 8 0 0 0 0 0 0 0
0 0 0 0 8 0 0 0 0 0 0 8 0 0 0 0 0 0 0
0 0 0 0 8 0 0 0 0 0 0 8 0 0 0 0 0 0 0
0 0 0 0 8 0 0 0 0 0 0 8 0 0 0 0 0 0 0

Output:
0 0 0 0 8 2 2 2 2 2 2 8 0 0 0 0 0 0 0
0 0 0 0 8 2 2 2 2 2 2 8 0 0 0 0 0 0 0
8 8 8 8 8 8 8 8 8 8 3 8 8 8 8 8 8 8 8
4 4 4 4 8 6 6 6 6 6 6 8 3 3 3 3 3 3 3
4 2 4 4 8 6 6 3 6 6 6 8 3 3 3 3 3 3 3
4 4 4 4 8 6 6 6 6 6 6 8 3 3 3 3 3 2 3
4 4 4 3 8 6 6 6 6 6 6 8 3 3 3 3 3 3 3
8 8 8 8 8 8 8 8 8 8 8 4 8 8 8 8 8 8 8
0 0 0 0 8 1 1 1 1 1 1 8 0 0 0 0 0 0 0
4 0 0 0 2 1 1 1 1 1 1 8 0 0 0 0 0 0 0
0 0 0 0 8 1 1 1 1 1 1 8 0 0 0 0 0 0 0
0 0 0 4 8 1 1 1 1 1 1 8 0 0 0 0 0 0 0
4 0 0 0 8 1 1 1 1 1 1 8 0 0 0 0 0 4 0
3 3 0 0 8 1 1 1 1 1 1 8 0 6 0 0 0 0 0
0 2 0 0 8 1 1 1 1 1 1 8 0 0 0 0 0 0 0
0 0 0 0 8 1 1 1 1 1 1 8 0 0 0 0 0 0 0
0 0 0 0 8 1 1 1 1 8 1 8 0 0 0 0 0 0 0
0 0 0 0 8 1 1 1 1 1 1 8 0 0 0 0 0 0 2


Example4
Input:
0 0 0 0 8 0 0 0 0 0 0 8 0 0 0 0 0 0 0
0 0 0 0 8 0 0 0 0 0 0 8 0 0 0 0 0 0 0
8 8 8 8 8 8 8 8 8 8 8 8 8 8 8 8 8 8 8
0 0 0 0 8 0 0 0 0 0 0 8 0 0 0 0 0 0 0
0 0 0 0 8 0 0 0 0 0 0 8 0 0 0 0 0 0 0
0 0 0 0 8 0 0 0 0 0 0 8 0 0 0 0 0 0 0
0 0 0 0 8 0 0 0 0 0 0 8 0 0 0 0 0 0 0
8 8 8 8 8 8 8 8 8 8 8 8 8 8 8 8 8 8 8
0 0 0 0 8 0 0 0 0 0 0 8 0 0 0 0 0 0 0
0 0 0 0 8 0 0 0 0 0 0 8 0 0 0 0 0 0 0
0 0 0 0 8 0 0 0 0 0 0 8 0 0 0 0 0 0 0
0 0 0 0 8 0 0 0 0 0 0 8 0 0 0 0 0 0 0
0 0 0 0 8 0 0 0 0 0 0 8 0 0 0 0 0 0 0
0 0 0 0 8 0 0 0 0 0 0 8 0 0 0 0 0 0 0
0 0 0 0 8 0 0 0 0 0 0 8 0 0 0 0 0 0 0
0 0 0 0 8 0 0 0 0 0 0 8 0 0 0 0 0 0 0
0 0 0 0 8 0 0 0 0 0 0 8 0 0 0 0 0 0 0
0 0 0 0 8 0 0 0 0 0 0 8 0 0 0 0 0 0 0

Output:
0 0 0 0 3 2 2 2 2 2 8 8 0 0 0 0 0 0 0
0 0 0 0 8 2 2 2 2 2 2 8 0 0 0 0 0 0 0
8 8 8 8 8 8 8 8 8 8 8 8 8 8 8 8 8 8 8
4 4 4 4 8 6 6 4 6 6 6 8 3 3 3 3 3 3 3
4 4 4 4 8 6 6 6 6 6 6 8 3 3 3 3 4 3 3
4 4 4 4 8 6 6 6 6 6 6 8 3 3 3 3 3 3 3
4 4 4 3 8 6 6 6 6 3 6 8 3 3 3 3 3 3 3
8 8 8 8 8 8 8 8 8 8 8 8 8 8 8 8 8 8 8
0 4 0 0 8 1 1 1 1 1 1 8 1 0 0 0 0 0 0
0 0 0 0 8 1 1 1 1 1 1 8 0 0 0 0 0 0 0
0 0 0 0 8 1 1 1 1 1 2 8 0 0 0 0 0 0 0
0 0 0 0 8 1 2 4 1 1 1 2 0 0 0 0 0 0 0
0 0 0 0 8 1 1 1 1 1 1 8 0 0 0 0 0 0 0
0 0 0 0 8 1 6 1 1 1 1 8 0 0 0 0 8 4 2
0 0 0 0 8 1 1 1 1 1 1 8 0 0 0 0 0 1 0
0 0 0 0 8 1 1 1 1 1 1 8 0 0 0 0 0 0 0
0 0 0 0 8 1 1 1 1 1 1 8 0 0 0 0 0 0 0
0 0 0 0 8 1 1 1 1 1 1 8 0 0 0 0 0 0 0


Example5
Input:
0 0 0 0 8 0 0 0 0 0 0 8 0 0 0 0 0 0 0
0 0 0 0 8 0 0 0 0 0 0 8 0 0 0 0 0 0 0
8 8 8 8 8 8 8 8 8 8 8 8 8 8 8 8 8 8 8
0 0 0 0 8 0 0 0 0 0 0 8 0 0 0 0 0 0 0
0 0 0 0 8 0 0 0 0 0 0 8 0 0 0 0 0 0 0
0 0 0 0 8 0 0 0 0 0 0 8 0 0 0 0 0 0 0
0 0 0 0 8 0 0 0 0 0 0 8 0 0 0 0 0 0 0
8 8 8 8 8 8 8 8 8 8 8 8 8 8 8 8 8 8 8
0 0 0 0 8 0 0 0 0 0 0 8 0 0 0 0 0 0 0
0 0 0 0 8 0 0 0 0 0 0 8 0 0 0 0 0 0 0
0 0 0 0 8 0 0 0 0 0 0 8 0 0 0 0 0 0 0
0 0 0 0 8 0 0 0 0 0 0 8 0 0 0 0 0 0 0
0 0 0 0 8 0 0 0 0 0 0 8 0 0 0 0 0 0 0
0 0 0 0 8 0 0 0 0 0 0 8 0 0 0 0 0 0 0
0 0 0 0 8 0 0 0 0 0 0 8 0 0 0 0 0 0 0
0 0 0 0 8 0 0 0 0 0 0 8 0 0 0 0 0 0 0
0 0 0 0 8 0 0 0 0 0 0 8 0 0 0 0 0 0 0
0 0 0 0 8 0 0 0 0 0 0 8 0 0 0 0 0 0 0

Output:
0 0 0 0 8 2 2 2 2 2 2 8 6 0 0 0 0 0 0
0 0 0 8 8 2 2 2 2 2 2 8 0 0 0 0 0 0 0
8 8 8 8 8 8 8 8 8 8 8 8 8 8 8 8 8 8 8
4 4 4 4 8 6 6 6 6 0 6 8 3 3 3 3 3 3 3
4 4 4 4 8 6 1 4 6 6 6 8 3 3 3 3 3 3 3
4 4 4 4 8 4 6 6 6 6 6 8 3 3 3 1 3 3 3
4 4 4 3 3 6 6 6 6 6 6 8 3 3 3 3 3 3 3
8 8 8 8 8 8 8 8 8 8 8 8 8 8 8 8 8 8 8
0 0 0 0 8 1 1 1 1 1 1 8 0 0 0 0 0 0 0
0 3 0 0 8 1 1 1 1 1 1 8 3 0 0 0 0 0 0
0 0 0 0 8 1 1 1 2 1 1 8 0 0 0 0 0 0 0
0 0 0 0 3 1 1 1 1 1 1 8 0 0 0 0 0 0 0
0 0 0 0 8 1 1 1 1 1 1 8 0 0 0 0 0 0 0
0 0 0 0 2 1 1 1 1 1 1 8 0 0 0 0 0 0 0
0 0 0 0 8 1 1 1 1 1 1 8 0 0 0 0 0 0 0
0 0 0 0 8 1 1 1 1 1 1 8 0 0 0 0 0 0 0
0 0 0 1 2 1 1 1 1 1 1 8 0 0 0 0 0 0 0
0 0 0 0 8 1 1 1 1 1 1 8 0 0 0 0 0 1 0


Example6
Input:
0 0 0 0 8 0 0 0 0 0 0 8 0 0 0 0 0 0 0
0 0 0 0 8 0 0 0 0 0 0 8 0 0 0 0 0 0 0
8 8 8 8 8 8 8 8 8 8 8 8 8 8 8 8 8 8 8
0 0 0 0 8 0 0 0 0 0 0 8 0 0 0 0 0 0 0
0 0 0 0 8 0 0 0 0 0 0 8 0 0 0 0 0 0 0
0 0 0 0 8 0 0 0 0 0 0 8 0 0 0 0 0 0 0
0 0 0 0 8 0 0 0 0 0 0 8 0 0 0 0 0 0 0
8 8 8 8 8 8 8 8 8 8 8 8 8 8 8 8 8 8 8
0 0 0 0 8 0 0 0 0 0 0 8 0 0 0 0 0 0 0
0 0 0 0 8 0 0 0 0 0 0 8 0 0 0 0 0 0 0
0 0 0 0 8 0 0 0 0 0 0 8 0 0 0 0 0 0 0
0 0 0 0 8 0 0 0 0 0 0 8 0 0 0 0 0 0 0
0 0 0 0 8 0 0 0 0 0 0 8 0 0 0 0 0 0 0
0 0 0 0 8 0 0 0 0 0 0 8 0 0 0 0 0 0 0
0 0 0 0 8 0 0 0 0 0 0 8 0 0 0 0 0 0 0
0 0 0 0 8 0 0 0 0 0 0 8 0 0 0 0 0 0 0
0 0 0 0 8 0 0 0 0 0 0 8 0 0 0 0 0 0 0
0 0 0 0 8 0 0 0 0 0 0 8 0 0 0 0 0 0 0

Output:
0 0 0 0 8 2 2 2 2 2 2 8 0 0 0 0 0 0 4
0 0 4 0 8 2 2 2 2 2 2 8 3 0 0 0 0 0 0
8 8 8 8 8 8 8 8 8 8 8 8 8 8 8 8 8 8 8
4 4 4 4 8 6 6 6 6 6 6 8 0 3 4 3 3 3 3
4 4 4 0 8 6 6 6 6 6 6 8 3 3 3 3 3 3 3
4 4 4 4 8 6 6 6 6 6 6 8 3 3 3 3 3 3 3
8 4 4 4 8 6 3 1 6 6 6 8 3 3 3 3 3 3 3
8 8 8 8 8 0 8 8 8 8 8 8 8 8 8 8 8 8 8
0 0 0 0 8 1 1 1 1 1 1 8 0 0 0 0 0 0 0
0 0 0 0 8 1 1 1 1 1 1 8 0 0 0 0 0 0 0
0 0 0 0 8 1 1 1 1 1 1 8 0 0 0 0 0 0 0
0 0 0 0 8 1 1 1 1 1 1 8 0 0 0 0 0 0 0
0 0 0 0 0 1 1 1 1 1 1 8 0 0 0 0 0 0 0
0 0 0 0 8 1 1 1 1 1 6 8 0 0 0 0 0 0 0
0 0 0 0 8 1 1 1 1 1 1 8 0 0 0 8 0 0 0
0 0 0 0 8 1 1 1 1 1 1 8 4 0 0 0 0 0 0
0 0 0 0 8 1 1 1 1 1 1 8 0 0 0 8 0 0 0
0 0 0 0 8 1 1 8 1 1 1 8 0 0 0 0 1 0 0


Example7
Input:
0 0 0 0 8 0 0 0 0 0 0 8 0 0 0 0 0 0 0
0 0 0 0 8 0 0 0 0 0 0 8 0 0 0 0 0 0 0
8 8 8 8 8 8 8 8 8 8 8 8 8 8 8 8 8 8 8
0 0 0 0 8 0 0 0 0 0 0 8 0 0 0 0 0 0 0
0 0 0 0 8 0 0 0 0 0 0 8 0 0 0 0 0 0 0
0 0 0 0 8 0 0 0 0 0 0 8 0 0 0 0 0 0 0
0 0 0 0 8 0 0 0 0 0 0 8 0 0 0 0 0 0 0
8 8 8 8 8 8 8 8 8 8 8 8 8 8 8 8 8 8 8
0 0 0 0 8 0 0 0 0 0 0 8 0 0 0 0 0 0 0
0 0 0 0 8 0 0 0 0 0 0 8 0 0 0 0 0 0 0
0 0 0 0 8 0 0 0 0 0 0 8 0 0 0 0 0 0 0
0 0 0 0 8 0 0 0 0 0 0 8 0 0 0 0 0 0 0
0 0 0 0 8 0 0 0 0 0 0 8 0 0 0 0 0 0 0
0 0 0 0 8 0 0 0 0 0 0 8 0 0 0 0 0 0 0
0 0 0 0 8 0 0 0 0 0 0 8 0 0 0 0 0 0 0
0 0 0 0 8 0 0 0 0 0 0 8 0 0 0 0 0 0 0
0 0 0 0 8 0 0 0 0 0 0 8 0 0 0 0 0 0 0
0 0 0 0 8 0 0 0 0 0 0 8 0 0 0 0 0 0 0

Output:
0 0 0 0 8 2 2 2 2 2 2 8 0 0 0 4 0 0 0
0 0 0 0 0 2 2 2 2 2 2 8 0 0 0 0 0 0 0
2 8 8 8 8 8 8 4 8 8 1 8 8 8 8 8 8 8 8
4 3 4 0 8 6 6 6 6 6 6 8 3 3 3 3 3 3 3
4 4 4 4 8 6 6 6 6 6 6 8 3 3 4 3 3 3 3
4 4 4 4 8 6 6 6 6 6 6 8 3 3 3 3 3 3 3
4 4 4 4 8 6 6 6 6 6 6 8 3 3 3 3 3 3 3
8 8 8 8 8 8 8 8 8 8 0 8 8 8 8 8 8 8 8
0 0 0 0 8 1 1 1 1 1 2 8 0 0 0 3 0 0 0
0 0 0 0 8 1 1 1 1 1 1 8 0 0 0 0 0 0 0
0 0 0 0 8 1 1 1 1 1 1 8 0 0 0 0 0 0 0
0 0 0 0 8 1 1 1 1 1 1 8 4 0 0 1 0 0 0
0 0 0 0 8 1 1 1 1 1 1 8 0 0 0 0 0 0 0
1 0 0 0 8 1 1 1 1 1 1 8 0 0 0 0 0 0 0
0 0 0 0 8 1 1 1 1 1 1 8 0 0 0 0 0 0 0
0 0 0 0 8 1 1 1 1 1 2 8 0 0 0 0 0 0 0
0 0 0 0 4 1 1 1 1 1 1 8 0 0 0 0 0 3 0
0 0 0 0 8 1 1 1 1 1 1 8 0 0 0 0 0 0 0


Example8
Input:
0 0 0 0 8 0 0 0 0 0 0 8 0 0 0 0 0 0 0
0 0 0 0 8 0 0 0 0 0 0 8 0 0 0 0 0 0 0
8 8 8 8 8 8 8 8 8 8 8 8 8 8 8 8 8 8 8
0 0 0 0 8 0 0 0 0 0 0 8 0 0 0 0 0 0 0
0 0 0 0 8 0 0 0 0 0 0 8 0 0 0 0 0 0 0
0 0 0 0 8 0 0 0 0 0 0 8 0 0 0 0 0 0 0
0 0 0 0 8 0 0 0 0 0 0 8 0 0 0 0 0 0 0
8 8 8 8 8 8 8 8 8 8 8 8 8 8 8 8 8 8 8
0 0 0 0 8 0 0 0 0 0 0 8 0 0 0 0 0 0 0
0 0 0 0 8 0 0 0 0 0 0 8 0 0 0 0 0 0 0
0 0 0 0 8 0 0 0 0 0 0 8 0 0 0 0 0 0 0
0 0 0 0 8 0 0 0 0 0 0 8 0 0 0 0 0 0 0
0 0 0 0 8 0 0 0 0 0 0 8 0 0 0 0 0 0 0
0 0 0 0 8 0 0 0 0 0 0 8 0 0 0 0 0 0 0
0 0 0 0 8 0 0 0 0 0 0 8 0 0 0 0 0 0 0
0 0 0 0 8 0 0 0 0 0 0 8 0 0 0 0 0 0 0
0 0 0 0 8 0 0 0 0 0 0 8 0 0 0 0 0 0 0
0 0 0 0 8 0 0 0 0 0 0 8 0 0 0 0 0 0 0

Output:
0 0 0 0 8 2 2 2 2 2 2 8 0 0 0 0 0 0 2
0 0 0 0 0 2 2 2 2 2 2 8 6 0 0 0 0 0 0
8 8 8 8 8 8 8 8 8 8 8 8 8 8 8 8 8 8 8
4 4 3 4 8 6 6 6 6 6 6 8 3 3 3 3 3 3 3
4 4 4 4 8 6 6 6 6 6 6 8 3 3 3 4 3 3 3
4 4 1 4 8 6 6 6 6 8 6 8 3 3 3 3 3 3 3
4 4 4 4 8 6 8 6 6 6 6 8 3 3 3 3 3 3 3
8 8 8 8 8 8 8 8 8 8 2 8 8 8 8 8 8 8 8
0 0 0 0 8 1 1 3 1 1 1 8 0 0 0 0 0 0 0
0 0 0 0 8 1 1 1 1 1 1 8 0 0 0 0 0 0 0
0 0 0 0 8 1 1 1 1 1 1 8 0 0 0 0 0 3 0
0 0 0 0 8 1 1 1 1 1 1 8 0 0 0 0 0 0 0
0 0 0 0 8 1 1 1 1 1 1 8 0 0 0 2 0 0 0
0 0 0 0 8 1 1 1 1 1 1 8 0 2 0 0 0 0 0
0 0 0 0 8 1 1 1 1 1 1 8 0 0 0 0 0 0 0
0 0 0 0 8 1 1 1 1 1 1 8 3 0 0 0 0 0 0
0 0 0 0 8 1 1 1 1 1 1 8 0 1 0 0 0 0 3
0 0 0 0 8 1 0 1 1 1 1 8 0 0 0 0 0 0 0


Example9
Input:
0 0 0 0 8 0 0 0 0 0 0 8 0 0 0 0 0 0 0
0 0 0 0 8 0 0 0 0 0 0 8 0 0 0 0 0 0 0
8 8 8 8 8 8 8 8 8 8 8 8 8 8 8 8 8 8 8
0 0 0 0 8 0 0 0 0 0 0 8 0 0 0 0 0 0 0
0 0 0 0 8 0 0 0 0 0 0 8 0 0 0 0 0 0 0
0 0 0 0 8 0 0 0 0 0 0 8 0 0 0 0 0 0 0
0 0 0 0 8 0 0 0 0 0 0 8 0 0 0 0 0 0 0
8 8 8 8 8 8 8 8 8 8 8 8 8 8 8 8 8 8 8
0 0 0 0 8 0 0 0 0 0 0 8 0 0 0 0 0 0 0
0 0 0 0 8 0 0 0 0 0 0 8 0 0 0 0 0 0 0
0 0 0 0 8 0 0 0 0 0 0 8 0 0 0 0 0 0 0
0 0 0 0 8 0 0 0 0 0 0 8 0 0 0 0 0 0 0
0 0 0 0 8 0 0 0 0 0 0 8 0 0 0 0 0 0 0
0 0 0 0 8 0 0 0 0 0 0 8 0 0 0 0 0 0 0
0 0 0 0 8 0 0 0 0 0 0 8 0 0 0 0 0 0 0
0 0 0 0 8 0 0 0 0 0 0 8 0 0 0 0 0 0 0
0 0 0 0 8 0 0 0 0 0 0 8 0 0 0 0 0 0 0
0 0 0 0 8 0 0 0 0 0 0 8 0 0 0 0 0 0 0

Output:
0 0 0 0 8 2 2 2 2 2 2 8 0 0 0 0 0 0 0
3 0 0 0 8 2 1 2 2 2 2 8 0 0 0 0 0 0 0
8 8 8 8 8 8 8 8 8 8 8 8 8 8 8 8 0 8 8
4 4 4 4 8 6 6 6 6 6 2 8 3 3 3 6 3 3 3
4 4 4 4 8 6 6 6 6 6 6 8 3 3 3 3 3 3 3
4 4 4 4 8 6 6 6 6 6 6 8 3 3 3 3 3 3 3
4 4 4 4 8 6 6 6 6 6 6 8 3 3 3 3 3 3 3
8 8 8 8 8 4 8 8 8 8 8 8 8 8 8 8 8 3 8
0 0 0 0 8 1 1 8 1 1 1 8 0 0 0 0 0 0 0
0 0 0 4 8 1 1 1 8 1 1 8 0 0 0 0 0 0 0
0 0 0 0 8 1 1 1 1 1 1 8 0 0 0 0 0 0 0
0 0 0 0 8 1 1 1 1 1 1 0 0 0 4 0 0 0 0
0 0 0 0 8 1 1 1 1 1 1 8 0 0 0 0 0 0 0
0 0 6 0 8 1 1 1 1 1 1 8 0 0 0 0 0 0 0
0 0 0 0 8 1 1 1 1 1 2 8 0 0 0 0 0 0 0
0 0 0 0 8 1 1 1 1 1 1 8 0 0 0 0 8 0 4
0 0 0 0 8 1 1 1 1 1 1 0 0 0 0 0 0 0 0
0 0 0 0 8 1 1 1 1 1 1 8 0 0 0 0 0 0 0
Example10
Input:
0 0 8 0 0 0 0 0 0 8 0 0 0 0
0 0 8 0 0 0 0 0 0 8 0 0 0 0
0 0 8 0 0 0 0 0 0 8 0 0 0 0
0 0 8 0 0 0 0 0 0 8 0 0 0 0
8 8 8 8 8 8 8 8 8 8 8 8 8 8
0 0 8 0 0 0 0 0 0 8 0 0 0 0
0 0 8 0 0 0 0 0 0 8 0 0 0 0
8 8 8 8 8 8 8 8 8 8 8 8 8 8
0 0 8 0 0 0 0 0 0 8 0 0 0 0
0 0 8 0 0 0 0 0 0 8 0 0 0 0
0 0 8 0 0 0 0 0 0 8 0 0 0 0
0 0 8 0 0 0 0 0 0 8 0 0 0 0

Output:
0 0 8 2 2 2 2 2 2 8 0 0 0 0
0 0 8 2 2 2 2 2 2 8 0 0 0 0
0 0 8 2 2 2 2 2 2 8 0 0 0 0
0 0 8 2 2 2 2 2 2 8 0 0 0 0
8 8 8 8 8 8 8 8 8 8 8 3 8 8
4 4 8 6 6 1 6 6 6 8 3 3 3 3
4 4 6 6 6 6 6 6 6 8 3 3 3 3
8 8 8 8 8 8 8 8 8 8 8 8 8 8
0 0 8 1 8 1 1 3 1 8 0 0 0 0
0 0 8 1 1 1 1 1 1 8 0 0 0 0
0 0 8 1 1 1 1 1 1 8 0 0 0 0
0 1 8 3 1 1 2 1 1 8 0 0 0 0


Example11
Input:
0 0 8 0 0 0 0 0 0 8 0 0 0 0
0 0 8 0 0 0 0 0 0 8 0 0 0 0
0 0 8 0 0 0 0 0 0 8 0 0 0 0
0 0 8 0 0 0 0 0 0 8 0 0 0 0
8 8 8 8 8 8 8 8 8 8 8 8 8 8
0 0 8 0 0 0 0 0 0 8 0 0 0 0
0 0 8 0 0 0 0 0 0 8 0 0 0 0
8 8 8 8 8 8 8 8 8 8 8 8 8 8
0 0 8 0 0 0 0 0 0 8 0 0 0 0
0 0 8 0 0 0 0 0 0 8 0 0 0 0
0 0 8 0 0 0 0 0 0 8 0 0 0 0
0 0 8 0 0 0 0 0 0 8 0 0 0 0

Output:
0 0 8 2 2 2 2 2 2 8 0 0 0 4
0 0 8 2 2 2 2 2 2 8 0 0 0 0
0 0 8 1 2 2 2 2 2 8 0 0 0 0
0 0 8 2 2 2 2 2 8 8 0 0 0 0
8 8 8 8 8 8 8 8 8 8 8 8 8 8
4 4 8 6 6 6 6 6 6 8 3 3 3 3
4 4 8 6 6 6 6 6 6 8 3 3 3 3
8 8 8 8 8 1 8 8 8 8 8 8 8 8
0 0 8 1 1 1 1 1 1 8 0 0 0 0
0 0 4 1 0 1 1 1 1 8 0 0 0 0
0 0 8 1 1 1 1 1 1 8 0 8 0 0
0 0 8 1 1 1 1 1 1 8 0 0 0 3


Example12
Input:
0 0 8 0 0 0 0 0 0 8 0 0 0 0
0 0 8 0 0 0 0 0 0 8 0 0 0 0
0 0 8 0 0 0 0 0 0 8 0 0 0 0
0 0 8 0 0 0 0 0 0 8 0 0 0 0
8 8 8 8 8 8 8 8 8 8 8 8 8 8
0 0 8 0 0 0 0 0 0 8 0 0 0 0
0 0 8 0 0 0 0 0 0 8 0 0 0 0
8 8 8 8 8 8 8 8 8 8 8 8 8 8
0 0 8 0 0 0 0 0 0 8 0 0 0 0
0 0 8 0 0 0 0 0 0 8 0 0 0 0
0 0 8 0 0 0 0 0 0 8 0 0 0 0
0 0 8 0 0 0 0 0 0 8 0 0 0 0

Output:
0 0 3 2 2 2 2 2 2 8 0 0 0 0
0 0 8 2 2 2 2 2 2 8 0 0 0 0
0 0 8 2 6 2 2 2 2 8 0 0 0 0
0 0 8 2 2 2 2 2 2 8 0 0 0 0
8 8 8 8 8 8 8 8 8 8 8 8 8 8
4 4 8 6 6 6 6 6 6 8 3 3 3 3
4 4 8 6 6 6 6 6 6 8 3 3 4 3
8 8 8 8 8 8 8 8 2 8 8 8 8 4
0 6 8 1 1 1 1 1 1 1 0 0 0 0
0 0 8 1 1 1 1 1 1 8 0 0 0 0
0 0 0 1 1 1 1 1 1 8 0 0 0 0
0 0 8 1 1 1 1 1 1 8 0 0 0 0


Example13
Input:
0 0 8 0 0 0 0 0 0 8 0 0 0 0
0 0 8 0 0 0 0 0 0 8 0 0 0 0
0 0 8 0 0 0 0 0 0 8 0 0 0 0
0 0 8 0 0 0 0 0 0 8 0 0 0 0
8 8 8 8 8 8 8 8 8 8 8 8 8 8
0 0 8 0 0 0 0 0 0 8 0 0 0 0
0 0 8 0 0 0 0 0 0 8 0 0 0 0
8 8 8 8 8 8 8 8 8 8 8 8 8 8
0 0 8 0 0 0 0 0 0 8 0 0 0 0
0 0 8 0 0 0 0 0 0 8 0 0 0 0
0 0 8 0 0 0 0 0 0 8 0 0 0 0
0 0 8 0 0 0 0 0 0 8 0 0 0 0

Output:
0 0 8 2 2 2 2 2 2 8 0 0 0 0
0 2 8 2 2 8 2 2 2 8 0 0 0 2
0 0 8 2 2 2 2 2 2 8 0 0 0 0
0 0 8 2 2 2 2 2 2 8 0 0 0 0
8 8 8 8 8 8 8 8 8 8 8 8 8 8
4 4 8 6 6 6 6 6 6 8 3 3 3 3
4 4 8 2 6 6 3 6 6 8 3 3 3 3
8 8 8 8 8 8 8 0 8 8 8 8 8 8
0 0 8 1 1 1 1 1 1 8 0 0 1 0
0 0 8 1 1 1 1 1 1 8 0 0 0 0
0 0 8 1 1 1 1 1 1 8 0 0 0 0
0 0 8 1 6 1 1 1 1 8 0 0 0 0


Example14
Input:
0 0 8 0 0 0 0 0 0 8 0 0 0 0
0 0 8 0 0 0 0 0 0 8 0 0 0 0
0 0 8 0 0 0 0 0 0 8 0 0 0 0
0 0 8 0 0 0 0 0 0 8 0 0 0 0
8 8 8 8 8 8 8 8 8 8 8 8 8 8
0 0 8 0 0 0 0 0 0 8 0 0 0 0
0 0 8 0 0 0 0 0 0 8 0 0 0 0
8 8 8 8 8 8 8 8 8 8 8 8 8 8
0 0 8 0 0 0 0 0 0 8 0 0 0 0
0 0 8 0 0 0 0 0 0 8 0 0 0 0
0 0 8 0 0 0 0 0 0 8 0 0 0 0
0 0 8 0 0 0 0 0 0 8 0 0 0 0

Output:
0 0 8 2 2 2 2 2 2 8 0 0 0 0
0 0 8 2 2 2 2 2 2 8 0 0 0 0
0 0 8 2 2 2 2 2 2 8 0 0 0 0
0 0 8 2 2 2 2 2 2 8 0 0 0 0
8 8 8 8 8 8 1 8 8 8 8 8 8 8
0 4 8 6 6 6 6 6 6 8 3 3 3 3
4 4 8 6 6 6 6 6 6 8 3 3 3 3
8 8 8 8 8 8 8 8 8 8 8 4 8 8
0 0 8 1 1 1 1 1 1 8 0 0 2 0
6 0 8 3 1 1 1 1 1 4 0 0 0 0
0 0 8 1 1 1 1 1 1 8 0 0 0 0
0 0 8 1 1 1 1 1 1 8 0 0 0 2


Example15
Input:
0 0 8 0 0 0 0 0 0 8 0 0 0 0
0 0 8 0 0 0 0 0 0 8 0 0 0 0
0 0 8 0 0 0 0 0 0 8 0 0 0 0
0 0 8 0 0 0 0 0 0 8 0 0 0 0
8 8 8 8 8 8 8 8 8 8 8 8 8 8
0 0 8 0 0 0 0 0 0 8 0 0 0 0
0 0 8 0 0 0 0 0 0 8 0 0 0 0
8 8 8 8 8 8 8 8 8 8 8 8 8 8
0 0 8 0 0 0 0 0 0 8 0 0 0 0
0 0 8 0 0 0 0 0 0 8 0 0 0 0
0 0 8 0 0 0 0 0 0 8 0 0 0 0
0 0 8 0 0 0 0 0 0 8 0 0 0 0

Output:
0 0 2 4 2 2 2 8 2 8 0 0 0 0
0 0 8 2 2 2 2 2 2 8 0 0 0 3
0 0 8 2 2 2 2 2 2 8 0 0 0 0
0 0 8 2 2 2 2 2 2 8 0 0 0 0
8 8 8 8 8 8 8 8 8 8 6 8 8 8
4 4 8 6 6 6 6 6 6 8 3 3 3 3
4 4 8 6 6 6 6 6 6 8 3 3 3 3
8 8 8 8 8 8 8 8 8 8 8 8 8 8
0 8 8 1 1 1 1 1 1 8 0 0 0 0
0 0 8 1 1 1 1 1 1 8 0 0 0 0
0 0 8 1 1 1 1 1 1 8 6 0 0 0
0 0 8 1 1 1 1 4 1 8 0 0 0 0


Example16
Input:
0 0 8 0 0 0 0 0 0 8 0 0 0 0
0 0 8 0 0 0 0 0 0 8 0 0 0 0
0 0 8 0 0 0 0 0 0 8 0 0 0 0
0 0 8 0 0 0 0 0 0 8 0 0 0 0
8 8 8 8 8 8 8 8 8 8 8 8 8 8
0 0 8 0 0 0 0 0 0 8 0 0 0 0
0 0 8 0 0 0 0 0 0 8 0 0 0 0
8 8 8 8 8 8 8 8 8 8 8 8 8 8
0 0 8 0 0 0 0 0 0 8 0 0 0 0
0 0 8 0 0 0 0 0 0 8 0 0 0 0
0 0 8 0 0 0 0 0 0 8 0 0 0 0
0 0 8 0 0 0 0 0 0 8 0 0 0 0

Output:
0 0 8 2 2 2 2 2 2 8 0 0 0 0
0 0 2 2 2 2 2 2 2 8 0 8 0 0
0 0 8 2 2 2 2 2 2 8 0 0 0 0
0 0 8 2 2 2 2 2 2 8 0 0 0 0
8 8 8 8 8 8 8 8 1 8 8 8 8 8
4 4 8 6 6 6 2 6 6 8 0 3 3 3
4 4 8 6 6 6 6 6 6 8 3 3 3 3
8 8 8 4 8 8 8 8 8 8 8 8 8 8
0 0 8 1 1 1 1 1 1 8 0 0 0 0
0 0 8 1 1 1 1 1 1 8 0 0 0 0
0 0 8 1 1 1 0 1 1 8 0 0 0 0
0 0 8 1 1 1 0 1 1 8 0 0 0 0


Example17
Input:
0 0 8 0 0 0 0 0 0 8 0 0 0 0
0 0 8 0 0 0 0 0 0 8 0 0 0 0
0 0 8 0 0 0 0 0 0 8 0 0 0 0
0 0 8 0 0 0 0 0 0 8 0 0 0 0
8 8 8 8 8 8 8 8 8 8 8 8 8 8
0 0 8 0 0 0 0 0 0 8 0 0 0 0
0 0 8 0 0 0 0 0 0 8 0 0 0 0
8 8 8 8 8 8 8 8 8 8 8 8 8 8
0 0 8 0 0 0 0 0 0 8 0 0 0 0
0 0 8 0 0 0 0 0 0 8 0 0 0 0
0 0 8 0 0 0 0 0 0 8 0 0 0 0
0 0 8 0 0 0 0 0 0 8 0 0 0 0

Output:
0 0 8 2 2 2 2 6 2 8 0 3 0 0
0 0 8 2 2 2 3 2 2 8 0 0 0 0
0 0 8 2 2 2 2 2 2 8 0 0 0 0
0 0 8 1 2 2 2 2 2 8 0 0 0 0
8 8 8 8 8 8 8 8 4 8 8 8 8 8
4 4 8 6 6 6 6 6 6 8 4 3 3 3
4 4 8 6 6 6 6 6 6 8 3 0 3 3
8 8 8 8 8 8 8 8 8 8 8 8 8 8
0 0 8 1 1 1 1 1 1 8 0 0 0 0
0 0 8 1 1 1 1 1 1 8 0 0 0 0
0 0 8 1 1 1 1 1 1 8 1 0 0 0
0 0 8 1 1 1 1 1 1 8 0 0 0 0


Example18
Input:
0 0 8 0 0 0 0 0 0 8 0 0 0 0
0 0 8 0 0 0 0 0 0 8 0 0 0 0
0 0 8 0 0 0 0 0 0 8 0 0 0 0
0 0 8 0 0 0 0 0 0 8 0 0 0 0
8 8 8 8 8 8 8 8 8 8 8 8 8 8
0 0 8 0 0 0 0 0 0 8 0 0 0 0
0 0 8 0 0 0 0 0 0 8 0 0 0 0
8 8 8 8 8 8 8 8 8 8 8 8 8 8
0 0 8 0 0 0 0 0 0 8 0 0 0 0
0 0 8 0 0 0 0 0 0 8 0 0 0 0
0 0 8 0 0 0 0 0 0 8 0 0 0 0
0 0 8 0 0 0 0 0 0 8 0 0 0 0

Output:
0 0 8 2 2 2 2 2 2 8 0 0 0 0
0 0 8 2 2 2 2 2 2 8 0 0 0 0
0 0 8 1 2 2 2 2 2 8 0 3 0 0
0 0 8 2 2 2 2 2 2 8 0 0 0 0
8 8 8 8 8 8 8 8 8 8 8 8 8 8
4 4 8 6 6 6 6 6 6 8 3 3 3 3
4 4 8 6 6 6 6 6 6 8 3 3 3 3
4 8 8 8 8 8 8 8 8 0 1 8 8 8
0 0 8 1 6 1 1 1 1 8 0 0 0 0
0 0 8 1 1 1 1 1 1 8 0 0 0 0
0 0 8 1 1 1 1 1 1 1 0 0 0 0
0 0 8 1 1 1 1 8 1 8 0 0 0 0

Below is a test input grid. 
Predict the corresponding output. 


Input:
0 0 0 8 0 0 0 0 8 0 0 0 0 0 0
0 0 0 8 0 0 0 0 8 0 0 0 0 0 0
0 0 0 8 0 0 0 0 8 0 0 0 0 0 0
0 0 0 8 0 0 0 0 8 0 0 0 0 0 0
0 0 0 8 0 0 0 0 8 0 0 0 0 0 0
0 0 0 8 0 0 0 0 8 0 0 0 0 0 0
8 8 8 8 8 8 8 8 8 8 8 8 8 8 8
0 0 0 8 0 0 0 0 8 0 0 0 0 0 0
0 0 0 8 0 0 0 0 8 0 0 0 0 0 0
0 0 0 8 0 0 0 0 8 0 0 0 0 0 0
0 0 0 8 0 0 0 0 8 0 0 0 0 0 0
0 0 0 8 0 0 0 0 8 0 0 0 0 0 0
0 0 0 8 0 0 0 0 8 0 0 0 0 0 0
8 8 8 8 8 8 8 8 8 8 8 8 8 8 8
0 0 0 8 0 0 0 0 8 0 0 0 0 0 0
0 0 0 8 0 0 0 0 8 0 0 0 0 0 0
0 0 0 8 0 0 0 0 8 0 0 0 0 0 0
\end{lstlisting}
\end{tcolorbox}

\end{paracol}
\begin{paracol}{2} 

\begin{tcolorbox}[colback=cyan!10!white, colframe=black,
    boxrule=0.6pt, arc=1mm, breakable, enhanced,
    title=\centering \textbf{9-Noisy Input Example With New Prompt}]
\scriptsize
\begin{lstlisting}
Find the common rule that maps 
an input grid to an output grid,
given the examples below. 
Note that random noise has 
been added to the input 
grids, meaning that different 
noisy input grids may map 
to the same output. 
In Examples 1 to 9, each 
example contains a noisy i
nput grid that maps to the 
same output grid. 
In Examples 10 to 18, 
each example contains 
a noisy input grid that 
maps to its respective output grid.

Example 1:

Input:
0 0 0 0 8 0 2 0 0 0 0 4 0 0 0 0 0 4 6
0 0 0 0 8 0 0 0 0 0 0 8 8 6 0 0 0 0 4
2 8 8 0 8 8 8 8 8 8 8 8 8 8 8 8 8 8 8
0 0 0 0 8 8 0 0 0 0 0 8 0 0 0 0 0 0 0
0 3 0 4 8 0 0 0 0 0 0 2 0 0 0 0 0 0 0
0 0 0 0 4 0 0 6 0 0 0 8 0 0 0 0 0 0 0
0 0 0 0 8 0 0 0 0 0 0 8 0 0 0 2 0 0 1
8 8 8 8 6 8 8 0 8 8 8 8 8 8 8 6 8 8 8
2 0 0 0 8 0 0 0 1 0 1 8 0 0 1 0 0 0 0
0 0 2 0 8 0 0 0 0 0 2 8 0 0 0 0 0 0 0
0 0 0 0 1 8 0 0 1 0 0 8 0 0 0 0 0 0 0
0 0 0 0 8 0 0 0 2 0 0 8 0 0 0 0 0 0 0
2 1 0 0 8 0 0 0 0 0 0 0 0 0 0 0 0 0 0
0 0 0 0 0 0 1 0 0 0 0 8 0 0 0 0 0 0 0
8 0 0 0 8 2 0 0 0 0 0 8 0 0 0 0 0 0 0
0 0 0 0 8 0 0 0 2 0 0 8 0 0 8 0 2 0 0
0 2 0 0 8 0 0 0 0 0 0 8 0 0 0 0 0 0 0
0 0 0 0 8 0 0 0 0 0 0 8 0 0 0 0 0 0 2

Output:
0 0 0 0 8 2 2 2 2 2 2 8 0 0 0 0 0 0 0
0 0 0 0 8 2 2 2 2 2 2 8 0 0 0 0 0 0 0
8 8 8 8 8 8 8 8 8 8 8 8 8 8 8 8 8 8 8
4 4 4 4 8 6 6 6 6 6 6 8 3 3 3 3 3 3 3
4 4 4 4 8 6 6 6 6 6 6 8 3 3 3 3 3 3 3
4 4 4 4 8 6 6 6 6 6 6 8 3 3 3 3 3 3 3
4 4 4 4 8 6 6 6 6 6 6 8 3 3 3 3 3 3 3
8 8 8 8 8 8 8 8 8 8 8 8 8 8 8 8 8 8 8
0 0 0 0 8 1 1 1 1 1 1 8 0 0 0 0 0 0 0
0 0 0 0 8 1 1 1 1 1 1 8 0 0 0 0 0 0 0
0 0 0 0 8 1 1 1 1 1 1 8 0 0 0 0 0 0 0
0 0 0 0 8 1 1 1 1 1 1 8 0 0 0 0 0 0 0
0 0 0 0 8 1 1 1 1 1 1 8 0 0 0 0 0 0 0
0 0 0 0 8 1 1 1 1 1 1 8 0 0 0 0 0 0 0
0 0 0 0 8 1 1 1 1 1 1 8 0 0 0 0 0 0 0
0 0 0 0 8 1 1 1 1 1 1 8 0 0 0 0 0 0 0
0 0 0 0 8 1 1 1 1 1 1 8 0 0 0 0 0 0 0
0 0 0 0 8 1 1 1 1 1 1 8 0 0 0 0 0 0 0


Example 2:

Input:
0 0 0 0 8 0 0 0 0 1 0 8 0 0 0 0 0 0 0
0 1 0 0 8 0 0 0 2 0 0 8 0 0 0 1 8 1 0
8 8 8 2 8 8 8 8 2 8 8 8 8 8 8 8 8 8 8
2 0 0 0 8 0 0 0 0 0 0 8 4 0 0 4 0 0 2
0 0 0 0 8 0 0 0 1 0 0 8 0 0 0 0 0 0 0
0 0 0 0 8 0 0 0 0 0 0 8 0 0 0 0 0 4 0
0 1 0 4 4 4 0 0 0 0 0 8 8 0 0 0 0 0 0
8 4 8 3 8 8 8 8 8 4 8 8 8 8 8 4 1 8 8
0 0 0 0 2 0 0 0 0 0 0 8 0 0 0 0 0 0 0
0 0 0 0 8 0 0 0 0 0 8 8 0 0 0 0 8 8 0
0 0 0 0 8 1 0 0 0 0 0 8 8 0 0 0 0 0 0
0 0 0 0 8 1 0 0 0 0 0 8 0 0 0 0 0 0 0
0 3 0 0 1 0 0 0 0 0 0 6 0 4 0 2 0 0 0
0 0 0 0 8 2 0 0 0 0 0 8 0 0 0 0 0 0 8
0 0 0 0 8 4 0 0 0 0 0 8 0 0 0 4 0 0 0
0 0 0 0 8 0 0 0 0 8 0 8 0 0 0 0 0 0 0
0 0 0 0 8 0 0 0 0 0 0 8 0 0 0 0 0 0 0
0 0 0 0 8 0 0 0 0 0 0 1 0 0 0 0 0 0 0

Output:
0 0 0 0 8 2 2 2 2 2 2 8 0 0 0 0 0 0 0
0 0 0 0 8 2 2 2 2 2 2 8 0 0 0 0 0 0 0
8 8 8 8 8 8 8 8 8 8 8 8 8 8 8 8 8 8 8
4 4 4 4 8 6 6 6 6 6 6 8 3 3 3 3 3 3 3
4 4 4 4 8 6 6 6 6 6 6 8 3 3 3 3 3 3 3
4 4 4 4 8 6 6 6 6 6 6 8 3 3 3 3 3 3 3
4 4 4 4 8 6 6 6 6 6 6 8 3 3 3 3 3 3 3
8 8 8 8 8 8 8 8 8 8 8 8 8 8 8 8 8 8 8
0 0 0 0 8 1 1 1 1 1 1 8 0 0 0 0 0 0 0
0 0 0 0 8 1 1 1 1 1 1 8 0 0 0 0 0 0 0
0 0 0 0 8 1 1 1 1 1 1 8 0 0 0 0 0 0 0
0 0 0 0 8 1 1 1 1 1 1 8 0 0 0 0 0 0 0
0 0 0 0 8 1 1 1 1 1 1 8 0 0 0 0 0 0 0
0 0 0 0 8 1 1 1 1 1 1 8 0 0 0 0 0 0 0
0 0 0 0 8 1 1 1 1 1 1 8 0 0 0 0 0 0 0
0 0 0 0 8 1 1 1 1 1 1 8 0 0 0 0 0 0 0
0 0 0 0 8 1 1 1 1 1 1 8 0 0 0 0 0 0 0
0 0 0 0 8 1 1 1 1 1 1 8 0 0 0 0 0 0 0


Example 3:

Input:
0 0 0 0 8 0 0 0 0 0 0 8 0 6 6 0 0 0 0
0 0 0 0 4 0 0 0 0 0 4 8 0 0 0 0 0 0 0
8 8 8 8 8 8 0 8 8 2 8 8 8 8 3 8 8 8 0
0 0 0 0 4 0 0 0 0 0 8 8 0 0 0 0 0 0 0
0 0 0 0 8 0 0 0 0 0 0 8 0 4 6 0 0 0 0
0 1 0 0 8 0 8 0 0 0 4 8 0 0 0 1 0 0 0
0 0 0 0 8 0 8 0 0 1 0 8 0 0 0 0 0 0 0
8 8 8 8 8 8 8 8 8 3 8 6 8 8 8 8 8 8 8
0 0 0 0 2 0 4 0 0 0 0 8 0 0 0 0 2 6 0
0 0 0 0 8 4 0 0 0 0 0 8 0 0 0 0 0 6 0
0 0 0 0 8 0 0 0 0 4 0 8 0 0 0 0 0 0 0
0 0 0 8 8 0 1 0 0 0 0 8 0 0 0 0 0 0 0
0 0 2 0 8 0 0 0 0 0 0 8 0 4 0 0 3 0 0
0 0 0 0 8 0 0 0 1 0 0 8 0 0 0 0 0 0 0
0 0 0 0 8 0 0 0 0 0 0 8 0 0 0 8 0 0 3
0 0 0 0 8 0 8 0 0 0 0 8 0 0 0 0 0 0 0
0 0 0 0 3 0 6 0 0 1 0 8 2 0 0 0 0 0 0
0 0 0 0 8 8 4 0 0 0 0 8 0 0 0 0 0 0 0

Output:
0 0 0 0 8 2 2 2 2 2 2 8 0 0 0 0 0 0 0
0 0 0 0 8 2 2 2 2 2 2 8 0 0 0 0 0 0 0
8 8 8 8 8 8 8 8 8 8 8 8 8 8 8 8 8 8 8
4 4 4 4 8 6 6 6 6 6 6 8 3 3 3 3 3 3 3
4 4 4 4 8 6 6 6 6 6 6 8 3 3 3 3 3 3 3
4 4 4 4 8 6 6 6 6 6 6 8 3 3 3 3 3 3 3
4 4 4 4 8 6 6 6 6 6 6 8 3 3 3 3 3 3 3
8 8 8 8 8 8 8 8 8 8 8 8 8 8 8 8 8 8 8
0 0 0 0 8 1 1 1 1 1 1 8 0 0 0 0 0 0 0
0 0 0 0 8 1 1 1 1 1 1 8 0 0 0 0 0 0 0
0 0 0 0 8 1 1 1 1 1 1 8 0 0 0 0 0 0 0
0 0 0 0 8 1 1 1 1 1 1 8 0 0 0 0 0 0 0
0 0 0 0 8 1 1 1 1 1 1 8 0 0 0 0 0 0 0
0 0 0 0 8 1 1 1 1 1 1 8 0 0 0 0 0 0 0
0 0 0 0 8 1 1 1 1 1 1 8 0 0 0 0 0 0 0
0 0 0 0 8 1 1 1 1 1 1 8 0 0 0 0 0 0 0
0 0 0 0 8 1 1 1 1 1 1 8 0 0 0 0 0 0 0
0 0 0 0 8 1 1 1 1 1 1 8 0 0 0 0 0 0 0


Example 4:

Input:
0 0 0 0 8 3 0 0 0 0 0 8 0 0 0 0 0 0 0
0 0 0 0 8 0 3 0 0 0 0 8 0 0 0 1 0 0 8
3 8 8 8 8 8 8 8 4 8 8 8 8 3 0 8 8 8 8
0 0 0 2 8 0 0 0 0 0 0 8 0 0 0 0 0 6 0
0 0 0 0 8 3 0 0 0 0 0 1 0 8 0 0 0 0 0
0 0 0 0 6 0 0 0 0 0 0 8 0 0 0 0 0 0 0
0 3 0 0 8 0 0 0 0 0 0 8 0 0 0 0 2 0 0
8 8 8 8 8 8 8 0 8 8 8 4 8 8 8 8 8 8 8
0 0 0 0 0 0 0 2 0 0 0 8 0 6 0 0 0 0 0
0 0 0 0 8 0 0 0 0 0 0 8 0 0 0 0 0 0 0
0 0 0 0 2 0 0 0 0 0 0 8 4 0 3 0 8 0 0
0 0 0 0 8 0 0 0 0 0 0 8 4 0 0 0 0 2 0
0 0 0 3 4 0 4 0 0 0 0 8 0 0 3 0 0 3 0
0 0 8 0 8 0 0 8 0 0 0 8 0 0 0 0 0 0 0
0 0 0 0 8 0 0 0 0 0 0 8 0 0 0 0 0 0 0
0 0 0 0 8 0 0 0 0 0 0 6 0 0 8 0 3 0 8
0 0 0 2 8 0 0 0 0 0 0 8 1 0 0 0 0 0 3
0 0 6 0 8 0 0 0 0 0 0 8 0 0 0 0 0 0 0

Output:
0 0 0 0 8 2 2 2 2 2 2 8 0 0 0 0 0 0 0
0 0 0 0 8 2 2 2 2 2 2 8 0 0 0 0 0 0 0
8 8 8 8 8 8 8 8 8 8 8 8 8 8 8 8 8 8 8
4 4 4 4 8 6 6 6 6 6 6 8 3 3 3 3 3 3 3
4 4 4 4 8 6 6 6 6 6 6 8 3 3 3 3 3 3 3
4 4 4 4 8 6 6 6 6 6 6 8 3 3 3 3 3 3 3
4 4 4 4 8 6 6 6 6 6 6 8 3 3 3 3 3 3 3
8 8 8 8 8 8 8 8 8 8 8 8 8 8 8 8 8 8 8
0 0 0 0 8 1 1 1 1 1 1 8 0 0 0 0 0 0 0
0 0 0 0 8 1 1 1 1 1 1 8 0 0 0 0 0 0 0
0 0 0 0 8 1 1 1 1 1 1 8 0 0 0 0 0 0 0
0 0 0 0 8 1 1 1 1 1 1 8 0 0 0 0 0 0 0
0 0 0 0 8 1 1 1 1 1 1 8 0 0 0 0 0 0 0
0 0 0 0 8 1 1 1 1 1 1 8 0 0 0 0 0 0 0
0 0 0 0 8 1 1 1 1 1 1 8 0 0 0 0 0 0 0
0 0 0 0 8 1 1 1 1 1 1 8 0 0 0 0 0 0 0
0 0 0 0 8 1 1 1 1 1 1 8 0 0 0 0 0 0 0
0 0 0 0 8 1 1 1 1 1 1 8 0 0 0 0 0 0 0


Example 5:

Input:
0 0 0 0 8 0 4 0 0 0 0 8 0 0 0 0 0 0 0
0 0 0 0 8 2 0 0 0 0 0 8 1 0 0 0 0 0 0
6 8 8 8 0 8 8 8 4 8 8 8 8 8 8 8 8 8 8
0 0 0 1 8 0 0 0 0 0 0 8 0 0 0 0 0 0 4
0 0 0 0 8 0 0 0 0 0 0 8 0 0 0 0 0 0 4
0 0 2 0 8 0 0 0 0 0 0 8 0 0 0 0 0 0 0
0 0 0 3 8 0 2 0 0 0 0 8 0 6 0 0 0 0 0
8 8 8 8 8 3 2 8 8 8 2 8 6 8 8 8 8 8 8
0 0 0 0 8 0 0 0 0 0 8 8 3 0 0 0 2 0 0
0 0 0 0 8 0 0 0 0 0 0 8 0 0 0 0 0 0 0
0 0 3 0 6 0 0 0 0 0 0 8 0 0 0 0 0 0 0
0 0 0 4 8 0 3 3 0 0 0 1 0 0 0 0 0 0 0
0 8 0 0 8 0 0 8 0 0 6 8 0 0 0 0 0 0 0
8 2 0 0 8 0 0 0 0 0 0 8 0 0 0 4 0 0 0
0 0 0 0 8 1 0 0 0 0 0 8 8 0 0 4 0 0 0
0 0 0 0 0 0 0 0 0 0 0 3 0 0 0 0 8 0 0
0 0 0 0 8 0 0 0 0 0 0 8 8 0 0 0 0 0 0
0 0 0 0 8 0 0 0 3 0 0 4 0 0 0 0 0 1 0

Output:
0 0 0 0 8 2 2 2 2 2 2 8 0 0 0 0 0 0 0
0 0 0 0 8 2 2 2 2 2 2 8 0 0 0 0 0 0 0
8 8 8 8 8 8 8 8 8 8 8 8 8 8 8 8 8 8 8
4 4 4 4 8 6 6 6 6 6 6 8 3 3 3 3 3 3 3
4 4 4 4 8 6 6 6 6 6 6 8 3 3 3 3 3 3 3
4 4 4 4 8 6 6 6 6 6 6 8 3 3 3 3 3 3 3
4 4 4 4 8 6 6 6 6 6 6 8 3 3 3 3 3 3 3
8 8 8 8 8 8 8 8 8 8 8 8 8 8 8 8 8 8 8
0 0 0 0 8 1 1 1 1 1 1 8 0 0 0 0 0 0 0
0 0 0 0 8 1 1 1 1 1 1 8 0 0 0 0 0 0 0
0 0 0 0 8 1 1 1 1 1 1 8 0 0 0 0 0 0 0
0 0 0 0 8 1 1 1 1 1 1 8 0 0 0 0 0 0 0
0 0 0 0 8 1 1 1 1 1 1 8 0 0 0 0 0 0 0
0 0 0 0 8 1 1 1 1 1 1 8 0 0 0 0 0 0 0
0 0 0 0 8 1 1 1 1 1 1 8 0 0 0 0 0 0 0
0 0 0 0 8 1 1 1 1 1 1 8 0 0 0 0 0 0 0
0 0 0 0 8 1 1 1 1 1 1 8 0 0 0 0 0 0 0
0 0 0 0 8 1 1 1 1 1 1 8 0 0 0 0 0 0 0


Example 6:

Input:
0 0 0 0 0 0 0 0 0 0 0 8 0 0 0 4 0 0 0
0 0 0 0 8 0 0 6 0 0 0 8 0 1 0 0 0 0 0
8 8 8 8 8 8 8 8 8 8 8 8 0 8 4 8 8 8 8
0 0 0 0 8 0 0 0 0 0 0 3 0 0 0 6 0 0 0
1 0 0 8 8 0 0 0 0 0 0 8 0 0 0 0 0 0 0
0 0 0 0 8 0 0 3 0 8 0 8 0 0 0 0 0 0 0
0 0 6 8 8 0 0 0 6 0 0 8 0 0 0 0 0 0 0
8 8 8 3 8 8 8 8 8 8 6 8 8 8 8 8 8 8 8
0 0 0 0 8 8 0 0 0 0 0 8 8 6 0 0 0 0 4
0 2 0 0 8 0 0 8 6 0 0 8 0 0 0 0 0 0 0
0 0 0 0 8 0 2 0 0 3 0 8 0 4 0 0 0 0 0
0 0 0 0 8 4 0 0 0 0 0 8 0 0 0 0 0 0 0
8 0 3 0 3 0 0 0 0 0 0 8 0 0 0 6 0 0 0
0 0 6 0 8 0 0 0 0 0 0 8 0 0 0 0 0 0 0
0 0 0 0 8 0 0 8 0 0 0 8 6 0 0 0 0 0 0
0 0 0 0 8 0 4 0 0 0 0 8 0 1 0 0 1 0 0
0 2 0 0 8 0 0 3 0 0 2 8 0 0 0 0 0 0 0
0 0 0 0 8 0 0 0 0 0 0 8 0 0 0 6 0 0 0

Output:
0 0 0 0 8 2 2 2 2 2 2 8 0 0 0 0 0 0 0
0 0 0 0 8 2 2 2 2 2 2 8 0 0 0 0 0 0 0
8 8 8 8 8 8 8 8 8 8 8 8 8 8 8 8 8 8 8
4 4 4 4 8 6 6 6 6 6 6 8 3 3 3 3 3 3 3
4 4 4 4 8 6 6 6 6 6 6 8 3 3 3 3 3 3 3
4 4 4 4 8 6 6 6 6 6 6 8 3 3 3 3 3 3 3
4 4 4 4 8 6 6 6 6 6 6 8 3 3 3 3 3 3 3
8 8 8 8 8 8 8 8 8 8 8 8 8 8 8 8 8 8 8
0 0 0 0 8 1 1 1 1 1 1 8 0 0 0 0 0 0 0
0 0 0 0 8 1 1 1 1 1 1 8 0 0 0 0 0 0 0
0 0 0 0 8 1 1 1 1 1 1 8 0 0 0 0 0 0 0
0 0 0 0 8 1 1 1 1 1 1 8 0 0 0 0 0 0 0
0 0 0 0 8 1 1 1 1 1 1 8 0 0 0 0 0 0 0
0 0 0 0 8 1 1 1 1 1 1 8 0 0 0 0 0 0 0
0 0 0 0 8 1 1 1 1 1 1 8 0 0 0 0 0 0 0
0 0 0 0 8 1 1 1 1 1 1 8 0 0 0 0 0 0 0
0 0 0 0 8 1 1 1 1 1 1 8 0 0 0 0 0 0 0
0 0 0 0 8 1 1 1 1 1 1 8 0 0 0 0 0 0 0


Example 7:

Input:
0 8 0 0 8 0 0 0 0 0 0 1 3 0 0 0 0 0 0
0 0 0 0 8 1 0 0 1 0 0 1 0 0 0 0 0 0 0
8 8 8 8 8 8 8 8 8 8 8 8 8 8 8 8 8 8 8
0 0 0 0 8 0 0 0 0 0 0 8 0 0 0 0 0 0 0
2 0 0 0 8 0 0 1 0 0 3 6 0 0 0 0 0 0 0
0 0 6 0 8 0 0 0 0 0 0 8 0 0 2 0 0 0 0
0 1 0 0 8 0 0 0 0 0 0 4 0 0 0 0 0 0 0
8 8 8 8 3 8 8 0 8 8 8 8 8 8 8 8 2 0 8
0 0 0 6 8 0 0 0 0 0 0 8 0 0 0 0 0 0 0
0 0 0 0 8 0 0 0 0 0 0 8 0 0 0 0 0 0 0
0 0 0 0 8 0 0 0 0 2 0 8 0 0 0 0 0 0 0
0 0 8 0 8 0 0 1 0 0 0 8 2 0 0 0 0 0 0
0 0 0 0 8 0 0 0 0 0 0 8 2 0 0 0 0 0 0
0 0 1 0 1 2 0 0 0 0 4 8 0 0 0 0 0 0 0
0 6 0 0 8 0 0 0 0 0 0 8 0 0 6 0 0 0 8
0 3 0 6 8 0 0 0 0 0 0 8 6 0 0 0 0 1 0
0 0 0 0 8 0 0 0 0 6 3 8 0 0 0 0 2 0 0
0 0 0 6 8 0 0 0 0 4 0 8 0 0 0 0 8 0 3

Output:
0 0 0 0 8 2 2 2 2 2 2 8 0 0 0 0 0 0 0
0 0 0 0 8 2 2 2 2 2 2 8 0 0 0 0 0 0 0
8 8 8 8 8 8 8 8 8 8 8 8 8 8 8 8 8 8 8
4 4 4 4 8 6 6 6 6 6 6 8 3 3 3 3 3 3 3
4 4 4 4 8 6 6 6 6 6 6 8 3 3 3 3 3 3 3
4 4 4 4 8 6 6 6 6 6 6 8 3 3 3 3 3 3 3
4 4 4 4 8 6 6 6 6 6 6 8 3 3 3 3 3 3 3
8 8 8 8 8 8 8 8 8 8 8 8 8 8 8 8 8 8 8
0 0 0 0 8 1 1 1 1 1 1 8 0 0 0 0 0 0 0
0 0 0 0 8 1 1 1 1 1 1 8 0 0 0 0 0 0 0
0 0 0 0 8 1 1 1 1 1 1 8 0 0 0 0 0 0 0
0 0 0 0 8 1 1 1 1 1 1 8 0 0 0 0 0 0 0
0 0 0 0 8 1 1 1 1 1 1 8 0 0 0 0 0 0 0
0 0 0 0 8 1 1 1 1 1 1 8 0 0 0 0 0 0 0
0 0 0 0 8 1 1 1 1 1 1 8 0 0 0 0 0 0 0
0 0 0 0 8 1 1 1 1 1 1 8 0 0 0 0 0 0 0
0 0 0 0 8 1 1 1 1 1 1 8 0 0 0 0 0 0 0
0 0 0 0 8 1 1 1 1 1 1 8 0 0 0 0 0 0 0


Example 8:

Input:
0 0 0 0 8 0 0 0 0 0 0 3 0 0 0 0 0 3 0
0 0 0 0 8 0 0 0 0 0 0 8 0 1 0 0 0 0 0
8 8 8 8 2 8 8 8 8 8 8 2 3 8 8 3 1 8 8
0 0 0 0 8 0 0 0 0 0 0 0 0 0 0 0 6 0 0
0 0 0 0 8 0 0 0 0 0 0 8 0 0 0 0 0 0 0
0 4 2 0 8 0 8 0 0 0 0 8 0 0 0 0 0 0 0
0 2 0 0 8 0 0 0 0 0 0 8 0 0 0 0 0 0 1
8 8 8 8 8 8 1 8 8 8 6 8 8 8 8 8 8 8 8
0 0 0 0 8 0 0 0 0 0 0 8 0 0 0 0 0 0 0
4 0 0 0 8 0 8 0 3 0 0 8 6 0 0 0 0 0 0
0 0 0 0 8 0 3 0 0 0 1 8 0 0 0 0 2 4 0
2 0 0 0 8 0 0 0 0 0 0 3 0 0 3 0 0 0 0
2 0 0 0 8 0 0 0 0 0 0 8 0 0 0 2 1 3 0
0 0 0 0 8 0 0 2 0 0 0 8 4 0 0 0 0 0 0
0 0 0 4 8 0 0 0 0 0 0 8 0 0 0 3 0 0 0
4 0 0 0 8 0 0 0 0 4 0 8 1 0 0 0 0 0 0
8 0 0 0 6 0 0 0 0 0 0 8 0 0 0 0 0 0 0
0 0 0 0 8 0 0 0 0 0 0 2 0 0 0 0 0 0 0

Output:
0 0 0 0 8 2 2 2 2 2 2 8 0 0 0 0 0 0 0
0 0 0 0 8 2 2 2 2 2 2 8 0 0 0 0 0 0 0
8 8 8 8 8 8 8 8 8 8 8 8 8 8 8 8 8 8 8
4 4 4 4 8 6 6 6 6 6 6 8 3 3 3 3 3 3 3
4 4 4 4 8 6 6 6 6 6 6 8 3 3 3 3 3 3 3
4 4 4 4 8 6 6 6 6 6 6 8 3 3 3 3 3 3 3
4 4 4 4 8 6 6 6 6 6 6 8 3 3 3 3 3 3 3
8 8 8 8 8 8 8 8 8 8 8 8 8 8 8 8 8 8 8
0 0 0 0 8 1 1 1 1 1 1 8 0 0 0 0 0 0 0
0 0 0 0 8 1 1 1 1 1 1 8 0 0 0 0 0 0 0
0 0 0 0 8 1 1 1 1 1 1 8 0 0 0 0 0 0 0
0 0 0 0 8 1 1 1 1 1 1 8 0 0 0 0 0 0 0
0 0 0 0 8 1 1 1 1 1 1 8 0 0 0 0 0 0 0
0 0 0 0 8 1 1 1 1 1 1 8 0 0 0 0 0 0 0
0 0 0 0 8 1 1 1 1 1 1 8 0 0 0 0 0 0 0
0 0 0 0 8 1 1 1 1 1 1 8 0 0 0 0 0 0 0
0 0 0 0 8 1 1 1 1 1 1 8 0 0 0 0 0 0 0
0 0 0 0 8 1 1 1 1 1 1 8 0 0 0 0 0 0 0


Example 9:

Input:
0 0 0 0 8 0 0 0 0 0 0 2 0 0 0 0 4 8 0
0 0 0 0 8 0 0 0 0 0 0 8 0 0 0 0 0 0 0
8 0 8 8 8 8 8 6 6 8 8 8 8 8 8 8 8 8 8
0 0 6 0 8 0 0 0 0 0 3 6 0 0 0 2 0 0 0
0 0 0 0 8 0 0 0 0 0 1 2 0 0 3 0 6 0 0
0 4 0 0 8 0 0 0 0 0 0 8 3 0 0 0 0 0 0
0 0 0 0 1 0 8 0 0 0 0 8 0 0 0 0 0 0 0
8 8 8 8 8 8 8 8 8 8 8 8 8 8 8 8 4 8 8
0 6 4 2 8 0 0 0 8 0 0 8 0 0 0 0 0 0 0
3 0 0 0 8 0 0 0 0 0 0 8 0 0 0 0 0 0 0
0 0 0 0 8 0 0 8 0 0 6 8 0 0 0 0 0 0 6
3 0 3 0 8 0 0 0 0 0 0 8 0 0 0 0 0 0 0
0 0 0 0 1 0 0 0 6 0 0 8 2 0 0 0 0 0 0
0 0 0 3 8 0 0 0 0 0 0 8 0 0 0 0 8 0 0
2 0 0 0 8 0 0 0 0 0 0 8 0 1 0 0 3 0 0
0 0 0 0 8 0 0 0 0 0 0 4 0 0 0 0 0 0 0
0 0 0 0 3 0 0 0 0 0 0 8 0 0 0 0 0 0 3
0 0 0 0 8 0 0 0 0 0 1 8 0 0 0 1 0 0 0

Output:
0 0 0 0 8 2 2 2 2 2 2 8 0 0 0 0 0 0 0
0 0 0 0 8 2 2 2 2 2 2 8 0 0 0 0 0 0 0
8 8 8 8 8 8 8 8 8 8 8 8 8 8 8 8 8 8 8
4 4 4 4 8 6 6 6 6 6 6 8 3 3 3 3 3 3 3
4 4 4 4 8 6 6 6 6 6 6 8 3 3 3 3 3 3 3
4 4 4 4 8 6 6 6 6 6 6 8 3 3 3 3 3 3 3
4 4 4 4 8 6 6 6 6 6 6 8 3 3 3 3 3 3 3
8 8 8 8 8 8 8 8 8 8 8 8 8 8 8 8 8 8 8
0 0 0 0 8 1 1 1 1 1 1 8 0 0 0 0 0 0 0
0 0 0 0 8 1 1 1 1 1 1 8 0 0 0 0 0 0 0
0 0 0 0 8 1 1 1 1 1 1 8 0 0 0 0 0 0 0
0 0 0 0 8 1 1 1 1 1 1 8 0 0 0 0 0 0 0
0 0 0 0 8 1 1 1 1 1 1 8 0 0 0 0 0 0 0
0 0 0 0 8 1 1 1 1 1 1 8 0 0 0 0 0 0 0
0 0 0 0 8 1 1 1 1 1 1 8 0 0 0 0 0 0 0
0 0 0 0 8 1 1 1 1 1 1 8 0 0 0 0 0 0 0
0 0 0 0 8 1 1 1 1 1 1 8 0 0 0 0 0 0 0
0 0 0 0 8 1 1 1 1 1 1 8 0 0 0 0 0 0 0


Example 10:

Input:
0 0 8 0 0 0 0 0 0 8 0 0 0 0
0 0 3 0 2 0 0 0 0 8 0 0 0 0
0 0 8 0 0 0 0 0 0 2 2 8 0 0
0 0 8 0 0 0 0 0 0 8 0 0 0 0
8 8 8 8 8 6 8 8 8 8 8 8 8 8
0 0 8 0 8 0 0 0 4 8 0 0 0 0
0 0 8 0 0 2 0 0 0 8 0 0 1 0
8 8 8 8 4 8 8 8 8 8 8 8 8 8
0 0 8 0 2 0 6 0 0 8 0 0 0 0
0 0 8 0 0 1 0 0 0 8 0 0 0 0
0 0 8 3 0 0 0 0 0 8 0 4 0 6
0 0 8 3 0 3 6 0 0 8 0 4 0 0

Output:
0 0 8 2 2 2 2 2 2 8 0 0 0 0
0 0 8 2 2 2 2 2 2 8 0 0 0 0
0 0 8 2 2 2 2 2 2 8 0 0 0 0
0 0 8 2 2 2 2 2 2 8 0 0 0 0
8 8 8 8 8 8 8 8 8 8 8 8 8 8
4 4 8 6 6 6 6 6 6 8 3 3 3 3
4 4 8 6 6 6 6 6 6 8 3 3 3 3
8 8 8 8 8 8 8 8 8 8 8 8 8 8
0 0 8 1 1 1 1 1 1 8 0 0 0 0
0 0 8 1 1 1 1 1 1 8 0 0 0 0
0 0 8 1 1 1 1 1 1 8 0 0 0 0
0 0 8 1 1 1 1 1 1 8 0 0 0 0


Example 11:

Input:
0 0 8 0 0 0 0 2 0 8 6 0 0 0
0 0 8 0 0 4 0 0 0 8 0 2 0 0
0 0 8 0 0 0 2 0 0 8 0 0 0 0
0 6 8 1 0 0 0 0 0 8 0 0 0 0
8 8 8 8 8 8 4 8 8 8 8 8 8 8
0 0 8 0 0 0 0 0 0 8 0 4 0 0
0 0 8 0 0 0 0 0 4 8 0 4 1 3
8 8 8 8 8 8 8 8 8 8 8 3 8 8
0 0 8 6 0 0 0 6 0 2 0 0 0 0
0 0 8 0 0 0 0 4 0 8 1 0 0 0
0 0 8 0 0 0 0 0 0 8 8 0 8 0
0 0 8 0 0 0 0 0 0 8 0 0 0 0

Output:
0 0 8 2 2 2 2 2 2 8 0 0 0 0
0 0 8 2 2 2 2 2 2 8 0 0 0 0
0 0 8 2 2 2 2 2 2 8 0 0 0 0
0 0 8 2 2 2 2 2 2 8 0 0 0 0
8 8 8 8 8 8 8 8 8 8 8 8 8 8
4 4 8 6 6 6 6 6 6 8 3 3 3 3
4 4 8 6 6 6 6 6 6 8 3 3 3 3
8 8 8 8 8 8 8 8 8 8 8 8 8 8
0 0 8 1 1 1 1 1 1 8 0 0 0 0
0 0 8 1 1 1 1 1 1 8 0 0 0 0
0 0 8 1 1 1 1 1 1 8 0 0 0 0
0 0 8 1 1 1 1 1 1 8 0 0 0 0


Example 12:

Input:
0 0 8 0 0 2 0 6 0 8 0 1 0 0
0 0 8 0 0 0 0 0 0 8 0 0 0 3
0 0 8 0 0 3 0 0 0 8 0 0 0 0
0 0 8 0 0 0 0 0 0 8 0 4 0 0
3 8 8 8 8 8 8 3 8 8 3 6 8 8
0 0 8 0 0 0 0 0 0 8 1 0 2 0
0 0 1 0 0 0 0 0 0 1 0 0 0 0
8 8 8 8 8 8 8 8 8 8 8 8 8 8
0 0 0 8 0 0 0 0 0 8 0 0 0 0
0 0 8 0 0 3 0 0 0 8 0 2 0 0
0 0 8 0 0 4 0 0 2 0 0 0 0 0
0 0 8 0 0 0 0 0 0 8 0 0 0 0

Output:
0 0 8 2 2 2 2 2 2 8 0 0 0 0
0 0 8 2 2 2 2 2 2 8 0 0 0 0
0 0 8 2 2 2 2 2 2 8 0 0 0 0
0 0 8 2 2 2 2 2 2 8 0 0 0 0
8 8 8 8 8 8 8 8 8 8 8 8 8 8
4 4 8 6 6 6 6 6 6 8 3 3 3 3
4 4 8 6 6 6 6 6 6 8 3 3 3 3
8 8 8 8 8 8 8 8 8 8 8 8 8 8
0 0 8 1 1 1 1 1 1 8 0 0 0 0
0 0 8 1 1 1 1 1 1 8 0 0 0 0
0 0 8 1 1 1 1 1 1 8 0 0 0 0
0 0 8 1 1 1 1 1 1 8 0 0 0 0


Example 13:

Input:
0 0 8 0 0 0 0 3 4 8 0 0 0 1
0 0 8 0 0 0 0 2 0 8 0 0 0 0
0 3 8 0 0 0 0 0 0 8 0 6 0 0
0 0 8 0 4 0 0 0 0 8 0 3 0 1
8 8 8 8 8 2 8 8 8 8 8 8 8 8
0 0 8 0 0 0 0 0 0 8 0 0 1 0
0 0 8 0 3 0 0 0 4 8 0 8 0 0
8 8 8 8 8 3 8 8 8 8 8 8 8 8
0 0 8 8 0 0 0 0 0 8 4 0 0 0
0 0 8 0 0 0 0 0 0 8 1 6 0 0
0 0 2 0 0 0 0 0 0 8 0 0 0 0
0 0 8 0 4 0 0 0 0 8 0 0 0 0

Output:
0 0 8 2 2 2 2 2 2 8 0 0 0 0
0 0 8 2 2 2 2 2 2 8 0 0 0 0
0 0 8 2 2 2 2 2 2 8 0 0 0 0
0 0 8 2 2 2 2 2 2 8 0 0 0 0
8 8 8 8 8 8 8 8 8 8 8 8 8 8
4 4 8 6 6 6 6 6 6 8 3 3 3 3
4 4 8 6 6 6 6 6 6 8 3 3 3 3
8 8 8 8 8 8 8 8 8 8 8 8 8 8
0 0 8 1 1 1 1 1 1 8 0 0 0 0
0 0 8 1 1 1 1 1 1 8 0 0 0 0
0 0 8 1 1 1 1 1 1 8 0 0 0 0
0 0 8 1 1 1 1 1 1 8 0 0 0 0


Example 14:

Input:
0 0 8 0 0 0 0 0 0 8 0 0 2 1
0 0 8 0 0 0 0 0 0 8 0 0 0 0
0 4 8 0 3 0 8 0 0 8 0 8 0 0
0 0 8 0 0 0 0 0 3 8 0 0 0 0
8 8 8 8 8 8 8 8 8 8 8 4 8 8
6 0 8 0 0 3 0 0 0 8 0 0 0 3
0 1 8 0 6 0 0 0 1 8 0 0 0 0
8 8 8 0 8 8 8 8 8 8 8 8 8 8
0 0 8 2 0 0 0 0 0 2 0 0 0 0
0 0 8 0 0 0 0 0 0 8 0 0 0 0
0 0 8 0 0 0 0 0 3 1 0 0 0 0
0 4 8 0 0 0 0 0 0 8 0 0 8 0

Output:
0 0 8 2 2 2 2 2 2 8 0 0 0 0
0 0 8 2 2 2 2 2 2 8 0 0 0 0
0 0 8 2 2 2 2 2 2 8 0 0 0 0
0 0 8 2 2 2 2 2 2 8 0 0 0 0
8 8 8 8 8 8 8 8 8 8 8 8 8 8
4 4 8 6 6 6 6 6 6 8 3 3 3 3
4 4 8 6 6 6 6 6 6 8 3 3 3 3
8 8 8 8 8 8 8 8 8 8 8 8 8 8
0 0 8 1 1 1 1 1 1 8 0 0 0 0
0 0 8 1 1 1 1 1 1 8 0 0 0 0
0 0 8 1 1 1 1 1 1 8 0 0 0 0
0 0 8 1 1 1 1 1 1 8 0 0 0 0


Example 15:

Input:
0 0 8 0 0 0 0 0 0 1 0 0 0 0
0 0 8 0 0 0 0 4 0 8 0 0 0 0
0 0 8 0 0 1 0 0 0 8 0 0 2 0
0 8 8 1 0 4 0 0 0 4 0 0 0 0
8 8 8 8 8 8 8 8 8 8 8 0 8 8
0 0 8 0 0 2 0 0 0 8 0 0 0 0
8 0 8 0 0 0 0 0 0 8 0 0 0 0
8 8 8 8 8 8 8 8 8 8 4 8 8 4
0 0 8 0 0 0 0 0 0 8 0 0 0 1
0 6 8 3 0 0 0 0 0 8 0 0 0 0
4 0 8 0 0 0 0 0 0 8 0 0 0 0
0 6 8 8 3 0 0 0 0 8 0 0 8 0

Output:
0 0 8 2 2 2 2 2 2 8 0 0 0 0
0 0 8 2 2 2 2 2 2 8 0 0 0 0
0 0 8 2 2 2 2 2 2 8 0 0 0 0
0 0 8 2 2 2 2 2 2 8 0 0 0 0
8 8 8 8 8 8 8 8 8 8 8 8 8 8
4 4 8 6 6 6 6 6 6 8 3 3 3 3
4 4 8 6 6 6 6 6 6 8 3 3 3 3
8 8 8 8 8 8 8 8 8 8 8 8 8 8
0 0 8 1 1 1 1 1 1 8 0 0 0 0
0 0 8 1 1 1 1 1 1 8 0 0 0 0
0 0 8 1 1 1 1 1 1 8 0 0 0 0
0 0 8 1 1 1 1 1 1 8 0 0 0 0


Example 16:

Input:
0 0 8 0 0 0 0 0 0 8 0 1 0 0
0 0 0 0 0 0 0 0 0 8 0 0 0 0
0 0 0 0 0 0 0 0 0 8 0 0 0 0
0 0 8 0 0 8 0 0 0 8 0 0 0 0
8 8 8 8 8 4 8 8 8 8 8 8 8 8
0 0 8 1 0 0 4 0 0 8 0 0 0 0
0 0 8 1 0 0 0 0 0 8 0 8 0 0
1 8 8 8 0 8 0 8 8 8 8 8 8 8
6 0 8 3 0 0 0 0 0 8 0 0 0 3
0 0 8 0 0 0 0 0 0 8 0 0 0 0
3 0 8 0 0 0 0 0 0 0 0 1 0 6
0 0 8 0 0 0 0 8 0 6 0 0 0 0

Output:
0 0 8 2 2 2 2 2 2 8 0 0 0 0
0 0 8 2 2 2 2 2 2 8 0 0 0 0
0 0 8 2 2 2 2 2 2 8 0 0 0 0
0 0 8 2 2 2 2 2 2 8 0 0 0 0
8 8 8 8 8 8 8 8 8 8 8 8 8 8
4 4 8 6 6 6 6 6 6 8 3 3 3 3
4 4 8 6 6 6 6 6 6 8 3 3 3 3
8 8 8 8 8 8 8 8 8 8 8 8 8 8
0 0 8 1 1 1 1 1 1 8 0 0 0 0
0 0 8 1 1 1 1 1 1 8 0 0 0 0
0 0 8 1 1 1 1 1 1 8 0 0 0 0
0 0 8 1 1 1 1 1 1 8 0 0 0 0


Example 17:

Input:
0 0 8 0 0 0 0 0 0 8 0 0 8 0
4 1 8 0 0 0 0 4 0 8 3 8 0 0
0 1 8 0 0 0 0 0 0 8 0 0 0 0
0 0 8 0 8 0 0 0 0 8 0 0 0 0
8 8 0 8 8 8 8 8 4 6 8 8 0 8
0 0 8 0 0 0 0 0 0 6 0 0 0 0
0 0 8 0 0 0 6 1 0 8 0 0 0 0
8 8 8 8 8 8 8 8 8 8 8 8 1 8
0 0 8 0 6 0 0 0 0 8 8 2 0 0
0 0 8 0 0 0 0 0 0 8 0 0 2 0
0 0 8 0 0 0 0 0 0 8 0 0 0 0
2 0 8 0 0 0 0 0 0 8 0 0 0 0

Output:
0 0 8 2 2 2 2 2 2 8 0 0 0 0
0 0 8 2 2 2 2 2 2 8 0 0 0 0
0 0 8 2 2 2 2 2 2 8 0 0 0 0
0 0 8 2 2 2 2 2 2 8 0 0 0 0
8 8 8 8 8 8 8 8 8 8 8 8 8 8
4 4 8 6 6 6 6 6 6 8 3 3 3 3
4 4 8 6 6 6 6 6 6 8 3 3 3 3
8 8 8 8 8 8 8 8 8 8 8 8 8 8
0 0 8 1 1 1 1 1 1 8 0 0 0 0
0 0 8 1 1 1 1 1 1 8 0 0 0 0
0 0 8 1 1 1 1 1 1 8 0 0 0 0
0 0 8 1 1 1 1 1 1 8 0 0 0 0


Example 18:

Input:
0 0 8 0 0 0 0 0 0 8 0 0 0 0
0 0 8 0 0 0 0 0 0 8 0 0 0 0
6 0 8 0 0 0 0 0 0 8 0 0 0 0
3 0 4 3 0 0 0 0 0 8 0 8 0 0
2 8 8 4 4 8 8 8 8 8 8 8 8 8
0 0 8 0 0 0 0 0 0 4 2 0 0 0
2 0 8 0 0 6 8 0 0 8 0 0 0 0
8 8 8 8 8 4 8 8 8 8 8 8 3 8
0 0 8 2 0 0 0 0 0 8 0 0 0 0
0 0 8 0 0 0 0 0 2 8 0 0 0 1
0 0 8 0 0 2 0 0 0 8 0 0 0 0
0 0 4 0 0 0 0 0 0 8 0 0 0 1

Output:
0 0 8 2 2 2 2 2 2 8 0 0 0 0
0 0 8 2 2 2 2 2 2 8 0 0 0 0
0 0 8 2 2 2 2 2 2 8 0 0 0 0
0 0 8 2 2 2 2 2 2 8 0 0 0 0
8 8 8 8 8 8 8 8 8 8 8 8 8 8
4 4 8 6 6 6 6 6 6 8 3 3 3 3
4 4 8 6 6 6 6 6 6 8 3 3 3 3
8 8 8 8 8 8 8 8 8 8 8 8 8 8
0 0 8 1 1 1 1 1 1 8 0 0 0 0
0 0 8 1 1 1 1 1 1 8 0 0 0 0
0 0 8 1 1 1 1 1 1 8 0 0 0 0
0 0 8 1 1 1 1 1 1 8 0 0 0 0


Below is a test input grid. 
Predict the corresponding output. 

Input:
0 0 0 8 0 0 0 0 8 0 0 0 0 0 0
0 0 0 8 0 0 0 0 8 0 0 0 0 0 0
0 0 0 8 0 0 0 0 8 0 0 0 0 0 0
0 0 0 8 0 0 0 0 8 0 0 0 0 0 0
0 0 0 8 0 0 0 0 8 0 0 0 0 0 0
0 0 0 8 0 0 0 0 8 0 0 0 0 0 0
8 8 8 8 8 8 8 8 8 8 8 8 8 8 8
0 0 0 8 0 0 0 0 8 0 0 0 0 0 0
0 0 0 8 0 0 0 0 8 0 0 0 0 0 0
0 0 0 8 0 0 0 0 8 0 0 0 0 0 0
0 0 0 8 0 0 0 0 8 0 0 0 0 0 0
0 0 0 8 0 0 0 0 8 0 0 0 0 0 0
0 0 0 8 0 0 0 0 8 0 0 0 0 0 0
8 8 8 8 8 8 8 8 8 8 8 8 8 8 8
0 0 0 8 0 0 0 0 8 0 0 0 0 0 0
0 0 0 8 0 0 0 0 8 0 0 0 0 0 0
0 0 0 8 0 0 0 0 8 0 0 0 0 0 0
\end{lstlisting}
\end{tcolorbox}

\switchcolumn 

\begin{tcolorbox}[colback=cyan!10!white, colframe=black,
    boxrule=0.6pt, arc=1mm, breakable, enhanced,
    title=\centering \textbf{9-Noisy Output Example With New Prompt}]
\scriptsize
\begin{lstlisting}
Find the common rule that maps an 
input grid to an output grid, 
given the examples below. 
Note that random noise has 
been added to the output 
grids, meaning that different noisy 
output grids may map 
to the same input. 
In Examples 1 to 9, each example 
contains 
a noisy output grid that maps to 
the same input grid. In Examples 
10 to 18, each example 
contains a noisy output grid 
that maps to its respective input grid.

Example1
Input:
0 0 0 0 8 0 0 0 0 0 0 8 0 0 0 0 0 0 0
0 0 0 0 8 0 0 0 0 0 0 8 0 0 0 0 0 0 0
8 8 8 8 8 8 8 8 8 8 8 8 8 8 8 8 8 8 8
0 0 0 0 8 0 0 0 0 0 0 8 0 0 0 0 0 0 0
0 0 0 0 8 0 0 0 0 0 0 8 0 0 0 0 0 0 0
0 0 0 0 8 0 0 0 0 0 0 8 0 0 0 0 0 0 0
0 0 0 0 8 0 0 0 0 0 0 8 0 0 0 0 0 0 0
8 8 8 8 8 8 8 8 8 8 8 8 8 8 8 8 8 8 8
0 0 0 0 8 0 0 0 0 0 0 8 0 0 0 0 0 0 0
0 0 0 0 8 0 0 0 0 0 0 8 0 0 0 0 0 0 0
0 0 0 0 8 0 0 0 0 0 0 8 0 0 0 0 0 0 0
0 0 0 0 8 0 0 0 0 0 0 8 0 0 0 0 0 0 0
0 0 0 0 8 0 0 0 0 0 0 8 0 0 0 0 0 0 0
0 0 0 0 8 0 0 0 0 0 0 8 0 0 0 0 0 0 0
0 0 0 0 8 0 0 0 0 0 0 8 0 0 0 0 0 0 0
0 0 0 0 8 0 0 0 0 0 0 8 0 0 0 0 0 0 0
0 0 0 0 8 0 0 0 0 0 0 8 0 0 0 0 0 0 0
0 0 0 0 8 0 0 0 0 0 0 8 0 0 0 0 0 0 0

Output:
0 0 0 0 1 2 2 2 2 2 2 8 0 0 0 0 0 0 1
0 1 0 0 0 2 2 2 2 2 2 8 4 0 0 0 0 4 0
8 8 8 8 8 8 8 8 8 8 8 8 1 8 8 8 8 8 8
4 4 4 4 8 6 6 6 6 0 6 8 3 3 3 3 3 3 3
4 4 0 8 8 6 6 6 1 2 1 8 3 3 3 6 3 3 3
8 4 4 4 8 6 6 6 6 6 6 8 3 3 3 3 3 3 3
4 4 4 4 8 6 3 6 6 2 6 8 8 3 3 3 3 3 3
8 8 8 8 8 8 8 8 8 0 1 8 8 8 8 8 8 8 8
0 0 0 8 8 1 1 1 1 1 1 8 3 0 0 0 0 0 0
0 3 4 0 8 1 4 1 1 1 1 0 0 3 0 0 2 0 0
0 0 0 0 8 1 1 1 1 1 1 0 0 0 0 0 0 0 0
3 0 0 0 8 1 1 1 1 1 1 8 0 0 0 0 0 0 0
0 0 0 0 8 1 1 1 1 1 1 8 0 0 0 0 0 0 0
0 1 0 0 8 1 1 1 1 1 1 8 0 8 3 0 0 0 0
0 6 0 0 8 1 1 0 1 1 1 8 0 0 0 0 0 0 0
0 8 0 0 8 1 1 1 1 1 1 8 0 0 0 0 0 0 0
0 0 0 4 8 1 1 4 1 1 1 8 0 0 3 0 0 0 0
0 0 8 0 8 8 1 1 1 1 1 8 0 0 0 3 0 0 0


Example2
Input:
0 0 0 0 8 0 0 0 0 0 0 8 0 0 0 0 0 0 0
0 0 0 0 8 0 0 0 0 0 0 8 0 0 0 0 0 0 0
8 8 8 8 8 8 8 8 8 8 8 8 8 8 8 8 8 8 8
0 0 0 0 8 0 0 0 0 0 0 8 0 0 0 0 0 0 0
0 0 0 0 8 0 0 0 0 0 0 8 0 0 0 0 0 0 0
0 0 0 0 8 0 0 0 0 0 0 8 0 0 0 0 0 0 0
0 0 0 0 8 0 0 0 0 0 0 8 0 0 0 0 0 0 0
8 8 8 8 8 8 8 8 8 8 8 8 8 8 8 8 8 8 8
0 0 0 0 8 0 0 0 0 0 0 8 0 0 0 0 0 0 0
0 0 0 0 8 0 0 0 0 0 0 8 0 0 0 0 0 0 0
0 0 0 0 8 0 0 0 0 0 0 8 0 0 0 0 0 0 0
0 0 0 0 8 0 0 0 0 0 0 8 0 0 0 0 0 0 0
0 0 0 0 8 0 0 0 0 0 0 8 0 0 0 0 0 0 0
0 0 0 0 8 0 0 0 0 0 0 8 0 0 0 0 0 0 0
0 0 0 0 8 0 0 0 0 0 0 8 0 0 0 0 0 0 0
0 0 0 0 8 0 0 0 0 0 0 8 0 0 0 0 0 0 0
0 0 0 0 8 0 0 0 0 0 0 8 0 0 0 0 0 0 0
0 0 0 0 8 0 0 0 0 0 0 8 0 0 0 0 0 0 0

Output:
0 0 0 3 6 2 2 1 2 2 2 0 0 0 0 0 0 0 0
0 0 0 0 8 2 2 3 2 2 2 8 0 0 0 0 0 0 0
8 8 3 8 3 8 8 8 8 8 8 8 8 8 8 6 6 8 8
1 4 4 2 6 6 6 6 1 6 6 8 3 2 3 3 3 3 2
4 4 4 4 8 6 6 0 6 6 6 8 3 3 3 3 3 3 3
4 4 4 4 8 6 6 6 8 6 6 8 3 3 3 3 3 3 3
4 4 6 4 8 6 1 6 6 6 6 8 3 3 3 3 3 3 3
4 8 8 8 8 8 8 8 8 8 8 8 8 8 8 8 8 8 8
0 0 0 0 4 1 1 1 1 1 1 8 0 6 1 0 6 0 0
0 0 0 0 8 1 1 1 1 1 1 8 4 0 0 0 0 4 0
0 3 0 0 8 1 1 1 1 1 1 8 0 0 0 0 4 0 0
0 0 0 2 8 1 0 1 1 1 1 8 1 6 0 0 0 0 0
0 3 0 0 8 1 1 1 1 1 1 8 0 0 0 0 0 0 0
0 0 0 1 8 1 1 1 1 1 1 8 0 0 0 0 0 0 0
0 6 0 0 8 1 1 1 1 1 1 8 0 0 0 0 0 2 0
4 0 0 0 8 1 1 1 1 1 1 8 0 0 0 0 0 0 0
0 0 0 0 8 1 1 1 1 1 0 8 0 0 0 0 0 0 0
0 0 0 0 8 2 1 8 3 1 1 8 0 0 0 0 3 0 0


Example3
Input:
0 0 0 0 8 0 0 0 0 0 0 8 0 0 0 0 0 0 0
0 0 0 0 8 0 0 0 0 0 0 8 0 0 0 0 0 0 0
8 8 8 8 8 8 8 8 8 8 8 8 8 8 8 8 8 8 8
0 0 0 0 8 0 0 0 0 0 0 8 0 0 0 0 0 0 0
0 0 0 0 8 0 0 0 0 0 0 8 0 0 0 0 0 0 0
0 0 0 0 8 0 0 0 0 0 0 8 0 0 0 0 0 0 0
0 0 0 0 8 0 0 0 0 0 0 8 0 0 0 0 0 0 0
8 8 8 8 8 8 8 8 8 8 8 8 8 8 8 8 8 8 8
0 0 0 0 8 0 0 0 0 0 0 8 0 0 0 0 0 0 0
0 0 0 0 8 0 0 0 0 0 0 8 0 0 0 0 0 0 0
0 0 0 0 8 0 0 0 0 0 0 8 0 0 0 0 0 0 0
0 0 0 0 8 0 0 0 0 0 0 8 0 0 0 0 0 0 0
0 0 0 0 8 0 0 0 0 0 0 8 0 0 0 0 0 0 0
0 0 0 0 8 0 0 0 0 0 0 8 0 0 0 0 0 0 0
0 0 0 0 8 0 0 0 0 0 0 8 0 0 0 0 0 0 0
0 0 0 0 8 0 0 0 0 0 0 8 0 0 0 0 0 0 0
0 0 0 0 8 0 0 0 0 0 0 8 0 0 0 0 0 0 0
0 0 0 0 8 0 0 0 0 0 0 8 0 0 0 0 0 0 0

Output:
0 0 0 0 8 3 2 2 2 2 2 8 0 0 0 0 6 0 0
0 0 0 0 8 2 2 2 2 2 2 8 0 0 0 0 0 4 0
8 8 8 3 8 8 8 8 8 8 8 8 8 0 8 8 4 8 8
4 4 4 6 8 6 6 6 6 6 6 8 3 3 3 3 3 3 3
4 4 4 4 8 6 6 0 6 6 6 8 3 3 3 3 1 3 3
2 4 4 4 8 6 0 6 6 6 6 8 3 3 3 3 3 3 3
4 4 1 4 8 6 6 6 6 6 1 8 3 3 3 3 2 3 2
8 8 8 1 8 8 8 8 8 8 8 8 8 8 6 8 8 8 8
0 0 0 0 8 1 1 1 1 1 1 8 0 0 0 0 0 0 0
0 0 6 0 8 0 1 1 1 1 1 8 0 0 3 0 0 0 0
0 0 0 0 8 1 8 1 1 1 1 8 0 6 0 0 3 0 0
0 0 0 0 0 1 1 1 1 8 1 8 0 0 0 0 2 8 0
4 0 0 0 1 1 1 1 1 1 1 8 0 0 0 0 0 0 0
0 0 0 0 8 1 1 1 1 1 1 8 0 0 0 0 0 0 0
0 0 0 0 3 1 1 1 1 1 1 8 0 0 8 0 4 0 0
1 0 0 0 8 1 1 1 1 1 1 8 0 0 4 0 0 0 0
1 0 0 0 8 1 8 0 1 1 1 8 6 0 0 0 0 0 0
0 0 0 0 8 1 1 3 1 1 1 8 0 0 6 0 1 4 0


Example4
Input:
0 0 0 0 8 0 0 0 0 0 0 8 0 0 0 0 0 0 0
0 0 0 0 8 0 0 0 0 0 0 8 0 0 0 0 0 0 0
8 8 8 8 8 8 8 8 8 8 8 8 8 8 8 8 8 8 8
0 0 0 0 8 0 0 0 0 0 0 8 0 0 0 0 0 0 0
0 0 0 0 8 0 0 0 0 0 0 8 0 0 0 0 0 0 0
0 0 0 0 8 0 0 0 0 0 0 8 0 0 0 0 0 0 0
0 0 0 0 8 0 0 0 0 0 0 8 0 0 0 0 0 0 0
8 8 8 8 8 8 8 8 8 8 8 8 8 8 8 8 8 8 8
0 0 0 0 8 0 0 0 0 0 0 8 0 0 0 0 0 0 0
0 0 0 0 8 0 0 0 0 0 0 8 0 0 0 0 0 0 0
0 0 0 0 8 0 0 0 0 0 0 8 0 0 0 0 0 0 0
0 0 0 0 8 0 0 0 0 0 0 8 0 0 0 0 0 0 0
0 0 0 0 8 0 0 0 0 0 0 8 0 0 0 0 0 0 0
0 0 0 0 8 0 0 0 0 0 0 8 0 0 0 0 0 0 0
0 0 0 0 8 0 0 0 0 0 0 8 0 0 0 0 0 0 0
0 0 0 0 8 0 0 0 0 0 0 8 0 0 0 0 0 0 0
0 0 0 0 8 0 0 0 0 0 0 8 0 0 0 0 0 0 0
0 0 0 0 8 0 0 0 0 0 0 8 0 0 0 0 0 0 0

Output:
0 0 0 0 8 2 2 2 2 2 3 8 0 0 6 0 0 1 0
0 0 0 0 8 2 0 2 2 2 4 8 0 8 0 0 0 0 0
1 8 8 8 4 8 8 8 8 3 8 8 3 8 8 8 8 8 8
4 4 4 4 8 6 6 6 6 6 3 8 3 3 3 0 3 3 3
4 4 4 4 8 6 6 6 6 6 6 8 3 3 3 3 3 3 3
4 4 4 4 8 6 6 6 6 6 6 8 3 3 3 3 3 3 3
4 2 8 4 2 6 6 6 6 6 8 8 3 3 3 3 3 3 3
8 8 8 8 8 8 8 8 2 8 1 8 8 8 8 8 8 8 4
0 0 0 0 8 1 1 1 1 1 1 8 0 0 0 0 0 0 3
0 0 0 0 2 1 1 1 1 1 1 8 0 0 0 0 1 0 0
0 8 0 0 8 1 1 1 1 1 1 8 0 0 1 0 0 6 0
8 0 0 0 1 1 1 1 1 1 1 8 0 0 0 8 0 4 0
0 0 0 0 8 1 1 1 1 3 1 8 0 0 0 0 0 0 0
3 0 0 0 8 1 1 1 6 3 3 8 3 0 0 0 0 0 0
8 0 0 0 8 1 8 8 1 1 1 8 0 0 2 0 0 0 0
0 0 0 0 8 1 1 1 1 1 3 8 0 0 0 0 0 0 0
0 0 0 0 8 1 1 1 1 1 1 8 0 0 0 0 0 0 0
0 0 0 3 8 1 1 1 1 1 8 8 0 0 0 0 0 0 0


Example5
Input:
0 0 0 0 8 0 0 0 0 0 0 8 0 0 0 0 0 0 0
0 0 0 0 8 0 0 0 0 0 0 8 0 0 0 0 0 0 0
8 8 8 8 8 8 8 8 8 8 8 8 8 8 8 8 8 8 8
0 0 0 0 8 0 0 0 0 0 0 8 0 0 0 0 0 0 0
0 0 0 0 8 0 0 0 0 0 0 8 0 0 0 0 0 0 0
0 0 0 0 8 0 0 0 0 0 0 8 0 0 0 0 0 0 0
0 0 0 0 8 0 0 0 0 0 0 8 0 0 0 0 0 0 0
8 8 8 8 8 8 8 8 8 8 8 8 8 8 8 8 8 8 8
0 0 0 0 8 0 0 0 0 0 0 8 0 0 0 0 0 0 0
0 0 0 0 8 0 0 0 0 0 0 8 0 0 0 0 0 0 0
0 0 0 0 8 0 0 0 0 0 0 8 0 0 0 0 0 0 0
0 0 0 0 8 0 0 0 0 0 0 8 0 0 0 0 0 0 0
0 0 0 0 8 0 0 0 0 0 0 8 0 0 0 0 0 0 0
0 0 0 0 8 0 0 0 0 0 0 8 0 0 0 0 0 0 0
0 0 0 0 8 0 0 0 0 0 0 8 0 0 0 0 0 0 0
0 0 0 0 8 0 0 0 0 0 0 8 0 0 0 0 0 0 0
0 0 0 0 8 0 0 0 0 0 0 8 0 0 0 0 0 0 0
0 0 0 0 8 0 0 0 0 0 0 8 0 0 0 0 0 0 0

Output:
0 0 0 0 8 2 2 2 2 2 2 8 0 0 0 0 0 0 0
0 0 0 4 3 2 2 2 2 2 2 8 0 0 0 0 1 0 0
8 8 8 8 8 8 8 6 8 8 8 8 8 8 8 8 8 6 8
4 4 4 4 8 6 6 6 6 6 6 8 3 3 3 3 3 2 3
4 4 4 4 8 6 6 6 6 6 6 8 3 3 3 3 3 3 3
4 4 1 4 8 6 6 6 2 6 6 8 3 3 3 3 3 0 3
4 4 8 4 1 6 0 6 8 6 6 8 3 3 3 6 3 3 3
8 8 8 8 8 8 8 8 8 8 8 8 8 8 0 8 8 8 8
0 2 0 0 8 6 1 1 1 1 1 8 0 0 0 0 0 0 0
0 0 0 8 0 1 1 1 1 1 1 8 0 0 0 0 0 0 4
8 0 0 0 8 3 1 4 1 1 8 8 0 0 0 0 0 0 0
0 8 0 0 8 1 1 1 1 1 1 8 0 0 0 0 0 0 0
0 0 0 0 8 1 1 1 1 1 1 8 0 0 0 0 0 0 2
0 0 0 0 8 1 8 2 3 1 1 8 0 3 2 8 0 0 0
0 0 0 0 8 1 1 8 1 1 1 8 0 0 0 0 0 6 6
3 0 0 0 8 1 1 1 1 1 1 8 0 0 0 0 0 0 0
0 0 0 0 8 1 3 1 1 1 1 8 0 0 0 6 0 0 4
0 0 0 0 8 8 1 1 1 1 6 8 3 0 0 0 0 0 0


Example6
Input:
0 0 0 0 8 0 0 0 0 0 0 8 0 0 0 0 0 0 0
0 0 0 0 8 0 0 0 0 0 0 8 0 0 0 0 0 0 0
8 8 8 8 8 8 8 8 8 8 8 8 8 8 8 8 8 8 8
0 0 0 0 8 0 0 0 0 0 0 8 0 0 0 0 0 0 0
0 0 0 0 8 0 0 0 0 0 0 8 0 0 0 0 0 0 0
0 0 0 0 8 0 0 0 0 0 0 8 0 0 0 0 0 0 0
0 0 0 0 8 0 0 0 0 0 0 8 0 0 0 0 0 0 0
8 8 8 8 8 8 8 8 8 8 8 8 8 8 8 8 8 8 8
0 0 0 0 8 0 0 0 0 0 0 8 0 0 0 0 0 0 0
0 0 0 0 8 0 0 0 0 0 0 8 0 0 0 0 0 0 0
0 0 0 0 8 0 0 0 0 0 0 8 0 0 0 0 0 0 0
0 0 0 0 8 0 0 0 0 0 0 8 0 0 0 0 0 0 0
0 0 0 0 8 0 0 0 0 0 0 8 0 0 0 0 0 0 0
0 0 0 0 8 0 0 0 0 0 0 8 0 0 0 0 0 0 0
0 0 0 0 8 0 0 0 0 0 0 8 0 0 0 0 0 0 0
0 0 0 0 8 0 0 0 0 0 0 8 0 0 0 0 0 0 0
0 0 0 0 8 0 0 0 0 0 0 8 0 0 0 0 0 0 0
0 0 0 0 8 0 0 0 0 0 0 8 0 0 0 0 0 0 0

Output:
0 0 0 4 8 2 2 2 8 2 2 8 0 3 8 0 0 1 0
0 0 1 0 8 2 2 2 6 2 6 8 3 0 0 8 0 0 0
8 8 8 8 8 8 8 3 8 8 8 8 8 8 8 8 8 8 8
4 3 4 4 8 1 0 6 6 6 6 8 3 3 3 3 3 3 3
8 0 4 4 8 6 6 6 0 6 6 8 3 3 3 3 3 3 3
4 4 4 4 8 6 6 6 6 6 6 8 3 3 2 3 3 6 3
4 4 6 0 8 6 6 6 6 6 6 8 3 3 3 3 2 3 3
8 8 8 8 8 8 8 8 0 8 8 8 8 8 8 8 8 8 2
0 0 0 8 8 1 1 1 1 1 1 1 0 0 0 0 0 0 0
2 0 0 1 8 1 1 1 1 1 1 8 0 0 0 0 0 0 0
0 0 0 0 8 1 1 1 1 1 1 8 0 0 0 0 0 0 0
0 0 0 0 8 1 1 1 1 1 1 8 0 0 0 0 0 0 8
0 0 0 0 8 1 1 1 1 1 4 8 0 0 0 0 0 0 0
0 0 0 0 4 1 1 1 4 1 1 8 0 1 0 0 0 0 0
0 0 0 0 8 1 1 1 1 1 1 0 0 8 6 0 0 0 0
0 0 8 0 8 1 1 1 1 1 1 6 0 0 0 0 0 0 0
0 0 4 0 8 1 1 1 1 1 1 8 0 0 0 0 0 0 0
0 3 6 0 8 4 1 1 1 1 1 8 0 0 0 0 0 0 0


Example7
Input:
0 0 0 0 8 0 0 0 0 0 0 8 0 0 0 0 0 0 0
0 0 0 0 8 0 0 0 0 0 0 8 0 0 0 0 0 0 0
8 8 8 8 8 8 8 8 8 8 8 8 8 8 8 8 8 8 8
0 0 0 0 8 0 0 0 0 0 0 8 0 0 0 0 0 0 0
0 0 0 0 8 0 0 0 0 0 0 8 0 0 0 0 0 0 0
0 0 0 0 8 0 0 0 0 0 0 8 0 0 0 0 0 0 0
0 0 0 0 8 0 0 0 0 0 0 8 0 0 0 0 0 0 0
8 8 8 8 8 8 8 8 8 8 8 8 8 8 8 8 8 8 8
0 0 0 0 8 0 0 0 0 0 0 8 0 0 0 0 0 0 0
0 0 0 0 8 0 0 0 0 0 0 8 0 0 0 0 0 0 0
0 0 0 0 8 0 0 0 0 0 0 8 0 0 0 0 0 0 0
0 0 0 0 8 0 0 0 0 0 0 8 0 0 0 0 0 0 0
0 0 0 0 8 0 0 0 0 0 0 8 0 0 0 0 0 0 0
0 0 0 0 8 0 0 0 0 0 0 8 0 0 0 0 0 0 0
0 0 0 0 8 0 0 0 0 0 0 8 0 0 0 0 0 0 0
0 0 0 0 8 0 0 0 0 0 0 8 0 0 0 0 0 0 0
0 0 0 0 8 0 0 0 0 0 0 8 0 0 0 0 0 0 0
0 0 0 0 8 0 0 0 0 0 0 8 0 0 0 0 0 0 0

Output:
0 0 0 0 8 2 2 2 2 2 2 3 0 8 0 0 0 0 0
0 0 0 0 8 2 2 3 2 2 0 8 0 0 1 6 0 0 0
8 8 8 8 8 1 8 8 0 8 8 8 8 8 8 8 8 8 8
4 8 4 6 8 6 1 6 6 6 6 6 3 3 3 3 3 3 3
4 4 4 4 1 6 6 6 6 6 6 8 3 3 3 3 3 3 3
4 4 6 4 8 6 6 6 4 6 6 8 3 3 3 3 3 3 3
4 8 4 4 8 8 6 6 8 6 6 8 3 4 3 3 8 3 3
8 8 8 8 8 8 8 8 8 8 8 4 8 8 8 8 8 8 8
0 0 0 0 8 1 1 1 1 1 1 8 0 3 0 0 0 0 0
0 0 0 0 8 1 1 1 1 1 1 8 0 0 0 0 0 0 0
3 0 0 0 8 1 1 1 1 6 1 4 0 0 8 0 0 0 2
0 0 0 0 8 1 1 8 3 0 1 8 0 0 0 0 3 0 0
0 0 0 0 8 1 1 1 1 1 1 8 0 0 4 0 0 1 0
0 0 0 0 8 1 1 1 1 1 6 8 0 0 0 0 0 0 0
0 0 0 0 4 1 1 1 1 1 1 4 0 0 0 0 0 0 0
0 0 0 0 8 1 1 3 1 1 1 8 0 0 0 0 4 0 0
0 0 0 0 8 1 6 1 1 1 1 8 0 0 0 0 0 0 0
0 0 0 0 8 1 1 0 8 1 1 8 0 0 0 0 8 0 0


Example8
Input:
0 0 0 0 8 0 0 0 0 0 0 8 0 0 0 0 0 0 0
0 0 0 0 8 0 0 0 0 0 0 8 0 0 0 0 0 0 0
8 8 8 8 8 8 8 8 8 8 8 8 8 8 8 8 8 8 8
0 0 0 0 8 0 0 0 0 0 0 8 0 0 0 0 0 0 0
0 0 0 0 8 0 0 0 0 0 0 8 0 0 0 0 0 0 0
0 0 0 0 8 0 0 0 0 0 0 8 0 0 0 0 0 0 0
0 0 0 0 8 0 0 0 0 0 0 8 0 0 0 0 0 0 0
8 8 8 8 8 8 8 8 8 8 8 8 8 8 8 8 8 8 8
0 0 0 0 8 0 0 0 0 0 0 8 0 0 0 0 0 0 0
0 0 0 0 8 0 0 0 0 0 0 8 0 0 0 0 0 0 0
0 0 0 0 8 0 0 0 0 0 0 8 0 0 0 0 0 0 0
0 0 0 0 8 0 0 0 0 0 0 8 0 0 0 0 0 0 0
0 0 0 0 8 0 0 0 0 0 0 8 0 0 0 0 0 0 0
0 0 0 0 8 0 0 0 0 0 0 8 0 0 0 0 0 0 0
0 0 0 0 8 0 0 0 0 0 0 8 0 0 0 0 0 0 0
0 0 0 0 8 0 0 0 0 0 0 8 0 0 0 0 0 0 0
0 0 0 0 8 0 0 0 0 0 0 8 0 0 0 0 0 0 0
0 0 0 0 8 0 0 0 0 0 0 8 0 0 0 0 0 0 0

Output:
0 2 0 0 8 2 2 1 2 2 2 8 0 0 1 0 0 0 0
0 0 0 0 8 2 0 2 2 2 2 8 0 0 0 0 0 0 0
8 8 8 8 8 8 8 8 8 8 8 8 0 8 8 8 8 8 8
4 8 8 4 8 6 6 6 3 6 6 8 3 3 3 3 3 3 8
4 0 4 4 8 6 6 6 6 6 1 8 3 3 0 3 3 0 3
4 4 4 4 8 6 6 6 1 6 6 8 3 3 3 3 3 3 3
4 4 4 0 8 6 6 6 6 6 6 8 3 3 3 3 6 3 3
3 8 8 8 3 8 8 8 1 2 8 8 8 2 8 8 2 8 8
0 0 0 0 8 1 8 1 1 0 1 8 0 0 0 0 0 0 0
0 0 0 0 8 1 1 8 1 1 1 8 0 0 0 0 0 0 0
0 0 0 0 8 1 1 1 1 1 1 8 0 0 0 3 6 0 0
0 0 8 0 8 1 1 1 1 2 1 8 0 0 2 0 0 0 0
0 2 0 0 8 1 1 1 1 1 1 8 0 0 0 0 0 0 0
0 0 0 0 8 1 1 1 1 1 1 8 0 0 0 0 0 0 0
8 0 8 0 6 1 1 1 1 1 1 8 0 4 0 0 1 0 0
0 0 0 0 8 8 8 1 1 1 1 2 0 0 0 1 0 0 0
0 0 0 0 8 1 1 0 1 1 1 8 0 0 0 0 0 0 0
0 0 0 0 8 1 1 1 1 1 1 8 0 0 0 0 0 0 1


Example9
Input:
0 0 0 0 8 0 0 0 0 0 0 8 0 0 0 0 0 0 0
0 0 0 0 8 0 0 0 0 0 0 8 0 0 0 0 0 0 0
8 8 8 8 8 8 8 8 8 8 8 8 8 8 8 8 8 8 8
0 0 0 0 8 0 0 0 0 0 0 8 0 0 0 0 0 0 0
0 0 0 0 8 0 0 0 0 0 0 8 0 0 0 0 0 0 0
0 0 0 0 8 0 0 0 0 0 0 8 0 0 0 0 0 0 0
0 0 0 0 8 0 0 0 0 0 0 8 0 0 0 0 0 0 0
8 8 8 8 8 8 8 8 8 8 8 8 8 8 8 8 8 8 8
0 0 0 0 8 0 0 0 0 0 0 8 0 0 0 0 0 0 0
0 0 0 0 8 0 0 0 0 0 0 8 0 0 0 0 0 0 0
0 0 0 0 8 0 0 0 0 0 0 8 0 0 0 0 0 0 0
0 0 0 0 8 0 0 0 0 0 0 8 0 0 0 0 0 0 0
0 0 0 0 8 0 0 0 0 0 0 8 0 0 0 0 0 0 0
0 0 0 0 8 0 0 0 0 0 0 8 0 0 0 0 0 0 0
0 0 0 0 8 0 0 0 0 0 0 8 0 0 0 0 0 0 0
0 0 0 0 8 0 0 0 0 0 0 8 0 0 0 0 0 0 0
0 0 0 0 8 0 0 0 0 0 0 8 0 0 0 0 0 0 0
0 0 0 0 8 0 0 0 0 0 0 8 0 0 0 0 0 0 0

Output:
0 0 0 0 8 2 2 2 2 2 2 8 0 0 0 1 0 0 0
0 0 0 0 8 2 2 2 2 2 2 8 0 1 0 0 2 0 0
8 8 2 8 8 8 8 8 8 8 8 8 8 8 8 8 8 8 8
4 4 4 4 8 6 6 6 6 6 6 8 3 3 3 3 4 3 3
4 6 4 4 8 6 6 6 6 6 6 8 3 3 3 3 3 3 4
4 4 4 4 6 6 6 1 6 6 6 8 3 3 3 3 3 8 3
4 4 4 0 8 6 6 6 4 6 6 8 3 3 3 3 3 3 3
8 8 8 8 8 8 8 8 8 8 8 8 8 8 8 8 8 2 8
0 0 0 0 8 1 1 1 1 1 1 8 0 0 0 0 0 0 0
0 0 0 0 0 1 1 1 1 1 6 8 0 0 2 2 0 0 0
0 0 0 0 8 1 0 1 1 1 6 8 0 0 0 0 0 0 0
0 0 0 0 8 1 1 1 1 1 1 8 8 0 0 0 0 0 0
0 0 8 0 0 1 1 1 2 1 1 8 1 0 0 0 3 0 0
0 0 1 0 8 1 1 1 0 8 1 8 0 3 0 0 8 0 0
0 0 0 0 8 1 1 1 1 3 1 8 0 0 3 0 0 0 0
0 0 6 0 8 1 1 1 0 1 1 8 0 0 4 0 3 0 8
0 0 0 0 8 1 1 1 1 1 1 6 0 0 0 6 0 0 0
0 0 0 0 8 1 1 6 1 1 3 8 0 0 8 0 0 0 0
Example10
Input:
0 0 8 0 0 0 0 0 0 8 0 0 0 0
0 0 8 0 0 0 0 0 0 8 0 0 0 0
0 0 8 0 0 0 0 0 0 8 0 0 0 0
0 0 8 0 0 0 0 0 0 8 0 0 0 0
8 8 8 8 8 8 8 8 8 8 8 8 8 8
0 0 8 0 0 0 0 0 0 8 0 0 0 0
0 0 8 0 0 0 0 0 0 8 0 0 0 0
8 8 8 8 8 8 8 8 8 8 8 8 8 8
0 0 8 0 0 0 0 0 0 8 0 0 0 0
0 0 8 0 0 0 0 0 0 8 0 0 0 0
0 0 8 0 0 0 0 0 0 8 0 0 0 0
0 0 8 0 0 0 0 0 0 8 0 0 0 0

Output:
0 0 8 2 0 2 2 2 2 8 0 0 0 4
0 0 8 2 2 2 2 2 6 8 1 0 8 0
0 0 8 2 2 2 2 2 2 8 8 0 0 0
0 0 8 2 2 0 2 2 2 8 0 0 0 3
6 8 8 8 8 8 8 8 0 8 2 8 8 8
4 4 8 6 6 6 3 4 6 8 3 3 3 3
4 4 1 6 6 6 6 6 6 8 3 3 3 3
8 8 8 8 8 8 8 8 8 8 8 8 8 8
0 6 8 1 1 1 1 6 1 8 0 0 0 0
0 2 8 1 1 1 1 1 1 8 4 0 0 0
0 0 8 1 1 1 1 1 1 8 0 6 0 0
0 0 8 1 1 1 1 1 2 8 0 2 0 0


Example11
Input:
0 0 8 0 0 0 0 0 0 8 0 0 0 0
0 0 8 0 0 0 0 0 0 8 0 0 0 0
0 0 8 0 0 0 0 0 0 8 0 0 0 0
0 0 8 0 0 0 0 0 0 8 0 0 0 0
8 8 8 8 8 8 8 8 8 8 8 8 8 8
0 0 8 0 0 0 0 0 0 8 0 0 0 0
0 0 8 0 0 0 0 0 0 8 0 0 0 0
8 8 8 8 8 8 8 8 8 8 8 8 8 8
0 0 8 0 0 0 0 0 0 8 0 0 0 0
0 0 8 0 0 0 0 0 0 8 0 0 0 0
0 0 8 0 0 0 0 0 0 8 0 0 0 0
0 0 8 0 0 0 0 0 0 8 0 0 0 0

Output:
1 0 8 2 2 2 2 2 2 8 0 0 0 0
2 1 8 1 2 2 0 2 2 8 0 0 0 0
0 0 8 2 6 2 2 2 4 8 0 0 0 0
0 0 8 6 2 2 2 2 2 8 0 0 0 0
8 8 8 2 8 8 8 8 8 8 2 8 8 8
4 4 8 6 6 6 6 6 6 1 3 3 3 3
4 4 8 6 4 6 6 6 6 2 3 3 3 3
8 8 8 8 8 8 0 8 8 8 8 8 8 8
0 4 8 1 1 1 1 1 1 8 4 0 4 0
0 0 8 1 1 1 1 4 1 8 0 0 0 0
0 0 8 1 1 1 1 1 1 8 0 1 3 0
0 0 8 1 1 1 1 1 1 2 0 0 0 0


Example12
Input:
0 0 8 0 0 0 0 0 0 8 0 0 0 0
0 0 8 0 0 0 0 0 0 8 0 0 0 0
0 0 8 0 0 0 0 0 0 8 0 0 0 0
0 0 8 0 0 0 0 0 0 8 0 0 0 0
8 8 8 8 8 8 8 8 8 8 8 8 8 8
0 0 8 0 0 0 0 0 0 8 0 0 0 0
0 0 8 0 0 0 0 0 0 8 0 0 0 0
8 8 8 8 8 8 8 8 8 8 8 8 8 8
0 0 8 0 0 0 0 0 0 8 0 0 0 0
0 0 8 0 0 0 0 0 0 8 0 0 0 0
0 0 8 0 0 0 0 0 0 8 0 0 0 0
0 0 8 0 0 0 0 0 0 8 0 0 0 0

Output:
0 0 8 2 6 2 4 2 2 8 0 0 4 0
0 0 8 2 2 2 2 2 2 8 0 0 0 0
0 8 8 2 2 2 2 2 1 8 0 0 0 0
0 0 6 2 2 2 2 2 2 8 0 6 0 0
1 8 8 8 8 8 8 8 8 8 8 8 8 8
4 4 8 1 6 6 6 6 6 8 3 3 3 3
4 4 8 1 6 6 1 6 6 8 3 3 3 3
8 0 8 8 8 8 8 8 0 4 8 8 8 6
0 0 8 1 1 1 1 1 1 3 0 0 0 0
0 0 8 1 1 1 1 1 1 8 0 0 0 4
1 0 8 1 2 1 1 1 1 8 6 0 0 0
0 0 8 8 1 1 1 1 1 8 0 0 0 0


Example13
Input:
0 0 8 0 0 0 0 0 0 8 0 0 0 0
0 0 8 0 0 0 0 0 0 8 0 0 0 0
0 0 8 0 0 0 0 0 0 8 0 0 0 0
0 0 8 0 0 0 0 0 0 8 0 0 0 0
8 8 8 8 8 8 8 8 8 8 8 8 8 8
0 0 8 0 0 0 0 0 0 8 0 0 0 0
0 0 8 0 0 0 0 0 0 8 0 0 0 0
8 8 8 8 8 8 8 8 8 8 8 8 8 8
0 0 8 0 0 0 0 0 0 8 0 0 0 0
0 0 8 0 0 0 0 0 0 8 0 0 0 0
0 0 8 0 0 0 0 0 0 8 0 0 0 0
0 0 8 0 0 0 0 0 0 8 0 0 0 0

Output:
0 4 8 4 2 2 2 2 2 8 0 0 0 0
0 1 8 2 1 2 2 2 2 8 0 0 0 0
0 0 8 2 2 2 1 2 2 8 0 0 0 0
0 0 8 2 2 2 2 2 6 8 0 0 0 0
8 8 8 8 4 8 8 8 8 8 8 8 8 8
4 4 8 8 1 6 6 6 6 8 3 3 1 3
4 4 8 6 6 6 6 6 0 8 3 3 3 3
3 6 8 3 8 8 8 8 8 8 8 8 8 8
1 0 8 1 1 1 1 1 3 8 4 0 0 0
0 0 8 1 1 1 1 1 1 8 0 0 0 0
0 3 8 2 1 1 1 1 1 8 0 0 0 4
1 0 8 1 1 1 1 1 1 8 0 0 0 0


Example14
Input:
0 0 8 0 0 0 0 0 0 8 0 0 0 0
0 0 8 0 0 0 0 0 0 8 0 0 0 0
0 0 8 0 0 0 0 0 0 8 0 0 0 0
0 0 8 0 0 0 0 0 0 8 0 0 0 0
8 8 8 8 8 8 8 8 8 8 8 8 8 8
0 0 8 0 0 0 0 0 0 8 0 0 0 0
0 0 8 0 0 0 0 0 0 8 0 0 0 0
8 8 8 8 8 8 8 8 8 8 8 8 8 8
0 0 8 0 0 0 0 0 0 8 0 0 0 0
0 0 8 0 0 0 0 0 0 8 0 0 0 0
0 0 8 0 0 0 0 0 0 8 0 0 0 0
0 0 8 0 0 0 0 0 0 8 0 0 0 0

Output:
0 0 8 2 2 2 2 2 2 8 0 0 4 0
4 2 8 0 2 2 2 2 2 8 0 0 0 0
2 0 8 2 2 2 3 2 2 8 0 0 0 0
2 0 8 2 2 2 2 2 2 8 0 0 0 0
8 8 8 8 8 8 8 8 8 8 8 8 8 8
4 4 8 6 6 6 6 6 1 8 3 3 3 3
4 4 4 8 6 6 6 3 6 8 3 3 3 3
8 3 8 8 8 8 8 8 8 8 8 8 8 8
1 8 8 1 1 1 4 1 1 8 0 0 0 0
0 0 8 1 1 1 1 1 1 8 0 6 0 0
0 0 6 1 1 1 1 1 1 8 0 0 6 0
0 0 8 1 1 1 1 3 1 8 4 0 8 0


Example15
Input:
0 0 8 0 0 0 0 0 0 8 0 0 0 0
0 0 8 0 0 0 0 0 0 8 0 0 0 0
0 0 8 0 0 0 0 0 0 8 0 0 0 0
0 0 8 0 0 0 0 0 0 8 0 0 0 0
8 8 8 8 8 8 8 8 8 8 8 8 8 8
0 0 8 0 0 0 0 0 0 8 0 0 0 0
0 0 8 0 0 0 0 0 0 8 0 0 0 0
8 8 8 8 8 8 8 8 8 8 8 8 8 8
0 0 8 0 0 0 0 0 0 8 0 0 0 0
0 0 8 0 0 0 0 0 0 8 0 0 0 0
0 0 8 0 0 0 0 0 0 8 0 0 0 0
0 0 8 0 0 0 0 0 0 8 0 0 0 0

Output:
0 0 8 1 2 2 2 8 2 8 0 0 4 4
0 0 8 2 2 2 2 1 2 8 0 0 0 0
0 0 8 2 2 2 2 2 2 8 1 6 0 0
0 0 8 2 2 2 2 0 2 8 0 0 0 0
8 8 8 8 8 8 8 8 8 2 8 8 8 8
4 4 8 6 6 6 6 6 6 8 3 3 3 3
4 6 8 2 6 6 6 6 3 8 3 3 3 3
8 8 8 8 8 8 8 8 3 8 8 8 0 8
0 0 8 1 1 1 1 1 1 8 4 0 0 0
1 0 8 1 1 1 1 4 1 8 0 0 3 0
0 0 8 1 1 6 1 1 1 8 0 0 0 0
0 0 8 1 1 6 1 1 1 8 0 0 4 0


Example16
Input:
0 0 8 0 0 0 0 0 0 8 0 0 0 0
0 0 8 0 0 0 0 0 0 8 0 0 0 0
0 0 8 0 0 0 0 0 0 8 0 0 0 0
0 0 8 0 0 0 0 0 0 8 0 0 0 0
8 8 8 8 8 8 8 8 8 8 8 8 8 8
0 0 8 0 0 0 0 0 0 8 0 0 0 0
0 0 8 0 0 0 0 0 0 8 0 0 0 0
8 8 8 8 8 8 8 8 8 8 8 8 8 8
0 0 8 0 0 0 0 0 0 8 0 0 0 0
0 0 8 0 0 0 0 0 0 8 0 0 0 0
0 0 8 0 0 0 0 0 0 8 0 0 0 0
0 0 8 0 0 0 0 0 0 8 0 0 0 0

Output:
0 0 8 2 4 1 2 2 1 8 2 0 3 0
1 0 8 2 2 2 2 2 2 8 0 0 0 0
0 0 8 2 2 2 2 2 2 8 0 0 0 0
0 8 8 2 2 2 2 3 2 8 0 0 0 0
8 8 8 8 8 8 2 8 8 8 8 8 6 8
4 4 8 8 6 0 6 6 6 6 3 4 3 3
4 4 8 6 6 2 6 6 6 8 3 3 3 8
8 8 8 8 8 8 8 8 8 3 8 8 8 8
0 0 8 1 1 1 6 1 1 8 0 1 0 0
0 0 8 1 1 1 1 1 1 8 0 0 0 0
0 0 8 1 1 1 1 1 1 8 1 0 0 0
0 6 8 1 1 1 1 1 1 8 0 0 0 0


Example17
Input:
0 0 8 0 0 0 0 0 0 8 0 0 0 0
0 0 8 0 0 0 0 0 0 8 0 0 0 0
0 0 8 0 0 0 0 0 0 8 0 0 0 0
0 0 8 0 0 0 0 0 0 8 0 0 0 0
8 8 8 8 8 8 8 8 8 8 8 8 8 8
0 0 8 0 0 0 0 0 0 8 0 0 0 0
0 0 8 0 0 0 0 0 0 8 0 0 0 0
8 8 8 8 8 8 8 8 8 8 8 8 8 8
0 0 8 0 0 0 0 0 0 8 0 0 0 0
0 0 8 0 0 0 0 0 0 8 0 0 0 0
0 0 8 0 0 0 0 0 0 8 0 0 0 0
0 0 8 0 0 0 0 0 0 8 0 0 0 0

Output:
0 0 0 2 2 2 2 2 2 8 0 0 0 0
0 0 8 2 2 6 2 2 2 8 0 0 0 0
0 0 8 3 2 2 2 2 0 8 0 0 0 0
0 0 8 2 2 2 2 8 8 8 0 0 0 0
8 8 8 2 8 8 8 8 8 8 6 8 8 8
4 0 8 6 6 0 6 6 6 8 3 3 3 3
4 4 8 6 6 6 6 6 6 8 3 3 3 3
1 8 8 8 8 8 8 8 0 8 8 8 8 3
0 0 8 1 1 1 4 1 1 8 0 0 1 0
0 0 3 1 1 1 1 1 1 8 0 0 0 0
0 0 8 6 8 1 1 1 1 4 0 0 0 0
0 4 8 1 1 1 1 1 1 8 0 0 8 0


Example18
Input:
0 0 8 0 0 0 0 0 0 8 0 0 0 0
0 0 8 0 0 0 0 0 0 8 0 0 0 0
0 0 8 0 0 0 0 0 0 8 0 0 0 0
0 0 8 0 0 0 0 0 0 8 0 0 0 0
8 8 8 8 8 8 8 8 8 8 8 8 8 8
0 0 8 0 0 0 0 0 0 8 0 0 0 0
0 0 8 0 0 0 0 0 0 8 0 0 0 0
8 8 8 8 8 8 8 8 8 8 8 8 8 8
0 0 8 0 0 0 0 0 0 8 0 0 0 0
0 0 8 0 0 0 0 0 0 8 0 0 0 0
0 0 8 0 0 0 0 0 0 8 0 0 0 0
0 0 8 0 0 0 0 0 0 8 0 0 0 0

Output:
0 0 8 2 2 2 2 0 2 8 0 0 0 0
0 0 8 2 2 2 2 1 2 8 0 0 0 0
0 0 8 2 2 2 2 2 1 8 0 0 8 0
0 6 8 2 2 4 2 2 2 8 2 0 0 0
6 8 2 8 8 8 8 8 8 8 1 8 8 8
4 4 8 6 6 6 2 6 6 8 3 3 3 3
4 4 8 6 6 6 6 1 6 8 3 3 0 8
8 8 8 8 4 8 8 8 8 8 8 8 8 8
0 0 8 1 1 4 1 1 1 8 0 0 0 0
0 0 0 1 1 1 1 1 1 8 0 0 0 0
0 0 8 1 1 1 1 1 8 8 0 0 6 0
0 0 8 1 1 1 0 1 1 8 0 0 0 6
Below is a test input grid. 
Predict the corresponding output. 

Input:
0 0 0 8 0 0 0 0 8 0 0 0 0 0 0
0 0 0 8 0 0 0 0 8 0 0 0 0 0 0
0 0 0 8 0 0 0 0 8 0 0 0 0 0 0
0 0 0 8 0 0 0 0 8 0 0 0 0 0 0
0 0 0 8 0 0 0 0 8 0 0 0 0 0 0
0 0 0 8 0 0 0 0 8 0 0 0 0 0 0
8 8 8 8 8 8 8 8 8 8 8 8 8 8 8
0 0 0 8 0 0 0 0 8 0 0 0 0 0 0
0 0 0 8 0 0 0 0 8 0 0 0 0 0 0
0 0 0 8 0 0 0 0 8 0 0 0 0 0 0
0 0 0 8 0 0 0 0 8 0 0 0 0 0 0
0 0 0 8 0 0 0 0 8 0 0 0 0 0 0
0 0 0 8 0 0 0 0 8 0 0 0 0 0 0
8 8 8 8 8 8 8 8 8 8 8 8 8 8 8
0 0 0 8 0 0 0 0 8 0 0 0 0 0 0
0 0 0 8 0 0 0 0 8 0 0 0 0 0 0
0 0 0 8 0 0 0 0 8 0 0 0 0 0 0

\end{lstlisting}
\end{tcolorbox}
\end{paracol}
\end{document}